# PREDICTING LETHAL OUTCOME (CAUSE) AND UNDERSTANDING KEY BIOMARKERS LINKED WITH ACUTE MYOCARDIAL INFARCTION USING DEEP ARTIFICIAL NEURAL NETWORK AND ENSEMBLE OF MACHINE LEARNING METHODOLOGIES.

SAGNIK GHOSH

Master of Science (MSc), Thesis Report

MARCH 2022

## ABSTRACT

Cardiovascular disease is still one of the main causes of death around the world. Acute myocardial infarction (MI), or heart attack, claims millions of lives each year. MI happens when blood flow to the coronary arteries is blocked or reduced, which causes permanent damage to the heart muscle. Without treatment, this can lead to cardiac arrest, where the heart stops pumping blood to the organs, resulting in organ failure and death. Even survivors often face serious problems like heart failure, pulmonary edema, and asystole. Research shows that 5–10% of survivors die within the first year after an MI, and nearly half need to be hospitalized again. Early thrombolytic treatment leads to better outcomes, so there is a clear need for faster and more accurate ways to diagnose MI.

Right now, doctors usually review patient history and use their own experience to find the causes of MI. This process takes a lot of time and can be inconsistent. Detecting MI accurately and quickly can help patients take better care of themselves and prevent fatal events.

In this study, we introduce an automated model to predict deadly outcomes of MI and help doctors understand important biomarkers linked to its complications. This approach aims to make diagnosis clearer, faster, and more affordable. The process includes preparing the data, filling in missing values, and handling imbalanced data using SVM-SMOTE, ADASYN, and class-weighted methods. We use wrapper and embedded feature selection to find the most important variables, then scale the features for consistency. The model combines Logistic Regression, Random Forest, Light-GBM, and Bagging SVM, and is further improved with an artificial neural network to increase accuracy. We evaluate all models using precision, recall, and other key measures to find the best option for clinical use.

# TABLE OF CONTENTS

# LIST OF TABLES

## LIST OF FIGURES

## LIST OF ABBREVIATIONS

ACS ……............................Acute Coronary Syndrome
ACVI…..............................Acute Cerebrovascular Accident
AdaBoost …........................Adaptive Boosting
ADASYN…………………Adaptive Synthetic Sampling
AH….................................Arterial Hypertension
AI…...................................Artificial Intelligence
ANN…..............................Artificial Neural Network
ANOVA….........................Analysis of Varianc
AUC…..............................Area under the Curve
AV…................................Atrioventricular block
BIHS….............................Bangladesh Institute of Health Science
BMI…...............................Body Mass Index
BP…..................................Blood Pressure
CAD…..............................Coronary Artery Disease
CAN…..............................Cardiovascular Autonomic Neuropathy
Catboost …........................Category Boosting
CGB…...............................Category Gradient Boosting
CHD…..............................Coronary Heart Disease
CHF…...............................Chronic Cardiovascular Failure
CNN…..............................Convolutional Neural Network
COPD…............................Chronic Obstructive Pulmonary Disease
CVD….............................Cardiovascular Diseases
DT…................................Decision Tree
DWT…............................Discrete Wavelet Transform
ECG…..............................Electrocardiogram
ECG…..............................Electrocardiogram
ECG-VIEW II…..................Electrocardiogram Database
EDA…...............................Exploratory Data Analysis
ESR…...............................Erythrocyte Sedimentation Rate
FC…..................................The Functional Class of angina pectoris
FN…..................................False Negative
FP…...................................False Positive
GA…..................................Genetic Algorithm
GB…..................................Gigabyte
GTX….............................Giga Texel Shader Xtreme
HCM…............................Hypertrophic Cardiomyopathy
HF….................................Heart Failure
HF….................................Heart Failure
HRFLC…..........................Hybrid feature selection technique for prediction of cardiovascular disease

HRV…..............................Heart Rate Variability
ICU….................................Intensive Care Unit
IOT…..................................Internet of Things
KILLIP…..........................Killip Classification Score
KNN…...............................K-Nearest Neighbour
LASSO…...........................Least Absolute Shrinkage and Selection Operator
LBBB…............................Left Bundle Branch Block
LGBM…...........................Light Gradient Boosting Machine
LR…..................................Logistic Regression
MACE…............................Major Adverse Cardiovascular Events
MATLAB…......................Matrix Laboratory
MI…..................................Myocardial Infarction
ML…................................Machine Learning
MLP…...............................Multilayer Perceptron
MRI…................................Magnetic Resonance Imaging
NAN…...............................Not A Number
NHIRD…...........................National Health Insurance Research Database
NHLBI…............................National Heart, Lung, and Blood Institute
NSAID…............................Non-Steroidal Anti-inflammatory Drugs
PTBDB…...........................Physikalisch-Technische Bun desanstalt Diagnostic ECG Database
QRS…................................QRS complex in ECG IU is international unit
RAM…...............................Random Access Memory
RBBB….............................Right Bundle Branch Block
RELU…..............................Rectified Linear Activation Unit
RF..................................... .Random Forest
RFE…................................Recursive Feature Elimination
ROC…...............................Receiver Operating Characteristic
RUSBOOST…....................Random Under Sampling Boosting
SMOTE…...........................Synthetic Minority Oversampling Technique
SVM…...............................Support Vector Machines
TB…..................................Terabyte
TN…..................................True Negative
TP…...................................True Positive
WEKA…............................Waikato Environment for Knowledge Analysis
XGB…...............................Xtreme Gradient Boosting
XGBoost…........................Xtreme Gradient Boosting

# CHAPTER 1

# INTRODUCTION

## 1.1 Background of the Study

The World Health Organization claims that, around 17.9 million people died from CVDs in 2019 and among which 85% of all deaths occurred due to heart attacks or strokes or of severe MI. Around one third of this fatality occurs in people below 70 years of age. Greater than 75% Of CVD deaths occur in low- and middle-income countries. It is estimated that 23.6 million people will die from CVDs by 2030 if current trend continues. The death rate due to cardiovascular disease is increasing rapidly worldwide in recent years (World Health Organization, Cardiovascular diseases (CVDs), 2021). Heart attack or myocardial infarction or myocardial ischemia is generally occurring in an individual when oxygenated blood flow stops or decreases to the coronary arterial blood vessels (generally called myocardial tissues), causing damage to the cardiac muscle. If an area in coronary artery bursts, blood gets clotted at the site of the damage, and when this clot becomes larger in size, it completely or partially blocks the flow of oxygen rich blood to the myocardial tissues. An acute myocardial infarction can cause heart failure or cardiac arrest, in which the pumping action of the heart is damaged, causing transmission failure of enough or no blood to the vital organs, hence causing failure of those organs, which might lead to death (Signs and Symptoms of Coronary Heart Disease - National Institute of Health, NHLBI, 2016). CVDs (cardiovascular diseases) are group of heart and blood vessel problems, that include coronary heart disease, peripheral arterial disease, rheumatic heart disease, congenital heart disease, and deep vein thrombosis and pulmonary embolism. Fat deposits which are created on the inner walls of the blood arteries that supply blood to the heart or brain are the most common cause of heart attack. An unhealthy diet, physical inactivity, tobacco, and alcohol use are the most common risk factor for cardiac disease and stroke. Healthy diet, physical exercises and reduced use of tobacco or alcohol will slow down the risk of cardiovascular disease. Common symptoms of the heart attack or stroke may include pain in the middle of the chest, discomfort in the arms, back, jaws, and shoulder. Person suffering from CVDs may also have trouble breathing or shortness of breath (World Health Organization, Cardiovascular diseases (CVDs), 2021) . According to study (Piros et al., 2020) , in United States and other countries, CVDs not only create serious health problem but also put an immense load on the country's economic condition. Patients in low- and middle-income nations frequently lack access to primary health-care programmes that allow for the early detection and treatment of people with CVD risk factors. People with CVDs in low- and middle-income countries have reduced access to effective and equitable health care services that meet their needs. As a result, in many of these nations, disease identification occurs late, and people die of CVDs at a younger age. This study also mentioned that out of the top ten leading causes of death in the United States, deaths due to heart disease always topped the list. Surgical operation is required in some extreme cases after acute myocardial infarction.

The authors of a study (Keya et al., 2021) investigate about estimating the probability of a heart stroke in a cardiac patient. Researchers also analyse in this study that 65 years for men and 72 years for women is the average age at which an initial heart attack can occur. About 7,35,000 U.S. citizens died of acute myocardial infarction last year. Smoking, high blood pressure, high cholesterol levels, diabetes, obesity, and hypertension are some of the vital behavioural risk factors which are responsible for causing heart attacks in men and women. As a future scope, advanced ML, or deep learning techniques, can be used for the prediction of MI in a heart patient. A study (Kashirina et al., 2021) carries out based on some important features of cardiac patients like gender, age group, arterial hypertension, chronic obstructive pulmonary disease, diabetes mellitus, severity according to KILLIP, etc. to identify the vital risk factors for casualty in them after MI. Finally, researchers identify that the severity indicator on the KILLIP scale, age, the existence of chronic heart disease, and diabetes mellitus are some of the responsible factors that influence the mortality rate in a patient after acute myocardial infarction. The death rate due to heart stroke, or myocardial ischemia, is increasing rapidly in Bangladesh every day. Research carries out to predict MI diseases in heart patients using some unique and non-unique features related to patients' health who are suffering from the disease. Some unique risk factors of a cardiac patient include weakness, weight loss, and headaches for blood pressure that are used for this study. With the help of some medical reports, it is a challenging task to scrutinise the cause of MI in patients, hence the authors came up with a smart ML based system for predicting myocardial infarction. Future scope for this study can include the use of ANN, MRI and radiology image processing using CNN for better predictive power of the proposed model (Kayyum et al., 2020). A new idea in (Nasimov et al., 2021) is used to differentiate between MI and cardiomyopathy using deep learning methodologies. Cardiomyopathy is a heart ailment which occurs even in people who don't have any heart related disease like BP, coronary artery necrosis etc. Cardiomyopathy is heart disease, which is distinguished by enlarged, rigid or thick cardiac muscles. This study is implemented to accurately predict the MI and cardiomyopathy syndrome in a cardiac patient using their ECG and ultrasound images. Finally, as a future prospect, 12-lead ECG data, captured during 5-6 time periods a day, can be used to increase the prediction power of the proposed model.

In most of the MI cases, doctors take decision based on their knowledge and experience by examining the patients' medical report manually, which is a challenging and time-consuming task. Another demanding task for a physician is to determine all significant factors affecting the survival after acute MI and detects its lethal outcomes early in a cardiac patient. Therefore, purpose of the study is to provide an accurate automated model which can predict the lethal outcomes after myocardial infarction in a cardiac patient. Also, it will help the physician in understanding the vital biomarkers associated with acute MI for transparent diagnosis of its complications. This model will help the doctors and patients for carry out the preventive measures after acute MI in a timely and cost-effective manners.

### 1.2 Problem Statement

According to (Department of Health and Human Services and United States federal agency, 2021) , when cholesterol forms in the arteries of an individual with cardiovascular disease, the inside of the vessels narrows, reducing or blocking blood flow. Plaque has the potential to burst the vessels as well. In the United States, cardiovascular disease is the top cause of mortality for males, females, and people of most races and ethnicities. In the United States, one patient dies from cardiovascular disease every 36 seconds. Cardiovascular disease claims the lives of around 659,000 adults in the U.S. yearly, accounting for one out of every four deaths. Between 2016 and 2017, the United States spent $363 bn on myocardial infarction. This includes the costs of medical services, medications, and foregone productivity and effectiveness of mortality. In 40 seconds, somebody in the United States has a cardiac arrest. In the United States, over 805,000 individuals have a cardiac event each year. Therefore, it is very important to predict the heart attack or myocardial infarction or stroke or any cardiovascular disease at earliest in a cardiac patient in order to timely carry out the necessary care and preventive measures.

In a research study (Liang et al., 2021) , author has discussed about the heart failure in a patient after acute myocardial infarction. This study has been carried over on 9,985 patients from Taiwan's NHIRD. The study has been applied on two different age groups of patients with myocardial infarction, one group with less than 71 years (median age) of age and other group with greater than 71 years (median age) of age. The main aim of the research is to predict whether a patient with acute myocardial infarction would likely to suffer from heart failure and to find out the different factors which are responsible for transitioning towards HF after acute MI in a cardiac patient. Some supervised ML based models has been used for prediction and comparative study has been done. The limitation of this paper is that, if there are more features associated with a patient like living habits or their hospital visit information etc. then the accuracy of the classifying model can be enhanced further.

The authors in similar research (Zheng et al., 2021) have proposed a Stacking ML model to detect the presence of Major Adverse Cardiovascular Events in patients with intense cardiac ailment. This study has been carried over 13,104 patients' medical history which was collected from 52 different Korean medical institutions. The authors of this paper have implemented a smart intelligent stacking ML model to predict MACE in patients with acute coronary syndrome at an initial phase. As the dataset used in this study contains missing values, class imbalance problem and different anomalies, researchers, used different feature engineering and selection techniques to pre-process the data. The limitations that have been identified in this study like the dataset is limited to Korean population and this research cannot be extended to other regions, and secondly, authors should focus on implementing deep neural based model (ANN) for the MACE prediction as well. Also, instead of SMOTE Tomek hybrid technique of handling imbalance data, some enhanced version of SMOTE like borderline SMOTE or SVM-SMOTE can be use in the future study, which may improve the quality of the synthetically generated datapoints.

In another research (Piros et al., 2020), authors have discussed about prediction of myocardial infarction in Hungarian population. The origin of the dataset is Hungarian Myocardial Infarction Registry which contains 47,391 patients' information with MI. The primary aim of this study is to do a comparative analysis between two tree-based ML algorithms and conclude which machine learning models perform better in predicting 30-days and 1-year of mortality of individual admitted in hospital with intense MI. Authors have used missing values imputation methods using Fully Conditional Specification and Bayesian linear regression, and finally created 5 subsets of data on which further modelling techniques are applied. Some of the vital factors has also been determined from this study namely, age, cardiogenic shock, hyperlipidaemia, and level of creatinine for 30-days mortality and hyperlipidaemia, peripheral artery disease and percutan coronary intervention for 1-year of mortality. Hardware and time requirements for algorithms like RF is higher than DT, hence this can be improved as a future scope of this study and identified as one of the problems in this study.

Authors has done some investigation on myocardial infarction in (Kayyum et al., 2020) study, with the help of ML methods using some unique and non-unique features related to patients' health, suffering from MI. The data for the study consists of 345 instances with 26 features that has been collected from five different hospitals in Dhaka, Bangladesh. Some of the unique features are weakness, weight loss, headache for blood pressure etc. related to MI patients which has been used for this study. It is very complicated task to scrutinize the cause of MI in patients merely with the help of some medical reports. Therefore, researchers have aimed to develop a smart ML based system for predicting myocardial infarction. In this research, authors used two unsupervised filters namely, WEKA 3.8.3 (Waikato Environment for Knowledge Analysis) and randomized filters to replace the missing values with mean or mode for each feature. Bagging classifier, LR and RF algorithms has been used for prediction in this study. Limitations that have been identified and can be addressed as a future scope include the use of algorithms such as ANN with Neuron Fuzzy Inference System, CNN, and, based on the best performing model, an intelligent ML technique for MI prediction can be developed to identify cardiac arrest more efficiently.

In proposed study (Richards et al., 2021), authors have spoken about a novel framework called Logistic relapse that has been employed for analysis and prediction of myocardial casualty in patients. In this study researchers have used some open world dataset, along with medical data from different hospitals. The primary purpose of the research is to build a Logistic ML model which will efficiently detect MI strokes in a cardiac patient. Some of the vital pre-processing steps which authors have used in this study to improve the quality of the data, includes missing value treatment, data imbalance treatment, data normalization and most importantly used clustering techniques to label the unlabelled data collected manually from different patients. Author believes that this study will help the medical society to detect the presence of MI in a heart patient and will minimize the time required for manual detection of the disease by eradicating the human error to an extent. Authors has used SMOTE method for imbalance treatment in this study, but more improved imbalance treatment methods are available, which can be used for better model performance, this has been identified as one of the problems in this study.

Authors in study (Pavithra and Jayalakshmi, 2021), has done a relative examination of different ML classification methods to predict the acute coronary disease using an unprecedented method of Hybrid feature selection technique using HRFLC algorithm. The HRFLC algorithm is a combination of RF, AdaBoost, and Pearson statistical correlation coefficient technique. This study has been carried over 303 specimen whose record has been collected from an open-source repository. The main aim of the study is to select some vital features, those are responsible for CVD using HRFLC algorithm and finally, those features are passed to ML pipelines to efficiently predict the occurrence of heart disease in a cardiac patient. Limitations found in this study can be addressed by using ML-based imputation techniques like iterative or KNN imputers to impute missing value records rather than deleting them as a data pre-processing step.

An innovative scoring method has been proposed in a study (Wu et al., 2020) to predict MACE in a cardiac patient suffering from intense chest agony using ML techniques. Data of 1337 patients suffering from chest pain has been collected from Sanchong and Banqiao hospital in Taiwan, which has been used to train the machine learning models. The primary goal of this study is to detect the presence of MACE, acute MI, unstable agina pectoris and revascularization using automated ML methods. Using recursive feature elimination technique, Pearson correlation, logistic regression, and ridge regression methods, 5 attributes has been selected out of 37 features. SVM, RF and ANN algorithms has been applied on the selected invasive (creatinine and troponin I) and non-invasive (age, number of CAD risk factors, and QTc) variables. Finally, it has been observed that, based on this novel risk scoring system, full model performs better in predicting MACE in a heart patient than reduced model, which is faster on the other side. Authors could have use tree-based feature selection method like extra tree classifier or feature importance-based algorithms for selecting vital features for modelling, which might have provided a better accuracy than the methods that researcher has used in the study.

An automated detection of myocardial rupture using ML methods after intense MI has been discussed in another research (Azwari, 2021). The dataset contains record of around 1800 patients admitted to an inter district Russian hospital due to acute myocardial infarction. The main intention of this study is to find out the vital features and analyse their roles in causing the heart attacks. Missing value imputation and data normalization has been used as a pre-processing step. It is very interesting to know that the author in this study has used each feature without any feature selection methods applied, just to understand about the importance of every feature (105 input features) in predicting cardiac rupture after acute MI. Researcher conclude that not only few vital factors, but other numerous factors are responsible for acute MI in a cardiac patient which cannot be simply ignored as a part of feature selection techniques. In this paper, authors have used Min-Max scaling for scale transformation but as a future scope of this study, standardization technique can be used for scaling which might improve the predictive power of the ML model along with the use of other sophisticated technique of imbalance handling of target variable.

Imbalanced data is a common problem in medical data that has yet to be properly addressed by any of these previous works. Most prior studies on cardiovascular disease prediction, particularly for myocardial infarction, used datasets with severely unequal distribution of target class. Some research (Richards et al., 2021; Zheng et al., 2021) used a fairly crude method of treating imbalances, however more advanced imbalance treatment methods such as SMOTE, class weighted methodology, and others can be applied to improve performance of the model. Other similar publications (Azwari, 2021; Pavithra and Jayalakshmi, 2021) did not manage missing values adequately, and in order to address missing values, they erased the records. As a result of the potential for information loss, a more effective missing imputation technique must be adopted in order for the model to work efficiently. In addition, an effective feature selection technique can be used to discover the main biomarkers linked to myocardial infraction, which could aid the physician in identifying the critical variables that influence disease decision-making. The use of critical predictor factors and the development of a prediction model employing properly class-balanced data, properly missing value imputed data, will allow for a more positive decision-making process during myocardial infarction diagnosis.

## 1.3 Aim and Objectives

The primary aim of this study is to develop a model to predict the lethal outcomes (cause) and understanding the significant biomarkers linked with acute myocardial infarction in a cardiac patient. The identification of the fatal complications after intense myocardial ischemia using well-studied risk factors is necessary to carry out timely and cost-effective preventive measures. The research objectives are formulated based on the aim of this study, which are as follows:

- To analyse the relationship between the biomarkers through visualizations for improving the understandability of the diagnosis for physician and patients.

- To suggest a suitable missing value imputation and balancing technique that can be applied on the imbalance dataset.
- To identify the vital risk factors linked with severe myocardial infarction in a heart patient via suitable feature selection methods.
- To compare between the predictive models to determine the most accurate model for classifying the fatal complications after acute myocardial ischemia based on various evaluation metrices.

### 1.4 Research Questions

For each of the research objectives indicated below, the following research questions are proposed:

1. Will the identified biomarkers and demographics that are associated with a cardiac patient after acute myocardial infarction, improves the understandability of the diagnosis?
2. Which missing value imputation and balancing technique can be applied on the selected dataset?
3. Which predictive classification model can be implemented to identify the fatal outcome (cause) more accurately after intense myocardial ischemia based on various evaluation metrices?

### 1.5 Scope of the Study

This study is to propose a model to identify the lethal outcome (cause) and determine the vital factors linked with myocardial infarction in a cardiac patient. The objective of this research is to develop a model for determining the lethal outcome (cause) and vital factors associated with myocardial infarction in a heart patient. The current work is limited to presented myocardial clinical dataset, however, applying this model to other medical domains or datasets will necessitate a distinct training and modelling strategy. In addition, the implementation of this model for image datasets is beyond the scope of this project. For missing value imputation, iterative imputation technique will be used, as this technique uses integrated ML models to detect the missing values in continuous columns and provide more accurate prediction of those values.

For imbalance handling, SVM-SMOTE, ADASYN and class weighted method will be used, which is an enhanced variation of SMOTE and borderline SMOTE. SMOTE has a disadvantage of increasing the overlapping of the classes, as it does not consider majority class instance as a nearest neighbour, on the other hand the downside of borderline SMOTE is that more data is synthesized near the class boundary region. Hence, keeping all these disadvantages of SMOTE and borderline SMOTE in mind, the SVM-SMOTE, ADASYN and class weighted method of imbalance handling technique has been proposed, which is able to solve the above discredits to a certain extent.

For the feature selection method, RFE, Elastic Net and Extra Tree classifier will be implemented. The Recursive Feature Elimination (RFE) technique will be used, as it is very popular, easy to use, easy to configure, and more efficient in selecting the relevant features for predicting the target variable. Elastic Net regularization feature selection methods will be used, as LASSO technique use to randomly select one of the multicollinear variables and ignore the rest, on the other hand, RIDGE technique does not reduce the number of attributes, as it does not diminish the coefficient value completely to zero, only minimizes it. Inabilities of both LASSO and RIDGE techniques has been taken care off in elastic net method. Extra Tree classifier consist of inbuilt feature important mechanism and will be used as feature selection method, as it performs almost similar like random forest, but it is much faster compared to it, this is because Extra Tree randomly selects the cut points for every feature for optimum splitting of the node, rather than calculating Gini for every cut point and deciding the split based upon it.

For predictive modelling, LR, RF, LGBM, Bagging SVM classifier and Stacking classifier will be used. LR model will be implemented, as it is simple, and the model interpretation achieved by it is very appreciable. Different combinations of bagging and boosting ensemble architecture will be used individually along with its stacking blending approach, this will harness the capabilities of individual models and combines them together to provide an accurate prediction for a classification task. In addition, the modelling methods discussed above, ANN deep learning methodology will also be used in this study to provide a comparative and competitive analysis with the above discussed predictive models.

Usage of other missing value imputation, feature selection, data imbalance treatment and modelling approaches than the one discussed above is out of scope for this study.

## 1.6 Significance of the Study

In the urban world, MI, or acute myocardial infarction, precisely known as heart attack, is the deadly problem and one of the leading causes of death affecting millions of people. Acute MI cause an irrecoverable damage to the cardiac muscle due to lack of oxygen. Myocardial infarction can cause many serious complications to a cardiac patient like asystole, myocardial rupture, pulmonary edema etc. which are some of the lethal outcomes after MI in a heart patient. According to the World Health Organization, 17.9 million people died as a result of CVDs in 2019, death due to heart attacks, strokes, and severe MI accounted for 85% of all deaths. Around a third of these deaths occur in adults under the age of 70. Low- and middle-income nations account for more than 75% of CVD mortality. If current trends continue, it is anticipated that 23.6 million people will die from CVDs by 2030 (World Health Organization, Cardiovascular diseases (CVDs), 2021) . According to study (Piros et al., 2020), in United States and other countries, CVDs not only create serious health problem but also produce immense load on country's economic condition. They have also mentioned that out of top ten leading causes of death in United States, deaths due to heart disease always topped the list. Death due to heart attack is increasing rapidly day by day in Bangladesh, and in most of the MI cases, doctors take decision based on their knowledge and experience by examining the patients' medical report manually, which is a challenging and time-consuming task(Kayyum et al., 2020). The most challenging task for a medical person is to determine all significant factors affecting the survival after acute MI and detects its deadly outcomes early in a cardiac patient.

This study aims at providing an accurate automated model which can predict the lethal outcomes after myocardial infarction in a cardiac patient. Also, it will help the physician in understanding the vital biomarkers associated with acute MI for transparent diagnosis of its complications. This model will help the doctors and patients for carry out the preventive measures after acute MI in a timely and cost-effective manners.

There are several studies that have been done to create a machine learning model to detect the cardiovascular disease in a cardiac patient, but most of these involves an early detection of the disease, heart failure, chest pain or myocardial rupture in a heart patient. But there are several others fatal outcomes like, cardiogenic shock, pulmonary edema, asystole, thromboembolism etc. and factors those are linked with acute myocardial ischemia that need to be determined for an efficient diagnosis. Also, for a better prediction of the MI disease's deadly outcome, advanced predictive modelling techniques that complement the detection process and can be incorporated on top of existing methodologies should be implemented.

## 1.7 Structure of the Study

The following is the thesis's structure. Chapter 1 describes the research's history, including facts on cardiovascular illness, notably myocardial infarction, and its impact on society, as well as problem statements. In section 1.3, the study's goal and objectives are outlined. The research question is presented in section 1.4. The study's scope is described in section 1.5. In section 1.6, the study's significance is discussed.

By methodically evaluating the machine and deep learning models used in the prediction of cardiovascular disease, particularly myocardial infarction, Chapter 2 provides the necessary theoretical framework and underlines the challenges presented in Chapter 1. The benefits of data analytics and machine learning in the healthcare industry were highlighted in section 2.2. section 2.3 explains how data mining is used to diagnose chronic diseases. The purpose of section 2.4 is to present the impact of cardiovascular disease on society and its socioeconomic implications. Section 2.5 explains how data mining can aid in the prediction of cardiovascular disease. In section 2.6, some of the historical connected research publications on cardiovascular illness are discussed. In sections 2.7 and 2.8, the related research discussion and summary of the reviews are discussed and finished, respectively.

The research approach and theoretical foundation are discussed in Chapter 3. The data set description, several data pre-processing techniques, data transformation steps, and the imbalance handling technique are all described in section 3.2 of the research methodology. Section 3.3 discusses the machine learning / deep learning models that will be employed in this project, as well as the precise methods that will be used to predict acute myocardial infarction. Finally, section 3.4 contains a summary of the chapter.

Chapter 4 contains the analysis of the data and the design of the methodology that was discussed in chapter 3. The first section of the chapter will go through the dataset that was used in this dissertation. In section 4.2, the cardiac dataset that will be used to train machine learning models, as well as the number of features and records used in analysis, and specifics about each characteristic, will be discussed. The data must be pre-processed before being fed into classifiers, as shown in section 4.3. It also involves the removal of irrelevant variables and the selection of important features, as well as the transformation of categorical to numerical features and vice versa, the identification of missing values and the selection of an appropriate imputation technique, class imbalance analysis and treatment, the preparation of train and test datasets, and feature normalization. Section 4.4 will discuss exploratory data analysis, which includes Univariate, bivariate, and correlation/multivariate analysis to identify the relationship between the target variable and other dependent aspects. The six machine learning models, as well as their hyperparameter tuning, will be described in section 4.5 which will be applied to both imbalanced and class balanced datasets. Finally, in section 4.6, all the information presented in this chapter will be summarised.

In Chapter 5, the findings of the machine learning models used in this study will be discussed. Section 5.2 will go over the important biomarkers that can be discovered using the visualization and feature selection techniques. In section 5.3, the best approaches for dealing with class imbalance will be identified, and the assessment results of each modelling algorithm (accuracy, recall, f1-score, AUC, and so on) before and after class imbalance handling will be explained. The selection of the best performing classification model will be discussed in section 5.4, which will compare the assessment metrics collected from all six models. Section 5.5 will contain a summary of the entire chapter.

In Chapter 6, the study's conclusion and future recommendations will be discussed in detail, as well as a summary of the entire research. Section 6.2 will discuss how this study will be able to meet all its aims and objectives, as well as whether any gaps or modification scope still exists. If any future recommendations exist, they will be discussed in section 6.3. This is how the report's overall research has been organized and presented.

# CHAPTER 2

# LITERATURE REVIEW

## 2.1 Introduction

We live in an algorithmic age, in which machine learning and deep learning systems have revolutionised industries as diverse as manufacturing, transportation, and governance. This chapter will explain the previous studies that were carried out on visual analytics and data mining in healthcare. This section will include a summary of the literature, beginning with data analytics/machine learning in healthcare and progressing to different data mining techniques for the prediction of chronic diseases, then a brief introduction to different cardiovascular diseases/mortality facts, and ends with data mining in the prediction of cardiovascular diseases. Machine learning/Deep learning algorithms have become inseparable from our daily lives because of their widespread use in numerous fields.

Machine learning and deep learning algorithms are now having a great impact on healthcare as well. Intelligent software is expected to assist radiologists and physicians in examining patients soon, and machine learning will revolutionise medical research and practise. Healthcare insurers may use data mining to detect fraud and abuse, healthcare organisations can use data mining to make customer relationship management decisions, physicians can find effective therapies and best practises, and patients can get better and more economical healthcare. Several studies have been conducted on the use of data mining techniques for chronic disease diagnosis, such as cardiovascular disease, chronic kidney disease, thyroid cancer detection, lung cancer detection, and diabetics prediction that will be discussed in this section as well.

Aside from the above, machine learning approaches used on some of the most common types of cardiovascular disease, such as myocardial infarction, atherosclerosis disease, hypertrophic cardiomyopathy, and so on, will be thoroughly investigated to understand the current trends those are used to diagnose and predict these diseases. The creation of predictive models utilising data mining techniques and statistical methodologies will be the focus of this chapter. The final chapter will address the general findings on recent advances in visual analytics and data mining in healthcare, with an emphasis on cardiovascular diseases. Any previous studies' limitations and challenges will also be highlighted. Finally, the use of data mining to lower healthcare costs, find treatment plans and best practises, measure effectiveness, and eventually enhance the standard of patient care could be crucial in the future of healthcare.

## 2.2 Data Analytics and Machine Learning in Healthcare

According to big data as a service framework (Wang et al., 2017) medical data digitization changes the dimensions of data and expands the size of data, making data analytics more important. Volume, velocity, variability, and value are all attributes of big data. Big data analytics is not compatible with traditional data analysis methodologies. The tools and techniques used in big data analytics are unique. Medical information is private and comes in a variety of formats. They are as follows: electronic health records, claims data, patient-disease information, health surveys, and clinical trials. Patients with anomalies and symptoms that are plainly visible can be treated using current healthcare programs. Early detection of severe diseases aids in the treatment of the patient, lowering the danger. Otherwise, the patient will develop chronic illnesses and may even die. Annually, 59% of deaths are due to delayed detection of serious diseases.

Based on research paper (Bhardwaj et al., 2017), doctors used to treat patients solely on their symptoms, but now they're starting to diagnose and treat patients using a concept called as evidence-based medicine. This entails analyzing enormous volumes of data from clinical trials and other treatment pathways on a wide scale and making judgments based on the most up-to-date information. A physician can use big data technologies to look at national trends to see what course of therapy would be best for the patient and prescribe the appropriate meds. Individual data sets that would otherwise be worthless are combined to offer clinicians with the facts they need to make better, more holistic medical judgments.

Patients are given options based on their own diabetes diagnosis. It's a 5G smart diabetes testbed with wearable technology similar to smart clothes. It is customizable to the patient's needs, comfortable to wear, long-lasting, and cost-effective. The performance of the system is validated using an integration of SVM, ANN, and decision tree techniques (Chen et al., 2018). According to paper (Hossain and Muhammad, 2018), to determine the patient's medical status, an emotion-aware connected healthcare big data framework is being created. The patient's speech and image are examined, and their mood is determined depending on their emotion. SVM classifiers are used for voice and video processing with the Fourier transformation. This framework achieves a level of accuracy of 99.87%. To monitor the patient's health status, a health data visualisation tool is proposed (Galletta et al., 2019).

In healthcare's Big Data (Vuppalapati et al., 2016), the electronic health record (EHR) is a database that stores electronic health data about individual patients. Demographics, medical history, prescription and allergy information, immunisation status, laboratory test results, and personal information such as age and weight are all stored in the EHR. It aids in the management of patient data with high availability and protection against unauthorised access.

The major types of machine learning and deep learning that can be applied in healthcare applications are briefly outlined below (Simeone, 2018) :

1. Unsupervised Learning - Unsupervised learning approaches are machine learning approaches that utilizes unlabelled data. Data point grouping using a similarity measure is one of the most extensively used unsupervised learning methods.
2. Supervised Learning - The use of labelled training examples to create or map the link among inputs and outputs is referred to as supervised methods. The approach is termed classification if the output is discrete, and regression if the output is continuous.
3. Semi-Supervised Learning - When both labelled and unlabelled samples are available for training, such as a small amount of labelled data and a big number of unlabelled data, semi-supervised learning approaches are beneficial. Because obtaining a sufficient amount of labelled data for model training is difficult in healthcare, semi-supervised learning approaches can be particularly effective for a range of healthcare applications.
4. Reinforcement Learning - Reinforcement learning methods are those that learn a policy function from a series of observations, actions, and rewards in response to actions taken over time. Many healthcare applications could benefit from Reinforcement learning, and it has recently been employed for context-aware symptom assessment for illness diagnosis.

Machine Learning is a type of data analytics that automates model creation. Machine learning can identify hidden insights from data by learning to use specific algorithms; it's vital to emphasise that nobody tells the machines where to look in machine learning. Machine learning's iterative nature allows it to adjust its techniques and outputs when new scenarios and data are presented to it (Bhardwaj et al., 2017) . In research (Liu et al., 2015), which tells about FIG, which stands for Fisher criterion and genetic optimization, and it is used to better recognise diseases such as lung cancer. It is both efficient and effective in terms of computation. For detecting false positives in the scan, multiple streams of multi-view 2D convolution networks are used (Setio et al., 2016). When combined with a CT scan of the lungs, 3D CNN performs better. In terms of false positive detection, this technique has a sensitivity of 85 to 90%. Algorithm with high-order back propagation. Gamification has been found to be effective in predicting breast cancer. It's ideal for working with medical imaging databases. The use of 3D Rieszwavelets (Cirujeda et al., 2016), which perform better in highlighting inter and intra variations in lung CT, is employed. In terms of forecasting illness recurrence, it has an accuracy of 81.3 to 82.7%. This approach can predict NSCLC, the disease with the highest death rate. For detecting voice pathology, a machine learning algorithm can be utilised. Transfer learning is utilised in conjunction with the existing CNN method (Alhussein and Muhammad, 2018), resulting in an increase in accuracy of up to 97.5% in the Saarbrucken voice disorder database. Convolutional auto encoder was used to predict patient survival in lung cancer patients (Wang et al., 2018). The patients are divided into two groups, high risk and low risk, using Kaplan-Meier analysis and cox proportional hazards. For feature selection and model development, the LASSO-Cox model was utilised. Deep multi-view CNN considers (Hussein et al., 2017) high-level significant features for tumour detection such as calcification, sphericity, and texture.

## 2.3 Diagnosis of Chronic Diseases using Data Mining Techniques

Data mining is among the most helpful approaches for retrieving vital information from large sets of data by enterprises, researchers, and individuals. Knowledge Discovery in Databases is another name for data mining. Data cleansing, data integration, data selection, transformation of data, data analysis, pattern evaluation, and cognitive understanding are all part of the knowledge discovery process. In other words, data mining is the process of examining hidden knowledge from multiple viewpoints for classification into useful data, which is collected and compiled in specific areas such as data warehouses, authentic assessments, data mining algorithms, assisting strategic planning, and other data requirements, with the goal of ultimately cost-cutting and revenue creation (Umadevi and Marseline, 2018).

In healthcare, data mining offers a lot of potential for reforming the system. It employs data and analytics to gain a better understanding of the situation and to develop best practices that can improve health care services while lowering costs. Machine learning, cross dimensional databases, visualisation of data and stats are some of the data mining techniques used by analysts. Patients in each group can be predicted using data mining. The protocols ensure that patients receive aggressive treatment at the appropriate location and time. Healthcare insurers can also employ data mining to prevent fraud and misuse. According to research (Bahri et al., 2019), laboratory findings, prescriptions, visits, machine produced data, insurance information, and assessments data are just a few examples of the types of data produced in the medical industry. In fact, only clinical trial data in the United States topped 150 exabytes in 2011. Clinical information is expected to reach zettabytes or possibly yottabytes soon, according to projections. Prescription records, test results, and ECG data, according to the authors, account for 90% of all clinical evidence.

In a proposed study (Devika et al., 2019) , authors have used data mining classification methods and machine learning algorithms for detection of chronic kidney disease. Chronic kidney disease is characterized by an abnormal renal function or a gradual decline of kidney function over months or even years. People who are at risk of renal problems, such as those with hypertension, cardiac disease, or high insulin levels, or those who have a biological family with CKD, are frequently subjected to diagnostic. It differs from acute renal disease in that the decrease in kidney function must last longer than three months. As a result, one of the most important responsibilities is to detect chronic renal disease. In this study, data mining classification approaches have been used to forecast chronic kidney disease. Naive Bayes, K-Nearest Neighbor (KNN), and Random Forest classifiers were employed in this research. Their accuracy, precision, and F-measure have been used to evaluate their performance. Out of all these classification techniques, random forest has provided a better performance than the rest.

In another research (Saranya and Sasikala, 2020) , authors have used different data mining techniques for detecting malignant breast cancer cells. In today's world, breast cancer is the most common disease amongst women. Every year, 2.1 million women are affected, and it also causes the highest number of cancer-related deaths among women. The goal of advanced analysis processes is to provide timely access to cancer treatment while also reducing barriers to quality diagnosis solutions. As per the 2018 mortality statistics, 6,27,000 females died of breast cancer, accounting for around 15% of all cancer deaths in women. Authors has employed three algorithms in this research which are Naive Bayes, Decision Trees, and Deep Learning. On three distinct algorithms, the evaluation criteria and efficiency are assessed in the form of accuracy, sensitivity, and AUC, and statistics say that the deep algorithm (ANN) is the most efficient in predicting women with breast cancer than the other algorithms.

Thyroid diagnosis is a time taking and critical task. The typical method of diagnosing thyroid disease entails a medical assessment and blood test. The essential challenges, however, is to diagnose the condition at a preliminary phase with a high accuracy rate. In medical world In study (Begum and Parkavi, 2019) , authors have discussed about the automated thyroid prediction using different data mining techniques. The primary goal is the detection of thyroid diseases with greater accuracy at a preliminary phase. Thyroid illness is a condition that affects the thyroid gland; hence the task of classification is very crucial. The goal of this study is to predict thyroid disease using various classification methodologies, as well as to determine the TSH, T3, T4 connection with hyperthyroidism and hypothyroidism. Also find TSH, T3, T4 relationship with sexual identity in hyperthyroidism and hypothyroidism. This paper considers data mining methods such as KNN, Nave bayes, Support vector machine, and ID3. These diverse methodology outcomes are based on the model's quickness, precision, and efficiency, as well as the treatment's cost. These classifications of valuable information can also aid in the discovery of more cost-effective treatments for thyroid patients.

In a recent work (P.R. et al., 2019) , authors have proposed various data mining techniques for detection of lung cancer at the earliest stage. Lung cancer is a disorder in which the tissues in the lungs grow out of control. Although lung cancer cannot be eliminated, only the risk of developing it can be lowered. The most common cause of cancer-related death is lung cancer. Cancer of the lungs can start in the throat, trachea, or lungs. The unregulated proliferation and spread of certain cells from the lungs cause it. Lung cancer is more likely to be detected in those who have emphysema or have had previous chest difficulties. In Indian men, excessive use of nicotine, cigarettes, and beedis is the leading cause of lung cancer. In this paper, classification methods such as Naive Bayes, SVM, Decision Tree, and Logistic Regression have been used to investigate lung cancer. The main goal of this study is to examine the performance of different classifiers in order to diagnose lung cancer early. Because the SVM algorithm used a large number of dimensions to classify the sample, it has the best performance. This method allows for more precise lung cancer detection.

Another study (Sonar and JayaMalini, 2019) , focused on prediction of diabetes using different data mining techniques. Diabetes is a condition in which there is a lack of insulin in the blood, resulting in deficit. Frequent urination, thirst, and decreased appetite are all warning signs of elevated blood sugar levels. It will lead to a slew of problems if it isn't treated. This adversity resulted in the death of the individual. Heart illness, foot sores, and vision blurriness are all symptoms of severe challenges due to diabetes. The focus of this research is to create a system that can more accurately estimate a patient's diabetic overall risk. Classification methods such as Decision Tree, ANN, Naive Bayes, and SVM algorithms are used to construct different mining models. As a comparative study of various modelling techniques, the algorithms have precisions of 85% for Decision Tree, 77% for Naive Bayes, and 77.3% for SVM.

## 2.4 Cardiovascular Diseases and it’s Mortality Facts

Coronary artery disease is a condition in which the coronary arteries are narrowed or blocked, usually due to the build-up of fatty substances called plaques. Coronary artery disease is also called ischemic heart disease. Plaque is built of fat, cholesterol, and other blood-borne chemicals. Atherosclerosis is the name given to the build-up of plaque in the arteries.

Below are the most common types of cardiovascular disease according to

(World Health Organization, Cardiovascular diseases (CVDs), 2021):

1. Coronary heart disease
2. Cerebrovascular disease
3. Peripheral arterial disease
4. Rheumatic heart disease
5. Congenital heart disease
6. Deep vein thrombosis and pulmonary embolism
7. Myocardial Infarction

When blood flow to the brain is obstructed, a stroke or heart attack ensues. A blockage or a ruptured blood artery in the brain usually causes this disturbance of blood flow, which prevents oxygen from reaching the brain tissue. The oxygen-starved brain cells begin to die quickly if this happens, therefore prompt treatment is critical to a patient's recovery (New York State Government, 2021).

Most serious behavioural risk factors for cardiovascular disease and infarction are poor eating, lack of exercise, tobacco smoking, and excessive alcohol intake. As a result of behavioural risk factors, people may develop signs such as hypertension, hyperglycaemia, high blood triglycerides, and overweight or obesity. The underlying disease of the blood vessels frequently produces no symptoms. A cardiac event may be the first sign of a more serious problem. The most common symptom of a stroke is sudden paralysis of the face, limb, or legs on one side of the body. (World Health Organization, Cardiovascular diseases (CVDs), 2021).

Low- and middle-income nations account for at least three-quarters of all CVD deaths worldwide. Patients in low- and middle-income nations frequently lack access to primary treatment programmes that allow for the initial detection and treatment of people with CVD risk factors. The poorer individuals in low- and middle-income nations who bear the brunt of the consequences. Evidence is accumulating that CVDs and other noncommunicable illnesses lead to poverty at the household scale due to sudden health spending and significant out-of-pocket expenses. CVDs wreak havoc on the economies of low- and middle-income countries on a socioeconomic basis.

The most common cause of fatality around the world is cardiovascular diseases (CVDs). CVDs claimed the lives of 17.9 million individuals worldwide in 2019, accounting for 32% of all deaths. Heart attacks and strokes were responsible for 85% of these deaths. CVDs were accountable for 38% of the 17 million deaths worldwide caused by non - communicable disease in 2019. It has been estimated that 23.6 million people will die from CVDs by 2030 if current trend continues (World Health Organization, Cardiovascular diseases (CVDs), 2021).

## 2.5 Data Mining Methods in Prediction of Cardiovascular Diseases

Due to the large amount of patient data stored in healthcare system, data mining is becoming more popular. Even though medical systems utilize information systems with their own databases that produce vast quantities of patient data, data mining methods can be used to uncover valuable information that can be used to resolve a few key matters in health care, such as identifying the signs of certain diseases to aid diagnostics, determining the degree to which individual medications and therapies affect patients, and generally improving the quality of care.

As per study (Imamovic et al., 2020) , cardiovascular illnesses are a primary cause of death both globally and in Bosnia and Herzegovina, thus screening, inspection, and medication are crucial. Age, sexuality, heredity, tobacco, overweight, increased blood pressure, and lack of exercise are all factors that contribute to the development of coronary heart disease. One of the challenges in healthcare is identifying individuals who are at a high risk of getting a disease in a timely manner. This article discusses the use of a decision tree, neural network, and logistic regression to develop model that can predict death within a year through chronic heart failure. The model's aim is to estimate future health outcomes of patients, such as fatality, depending on data from previous occurrences. This study has been carried out using patient information recorded at the Mostar hospital's cardiovascular unit from 2011 to 2017. Different data preprocessing techniques has been implemented like missing value imputation and converting categorical to numerical variables. The research work has proposed three methods for predicting the death due to chronic heart failure: data tree mining, neural networks, and logistic regression. The performance of the classifier was assessed using generally used measures for model evaluation like AUC score, accuracy, F1 measures, recall etc. The study found that the deep neural network is somewhat more accurate than decision trees and logistic regression in both favorable and unfavorable sample categorization. Despite the fact that all models produced excellent outcomes.

According to similar research (Alex P. and Shaji, 2019) , changes in stress levels, insulin, heartbeat, and other factors can result in cardiac illnesses such as constricted or clogged blood arteries. Heart failure, aneurysms, arterial disease, heart attacks, strokes, and even sudden cardiac death can all be caused by it. Many types of heart disease can be discovered or identified using various medical tests when family health records and other factors are taken into account. However, predicting heart disease without any clinical checks is extremely challenging. The goal of this initiative is to diagnose various cardiac illnesses and take all necessary actions to prevent them at a reasonable charge as early as possible. The identification of cardiovascular disease appears to be a significant task of data mining tools. A doctor can diagnose cardiovascular disease using this information. The aim of this research is to compare different classification techniques for heart disease diagnosis. The "Jubilee Mission Medical College and Research Institute Thrissur" provided the real-time data set for conducting this study. Data was collected by conversing with patients one-on-one. The information was also gathered from the discharge summaries of the different patients. In this approach, a total of 20 qualities from almost 2,200 patients were gathered. In this study for prediction of cardiac illnesses, data mining techniques like SVM, Random Forest, KNN, and ANN were employed. The finest and most accurate result came from ANN. According to this study, when it comes to diagnosing cardiac disorders, the ANN algorithm provides reliable results.

(Deepika and Kalaiselvi, 2018), has discussed about data mining techniques for various disease prediction including heart disease. Data mining is an important aspect of the learning response process, wherein optimization algorithms are used to extract patterns. Coronary artery disease encompasses a wide range of disorders that affect the heart, including irregular heart rhythm and abnormal heart valves. Efficient and successful computerized heart disease prediction systems can be very useful in the healthcare industry for prediction of heart disease. A total of 573 entries were included in the cardiovascular disease data set for the study. With 303 entries in the training set and 270 records in the testing set. WEKA 3.6.6 tool has been used, and the missing data were filled in with the tool's filter. Artificial Neural Networks (ANN), Logistic Regression, Decision Tree, Naive Bayes, and Support Vector Machines were some of the approaches employed in this study. Artificial Neural Networks and Decision Tree produced good result than other proposed methods. Researcher has also found that the choice of data mining tool and optimization technique influence the sensitivity and precision of the proposed methods.

Finally, data mining has contributed extensively to the resolution of several real-world problems. Classification, cluster analysis, and regression are popular strategies in this field. A myriad of obstacles has emerged recently, including class imbalance, multi-label and multi-instance problems, poor quality and/or inconsistent data, and semi-supervised learning. Whenever these out-of-the-box circumstances occur in the realm of big data, it remains undiscovered research terrain, despite an increasing effort to push the boundaries. The recent trend is to confront both traditional and newer data mining problems in big data and improve the learning process. For high-dimensional real-world applications, data mining technology in evolutionary algorithms can find an extremely effective D-optimal design. To retrieve important features from high-dimensional data and provide an acceptable clustering result in a timely manner (Ding et al., 2019) . Big data techniques and technologies' advanced analytics, as well as the constant knowledge and deep insight that can be obtained from collected Big Data, are helpful for making predictions, recommendations, medical diagnosis, allocating resources, and personalized treatment strategies. This ability has the potential to improve the performance and consequences of healthcare(Bahri et al., 2019).
Based on the studies discussed above, one can affirm the numerous opportunities provided by Big Data, data mining techniques, and data analysis.

## 2.6 Related Research Publications

Because the dataset utilized in this work is quite new, there are few related research papers based on it. Hence this section provides the glimpse some of the most impactful research papers for predicting cardiovascular disease or similar medical fields.
In the past few years, the death rate due to different cardiovascular diseases has been rapidly increasing worldwide. According to World Health Organization, around 17.9 million people died from CVDs in 2019 and among which 85% of all deaths occurred due to heart attacks or strokes (World Health Organization, Cardiovascular diseases (CVDs), 2021). Therefore, it is very important to predict the heart attack or myocardial infarction or stroke or any cardiovascular disease at earliest in a cardiac patient in order to timely carry out the necessary care and preventive measures.

The proposed paper (Kumar et al., 2020) has been utilized to investigate the demonstration of machine learning classifiers in the prediction of cardiac ailments. Cardiovascular Illness refers to conditions characterized by limited or obstructed blood vessels, which can result in a cardiac arrest, chest pain, or myocardial infarction. The illness is predicted by the classification algorithm based on the patient's state of adverse reactions. Cardiovascular disease is a major cause of mortality in the United Kingdom. The researcher for this study worked with a cardiovascular disease dataset collected from the UCI (University of California at Irvine) archive; the data set contained 10 features and 304 cases. In the proposed study, five machine learning techniques have been applied to the collected dataset: Logistic Regression, Support Vector Machine, Decision Tree, Random Forest, and K-Nearest Neighbors. The suggested technique, which employed a random forest machine learning classification model, surpassed all the other classifiers under consideration in categorizing patients with cardiovascular problems. The gaps that are identified in this paper i.e., authors have removed the missing value records as a data pre-processing step in this study, but instead of removing it, they can impute those missing values using ML based imputation technique like KNN imputer or iterative imputation algorithms for better performance. Also, a sophisticated feature selection method can be used to identify the good predictor variable for the target.

In another study by (Dinesh et al., 2018) , authors have recommended a prognostication model for predicting whether a person has cardiovascular disease and to provide information or a diagnostic test on that basis. According to World Health data, more than 12 million people die each year as a result of heart disease. In India, it is a calamitous disease that causes even more disasters. The investigation of this illness is a complicated mechanism. This should be flawlessly and accurately measured. Because of a lack of specialists in some areas, patients are in a risky situation. Cardiologists are usually the ones who detect these. When these methods are blended with the health information framework, they are immensely helpful. This paper proposed machine learning approaches for predicting the uncertainty levels of heart disease based on characteristics present in patient. The Cleveland, Hungarian, Swiss, and Long Beach VA cardiovascular disease databases were used for this research. In the proposed study, data pre-processing techniques that has been implemented i.e., the removal of outliers, missing values imputation, redundant data removal, and default values imputation, if applicable. The authors have used Support Vector Machine, Gradient Boosting, Random Forest, Naive Bayes classifier, and logistic regression in this study to display a precise model for predicting cardiovascular events. Out of all proposed algorithms, logistic regression has proven to be a best predictive technique in this study with better model performance. The gap that is identified in this paper i.e., the use of ensemble classifiers using the mix of above algorithms can provide more accurate and improved performance.

Coronary heart disease is one of the primary causes of death. To identify this heart condition, a framework capable of predicting its presence and preventing coronary heart disease, which are on the rise, would be required. Study (Louridi et al., 2019) has focused on identifying such heart condition by automating the detection and assisting a doctor in determining whether or not a person has coronary artery disease. For this study, researcher has collected patient's data from four different institution namely, the Cleveland Clinic, the Hungarian Institute of Cardiology, the V.A. Medical Centre, and the University Hospital in Zurich. For missing value imputation, author has suggested that the mean values of each column be used to fill those values. The authors have used Support Vector Machine, KNN and Naive Bayes classifier, in this research for a comparative study. Out of which linear SVM produce a good performance based on some indicators such as accuracy, precision, f1-score, and recall. Authors in this study has used mean values technique as missing value imputation which is a very naïve method of imputation. But more sophisticated imputation technique like iterative imputation can be used for missing value fill up, this will also enhance the performance of the predictive model.

The primary cause of sudden cardiac arrest in youngsters is hypertrophic cardiomyopathy (HCM), a hereditary heart condition. HCM individuals are underdiagnosed and poorly treated, notwithstanding the well-known risk factors and existing treatment guidelines. Using electronic health record (EHR) data to construct machine learning algorithm could aid in bettering HCM diagnosis and thereby improving the quality of life of thousands of patients. In study (Farahani et al., 2021) , researcher has proposed a novel predictive analytic model that uses information gleaned from prior data on similar patients to assist clinicians in formulating a better diagnostic judgment. Conventional forms of systematically reviewing health records and clinical databases were time-consuming and prone to human mistakes, but machine learning technologies allow for reliable and efficient characterization of HCM patients using EHR data. The authors employed structured data from an electronic health record's HCM cohort of 11,562 individuals with a median age of 66 years to conduct this study. As HCM-related datasets usually extremely skewed in nature, and over- or under-sampling procedures, or a combination of the two, have successfully solved this issue in the above suggested model. Random forest classification algorithm has been utilized in this study, which produced a good overall performance. As, over- or under-sampling procedures of imbalance treatment can lead to overfit or under fitment of the machine learning model, hence a better imbalance handling method like SMOTE or class weighted technique can lead to improve and more accurate performance in prediction of the HCM patients.

In another similar research (Terrada et al., 2019) , authors have focused on detection of atherosclerosis disorder using machine learning algorithms. Numerous cardiovascular conditions exist, including coronary artery disease. An arteriosclerotic disease is another name for this illness. If remain unattended, this condition can lead to a thrombus, cardiac arrest, stroke, or cardiogenic shock, with other major consequences. The incidence of myocardial patients is increasing owing to the disregard of the earliest signs. If the patient is diagnosed at the correct time, it can prevent mortality. The Cleveland Clinic Foundation at the University of California Irvine provided the dataset for this research. There are 76 features in this dataset. The Artificial neural network and K-nearest neighbor were used in this study to develop a model for diagnosing atherosclerosis illness. To classify atherosclerosis condition, researchers used k-medoids and k-means clustering. Finally, on the validation set, the results of this work were evaluated using a variety of performance evaluation methodologies, including sensitivities, specificities, accuracy, and Matthews' correlation coefficient. Deep neural network has provided a better model performance among the rest. As a future prospect, researcher can use ensemble or boosting machine learning algorithms to assess the effectiveness of the suggested system. Also, in this study authors have dropped the records containing missing values, but this may lead to information loss, hence some sophisticated imputation technique like KNN imputation method can be used to fill up the missing values, this will also enhance the performance of the predictive model.

In (Terrada et al., 2020), researchers have proposed a supervised model for predicting atherosclerosis vascular disorder. This research revolves around a Medical Decision Support System for atherosclerosis illness that can take preventative measures based on the patient's medical history. The supervised machine intelligence techniques used in this support System. In an available database, researcher has employed Artificial Neural Networks and K-Nearest Neighbor to detect patients with or without atherosclerosis condition. On Cleveland heart disease, Hungarian, Swiss, and Long Beach VA databases, the method has been confirmed. The suggested model performance has been assessed using classification evaluation metrics such as accuracy, sensitivity, and specificity. The highest accuracy achieved in the research by the suggested ANN method was 96.4%, according to simulation results from four data sets. The results of the experiments were contrasted against Weighted Fuzzy Rules, Weighted Artificial Immune System, Neural networks ensemble and Hybrid Neural Network-Genetic, among other methodologies. This comparison revealed that the proposed solution had the higher precision based on commonly used classification’s performance metrics.

South Asian countries have a higher risk of heart disease than every other ethnic community at a young age. Predicting cardiac illness is difficult for healthcare professionals since it involves expertise and knowledge, and it is a difficult assignment to complete. This sector has a massive amount of data that can be used to draw beneficial conclusions based on its hidden data. As a result, improved data analytics approaches are employed to apply appropriate outcomes and make improved data judgements. (Islam et al., 2020), has implemented such improved modelling techniques like, logistic regression, decision trees, SVM, and Naive Bayes classification for predicting cardiovascular disease. Researcher has considered University of California Irvine dataset for this study. Logistic regression provided the best performance compared to all other modelling technique that was used in this study. The limitation that this paper has i.e., author can use other voting or stacking blending modelling technique for improved performance, also proper data imbalance handling technique like SMOTE could be implemented for betterment in the accuracy of the models.

In a similar study (Katarya and Srinivas, 2020) , authors have proposed some supervised machine learning approaches for detection and prediction of cardiac illness. To address cardiac disorders, healthcare facilities and other institutions are giving costly treatments and surgeries. As a result, anticipating cardiac disease in its initial stages will be beneficial to patients worldwide, allowing them to take required treatment before it becomes serious. Artificial neural networks, decision trees, random forests, support vector machines, nave Bayes, and the k-nearest neighbor method were among the supervised data mining algorithms utilized in this study for classification and prediction of cardiac disease. Some gaps that have been identified in this study i.e., some sophisticated feature selection technique like statistical approach, information value method could be implemented to select the vital predictive variable for the target variable, better missing value imputation method and ensemble of machine learning algorithms could be used for better predictive model performance. Therefore, this can be address as a future scope of this study.

Another study (Jinjri et al., 2021) , aim to use machine learning techniques to create and find a model that optimally identifies cardiac illness and detects the occurrence of such condition in patients. As a result, the five most advanced machine learning algorithms for classifying heart disease data are compared in this research namely, Support vector machine, K-nearest neighbor, Logistic regression, Decision tree, and Naive Bayes. The dataset that was used for this study was downloaded from the Kaggle repository. Different performance variables were used to study, analyze, and evaluate the algorithms' efficiency. The most effective approach that were identified for predicting cardiovascular illness were support vector machine and logistic regression, according to the findings.

Again, (Keya et al., 2021) has demonstrated various ML classification methods in estimating probability of heart attacks in cardiac patients. The authors in this paper have also provided some statistics on cardiac arrest, like, the average age at which the initial heart attack can occur are 65 years and 72 years for men and women respectively, also, around 7,35,000 U.S. citizens has died of cardiac arrest last year. Researchers has also analyzed some of the vital elements which are responsible for causing heart attacks in men and women which includes smoking, high blood pressure, cholesterol level, diabetes, obesity, hypertension etc. ML methods including RF, Bagging, DT, MLP and LR has been used for the prediction, among which LR outperforms others based on various metrices. Initial prediction of CVDs in a heart patient using feature selection and different ML methods has been discussed in (Rashme et al., 2021). Medical information of 70,000 patients has been used as a data source for the study. Missing value imputation, outlier analysis and extraction of two new features like BMI and BP, has been done as a data pre-processing steps. Interestingly, random forest classifier has been used for vital features selection strategy. After this, algorithms like KNN, SVM (Support Vector Machine), DT, XGB and LR has been used in the research for predicting CVD in a patient. XGB performs comparatively better than other ML methods. Finally, as a future scope of this study, authors have a plan to use MRI and radiology images for the same.

In another study (Kashirina et al., 2021), researchers have proposed some machine learning techniques to pinpoint some of the risk elements for casualty after MI. This study has been caried over on 1,1457 samples of patients diagnosed with MI and admitted to a medical center in Voronezh Region. This study has been carried out based on some important feature such as COPD, ACVI, CHF, AH etc. The purpose of the research is to build a robust and efficient ML model for predicting the chance of casualty following MI by marking the most critical factors and determine how those factors affect the MI condition in a patient. The researcher used Kaplan-Meier statistical risk prediction model along with three other machine learning methods namely, Cox regression, Logistic regression, and CatBoost gradient boosting. In most cases, data imbalance treatment needs to be applied to medical data, researchers have applied the same in this study as well and after balancing the data, CGB model provides the most accurate prediction. In case of risk factors identification Cox and LR model provides the best outcome. The researcher identifies some factors such as severity indicator on KILLIP scale, age, existence of chronic heart disease, diabetes mellitus etc. as important biomarkers which has influence on the mortality rate after acute myocardial infarction.

Moving forward, to another research (Diker et al., 2018), authors have proposed a smart technique based on Genetic algorithm and support vector machine for detecting MI from electrocardiogram signals. This study has been carried out on ECG signals obtained from 290 specimens those are collected from Physikalisch-Technische Bundesanstalt Diagnostic ECG

Database (PTBDB). Out of these patient's myocardial infarction ECG recordings, 148 recordings are labelled as abnormal and 52 marked as normal. The automated recognition of the cardiac abnormality from ECG signals is a complicated task, due to dissimilar undulations and variances of the signals. Also, the optical explications of ECG waveforms are hard to identify due to its small magnitudes and length. Hence the main purpose of this study is to recognize the ECG waveforms using morphological, time-domain and discrete wavelet transform (DWT) features which requires to classify the abnormal MI samples from normal. Different feature extraction and waveform transformation technique such as Morphological and Statistical Feature Extraction has been applied to ECG signals before feeding it to ML pipelines. Author uses two sophisticated machine learning algorithms namely, GA (used for feature selection methods as well) and SVM in this study. After dimensionality reduction using feature selection mechanism (GA), which reduces 23 features to 9 features, the overall classification performance of SVM model has been increased.

In (Praneetha et al., 2021) study of predicting cardiac ailment using ML methods, researchers have used Cleveland dataset which contains 303 specimen's clinical records from 4 different countries. The goal of this study is to deploy an efficient AI model to an IOT device and over the web, so that the users can use the prediction power of the model through a software application very conveniently across different regions. Different ML methods likes RF, SVM, DT, AdaBoost, KNN, Naive Bayes has been applied. Out of all those ML methods, KNN outperform others, finally it has been deployed over the web to predict the CVD precisely.

Another proposed study (Alkhodari et al., 2021), has discussed about examining the cardiovascular autonomic neuropathy in a diabetic patient using ML techniques. For this study, dataset has been collected from Bangladesh Institute of Health Sciences (BIHS) Hospital which includes 24-hour Holter ECG recording using Shimmer3 ECG Unit for 95 patients. The researchers in the study have projected a novel approach to explore the possibility of using HRV features monitored over 24-hours and feed it to machine learning model to predict the patient suffering from CAN. Using PhysioNet toolbox and MATLAB R2021a, 25 HRV features has been extracted from subject's ECG signals. After, HRV data augmentation and imbalance treatment, these extracted features have been applied to RF, One class SVM, RUSBOOST and CNN algorithms. The best performance is achieved by CNN followed by Random Forest. As a future scope of study, authors have suggested to do further investigation on larger population across the different country and more validation needs to be done to test the productivity of the above technique.

(Nasimov et al., 2021) has concocted a new idea to differentiate between MI and Cardiomyopathy using Deep learning methodologies. Cardiomyopathy is a heart ailment which occurs even in people who don't have any heart related disease like BP, coronary artery necrosis etc. Cardiomyopathy is heart disease, which is distinguished by enlarged, rigid or thick cardiac muscles. HCM is the most life-threatening form of cardiomyopathy syndrome, which leads to sudden cardiac death specifically in younger aged people. The primary goal of this study is to accurately predict the MI and cardiomyopathy syndrome in a cardiac patient using their ECG and ultrasound images, as these two diseases have very common symptoms and hard to differentiate from each other. ECG recordings (ECG-VIEW II database) from 3,71,401 patients has been used as a data source for this study. Custom CNN architecture with Leaky RELU activation has been used to extract features from 2-D ECG data which is finally passed to ANN to classify the specific heart condition. Authors has achieved more than 90% of accuracy in predicting the disease using the above approach. As a future prospect, researcher will use 12-lead ECG data, captured during 5-6 time periods a day to increase the prediction power of the proposed model.

## 2.7 Discussion

In this paragraph, previous cardiovascular disease research, especially myocardial infarction research, will be summarized in order to convey information on data mining approaches that have been investigated for heart disease detection and prediction. Because this research is focused on a specific type of cardiovascular disease, only data mining techniques used in different cardiac illness is described. The summaries of prior cardiac disease and diagnosis studies will be found in below table 2.1. ANN, SVM, Random Forest, K-NN, Logistic Regression and Cat Boost was the most widely utilized machine learning technique in both tables. Apart from heart disease diagnosis, these classification systems are quite popular in other areas of study. The majority of previous cardiovascular illness study have concentrated on building predictive model for survivorship at various time intervals, stages or type categorization, and a computerized system for heart disease early detection.

*Table 2.1 Summary of some past studies on cardiovascular diseases*

| Disease | Author & Year | Best Modelling Technique | Performance Measure | Scope of the Study |
|---|---|---|---|---|
| **Myocardial Infarction** | (Keya et al., 2021) | Logistic Regression | Accuracy=80% AUC score=87% | Measuring of different possibilities of heart attack |
| | (Kashirina et al., 2021) | Cat Boost Classifier | Accuracy = 88% Sensitivity = 70% | Mortality prediction after myocardial infarction |
| | (Diker et al., 2018) | SVM Classifier | Accuracy = 87% Sensitivity = 86% Specificity = 88% | Smart system to predict myocardial infarction from ECG |
| | (Liang et al., 2021) | Random Forest Classifier | Precision = 77% F1 score = 82% Accuracy = 80% | Prediction of heart failure after myocardial infarction |
| | (Richards et al., 2021) | Logistic Regression | Accuracy=68% AUC score=55% | Prediction of myocardial infarction |
| **Other forms of cardiovascular disease** | (Praneetha et al., 2021) | KNN Classifier | Accuracy=87% | Cardiovascular disorder prediction |
| | (Terrada et al., 2020) | ANN Classifier | Accuracy=96.4% | Atherosclerosis disease prediction |
| | (Farahani et al., 2021) | Random Forest Classifier | Recall = 70% F1 score = 72% Accuracy = 71% | Prediction of Hypertrophic Cardiomyopathy Patients |
| | (Louridi et al., 2019) | Linear SVM | Accuracy = 86.8% | Cardiovascular illness prediction |
| **Other forms of cardiovascular disease** | (Islam et al., 2020) | Logistic Regression | Accuracy = 86.2% | Cardiovascular illness forecast |
| | (Rashme et al., 2021) | XG Boost Classifier | Accuracy = 75.1%<br>F1 score = 72.8% | Early detection of cardiovascular illness |
| | (Jinjri et al., 2021) | SVM Classifier | Accuracy = 72.6%<br>Precision = 77.3% | Classification of cardiovascular disease |
| | (Dinesh et al., 2018) | Logistic Regression | Accuracy = 91.6% | Cardiovascular illness prediction |

Looking at different cardiovascular disease research, each one employs a similar data mining technique. The majority of studies on myocardial infarction have achieved effective prediction utilizing the Logistic regression, Random Forest, and SVM techniques. However, Random Forest appears to have generated a nearly flawless detection algorithm for Hypertrophic Cardiomyopathy study. ANN, SVM, and Cat Boost have been embraced as relatively improved prediction models for other forms of cardiac diseases such as Atherosclerosis disease. One of the reasons for the differences in classifier selection could be the varied features observed in the various datasets used in these experiments. The design of the detection model will alter depending on the importance of the variables in heart disease prediction. As a result, the majority of prediction models are data source specific. Even yet, the performance obtained by prediction models created from the same data varies. This could be due to the scope of the investigation, as different pre-processing stages will be used in various studies. Variability in the resultant prediction model can be caused by feature engineering, class balance, and composite data mining techniques. As a result, previous research has demonstrated that when comparing alternative data mining techniques, it is critical to use the same methodologies on the same dataset.

## 2.8 Summary

The use of data in healthcare analytics was examined in terms of prediction and diagnosis. The use of data mining methods in the treatment of several chronic illnesses was discussed in depth. The majority of earlier data analysis studies focused on different form of cardiovascular illness diagnosis. Several sorts of cardiovascular disease prediction utilizing data mining approaches were also incorporated, and classifiers were constructed to assess the survivability rate and to detect heart disease or myocardial infarction at early stage using risk criteria. The latest patterns in cardiac illness investigations were also described using the records where data mining methods were used. Because various datasets produce different performance metrics with varied levels of accuracy, the algorithms used to create prediction model will be more dataset-dependent and reliant on the study's purview. Also, various type of cardiovascular diseases and its fatality rates has been discussed in this study along with its behavioral and socio-economic risk. The discussion revealed that previous studies differed in terms of data pre-processing procedures and dataset types used, making it difficult to compare the effectiveness of data mining approaches fairly.

# CHAPTER 3

# RESEARCH METHODOLOGY

## 3.1 Introduction

This chapter covers the research technique as well as the theoretical foundation that underpins it. It explains each stage in the approach. Its purpose is to explain the technical terminology that will be utilized across the research. Each stage of the investigation will be explained in depth in the research paradigm. Data set definition, exploratory data analysis, data preparation, imbalance handling, feature selection, data mining methods, and evaluation of outcomes will all be covered in these parts. Oversampling, under sampling, SMOTE, SVM-SMOTE, ADASYN, class weighted method of imbalance handling and combinations of these are among the class balancing approaches discussed. Various feature selection approaches, such as recursive feature elimination, elastic net, and extra tree classifier, as well as combinations of these, are discussed. The supervised methodology, specifically classification approach, will be explained in the modelling technique portion, as well as its applications in myocardial infarction prediction. The confusion matrix and quality assessment metrics will be addressed in full in the evaluation of results section, with a focus on the technical jargon that will be utilized. The recommended classification techniques will next be explained, including the theoretical foundations and principals of six methods: Random Forest, ANN, Logistic regression, LGBM, Bagging with SVM, and Stacking blending classifiers.

## 3.2 Research Methodology

Heart attack or Myocardial infarction or Myocardial ischemia is generally occurring in an individual when oxygenated blood flow stops or decreases to the coronary arterial blood vessels which causes damage to the cardiac muscle. If an area in coronary artery bursts, blood gets clotted at the site of the damage and when this clot become larger in size, its completely or partially block the flow of oxygen rich blood to the myocardial tissues. An acute myocardial infarction can cause heart failure or cardiac arrest, in which the pumping action of the heart gets damaged causing transmission failure of enough or no blood to the vital organs, hence causing failure of those organs (Signs and Symptoms of Coronary Heart Disease - National Institute of Health, NHLBI, 2016).

The death rate due to cardiovascular disease is increasing rapidly worldwide in the recent years. According to World Health Organization, around 17.9 million people died from CVDs in 2019 and among which 85% of all deaths occurred due to heart attacks or strokes or of severe MI. It has been estimated that 23.6 million people will die from CVDs by 2030 if current trend continues (World Health Organization, Cardiovascular diseases (CVDs), 2021).

There is multiple research that has been performed to predict the complications of myocardial infarction in heart patient. In study (Kayyum et al., 2020), authors used mean imputation technique to replace the missing values. This kind of imputation is not a robust technique, as this will ignore feature correlation. In another similar study (Zheng et al., 2021), researchers used KNN imputation technique for handling the missing values. This is a robust method than a simple mean imputation technique, but it is also calculating the mean of its nearest neighbour to impute the null values which might sometime affect the accuracy of the model. In a related study (Richards et al., 2021) , authors used SMOTE technique of data imbalance handling to generate synthetic sample from minority class, this method is robust and efficient than random under and over sampling technique. But it has a disadvantage like, when generating synthetic sample, SMOTE does not take adjacent majority class instance into account. RIDGE regularization technique was used by researcher in (Wu et al., 2020) for feature selection has a disadvantage of its own, it does not reduce the number of attributes because it does not diminish the coefficient value completely to zero only minimizes it, hence, RIDGE regularization is not good for feature selection. To address the above-mentioned issues related to different studies, advance predictive modelling techniques along with more efficient and robust missing value imputation, imbalance treatment and feature selection technique will be implemented in this project.

The entire research workflow along with different activities is shown in below figure 3.1:

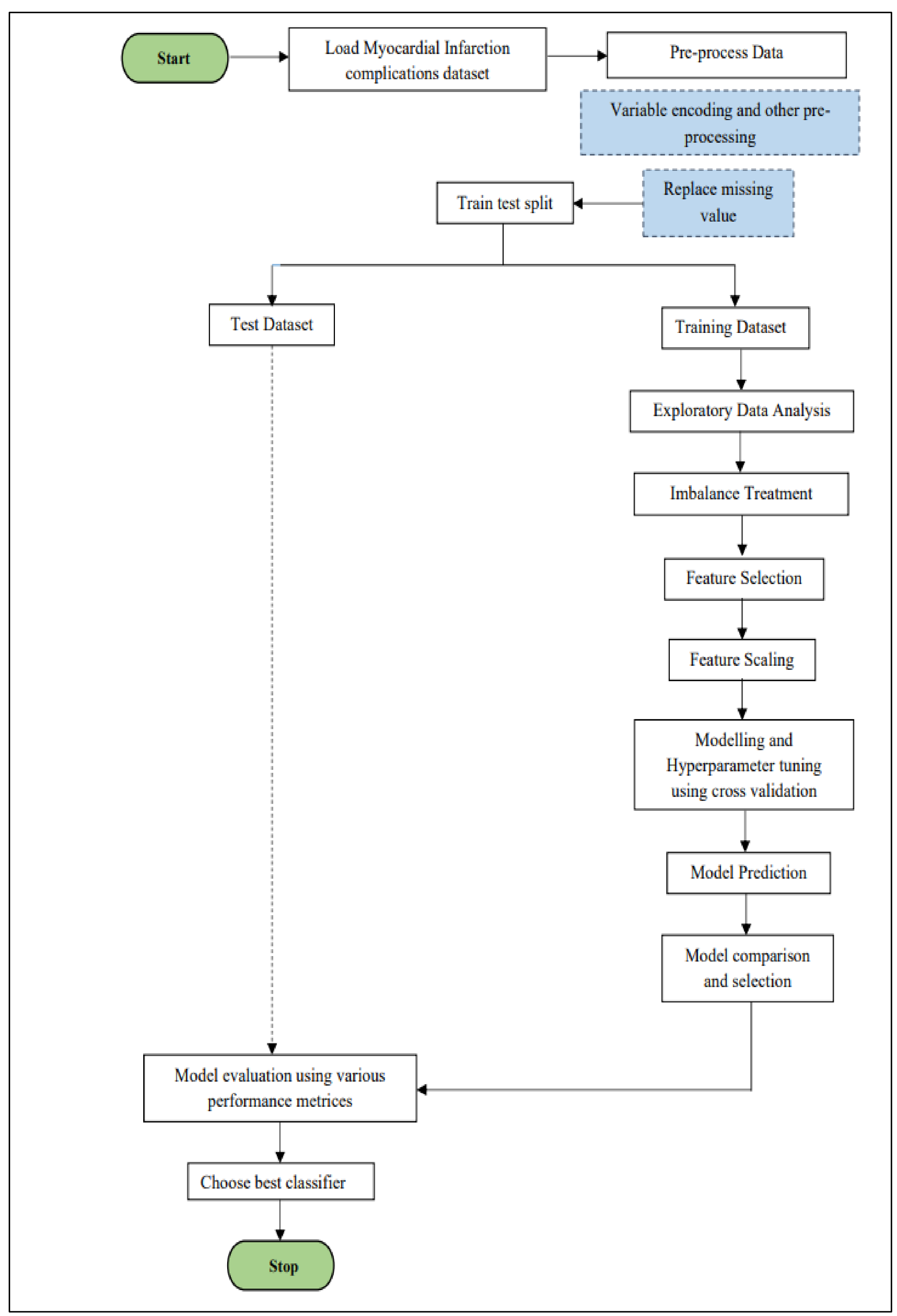

*Figure 3.1 Research Methodology Workflow*

### 3.2.1 Dataset Description

In disease analysis, the target data selection is critical since it impacts the prediction model's output. Massive amounts of medical information are created at a rapid rate in a variety of forms, necessitating the use of data mining methods to comprehend the data and extract useful information. For this study, the dataset has been collected from open world University of California, Irvine, repository (Mirkes et al., 2020). Presented dataset consists of medical records of 1700 patients suffering for myocardial ailments, which has been gathered from Berzon Krasnoyarsk Clinical Hospital, Russia. This dataset contains 113 attributes consisting of patient demographics and medical history, collected over three different time periods during hospitalization, after 48 hours and after elapse of 72 hours of hospital duration. There are 12 complication / outcomes after acute myocardial infarction have been presented in this dataset which makes this study of multiclass classification problem. As a medical dataset, it is imbalanced, contains many missing values and presence of outliers in some features, so pre-processing needs to be done on the raw dataset before compiling the predictive models.

Based on initial analysis, below table 3.1 consists of important attribute names along with its description / data types for better understanding of the all-predictor variables related to this presented dataset.

*Table 3.1 Some key attributes and its definition from presented dataset*

| Attribute Names | Attribute Data Type | Attribute Definition |
|---|---|---|
| **AGE** | Numerical | Age |
| **SEX** | Categorical | Gender |
| **INF_ANAM** | Categorical | The multitude of myocardial infarctions in anamnesis |
| **STENOK_AN** | Categorical | Anamnesis of exertional angina pectoris |
| **FK_STENOK** | Categorical | In the previous year, functional category of angina pectoris |
| **IBS_POST** | Categorical | Coronary heart disease (CHD) in recent weeks, days prior hospitalization |
| **IBS_NASL** | Categorical | Heredity on CHD |
| **GB** | Categorical | Presence of an essential hypertension |
| **SIM_GIPERT** | Categorical | Symptomatic hypertension |
| **DLIT_AG** | Categorical | The time span of arterial hypertension |
| **ZSN_A** | Categorical | Presence of chronic Heart failure in the anamnesis |
| **endocr_01** | Categorical | Diabetes mellitus in the anamnesis |
| **endocr_02** | Categorical | Obesity in the anamnesis |
| **endocr_03** | Categorical | Thyrotoxicosis in the anamnesis |
| **zab_leg_03** | Categorical | Anamnesis with chronic bronchitis |
| **zab_leg_06** | Categorical | Anamnesis with pulmonary tuberculosis |
| **S_AD_KBRIG** | Numerical | Per the Emergency Cardiology Group, systolic blood pressure |
| **D_AD_KBRIG** | Numerical | Per the Emergency Cardiology Group, diastolic blood pressure |
| **S_AD_ORIT** | Numerical | Per the intensive care unit, systolic blood pressure |
| **D_AD_ORIT** | Numerical | Per the intensive care unit, diastolic blood pressure |
| **O_L_POST** | Categorical | At the time of admission to the intensive care unit, the patient had pulmonary edema. |
| **K_SH_POST** | Categorical | At the time of admission to the intensive care unit, the patient was experiencing cardiogenic shock. |
| **MP_TP_POST** | Categorical | Atrial fibrillation paroxysms at the time of admission to the intensive care unit or at a prehospital phase |
| **SVT_POST** | Categorical | Supraventricular tachycardia paroxysms at the time of admission to the intensive care or at a pre-hospital phase |
| **GT_POST** | Categorical | At the time of hospitalization to the intensive care or at a pre-hospital stage, ventricular tachycardia paroxysms |
| **FIB_G_POST** | Categorical | Ventricular fibrillation at the time of ICU admission or in the pre-hospital setting |
| **ant_im** | Categorical | Left ventricular ECG changes in leads V1–V4 in the presence of an anterior myocardial infarction |
| **lat_im** | Categorical | Left ventricular ECG changes in leads V5–V6, I, AVL in the presence of a lateral myocardial infarction |
| **inf_im** | Categorical | Left ventricular ECG changes in leads III, AVF, and II in the presence of an inadequate myocardial infarction |
| **post_im** | Categorical | In the presence of a posterior myocardial infarction, the left ventricular ECG changes in leads V7–V9, and the reciprocity changes in leads V1–V3 |
| **IM_PG_P** | Categorical | A right ventricular myocardial infarction is present |
| **ritm_ecg_p_01** | Categorical | At the time of admission to the hospital, the ECG rhythm was sinus, with a heart rate of 60-90 beats per minute |
| **ritm_ecg_p_02** | Categorical | ECG rhythm at the time of hospitalisation – atrial fibrillation |

| **ritm_ecg_p_04** | Categorical | ECG rhythm at the time of hospitalization – atrial |
|---|---|---|
| **ritm_ecg_p_06** | Categorical | At the time of admission to hospital, the ECG rhythm was idioventricular |
| **ritm_ecg_p_07** | Categorical | At the time of admission to the hospital, the ECG rhythm was sinus with a heart rate of more than 90 beats per minute. |
| **ritm_ecg_p_08** | Categorical | At the time of admission to the hospital, the ECG rhythm was sinus with a heart rate of less than 60 beats per minute |
| **n_r_ecg_p_05** | Categorical | Atrial fibrillation paroxysms on ECG at the time of hospitalisation |
| **fibr_ter_01** | Categorical | Eliasum 750k IU fibrinolytic therapy |
| **fibr_ter_02** | Categorical | Eliasum 1m IU fibrinolytic therapy |
| **fibr_ter_03** | Categorical | Celiasum 3m IU fibrinolytic treatment |
| **GIPO_K** | Categorical | Hypokalemia (less than 4 mmol/L) |
| **K_BLOOD** | Numerical | Potassium content in serum |
| **GIPER_Na** | Categorical | More than 150 mmol/L sodium increase in serum |
| **Na_BLOOD** | Numerical | Sodium content in serum |
| **ALT_BLOOD** | Numerical | AlAT content in serum |
| **AST_BLOOD** | Numerical | AsAT content in serum |
| **KFK_BLOOD** | Numerical | CPK content in serum |
| **L_BLOOD** | Numerical | The number of white blood cells per litre is measured in billions. |

| **ROE** | Numerical | ESR (Erythrocyte Sedimentation Rate) is the rate at which erythrocytes settle |
|---|---|---|
| **TIME_B_S** | Categorical | Time it took from the start of the CHD attack to getting to the hospital |
| **R_AB_1_n** | Categorical | In the initial hours of the hospital stay, the agony returned. |
| **R_AB_2_n** | Categorical | The discomfort returned on the second day of the hospital stay |
| **R_AB_3_n** | Categorical | In the third day of the hospital stay, the pain returned. |
| **NA_KB** | Categorical | The Emergency Cardiology Team's use of opioid medications |
| **NOT_NA_KB** | Categorical | The Emergency Cardiology Team's Use of NSAIDs |
| **LID_KB** | Categorical | The Emergency Cardiology Team uses lidocaine |
| **NITR_S** | Categorical | In the ICU, liquid nitrates are used |
| **NA_R_3_n** | Categorical | In the third day of the hospital stay, opioid drugs were used in the ICU |
| **NOT_NA_3_n** | Categorical | In the third day of the hospital stay, NSAIDs were used in the ICU |
| **LID_S_n** | Categorical | In the ICU, lidocaine is used |
| **B_BLOK_S_n** | Categorical | In the intensive care unit, beta-blockers are used |
| **ANT_CA_S_n** | Categorical | Calcium channel blockers are commonly used in intensive care units |
| **GEPAR_S_n** | Categorical | In the ICU, heparin is used as an anticoagulant |

### 3.2.2 Exploratory Data Analysis

Exploratory data analysis (EDA) is a process of analysing the data by statistical and visualization technique for getting a proper understanding of dataset, recognizing the different data patterns, and getting a proper insight of the problem statement. In EDA, the statistical insight includes getting various statistical data like Mean, Standard Deviation, Median, Max Value, Min Value and data visualization includes converting raw data into maps or graphical forms to extract the useful insight from it. Analysing null entries, identifying duplicate rows/columns, analysis of outliers and other inconsistency in dataset is one of the most important steps of EDA.

In the dataset that is used for this study contains numerous missing values and inconsistencies in the presented medical records that needs to be identified during EDA process. EDA will also help to identify the distribution and relationship between the various features or biomarkers in this dataset, which can be used further at the time of feature engineering.

### 3.2.3 Data Preprocessing

The process of cleansing, preparing, formatting, and improving the quality of the raw data which makes it appropriate for machine learning task is known as data pre-processing. This is the first and foremost important step while creating a machine learning model. Any real-world data contains anomalies, missing value and may be in unusable format, this cannot be directly use in ML models, this makes the data pre-processing a vital step before applying ML algorithms which also improves the accuracy and the efficiency of the model. Also, encoding the categorical data into numbers is an important pre-processing step while building ML model.

#### 3.2.3.1 Missing Value Imputation

Real world dataset may contain missing or NAN values which can create a big problem while creating machine learning problem. Therefore, handling missing value is one of the vital tasks in this study as a data pre-processing step. One of the ways to handle missing values is that if a column contains missing value above a certain threshold (>45%) it is better to drop that column rather than imputing as this can introduce biasness into the model. In this presented dataset there are some columns which contains greater than 45% missing values, which needs to be handled specifically.

Another way to handle missing value is to calculate the mean of a particular column (which contains NAN values) and replace those NAN values with its mean value. In study (Kayyum et al., 2020), authors used mean imputation technique to replace the missing values. This kind of imputation is not a good and robust technique, as this will ignore feature correlation, means if a particular column which contain null values is correlated with another column, mean imputation method does not take this into consideration.

Another way of handling is missing value is using KNN imputer, which is a K-Nearest Neighbours algorithm. Given an instance of null value, KNN impute returns most similar neighbours and replace the missing element with mean or mode of those neighbours. In study (Zheng et al., 2021), researchers used KNN imputation technique for handling the missing values. This is a robust method than a simple mean imputation technique, but it is also calculating the mean of its nearest neighbour (based on some distant metrics) to impute the null values which might sometime affect the accuracy of the model depending on the problem statement.

In this specific study, an efficient and more robust imputation technique will be used namely, Iterative imputation for imputing the null values in case of continuous columns and simple mode imputation technique will be used in case of categorical column. These techniques are explained in detail in the below sections.

#### 3.2.3.1.1 Iterative Imputation Technique

Iterative imputation is a refined approach involves defining a model for imputing missing values by simulating each feature that predicts each missing values in an attribute as a function of all other feature recursively by evaluating the function values. Succeeding repetitions of predicting missing values using other features provide a refined estimate, which improves the quality of the null data being imputed. Below is the step-by-step interpretation of the functionality (Zhu et al., 2011) :

1. As it is used to impute null values in continuous column, a regressor is passed as a parameter into the iterative imputer transformation function.
2. The first attribute with null value is selected.
3. Then dataset is split into train and test set, where train set consists of all know value of the selected feature and test set consists of the missing values of the same feature.
4. Then a regressor is fit on all other features used as input and the feature contains null value as output.
5. The regressor predicts the missing value.
6. The iterative imputer function continues with the above-mentioned steps until all attributes are imputed.

7. The above 1-6 steps are single iteration. These steps carried out multiple times to provide a refined estimate of the missing data.

Although iterative imputation technique is a robust and efficient method of imputing missing value, but this can result into computationally more expensive depending on the volume of data, hence, simple machine learning model can be used as an estimator in iterative imputer function.

#### 3.2.3.1.2 Simple Imputation Technique

A handy and more comfortable strategy for missing data imputation is to impute all missing values with a statistically calculated value from the other values in that column. This strategy might often lead to good results depending on the state and quality of data when constructing any machine learning algorithms. Mean, median and mode of a column is commonly used statistical value to impute missing data. Mean and median imputation is used for numerical columns, whereas mode is used for both string and numerical columns.

In this specific study, mode imputation technique will be used to impute categorical missing data in a column. In a simple imputer function for mode imputation, missing data in a categorical column is replaced using most frequently used data value from the same column.

#### 3.2.3.2 Feature Scaling

Feature scaling is one of the vital pre-processing steps while building the machine learning model. This makes sure that all the independent features become scale free or scale independent or on the same scale, so that machine learning algorithm treats each feature with equal importance.

For gradient descent-based algorithm like Linear regression, Logistic regression, neural network, scaling ensure that all the features are updated with the same step size during gradient descent and converges more quickly towards minima. Furthermore, in a distance-based algorithm like KNN, SVM etc. where distance between the data points is used to determine the similarity, feature scaling plays an important role here. If both the features are on different scale, there is a probability that higher weightage is given to the attributes with higher magnitude. This can introduce biasness in ML algorithms which can finally affect the performance of the model.

There are two types of feature scaling techniques as follow:

1. Normalization
2. Standardization

Normalization feature scaling often known as Min-Max scaling. This technique ensures to re-scale the feature with distribution value between 0 and 1, that means all the data points for the feature will lie between 0 to 1. The general formula for this transformation is shown below equation 3.1:

$$x_{scaled} = \frac{x - x_{min}}{x_{max} - x_{min}} \tag{3.1}$$

Standardization is another feature scaling technique which is also knows as Z-score normalization technique where the values are centred around mean and unit standard deviation. This means the mean of the features become 0 and the final distribution after transformation have standard deviation of 1. The general formula for this transformation is shown below equation 3.2:

$$x_{scaled} = \frac{x - \mu}{\sigma} \tag{3.2}$$

In study (Azwari, 2021), researcher used normalization or Min-Max scaling for data transformation but during the implementation phase of this project, depending on data distribution, scaling will be implemented that means, Min-Max scaling will be used if the features in the presented dataset does not follow a normal or gaussian distribution whereas if the features follow a gaussian distribution then Standardization will be implemented for feature scaling.

### 3.2.4 Data Imbalance Treatment

Most classification analysis, face a severe problem of imbalance data where the classes are not equally present in the dataset. In a binary classification problem when the number of instances for one class is higher than the other or in multiclass classification problem, number of observations is not same for all classes, then this leads to a major problem of imbalance.

Generally medical data or records often face this problem of imbalance. For example, in the presented dataset for this study, consists of medical records of 1,700 patient suffering from myocardial infarction but out of these 1,700 patients, 396 patients suffered from chronic heart failure after acute MI and rest 1,306 patients yet not suffered from chronic heart failure. This is a classical problem of imbalance data where the presence of majority class is almost 3 times higher than minority class. Naturally, number of individual sufferings from a medical treatment or disease will be always less than the individuals who are medically fit.
If any predictive machine learning model developed using this imbalance dataset, that could be inaccurate and tend to produce unsatisfactory or misleading predictions. The developed model will be biased towards majority class while the minority class's observations will be treated as noise. Hence there is high probability of misclassification of the minority class than majority one. But, in the medical domain or the predictive model developed for a healthcare domain, the major focus will be on prediction of the minority class accurately. Therefore, imbalance data needs to be treated before passing it of ML pipelines.
There are majorly 4 data imbalance handling techniques commonly used, which are as follows:

1. Random under sampling
2. Random over sampling
3. SMOTE (Synthetic minority oversampling technique)
4. Class Weighted Method of imbalance handling

Random under sampling technique is basically used to balance the majority and minority class by randomly eliminating or deleting the majority class instances. This is done until both the classes balance out. Common disadvantage of this technique is, it basically randomly discards or deletes the majority classes instances and chance of losing vital information, this can also lead to underfitting issue.

Random over sampling technique is basically used to increase the minority class instances by randomly duplicating them to present a higher count of minority class. This is done until both majority and minority classes balance out. Common disadvantage of this technique is, it basically duplicates the same information multiple times, hence leads to overfitting issue.

The SMOTE imbalance treatment method uses a strategy of selecting a subset of data from a minority class and generates synthetic pseudo-observations based on instances selected from the subset. In a related study (Richards et al., 2021), author used SMOTE technique of data imbalance handling to generate synthetic sample from minority class, this method is robust and efficient than random under and over sampling technique. But it has a disadvantage like when generating synthetic sample, SMOTE does not take adjacent majority class instance into account. This can increase in overlapping of the classes and can induce noise.

As there are some disadvantages with SMOTE, hence ADASYN can be used in this study for imbalance handling. With one key exception, ADASYN (Adaptive Synthetic Sampling) is quite like SMOTE and evolved from it. The sample space will be skewed against points that are not in homogeneous neighbourhoods. ADASYN uses the kind='normal' SMOTE algorithm on points that are not in homogeneous areas. As a result, a hybrid of regular and borderline SMOTE arises. The primary issue with SMOTE's technique was that it couldn't establish inner point outer point bridges. Whether or not a heavy emphasis on anomalous points is appropriate depends on the application.

In this project, SVM-SMOTE imbalance handling method will be tried as well, which is also an enhanced and more efficient version of SMOTE, where SVM machine learning algorithm is used to detect the sample which to be use for generating new synthetic instance.

All the strategies for addressing imbalances outlined above generate synthetic data, which increases the observation in the minority class to match the majority class. There is a chance that real-time data or observations will differ from the synthetic observations on which the model was trained. As a result, model performance can suffer greatly under those situations. Hence, any type of oversampling or synthetic data generation approach of imbalance therapy is not recommended in such instances. Another approach of addressing imbalances, known as the class, weighted method, now comes to the rescue. In this method the existing training strategy can be tweaked to accommodate for the skewed distribution of classes. This can be accomplished by assigning the dominant and minority classes differing weights. During the training stage, the weight differences will influence the categorization of the classes. The goal is to penalize the minority class for misclassification by giving them a greater class weight while giving the majority class a lower weight.

#### 3.2.4.1 SVM-SMOTE Imbalance Handling

As SVM-SMOTE is an enhanced variation of the SMOTE, let’s discuss the step-by-step interpretation of the simple SMOTE functionality:

1. Select a minority class instance to start with.
2. Find its k nearest neighbours (k neighbours is specified as a parameter in the SMOTE () function) using any of the distance metric.
3. Draw a line joining the point chosen and its selected neighbour, then using the below formula new synthetic point is generated.

4. New synthetic sample: $x' = x + rand(0, 1) * |x - x_k|$, in which rand (0, 1) represents the random number between 0 and 1.
5. Repeat the above steps until both the class data is balanced

As discussed above, SMOTE has a disadvantage of increasing the overlapping of the classes as it does not consider majority class instance as a nearest neighbour. Borderline SMOTE algorithm, which is an improved variation of SMOTE, mitigates this by generating synthetic data only at the decision boundary between the two classes.

The downside of borderline SMOTE is that more data is synthesized near the class boundary region. This downside has been mitigated by SVM-SMOTE, unlike borderline SMOTE where random nearest-neighbours are used to determine misclassification, SVM-SMOTE involves an SVM algorithm in determining it, which makes it special as compared to the other techniques. In SVM-SMOTE, support vector points (after training with SVM classifiers) are responsible for determining the borderline area. Synthetic data is generated along the lines joining each minority class support vectors with several of its nearest neighbours. In this method of imbalance treatment, more data is synthesized away the class overlap region and more concentrated where the data points are separated (Tang et al., 2009) .

#### 3.2.4.2 ADASYN (Adaptive Synthetic Sampling) Imbalance Handling

The core concept of ADASYN is to generate a sufficient number of synthetic alternatives for each observation in the minority class. The concept of proper proportions in this case is determined by how difficult it is to remember the original observation. An instance from the minority class, in particular, is difficult to master if there are many instances from the dominant class with qualities identical to that observation. Let's discuss below, the step-by-step interpretation of the ADASYN functionality (Lu et al., 2020):

1. The ADASYN methodology operates by synthesizing various numbers of fresh sample data for less specimens while considering varied sample distributions. Finding a probability distribution is crucial. For finding the number of instances that should be synthesized per smaller class of instances, use the ri, put ri n function.
2. The K value nearest neighbour of xi in n region of space is obtained for each instance of a few classes, and its ration is ri=i/k,i=1, 2,.,m.
3. The quantity of most classes closes to the K of xi is Δi. As a result, ri ∈ (0,1].
4. Compute the status of most categories over each minority sample using r=ri/msi=1ri regularization and the probability distribution (ri=1).

5. xi per limited amount of data is used to calculate the number of synthesized samples: gi: gi=ri^+G.

As there are some disadvantages with SMOTE already discussed above, hence ADASYN will also be used in this study for imbalance handling.

#### 3.2.4.3 Class Weighted Method of Imbalance Handling

The under-sampling method eliminates data points from the majority class, leading to data loss, while up sampling or SMOTE adds synthetic data points to the minority class. Hence, class weight option can be used to address the asymmetry in the class in the given dataset during machine learning training.

A built-in setting "class weight" across most sklearn classifier modelling packages, as well as other boosting-based libraries like LightGBM, lets one maximize the scoring for the minority class. By standard, class weight=None, implying that both categories are given similar weights. Aside from that, one may either specify 'balanced' or pass a dictionary with manual weights for both categories. When the class weights parameter is set to 'balanced', the algorithm allocates class weights that are inversely proportionate to their respective frequencies.

The formula is wj=n_samples / (n_classes * n_samplesj), where the weight for each class is wj (j signifies the class), the total sample size or rows in the dataset is n samples, the entire number of unique classes in the target is n classesj, and the overall number of rows of the relevant class is n samplesj. This is how class weight = 'balanced' aids in providing the minority class greater weight and the dominant class lesser weights. Although supplying a value as 'balanced' produces good results in most circumstances, for extreme class imbalances, one can attempt manually allocating weights (Zhu et al., 2018).

All the procedure of class imbalance treatment that has been discussed above will be tried one by one and finally top performing imbalance handling method will be selected along with the selected machine learning model.

### 3.2.5 Feature Selection Techniques

The prediction power of the machine learning model is directly proportional to the number of attributes or features used to train the model. The model performance will be adversely affected if irrelevant features is used to build the ML models. The process of reducing the number of irrelevant features or the process of dimensionality reduction automatically or manually based on any statistical relationship is knowns as Feature Selection. This may improve the performance or the predictive power of the ML models. Reduction of overfitting, improving the accuracy of the model and reducing in the training time of the ML models, these are some of the important benefits of feature selection.

Three broad categories of feature selection methods are as follows:

1. Filter based feature selection technique
2. Wrapper based feature selection technique
3. Embedded feature selection technique

Filter based feature selection technique, which is also known as Univariate feature selection method, in which relevant attributes having strongest relationship with the target variable are selected based on statistical testing. Although this is faster method and computationally inexpensive, but this may not remove multicollinearity among independent variables. Some of the filter-based techniques are Pearson correlation, spearman rand correlation, Chi-square test, ANOVA (Analysis of Variance) etc.

Wrapper based feature selection technique is used to find the ideal combination of the predictor variable which maximize the model performance by evaluating all possible combination of features against a particular evaluation criterion that could be p-value, Adjusted R-squared in case of regression problem and precision, recall in case of classification problem. This selection method is built on a specific ML algorithm that are fit on a dataset. Some of the wrapper-based techniques are RFE (Recursive feature elimination), Forward feature selection, backward feature elimination etc. This is a well-suited approach of feature selection technique, but this can lead to overfitting and required high computational time than filter-based method.

When the qualities of both filter-based and wrapper-based feature selection technique combine, this gives the birth of a hybrid feature selection technique which is called as Embedded feature selection technique. This method is implemented by ML algorithms that have their own feature selection or feature importance methods in-built, this way it can take the advantage of its own attribute selection process. Some of the embedded feature selection techniques are LASSO, Elastic net, Tree based modelling etc. This method is faster than wrapper-based feature selection technique, less prone to overfitting and more accurate than filter method.

In study (Pavithra and Jayalakshmi, 2021), authors used combination of embedded (tree-based model) and filter based (Pearson correlation) feature selection technique for dimensionality reduction. In a similar study (Wu et al., 2020), researcher used wrapper-based feature selection technique namely RFE, to eliminate the irrelevant features which may impact the model performance adversely. On the other hand, tree based embedded attribute selection was used by authors in (Zheng et al., 2021), which produced better accuracy than other methods discussed above.

Out of all the feature section method discussed above, three of them will be implemented in this specific study for dimensionality reduction, namely, RFE from wrapper-based method and Extra-tree, Elastic net from embedded based method, then after model performance will be evaluated. The detail of these techniques along with their advantages over the other is discussed in the next section.

#### 3.2.5.1 Recursive Feature Elimination Method (RFE)

One of the most popular, easy to use, easy to configure and more efficient in selecting the relevant features based on feature importance or coefficient value for predicting the target variable is Recursive feature elimination technique. Above advantages of RFE makes it more ideal technique to use for feature section in this project. As various machine learning algorithm is used in the heart of this method and wrapped by RFE which is used to select relevant feature, hence RFE is a wrapper-based dimensionality reduction technique (Ustebay et al., 2018) .

Below is the high-level understanding how RFE works:

1. Select all the features in a training dataset.
2. Fits a ML model on all those features.
3. Score each feature and rank it by importance.
4. Discard the least important feature by score.
5. Refit the ML model with remaining features.
6. The above step repeats until the desired number of attributes remains.

The choice in the number of features to select and choice of the ML algorithm that helps to select the features are two important hyperparameter of the RFE feature selection method. These two parameters need to be configured well for optimal performance.

#### 3.2.5.2 Elastic Net Feature Selection Technique

Overfitting is a phenomenon that occurs when ML models works well with high accuracy on training set but poorly perform on unseen data. Regularization is a method of which is used to reduce overfitting problem by fitting the function properly on the training set. There are broadly three types of regularization technique used in machine learning as follow:

1. L1 regularization (LASSO)
2. L2 regularization (RIDGE)
3. Elastic Net regularization.

In this specific project Elastic Net regularization will be implemented. LASSO has some of the disadvantages like, when there are multicollinear variables, LASSO regularization randomly selects one of the variables and ignore the rest, which is not good for interpretation, also if the number of predictor variables is more than the number of observations (n), then LASSO select the maximum n predictor variables randomly and ignore others even if all the features are relevant. On the other hand, RIDGE regularization technique which was used by researcher in (Wu et al., 2020) also has a disadvantage that is, it does not reduce the number of attributes as it does not diminish the coefficient value completely to zero only minimizes it, hence, RIDGE regularization is not good for feature selection. These disadvantages of both RIDGE and LASSO makes its application unsuitable for this specific study. As in any medical domain project, interpretation of the model is very much necessary and reduction of the irrelevant feature is required at the same time, hence elastic net feature selection method is a good option for this project.

Elastic Net regularization method is a combination of both L1 (LASSO) and L2 (RIDGE) techniques. Both the penalty terms of LASSO and RIDGE are included in elastic net regularization. Incapability of both L1 and L2 techniques has been taken care off in elastic net method. In elastic net regularization, firstly, RIDGE regularization coefficient needs to be estimated that minimize the coefficient values and on top of it LASSO regularization is implemented which reduce the irrelevant coefficients value completely to zero. Therefore, these variables whose coefficient value is zero can be ignored as they will not participate in the prediction power of the model which makes this method ideal for feature selection mechanism (Challita et al., 2015).

Below is the Elastic Net regression function with penalty terms shown in equation 3.3:

$$L_{elasticnet}(\hat{\beta}) = \frac{\sum_{i=1}^{n}\left(y_i - x_i^T\hat{\beta}\right)^2}{2n} + \lambda\left(\frac{1-\alpha}{2}\sum_{j=1}^{m}\hat{\beta}_j^2 + \alpha\sum_{j=1}^{m}\left|\hat{B}_j\right|\right) \tag{3.3}$$

#### 3.2.5.3 Feature Importance using Extremely Randomized Trees

An ensemble machine learning technique which accumulates the resultant of numerous un-correlated decision tree collected in a forest to output the classification results, this technique is popularly known as Extremely Randomized Trees Classifier (Extra Trees Classifier). It is very similar to random forest classifier, only difference lies in how each of this decision tree builds in a forest.

Below is the high-level understanding of how each decision tree built in Extra tree forest:

1. Each decision tree in the extra tree forest is built with all the data available in the dataset.
2. Optimal split is determined by searching among the subset of selected features. This subset is selected by the equation sqrt (number of features).
3. Select the cut point randomly in order to select the node for optimum split.
4. Then based on the hyperparameter like max depth etc. each decision tree is built in the forest.

During the production or building of the extra tree forest, for every attribute, the normalized total reduction in Gini Index is calculated. This value is known as Gini Importance of a features. Then to perform features selection, based on Gini Index each attribute or feature is sorted in descending order and finally top important features based on problem statement are selected (Camana Acosta et al., 2020).

In the proposed study (Zheng et al., 2021) random forest tree-based feature selection technique was used by the researchers. But as random forest-based selection technique is quite computationally expensive and required more time to build each tree in a forest, hence total time taken by the process is quite long, whereas feature selection technique using extra tree perform almost similar like random forest, but it is much faster compared to it, this is because Extra Tree randomly selects the cut points for every feature for optimum splitting of the node rather than calculating Gini for every cut points and deciding the split based on it. Due to the above advantage of extra tree classifier, it will be a good candidate to use as feature selection method for this study.

## 3.3 Proposed Classification Methods

Selecting the appropriate predictive model or modelling techniques for classification to predict the lethal outcomes (cause) after acute myocardial infarction is an important task to achieve the better accuracy of the system. After applying suitable data pre-processing, feature engineering and feature selection technique on the presented dataset, this cleanse dataset needs to be pass to the various ML classification pipeline for predicting the class accurately. This section will discuss about different modelling techniques along with their advantages and why these methods should be used for this specific research.

### 3.3.1 Logistic Regression

Logistic Regression is one of the most popular supervised machine learning algorithms for predicting categorical dependent variable given a set of independent variables. Logistic regression is used to predict the categorical or discrete value which can be either 1 or 0, True or False, Yes or No etc. But instead of giving exact 1 or 0, it gives the probabilistic values which lies between 0 to 1 (minimum value is 0 and maximum value is 1). This method is use for classification problem statement. Instead of fitting line of best fit like linear regression, it fits a sigmoid curve (logistic function) as shown in below figure 3.2:

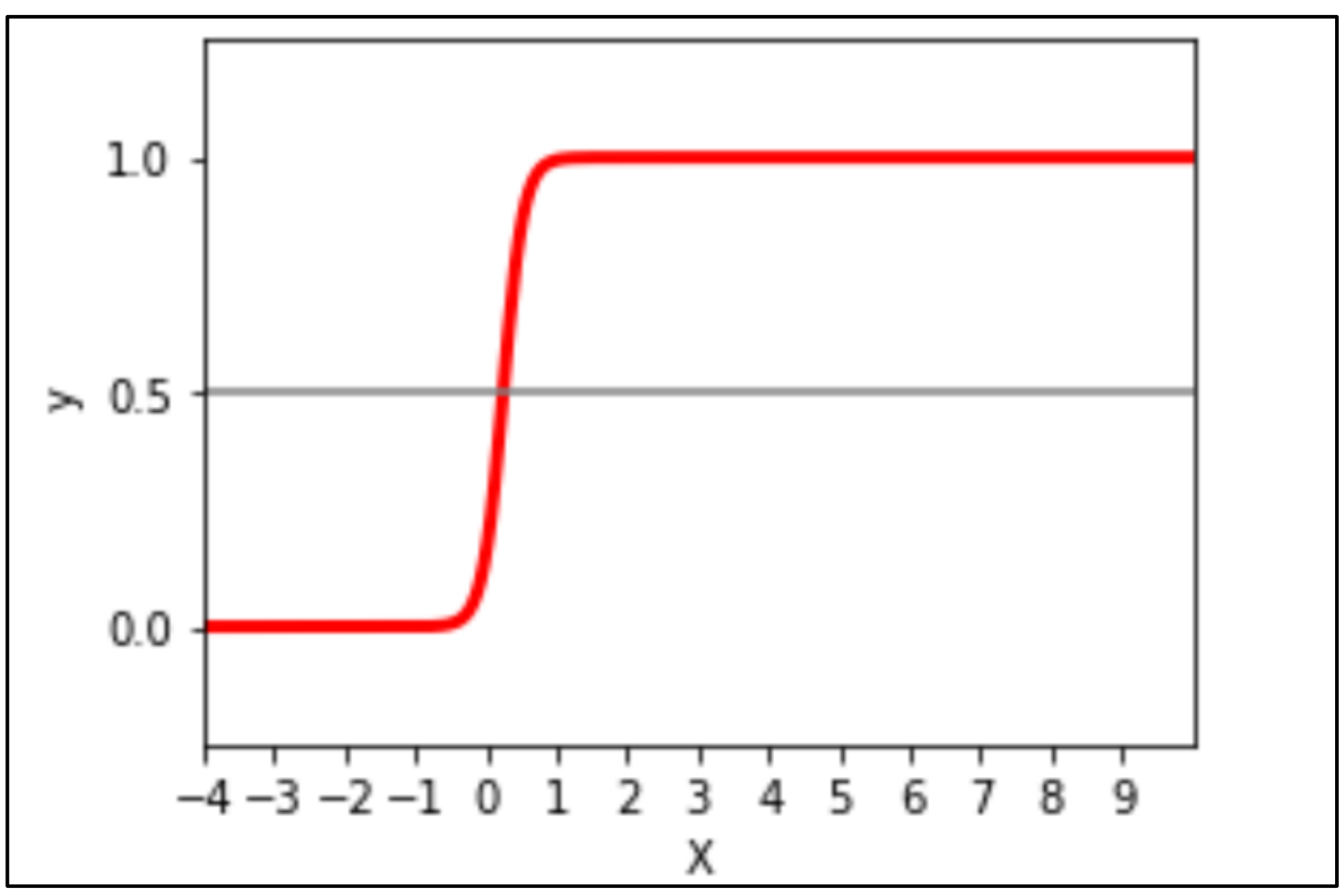

*Figure 3.2 Logistic Curve (Sigmoid)*

Th equation of the logistic function is shown below in equation 3.4:

$$f(x) = \frac{1}{1 + \mathrm{e}^{-x}} \tag{3.4}$$

The logistic function curve shown above indicates the likelihood of occurring a specific classification event like whether obese or not obese, whether diabetic or non-diabetic etc. This algorithm has the ability to give probability as an output which indicates how likely an event will occur and appropriately classify new data (Sivakumar et al., 2021)

In the research (Richards et al., 2021) , authors used Logistic regression technique for the analysis of myocardial infarction which has provided a descent accuracy for the predictive model. There are two important advantages of the Logistic model those are given below:

1. LR is one of the simple and popular machine learning algorithms
2. The model interpretably achieved with this algorithm is very much appreciable.

Being a medical domain project, the interpretation of the various features or biomarkers is necessary for diagnostic purpose. Therefore, keeping all the above advantages in mind, Logistic regression modelling will be implemented in the research as well.

### 3.3.2 Light Gradient Boosted Machine (LGBM)

LGBM is one of the popular gradient boosting frameworks tree-based learning algorithm. Gradient boosting belongs to ensemble class of machine learnings techniques that can be used for classification and regression problem statement. Models are fit with a differentiable loss function and gradient descent optimization algorithm is performed on it, hence the name gradient boosting. In boosting, ensemble is produced from decision trees. Decision trees are built one at a time and fits to correct the prediction error made by the prior models. LGBM is one of efficient and effective algorithm than other boosting mechanisms. As it is an extension of gradient boosting mechanism, below is the overall idea how gradient boosting works for regression:

1. Given a dataset, calculate the average value of the target column.
2. Now, calculate the residual which is the difference between actual and the predicted value (average value calculate above).
3. Construct each decision tree in ensemble by taking all independent variable and residual as target variable.
4. Now, predict the target label using all the decision tree in the ensemble.
5. Compute the new residual.
6. Repeat from step 3 to step 5 until the number estimators matches the specified value in hyperparameter (number of estimators)
7. After all the above steps are done, use all the trees to make the final prediction for the value of the target feature.

LGBM in many ways is special from ordinary gradient boost trees. The way LGBM builds the tree makes this algorithm beneficial than other tree-based models. It grows the tree vertically leaf-wise, whereas most other algorithms grow tree horizontally level-wise. Level-wise tree growth can result in unnecessary leaves and nodes, as the tree building does not stop until the maximum depth reached. In contrary, trees with leaf-wise growth choses the leaf with max delta loss to grow, which reduces more loss than level-wise algorithm and hence results in much higher accuracy. LGBM can take on large size of data and can required lower memory size as well, (Alzamzami et al., 2020).

Researcher in (Liang et al., 2021) used XGBoost algorithm for predicting heart failure in a patient after MI. XGBoost is fast and accurate ML method but LGBM on the other hand is much faster than XGBoost, can handle the sparse data better and accurate at the same time. Below are certain advantages of LGBM modelling.

1. 6 times faster than XGBoost and have higher efficiency.
2. Lower memory usage.
3. Provide better accuracy.
4. Can handle large dataset
5. Support GPU learning.

Therefore, keeping all the above advantages in mind, Light GBM modelling will be implemented in this specific research.

### 3.3.3 Random Forest Classifier

One of the most popular supervised machine learning algorithms that can be used for both classification and regression modelling is Random Forest machine learning technique. This belongs to the ensemble bagging class of machine learning techniques which is a process of combining multiple classifiers to solve a complex problem statement and can perform more accurately than an individual model. This is a combination of multiple decision tree which gets the prediction from each of the tree in a forest and finally choose the best solution using majority voting technique in case of classification. The greater number of trees in a forest means better accuracy of the model and reduce overfitting by minimizing the variance. Below are the high-level steps how the random forest works:

1. First select random samples of datapoints from training set, this sample is specifically known as Bootstrap samples (sampling with replacement).
2. Built decision tree with the selected subset of datapoints.
3. Choose the number of decision tree that needs to be build.
4. Repeat the steps from 1 and 2.
5. For new datapoints, find the prediction from each of the decision tree and assign the class to the new datapoint that wins the majority votes.

Random Forest is more robust and well performing machine learning algorithm (Xuan et al., 2021) .

Random forest modelling technique was used in various similar studies (Kayyum et al., 2020; Piros et al., 2020; Praneetha et al., 2021) for its high accuracy and stability. There are other advantages of random forest like robust to outlier, stable (low variance), and most importantly it has in-built feature important mechanism that can be used to identify vital features. For the above advantages, RF modelling technique will be a better choice for this specific project as well because interpretation of the vital biomarkers is one of the essential aspects for this study.

### 3.3.4 Bagging Classifier using SVM Modelling

A Bagging classifier is an ensemble of meta-estimator the is used to fit base classifiers each on random samples of the original dataset and for final prediction, it aggregates their individual prediction using majority voting method. Bagging ensemble technique is a way to reduce the overall variance of any individually used black box modelling technique, by introducing randomization into its fabrication, building procedure, and then turning it into an ensemble. By drawing samples randomly with replacement from training dataset, each base classifier is trained in a parallel manner. Training sample set for each base classifier and model training is independent of each other. Bagging reduces overfitting by reducing the variance with the help of majority voting methods (Wang et al., 2021) .

While developing an intelligent system for predicting myocardial infarction from ECG signal, Support vector machines was used for prediction model on selected features in study (Diker et al., 2018) , this modelling technique produces 87%, 86% accuracy and sensitivity respectively, which is good.

As, SVM works well with higher dimensional feature spaces and is memory efficient, but to avoid high variance and risk of overfitting, SVM will be implemented as a base classifier for bagging ensemble method in this specific research project that will help to reduce the risk of overfitting (majority voting).

The architecture of the SVM bagging classifier is shown in below figure 3.3:

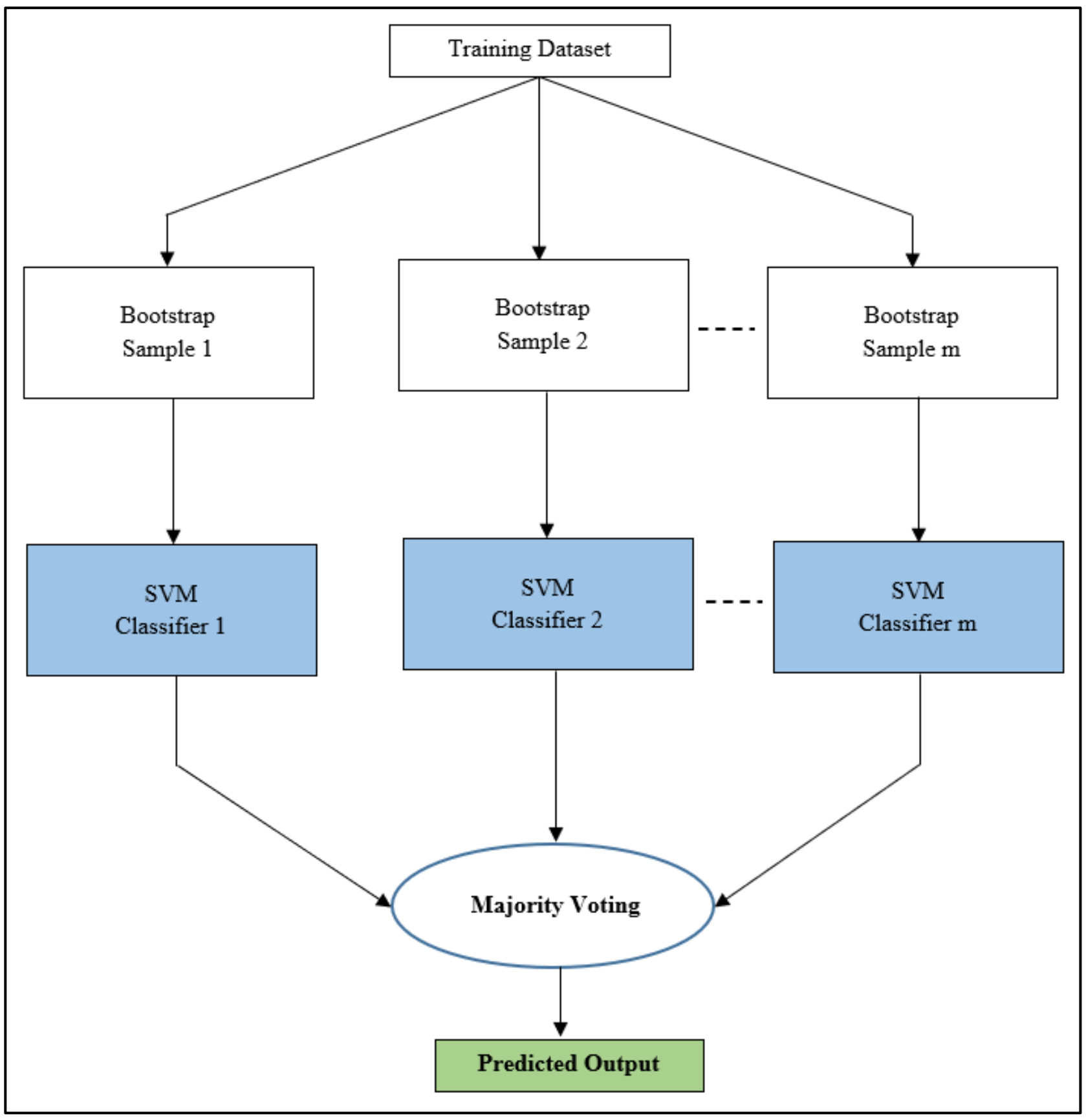

*Figure 3.3 SVM Bagging Classifier*

SVM is one of the popular and flexible supervised machine learning method that is used primarily for classification problem. SVM algorithm creates the best fit line or best decision boundary that can separate distinctly n-dimensional space into classes so that unseen data can easily be classified in correct category. This best decision boundary is called a hyperplane. This hyperplane is created by SVM chosen support vector points (these are the extreme points or boundary cases, which are closest to the hyperplane), hence the name Support Vector Machine (Eke et al., 2021). Below are the two important hyperparameter that needs to be tune for SVM algorithm:

1. Slack Variable: This is one of the most significant hyperparameters to consider when categorising non-linear data. When data points are not linearly separable, the slack variable permits some constraints to be broken by allowing certain training points to fall within the defined soft margin. This slack variable always takes a positive or zero value.
2. SVM kernel: Function that allows the SVM algorithm to transform non-linearly separable dataset into linearly separable one is called kernel or kernel function. There are three types of kernel function namely, linear kernel (use for linear decision boundary), polynomial kernel (use for non-linear polynomial decision boundary) and radial basis kernel (use for more complex and circular decision boundary).

### 3.3.5 Stacking Ensemble Learning

Stacking Generalization or Stacking is one of the machine learning algorithms. Stacking method combines all the predicted outputs from various ML models. In this ML technique, all the models involve in it are typically different and fits on the same dataset. It basically learns how to combine the best prediction from individual model. The stacking model consists of two or more first level model and one meta model (second level model) that combine the output or the prediction of the first level model. Below is the high-level working principal of the stacking ensemble modelling:

1. Each base or first level models are trained individually on whole training set and prediction are made from each base learner.
2. Then, along with the above predictions from each model and expected output together are passed to the meta- model (second label model), which act as input and output pair of the second level model.
3. Finally, predictions are made by the meta-models.

In case of stacking classifier, the output from the first level learner is the probability score which passed to the meta-learners for final combined prediction. Stacking modelling technique is appropriate when multiple ML models (in base learner) have different skill set in various ways and that has been proven on the same dataset. It is always advice to use a range of models those having different assumption on how to solve the prediction task as a base or first level models. On the other hand, meta models are often simple (like any liner model) just to provide a smooth interpretation of the prediction made by the base model (Petinrin and Saeed, 2019).

The architecture of stacking ensemble classifier is shown in below figure 3.4:

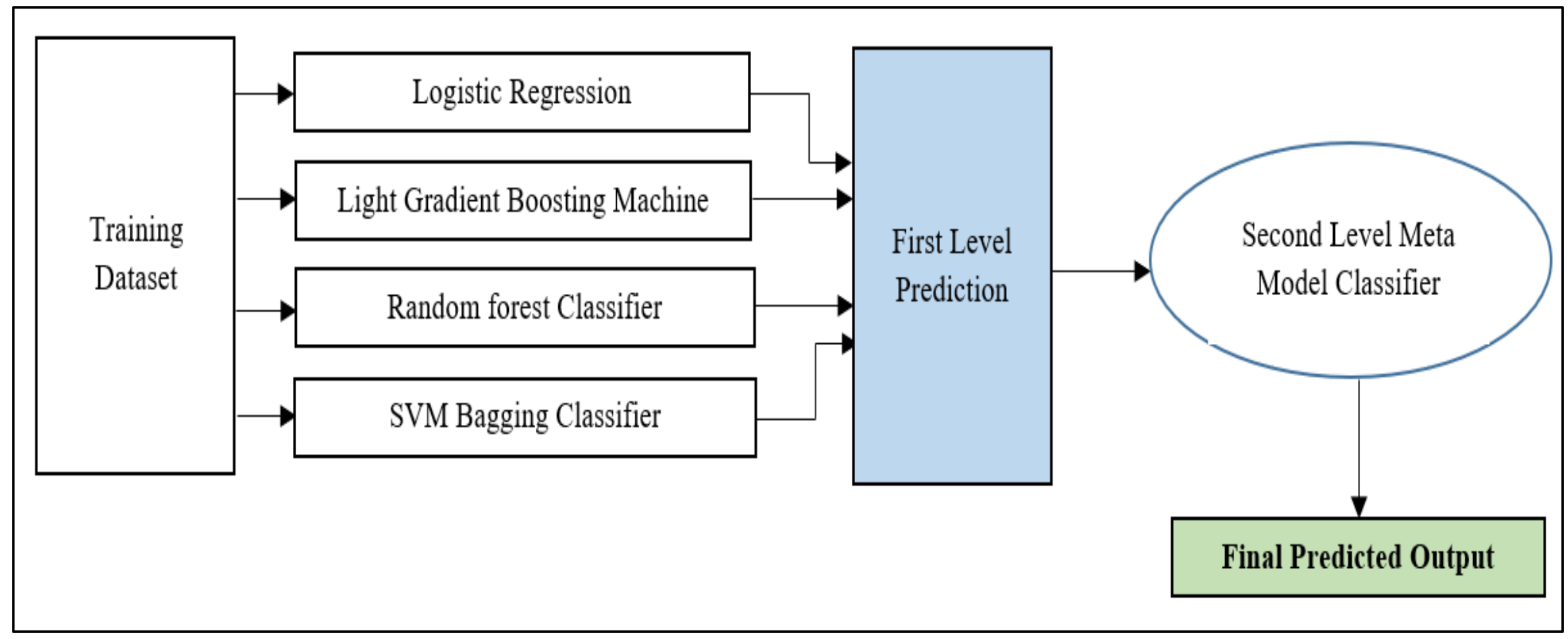

*Figure 3.4 Stacking Ensemble Classifier*

In the project, stacking model will be implemented which consists of all the above discussed individual model as a first level or base classifier and on top of it, a meta- classifier will be used. There are couple of advantages of stacking model that makes it ideal for this project like, it harness the abilities of range of well performing models that has already been discussed above on given classification task and it may provide better prediction than any individual model, also stacking improves the model accuracy.

### 3.3.6 Artificial Neural Network (ANN)

ANN is the biologically inspired computing system and sub-field of artificial intelligence. This modelling technique is based on the structure of human brain. As, the neurons are interconnected in human brains, ANN also have neurons those are interconnected with each other in different layers of the networks, and these are called nodes. ANN mimic the work of human brains so that the computer system can understand and make decision of their own just like human brain does. Artificial Neural Network is amazing parallel processor like human brain.

The architecture of ANN consists of below layers:

1. Input Layer – It basically accept the input from outside.
2. Hidden Layers – These are the layers which are in between of input and output layers. All the calculations are carried in these layers to find the hidden patterns from the dataset.
3. Output Layer – Different transformation are carried over the inputs in hidden layers and the result is communicated from this output layer.

Provided the inputs, ANN compute the weighted sum of those given inputs and include a bias term (b) in the calculation. Then an activation function is applied on it.

The architecture and formula in equation 3.5, for the ANN model computation is shown in below figure 3.5:

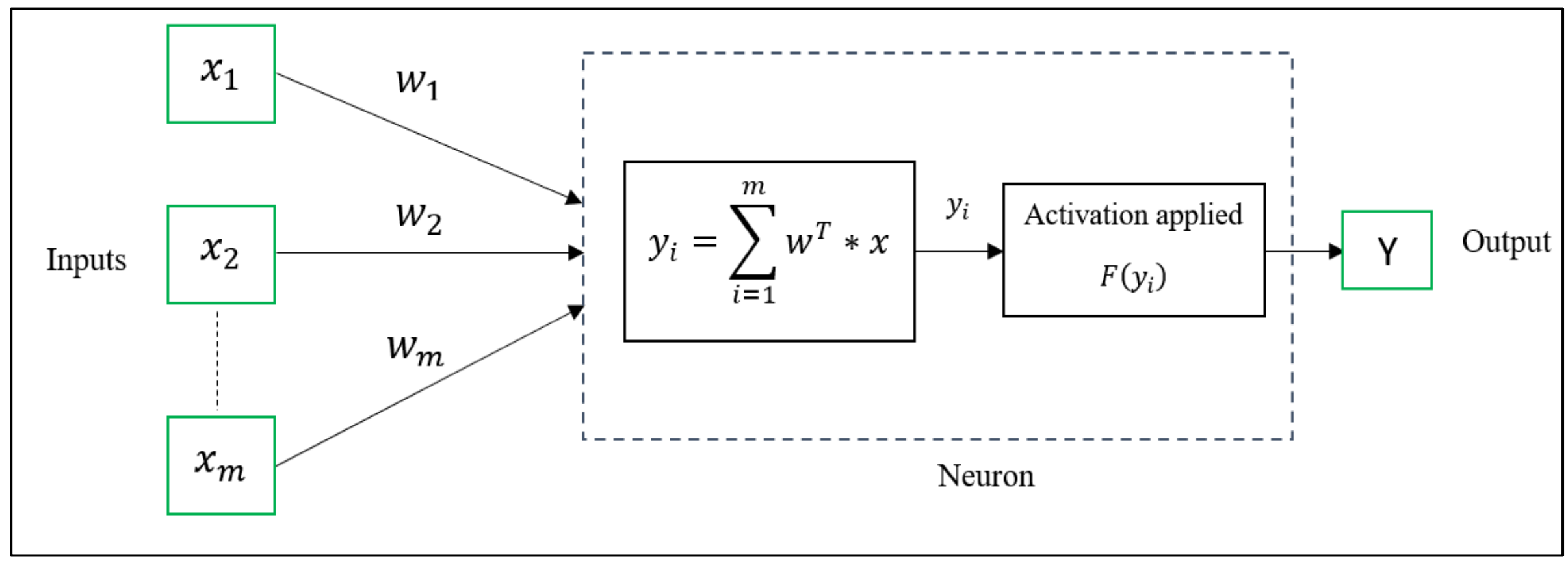

*Figure 3.5 Artificial Neural Network Architecture*

$$y_i = x_1 w_1 + x_2 w_2 \ldots + x_m w_m + b$$

$$y_i = \sum_{i}^{m} x_i w_i + b \tag{3.5}$$

The output after applying an activation function on the input shown in below equation 3.6.

$$Y = F(y_i) \tag{3.6}$$

In between the hidden layers, the weighted sum is passed to an activation function. This activation function determines whether a neuron will fire or not. Only the neuron those fired make the way to the output layers. Hence, in other word activation function define the output for that node (in between hidden layers) given the set of inputs from to that layer. There are different kind of activation function namely, linear, Tan hyperbolic, sigmoid, RELU etc. Below are the high-level steps on how an ANN works:

1. At start, assign random weights to all the linkages between the layers.
2. Find the activation rate of the hidden node using the input and the linkages.
3. Find the activation rate of the output nodes using the hidden nodes and the linkage to the output.
4. Recompute all the linkages between the hidden nodes and the output nodes based on error rate generated at the output node.
5. Stream down the error to the hidden nodes using the weight and the error found at the output node.
6. Recompute the weights between the hidden nodes and the input nodes.
7. Until the convergence criteria are met, repeat the above steps.
8. Compute the activation rate of the output nodes using the final linkage weights.

There are several advantages of the artificial neural networks like, it has good fault tolerance, amazing parallel processor, has distributed memory, produces almost accurate and quality results etc. (Moghanian et al., 2020).

Keeping all the above advantages in mind, specifically the quality and accuracy of the result, the ANN modelling will be implemented in this project. The lethal outcome prediction after MI needs to be accurate for the survival of the cardiac patients, hence it is very necessary for any modelling technique to be very much precise and accurate in nature. Hence the neural network is a very good candidate in terms of prediction accuracy for fatal disease prediction.

### 3.3.7 Evaluation Metrices

Evaluating the performance of different machine learning models is one of the important tasks during building a predictive model. There are different evaluation metrices those are used for evaluating different classification and regression task. Evaluating the model performance using appropriate evaluation metrices would improve the overall predictive power of the model. In case of classification problem, only depending on accuracy without evaluating other performance metrices is not a good idea to deploy the model in production on unseen data. This may lead to poor performance of the model in production environment.

Judging by past various literatures on chronic disease-related research (Devika et al., 2019; Sonar and JayaMalini, 2019; Saranya and Sasikala, 2020) and some on myocardial ischemia research (Kashirina et al., 2021; Richards et al., 2021), the popularly used metrics are predominantly deduced from the confusion matrix, which demonstrates the four outcomes (TP, TN, FP, FN); the measures include accuracy, precision, recall, F1-measure, and specificity. AUCROC and the area under the precision recall curve are two graphical display-based methods that are used to evaluate classifier performance (Liang et al., 2021). Technical terminologies, that will be explained below, are the performance metrices used to assess the performance of the classifier.

There are many evaluation metrices for classification task, some of the important metrices are discussed below:

- Accuracy – It is the ratio between the correct predictions and the total number of predictions. The formula is given below in equation 3.7:

$$\text{Accuarcy} = \frac{\text{TP} + \text{TN}}{\text{TP} + \text{TN} + \text{FP} + \text{FN}} \tag{3.7}$$

- Precision – It summarizes the fraction of cases assigned to the positive class that belongs to truly positive class. The formula is given below in equation 3.8:

$$\text{Precision} = \frac{\text{TP}}{\text{TP} + \text{FP}} \tag{3.8}$$

- Recall (Sensitivity) – It explains how many of the actual positive cases model is able to predict correctly. The formula is given below in equation 3.9:

$$\text{Recall} = \frac{\text{TP}}{\text{TP} + \text{FN}} \tag{3.9}$$

- Specificity – It basically complements Sensitivity; it explains how well the negative cases model is able to predict correctly. The formula is given below in equation 3.10:

$$\text{Specificity} = \frac{\text{TN}}{\text{FP} + \text{TN}} \tag{3.10}$$

- F1 Score – It is a combination of both precision and recall. This metric is evaluated when there is a need to maintain a balance between both precision and recall. The formula is given below in equation 3.11:

$$\text{F1Score} = 2 * \frac{\text{Precision} * \text{Recall}}{\text{Precision} + \text{Recall}} \tag{3.11}$$

- AUC-ROC – The Receiver Operator Characteristic (ROC) is a probability curve where TPR (True Positive Rate) is plotted against the FPR (False Positive Rate). This curve is measuring the capability of the classifier to differentiate between the classes properly. The greater is the area under the curve the better is the performance of the model.

The abbreviation of TP, FP, TN, FN are True positive, False positive, True negative and False negative respectively.

Studies show that the evaluation metrices like Recall, Precision, F1Score and AUC curve were usually used for classification problems in medical domains (Kayyum et al., 2020; Liang et al., 2021; Richards et al., 2021). Accuracy is not an appropriate statistic to judge because it will cause the model to be biased towards the majority class in the medical domain or healthcare data. In this specific project identification of all the positive class accurately is very important and impactful than identification of the negative class, therefore, Recall (Sensitivity) and AUCROC evaluation metric needs to be evaluated accurately along with the other classification metrices.

Finally, a comparative study will be done between all the proposed models and based on the above discussed evaluation metrices, best performing model will be selected for prediction of the acute myocardial infarction.

### 3.3.8 Hardware Requirements

Hardware requirements to execute this project are as follows:

1. RAM - 16 GB
2. Hard Disk - 1.2 TB
3. Graphics Card - NVIDIA GeForce GTX 1650 (4 GB) or above
4. Processor - Intel Core i7 or above

### 3.3.9 Software Requirements

The software requirements to execute this project are as follows:

1. Jupyter Notebook
2. Python version 3.8.5

Python libraries required to execute this project are as follows:

1. NumPy
2. Seaborn
3. Matplotlib
4. Pandas
5. Scikit-Learn
6. Statsmodels
7. Keras
8. TensorFlow
9. Imblearn
10. Scipy

## 3.4 Summary

The detail research methodology and dataset that will be used to predict lethal outcomes after acute myocardial infarction are briefly described in this chapter. Some extensively used performance metrices for measuring the performance of the suggested machine learning model for classification task have also been thoroughly discussed. The next section includes a full description of the dataset that will be used in this research. Following that, several data preparation processes, such as missing value handling, data scaling, and so on, that will be done on the raw data were discussed. To improve model performance, it's also necessary to deal with class imbalance, which has been briefed out in depth. Following that, another essential aspect, feature selection methods, was explained, which allows the machine learning system to train more quickly. It also simplifies a model's complexity and makes it easier to understand. When the proper subset is picked, a model's accuracy improves.

Various machine learning techniques, as well as ANN, which will be employed in this research, have been thoroughly discussed with diagram and its equations. The various types of ensembles learning that will be used in this thesis have been set out, including random forest, SVM bagging classifier, and stacking blending. All these algorithms, including logistic regression, LGBM, and random forest, have shown to produce better results in the past. As a result, in order to improve overall performance, the stacking ensemble of all these methods will be used in this thesis. Based on the proper evaluation metrices, best performing model will be selected for prediction.

This study requires a variety of hardware and software prerequisites, which have been specified. Because the deep learning model involves the use of GPU, which is readily available on the cloud and does not require a local setup, this study will also look into cloud infrastructure possibilities.

# CHAPTER 4

# ANALYSIS AND DESIGN

## 4.1 Introduction

The model building step is covered in this chapter, and the sub-chapters will provide a thorough explanation of the technique used in this research. The chapter will begin with a description of the dataset used in this thesis, plus numerous parameters that are included in it. All procedures associated with data pre-processing for further analysis will be provided. These processes include removing uninteresting variables using various feature selection approaches, as well as transforming the data into explanatory data by removing unrelated variables and converting it to categorical format. The distribution of the variables in the dataset will be determined through the detection of missing values and univariate analysis. Following that, missing values will be treated using the iterative imputation technique to make subsequent analysis easier and to reduce biases or inconsistency during analysis. The entire myocardial dataset will be subjected to both bivariate and multivariate exploratory data analysis to identify the significance of the association between the independent and dependent variables. Some of the independent factors have also been subjected to outlier treatment. The myocardial infraction dataset will then be divided into two sets: training and test.

Three class imbalance handling strategies will be applied to the dataset before to the use of data mining techniques to address the problem of class imbalance in the dataset. On the training dataset, the steps taken to treat class imbalance are oversampling (SMOTE), adaptive sampling (ADASYN), and class weighted balancing procedures will be detailed out. The model validation approach will be discussed, along with the process flow of the classifiers to explain the setup and any modifications in the parameters. Each classifier's outputs will be discussed in further depth. The measurements used to evaluate the performance of the classifiers using the confusion matrix will also be included in further sections.

### 4.2 Dataset Description

The dataset for this investigation was obtained from the open world repository of the University of California, Irvine (Mirkes et al., 2020). The presented dataset is made up of medical records from 1,700 patients with cardiac diseases who were treated at Berzon Krasnoyarsk Clinical Hospital in Russia. This dataset has 113 variables that include patient demographics and medical history and was collected across three time periods: during hospitalization, after 48 hours, and after 72 hours. This dataset contains 12 complications / outcomes after acute myocardial infarction, making this a multiclass classification task. Because the raw dataset is unbalanced, contains many missing values, and contains outliers in some of the features, it must be pre-processed before predictive models can be built. All the study's ethical difficulties and consequences were considered. Because the dataset was obtained from an open data source, the UC Irvine Machine Learning Repository website, there were no substantial ethical considerations in this study. Because this dataset is an open-source dataset that is available to the public for research, no ethical paperwork or declaration forms are necessary.

We have selected 56 essential features out of 113 independent variables after applying several feature selection methodologies. For a better understanding of the selected predictor variables offered in this study, see below table 4.1, which contains significant attribute names, descriptions, data types, and allowable values.

*Table 4.1 Selected predictor attributes from presented myocardial infarction dataset*

| Independent Feature Name | Feature Description / Permissible Value |
|---|---|
| AGE (Numerical) | Age (Range 26 - 92 years) |
| ALT_BLOOD (Numerical) | AlAT content in serum (Range 0.030 - 3.0) |
| ANT_CA_S_n (Categorical) | Calcium channel blockers are commonly used in intensive care units<br>**0 – no; 1 – yes** |
| ant_im (Categorical) | Left ventricular ECG changes in leads V1–V4 in the presence of an anterior myocardial infarction<br>**0 – there is no infarct in this location; 1 – QRS has no changes; 2 – QRS is like QR-complex; 3 – QRS is like Qr-complex; 4 – QRS is like QS-complex** |
| ASP_S_n (Categorical) | In the intensive care unit, acetylsalicylic acid is used<br>**0 – no;1 – yes** |
| B_BLOK_S_n (Categorical) | In the intensive care unit, beta-blockers are used<br>**0 – no;1 – yes** |
| D_AD_ORIT (Numerical) | Per the intensive care unit, diastolic blood pressure (Range 0.0 - 190.0) |
| DLIT_AG (Categorical) | The time span of arterial hypertension<br>**0 – there was no arterial hypertension; 1 – one year; 2 – two years; 3 – three years; 4 – four years; 5 – five years; 6 – 6-10 years; 7 – more than 10 years** |
| endocr_01 (Categorical) | Diabetes mellitus in the anamnesis<br>**0 – no; 1 – yes** |
| FK_STENOK (Categorical) | In the previous year, functional category of angina pectoris<br>**0 – there is no angina pectoris; 1 – I FC; 2 – IFC; 3 – III FC; 4 – IV FC** |
| GB (Categorical) | Presence of an essential hypertension<br>**0 – there is no essential hypertension; 1 – Stage 1; 2 – Stage 2; 3 – Stage 3** |
| GEPAR_S_n (Categorical) | In the ICU, heparin is used as an anticoagulant<br>**0 – no; 1 – yes** |
| GIPO_K (Categorical) | Hypokalaemia (less than 4 mmol/L)<br>**0 – no; 1 – yes** |
| IBS_POST (Categorical) | Coronary heart disease (CHD) in recent weeks, days prior hospitalization<br>**0 – there was no CHD; 1 – exertional angina pectoris; 2 – unstable angina pectoris** |
| IM_PG_P (Categorical) | A right ventricular myocardial infarction is present<br>**0 – no; 1 – yes** |
| INF_ANAM (Categorical) | The multitude of myocardial infarctions in anamnesis<br>**0 – zero; 1 – one; 2 – two; 3 – three and more** |
| inf_im (Categorical) | Left ventricular ECG changes in leads III, AVF, and II in the presence of an inadequate myocardial infarction<br>**0 – there is no infarct in this location; 1 – QRS has no changes; 2 – QRS is like QR-complex; 3 – QRS is like Qr-complex; 4 – QRS is like QS-complex** |
| K_BLOOD (Numerical) | Potassium content in serum (Range 2.30 - 8.20) |
| K_SH_POST (Categorical) | At the time of admission to the intensive care unit, the patient was experiencing cardiogenic shock<br>**0 – no; 1 – yes** |
| lat_im (Categorical) | Left ventricular ECG changes in leads V5–V6, I, AVL in the presence of a lateral myocardial infarction<br>**0 – there is no infarct in this location; 1 – QRS has no changes; 2 – QRS is like QR-complex; 3 – QRS is like Qr-complex; 4 – QRS is like QS-complex** |
| LID_KB (Categorical) | The Emergency Cardiology Team uses lidocaine<br>**0 – no; 1 – yes** |
| LID_S_n (Categorical) | In the ICU, lidocaine is used<br>**0 – no; 1 – yes** |
| MP_TP_POST (Categorical) | Atrial fibrillation paroxysms at the time of admission to the intensive care unit or at a prehospital phase<br>**0 – no; 1 – yes** |
| n_p_ecg_p_06 (Categorical) | At the time of admission to the hospital, the ECG revealed a third-degree AV block.<br>**0 – no; 1 – yes** |
| n_p_ecg_p_07 (Categorical) | At the time of admission to the hospital, the ECG showed LBBB (anterior branch)<br>**0 – no; 1 – yes** |
| n_p_ecg_p_12 (Categorical) | At the time of admission to the hospital, complete an ECG with RBBB<br>**0 – no; 1 – yes** |
| n_r_ecg_p_01 (Categorical) | At the time of admission to the hospital, the ECG revealed premature atrial contractions<br>**0 – no; 1 – yes** |
| n_r_ecg_p_03 (Categorical) | At the time of admission to the hospital, the ECG revealed premature ventricular contractions<br>**0 – no; 1 – yes** |
| n_r_ecg_p_04 (Categorical) | At the time of admission to the hospital, the ECG showed frequent premature ventricular contractions<br>**0 – no; 1 – yes** |
| n_r_ecg_p_05 (Categorical) | Atrial fibrillation paroxysms on ECG at the time of admission to the hospital |
| | **0 – no; 1 – yes** |
| NA_BLOOD (Numerical) | Sodium content in serum (Range 117.0 - 169.0) |
| NA_KB (Categorical) | The Emergency Cardiology Team's use of opioid medications<br>**0 – no; 1 – yes** |
| NA_R_1_n (Categorical) | In the early hours of a patient's stay in the hospital, opioid medications are commonly used in the ICU<br>**0 – no; 1 – once; 2 – twice; 3 – three times; 4 – four times** |
| NA_R_2_n (Categorical) | In the second day of the hospital stay, opioid drugs were used in the ICU<br>**0 – no; 1 – once; 2 – twice; 3 – three times** |
| NITR_S (Categorical) | In the ICU, liquid nitrates are used<br>**0 – no; 1 – yes** |
| NOT_NA_1_n (Categorical) | NSAIDs in the ICU during the first hours of hospitalisation<br>**0 – no; 1 – once; 2 – twice; 3 – three times; 4 – four or more times** |
| NOT_NA_3_n (Categorical) | Use of NSAIDs in the ICU in the third day of the hospital period<br>**0 – no; 1 – once; 2 – twice** |
| NOT_NA_KB (Categorical) | Use of NSAIDs by the Emergency Cardiology Team<br>**0 – no; 1 – yes** |
| nr_04 (Categorical) | In the anamnesis, a chronic form of atrial fibrillation<br>**0 – no; 1 – yes** |
| O_L_POST (Categorical) | The patient developed pulmonary edoema when he was admitted to the intensive care unit<br>**0 – no; 1 – yes** |
| R_AB_1_n (Categorical) | In the initial hours of the hospital stay, the agony returned<br>**0 – there is no relapse; 1 – only one; 2 – 2 times 3 – 3 or more times** |
| R_AB_2_n (Categorical) | The discomfort returned on the second day of the hospital stay |
| | **0 – there is no relapse; 1 – only one; 2 – 2 times 3 – 3 or more times** |
| R_AB_3_n (Categorical) | In the third day of the hospital stay, the pain returned **0 – there is no relapse; 1 – only one; 2 – 2 times 3 – 3 or more times** |
| ritm_ecg_p_01 (Categorical) | At the time of admittance to the hospital, the ECG rhythm was sinus (with a heart rate 60-90) **0 – no; 1 – yes** |
| ritm_ecg_p_02 (Categorical) | Atrial fibrillation was detected on the ECG at the time of admission to the hospital **0 – no; 1 – yes** |
| ritm_ecg_p_07 (Categorical) | At the time of admission to the hospital, the ECG rhythm was sinus with a heart rate of more than 90 beats per minute (tachycardia) **0 – no; 1 – yes** |
| S_AD_ORIT (Numerical) | According to the intensive care unit, systolic blood pressure is (Range 0.0 - 260.0) |
| SEX (Categorical) | Gender **0 – female; 1 – male** |
| STENOK_AN (Categorical) | Anamnesis of exercise-induced angina pectoris **0 – never; 1 – during the last year; 2 – one year ago; 3 – two years ago; 4 – three years ago; 5 – 4-5 years ago; 6 – more than 5 years ago** |
| TIME_B_S (Categorical) | Time it took from the start of the CHD attack to getting to the hospital **1 – less than 2 hours; 2 – 2-4 hours; 3 – 4-6 hours; 4 – 6-8 hours; 5 – 8-12 hours; 6 – 12-24 hours; 7 – more than 1 days; 8 – more than 2 days; 9 – more than 3 days** |
| TRENT_S_n (Categorical) | Trental in the Intensive Care Unit **0 – no; 1 – yes** |
| zab_leg_01 (Categorical) | Chronic bronchitis in the anamnesis **0 – no; 1 – yes** |
| zab_leg_02 (Categorical) | In the anamnesis, obstructive chronic bronchitis **0 – no; 1 – yes** |
| zab_leg_03 (Categorical) | Anamnesis of bronchial asthma |

| | **0 – no; 1 – yes** |
|---|---|
| ZSN_A (Categorical) | Chronic heart failure (HF) is present in the anamnesis. **0 – there is no chronic heart failure; 1 – I stage 2 – IIA stage (heart failure due to right ventricular systolic dysfunction); 3 – IIA stage (heart failure due to left ventricular systolic dysfunction); 4 – IIB stage (heart failure due to left and right ventricular systolic dysfunction)** |

There are 12 complications in this dataset, and out of which one has been chosen as a dependent/target variable, namely Lethal outcome (cause) (LET_IS). For a better understanding of the dependent variable that have been chosen for this study, see table 4.2, which includes the target attribute name, description, data type, and allowed value.

*Table 4.2 Selected target attribute from presented myocardial infarction dataset*

| **Target Feature Name** | **Feature Description / Permissible Value** |
|---|---|
| LET_IS (Categorical) | Lethal outcome (cause) **0 – unknown; 1 – cardiogenic shock; 2 – pulmonary edema; 3 – myocardial rupture; 4 – progress of congestive heart failure; 5 – thromboembolism; 6 – asystole; 7 – ventricular fibrillation** |

### 4.3 Data Preparation

Although the source document is in a structured manner, it is not clean and necessitates several pre-processing procedures before analysis can be performed to satisfy the study's objectives. Each pre-processing stage will be described in depth, including the objective of each step as well as the results of each procedure.

#### 4.3.1 Elimination of Variables

The 'ID' feature preserves the Record ID of each patient admitted to the hospital; this is only for the purpose of storing the patient number to uniquely identify a patient and has nothing to do with myocardial or cardiovascular illness. This parameter has been removed using panda's library with Python version 3.8.5 from the original dataset because it is not a myocardial or cardiovascular disease characteristic. In this research, the 'ID' variable serves no function or has no importance. As a result of this pre-processing procedure, the data now contains 112 variables with 1,700 instances that are linked to myocardial infarction.

### 4.3.2 Transformation of Variables

Out of 112 variables presented in the dataset, there are 21 categorical variables that needs to be converted to numerical for modelling purpose and 4 numerical variables that needs to be converted to categorical variable for exploratory data analysis and visualization purpose. All these conversions have been done using panda's library with Python version 3.8.5. Below subsection will provide the detail about the features before and after conversion.

#### 4.3.2.1 Categorical to Numerical Conversion

Below table 4.3 detailed about the dummification of the 21 categorical features to numerical format for classification modelling.

*Table 4.3 Dummification of categorical attributes from myocardial infarction dataset*

| Categorical Feature | Dummified Feature | Permissible Value |
|---|---|---|
| **INF_ANAM** | INF_ANAM_0 | 1 – zero multitude of myocardial infarctions in anamnesis; 0- Other value |
| | INF_ANAM_1 | 1 – one multitude of myocardial infarctions in anamnesis; 0- Other value |
| | INF_ANAM_2 | 1 – two multitude of myocardial infarctions in anamnesis; 0- Other value |
| | INF_ANAM_3 | 1 – three multitude of myocardial infarctions in anamnesis; 0- Other value |
| **STENOK_AN** | STENOK_AN_0 | 1 – no anamnesis of exertional angina pectoris; 0- Other value |
| | STENOK_AN_1 | 1 – during the last year anamnesis of exertional angina pectoris; 0- Other value |
| | STENOK_AN_2 | 1 – one year ago anamnesis of exertional angina pectoris; 0- Other value |
| | STENOK_AN_3 | 1 – two years ago anamnesis of exertional angina pectoris; 0- Other value |
| | STENOK_AN_4 | 1 – three years ago anamnesis of exertional angina pectoris; 0- Other value |
|  | STENOK_AN_5 | 1 – 4-5 years ago anamnesis of exertional angina pectoris; 0- Other value |
|  | STENOK_AN__6 | 1 – more than 5 years ago anamnesis of exertional angina pectoris; 0- Other value |
| **FK_STENOK** | FK_STENOK_0 | 1- in the previous year, No functional category of angina pectoris; 0- Other value |
|  | FK_STENOK_1 | 1- in the previous year, I functional category of angina pectoris; 0- Other value |
|  | FK_STENOK_2 | 1- in the previous year, II functional category of angina pectoris; 0- Other value |
|  | FK_STENOK_3 | 1- in the previous year, III functional category of angina pectoris; 0- Other value |
|  | FK_STENOK_4 | 1- in the previous year, IV functional category of angina pectoris; 0- Other value |
| **IBS_POST** | IBS_POST_0 | 1- no coronary heart disease (CHD) in recent weeks, days prior hospitalization; 0- Other value |
|  | IBS_POST_1 | 1- exertional angina pectoris coronary heart disease (CHD) in recent weeks, days prior hospitalization; 0- Other value |
|  | IBS_POST_2 | 1- unstable angina pectoris coronary heart disease (CHD) in recent weeks, days prior hospitalization; 0- Other value |
| **GB** | GB_0 | 1- no presence of an essential hypertension; 0- Other value |
|  | GB_1 | 1- stage 1 presence of an essential hypertension; 0- Other value |
|  | GB_2 | 1- stage 2 presence of an essential hypertension; 0- Other value |
|  | GB_3 | 1- stage 3 presence of an essential hypertension; 0- Other value |
| **DLIT_AG** | DLIT_AG_0 | 1- no Arterial hypertension duration; 0- Other value |
|  | DLIT_AG_1 | 1- one-year Arterial hypertension duration; 0- Other value |
|  | DLIT_AG_2 | 1- two-year Arterial hypertension duration; 0- Other value |
| | DLIT_AG_3 | 1- three-year Arterial hypertension duration; 0- Other value |
| | DLIT_AG_4 | 1- four years Arterial hypertension duration; 0- Other value |
| | DLIT_AG_5 | 1- five years Arterial hypertension duration; 0- Other value |
| | DLIT_AG_6 | 1- 6-10 years Arterial hypertension duration; 0- Other value |
| | DLIT_AG_7 | 1- more than 10 years Arterial hypertension duration; 0- Other value |
| **ZSN_A** | ZSN_A_0 | 1-no chronic Heart failure (HF) in the anamnesis; 0- Other value |
| | ZSN_A_1 | 1-I stage chronic Heart failure (HF) in the anamnesis; 0- Other value |
| | ZSN_A_2 | 1-IIA stage chronic Heart failure (HF) in the anamnesis; 0- Other value |
| | ZSN_A_3 | 1-IIA stage chronic Heart failure (HF) in the anamnesis; 0- Other value |
| | ZSN_A_4 | 1-IIB stage chronic Heart failure (HF) in the anamnesis; 0- Other value |
| **ant_im** | ant_im_0 | 1- there is no infarct in left ventricular anterior myocardial infarction; 0- Other value |
| | ant_im_1 | 1- QRS has no changes in left ventricular anterior myocardial infarction; 0- Other value |
| | ant_im_2 | 1- QRS is like QR-complex in left ventricular anterior myocardial infarction; 0- Other value |
| | ant_im_3 | 1- QRS is like Qr-complex in left ventricular anterior myocardial infarction; 0- Other value |
| | ant_im_4 | 1-QRS is like QS-complex in left ventricular anterior myocardial infarction; 0- Other value |
| **lat_im** | lat_im_0 | 1- there is no infarct in lateral left ventricular anterior myocardial infarction; 0- Other value |
| | lat_im_1 | 1- QRS has no changes in lateral left ventricular anterior myocardial infarction; 0- Other value |
| | lat_im_2 | 1- QRS is like QR-complex in lateral left ventricular anterior myocardial infarction; 0- Other value |
| | lat_im_3 | 1- QRS is like Qr-complex in lateral left ventricular anterior myocardial infarction; 0- Other value |
| | lat_im_4 | 1-QRS is like QS-complex in lateral left ventricular anterior myocardial infarction; 0- Other value |
| **inf_im** | inf_im_0 | 1- there is no infarct in inferior left ventricular anterior myocardial infarction; 0- Other value |
| | inf_im_1 | 1- QRS has no changes in inferior left ventricular anterior myocardial infarction; 0- Other value |
| | inf_im_2 | 1- QRS is like QR-complex in inferior left ventricular anterior myocardial infarction; 0- Other value |
| | inf_im_3 | 1- QRS is like Qr-complex in inferior left ventricular anterior myocardial infarction; 0- Other value |
| | inf_im_4 | 1-QRS is like QS-complex in inferior left ventricular anterior myocardial infarction; 0- Other value |
| **post_im** | post_im_0 | 1- there is no infarct in posterior left ventricular anterior myocardial infarction; 0- Other value |
| | post_im_1 | 1- QRS has no changes in posterior left ventricular anterior myocardial infarction; 0- Other value |
| | post_im_2 | 1- QRS is like QR-complex in posterior left ventricular anterior myocardial infarction; 0- Other value |
| | post_im_3 | 1- QRS is like Qr-complex in posterior left ventricular anterior myocardial infarction; 0- Other value |
| | post_im_4 | 1-QRS is like QS-complex in posterior left ventricular anterior myocardial infarction; 0-Other value |
| **TIME_B_S** | TIME_B_S_1 | 1 - less than 2 hours duration elapsed from the starting of the attack of CHD; 0- Other value |
| | TIME_B_S_2 | 1 - 2-4 hours duration elapsed from the starting of the attack of CHD; 0- Other value |
| | TIME_B_S_3 | 1 - 4-6 hours duration elapsed from the starting of the attack of CHD; 0- Other value |
| | TIME_B_S_4 | 1 - 6-8 hours duration elapsed from the starting of the attack of CHD; 0- Other value |
| | TIME_B_S_5 | 1 - 8-12 hours duration elapsed from the starting of the attack of CHD; 0- Other value |
| | TIME_B_S_6 | 1 - 12-24 hours duration elapsed from the starting of the attack of CHD; 0- Other value |
| | TIME_B_S_7 | 1 - more than 1 day’s duration elapsed from the starting of the attack of CHD: 0- Other value |
| | TIME_B_S_8 | 1 - more than 2 days duration elapsed from the starting of the attack of CHD: 0- Other value |
| | TIME_B_S_9 | 1 - more than 3 days duration elapsed from the starting of the attack of CHD: 0- Other value |
| **R_AB_1_n** | R_AB_1_n_0 | 1 – no reoccurrence of the pain in first hour; 0-Other value |
| | R_AB_1_n_1 | 1 - only one reoccurrence of the pain in first hour; 0- Other value |
| | R_AB_1_n_2 | 1 - 2 times reoccurrence of the pain in first hour; 0- Other value |
| | R_AB_1_n_3 | 1 - 3 or more times reoccurrence of the pain in first hour; 0- Other value |
| **R_AB_2_n** | R_AB_2_n_0 | 1 – No reoccurrence of the pain in second hour; 0- Other value |
| | R_AB_2_n_1 | 1 - only one reoccurrence of the pain in second hour; 0- Other value |
| | R_AB_2_n_2 | 1 - 2 times reoccurrence of the pain in second hour; 0- Other value |
|  | R_AB_2_n_3 | 1 - 3 or more times reoccurrence of the pain in second hour; 0- Other value |
| **R_AB_3_n** | R_AB_3_n_0 | 1 – no reoccurrence of the pain in third hour; 0- Other value |
|  | R_AB_3_n_1 | 1 - only one reoccurrence of the pain in third hour; 0- Other value |
|  | R_AB_3_n_2 | 1 - 2 times reoccurrence of the pain in third hour; 0- Other value |
|  | R_AB_3_n_3 | 1 - 3 or more times reoccurrence of the pain in third hour; 0- Other value |
| **NA_R_1_n** | NA_R_1_n_0 | 1 – no opioid drugs use in first hour in ICU; 0- Other value |
|  | NA_R_1_n_1 | 1 – once opioid drugs use in first hour in ICU; 0- Other value |
|  | NA_R_1_n_2 | 1 – twice opioid drugs use in first hour in ICU; 0- Other value |
|  | NA_R_1_n_3 | 1 – three times opioid drugs use in first hour in ICU; 0- Other value |
|  | NA_R_1_n_4 | 1 – four times opioid drugs use in first hour in ICU; 0- Other value |
| **NA_R_2_n** | NA_R_2_n_0 | 1 – no opioid drugs use in second day in ICU; 0- Other value |
|  | NA_R_2_n_1 | 1 – once opioid drugs use in second day in ICU; 0- Other value |
|  | NA_R_2_n_2 | 1 – twice opioid drugs use in second day in ICU; 0- Other value |
|  | NA_R_2_n_3 | 1 – three times opioid drugs use in second day in ICU; 0- Other value |
| **NA_R_3_n** | NA_R_3_n_0 | 1 – no opioid drugs use in third day in ICU; 0- Other value |
|  | NA_R_3_n_1 | 1 – once opioid drugs use in third day in ICU; 0- Other value |
|  | NA_R_3_n_2 | 1 – twice opioid drugs use in third day in ICU; 0- Other value |
| **NOT_NA_1_n** | NOT_NA_1_n_0 | 1 – no NSAIDs used in the ICU in the first hours; 0- Other value |
| | NOT_NA_1_n_1 | 1 – once NSAIDs used in the ICU in the first hours; 0- Other value |
| | NOT_NA_1_n_2 | 1 – twice NSAIDs used in the ICU in the first hours; 0- Other value |
| | NOT_NA_1_n_3 | 1 – three times NSAIDs used in the ICU in the first hours; 0- Other value |
| | NOT_NA_1_n_4 | 1 – four or more times NSAIDs used in the ICU in the first hours; 0- Other value |
| **NOT_NA_2_n** | NOT_NA_2_n_0 | 1 – no NSAIDs used in the ICU in the second day; 0- Other value |
| | NOT_NA_2_n_1 | 1 – once NSAIDs used in the ICU in the second day; 0- Other value |
| | NOT_NA_2_n_2 | 1 – twice NSAIDs used in the ICU in the second day; 0- Other value |
| | NOT_NA_2_n_3 | 1 – three times NSAIDs used in the ICU in the second day; 0- Other value |
| **NOT_NA_3_n** | NOT_NA_3_n_0 | 1 – no NSAIDs used in the ICU in the third day; 0- Other value |
| | NOT_NA_3_n_1 | 1 – once NSAIDs used in the ICU in the third day; 0- Other value |
| | NOT_NA_3_n_2 | 1 – twice NSAIDs used in the ICU in the third day; 0- Other value |

As a result of this dummification procedure, the data now contains 167 variables including target variable with 1,700 instances that are linked to myocardial infarction and has been considered for further processing.

#### 4.3.2.2 Numerical to Categorical Conversion

The Pandas library's binning method was used to turn 4 numerical features into categorical columns. This transformation can be used to see which value ranges are appropriate indicators for the target variable. The binning values for the transformed categorical features are described below in table 4.4.

*Table 4.4 Binning of numerical attributes from myocardial infarction dataset*

| Numerical Feature | Converted Column | Permissible Value |
|---|---|---|
| **AGE** | AGE_Interval | 10-20; Age between 10 to 20 years |
| | | 20-30; Age between 10 to 20 years |
| | | 30-40; Age between 10 to 20 years |
| | | 40-50; Age between 10 to 20 years |
| | | 50-60; Age between 10 to 20 years |
| | | 60-70; Age between 10 to 20 years |
| | | 70-80; Age between 10 to 20 years |
| | | 80-90; Age between 10 to 20 years |
| | | 90-above; Age between 10 to 20 years |
| **ROE** | ROE_Interval | 0-10; Erythrocyte sedimentation rate between 0 to 10 |
| | | 10-20; Erythrocyte sedimentation rate between `0 to 10 |
| | | 20-30; Erythrocyte sedimentation rate between 20 to 30 |
| | | 30-40; Erythrocyte sedimentation rate between 30 to 40 |
| | | 40-50; Erythrocyte sedimentation rate between 40 to 50 |
| | | 50-60; Erythrocyte sedimentation rate between 50 to 60 |
| | | 60-70; Erythrocyte sedimentation rate between 60 to 70 |
| | | 70-80; Erythrocyte sedimentation rate between 70 to 80 |
| | | 80-90; Erythrocyte sedimentation rate between 8 0 to 90 |
| | | 90-100; Erythrocyte sedimentation rate between 90 to 100 |
| | | 100-110; Erythrocyte sedimentation rate between 100 to 110 |
| | | 110-120; Erythrocyte sedimentation rate between 110 to 120 |
| | | 120-130; Erythrocyte sedimentation rate between 120 to 120 |
| | | 130-140; Erythrocyte sedimentation rate between 130 to 140 |
| | | 140-above; Erythrocyte sedimentation rate between 140 and above |
| **S_AD_ORIT** | S_AD_ORIT_Interval | 0-30; Systolic blood pressure between 0 -30 mmHg |
| | | 30-60; Systolic blood pressure between 30 -60 mmHg |
| | | 60-90; Systolic blood pressure between 60 -90 mmHg |
| | | 90-120; Systolic blood pressure between 90 -120 mmHg |
| | | 120-150; Systolic blood pressure between 120 - 150 mmHg |
| | | 150-180; Systolic blood pressure between150 - 180 mmHg |
| | | 180-210; Systolic blood pressure between 180 - 210 mmHg |
| | | 210-240; Systolic blood pressure between 210 - 240 mmHg |
| | | 240-above; Systolic blood pressure between 240 and above mmHg |
| **D_AD_ORIT** | D_AD_ORIT_Interval | 0-20; Diastolic blood pressure between 0 -20 mmHg |
| | | 20-40; Diastolic blood pressure between 20 -40 mmHg |
| | | 40-60; Diastolic blood pressure between 40 -60 mmHg |
| | | 60-80; Diastolic blood pressure between 60 -80 mmHg |
| | | 80-100; Diastolic blood pressure between 80 - 100 mmHg |
| | | 100-120; Diastolic blood pressure between 100 - 120 mmHg |
| | | 120-140; Diastolic blood pressure between 120 - 140 mmHg |
| | | 140-160; Diastolic blood pressure between 140 - 160 mmHg |
| | | 160-180; Diastolic blood pressure between 160 - 180 mmHg |
| | | 180-200; Diastolic blood pressure between 180 - 200 mmHg |
| | | 200-220; Diastolic blood pressure between 200 - 220 mmHg |
| | | 220-above; Diastolic blood pressure between 220 and above mmHg |

### 4.3.3 Identification of Missing Values

According to the presented research, most myocardial infarction-related factors in the sample have missing values in the provided dataset. If a value is missing, any analysis on those variables will be less efficient, and the conclusion may be slanted or skewed. Also, majority of the python libraries will throw error due to presence of null value in features during modelling phase. The missing value must be handled to avoid such problem. Missing or unknown value in the original presented dataset is represented by '?'. For further processing this '?' representation has been changed to NaN or null value. Then missing value analysis has been performed using pandas and missingno library with Python version 3.8.5.

Below table 4.5 listed down 110 features with corresponding percentage of missing value and missing count.

*Table 4.5 Missing value percentage from myocardial infarction dataset*

| Feature Name | Missing Value Count | Missing Value Percentage |
|---|---|---|
| KFK_BLOOD | 1696 | 99.76 |
| IBS_NASL | 1628 | 95.76 |
| S_AD_KBRIG | 1076 | 63.29 |
| D_AD_KBRIG | 1076 | 63.29 |
| NOT_NA_KB | 686 | 40.35 |
| LID_KB | 677 | 39.82 |
| NA_KB | 657 | 38.65 |

| GIPER_NA | 375 | 22.06 |
|---|---|---|
| NA_BLOOD | 375 | 22.06 |
| K_BLOOD | 371 | 21.82 |
| GIPO_K | 369 | 21.71 |
| AST_BLOOD | 285 | 16.76 |
| ALT_BLOOD | 284 | 16.71 |
| S_AD_ORIT | 267 | 15.71 |
| D_AD_ORIT | 267 | 15.71 |
| DLIT_AG | 248 | 14.59 |
| ROE | 203 | 11.94 |
| ritm_ecg_p_06 | 152 | 8.94 |
| ritm_ecg_p_08 | 152 | 8.94 |
| ritm_ecg_p_01 | 152 | 8.94 |
| ritm_ecg_p_07 | 152 | 8.94 |
| ritm_ecg_p_02 | 152 | 8.94 |
| ritm_ecg_p_04 | 152 | 8.94 |
| NA_R_3_n | 131 | 7.71 |
| NOT_NA_3_n | 131 | 7.71 |
| R_AB_3_n | 128 | 7.53 |
| TIME_B_S | 126 | 7.41 |
| L_BLOOD | 125 | 7.35 |
| n_r_ecg_p_08 | 115 | 6.76 |
| n_r_ecg_p_03 | 115 | 6.76 |
| n_r_ecg_p_04 | 115 | 6.76 |
| n_r_ecg_p_05 | 115 | 6.76 |
| n_r_ecg_p_06 | 115 | 6.76 |
| n_r_ecg_p_02 | 115 | 6.76 |
| n_r_ecg_p_09 | 115 | 6.76 |
| n_r_ecg_p_10 | 115 | 6.76 |
| n_p_ecg_p_01 | 115 | 6.76 |
| n_p_ecg_p_03 | 115 | 6.76 |
| n_p_ecg_p_04 | 115 | 6.76 |
| n_p_ecg_p_05 | 115 | 6.76 |
| n_p_ecg_p_06 | 115 | 6.76 |
| n_p_ecg_p_07 | 115 | 6.76 |

| n_p_ecg_p_08 | 115 | 6.76 |
|---|---|---|
| n_p_ecg_p_10 | 115 | 6.76 |
| n_p_ecg_p_11 | 115 | 6.76 |
| n_p_ecg_p_12 | 115 | 6.76 |
| n_p_ecg_p_09 | 115 | 6.76 |
| n_r_ecg_p_01 | 115 | 6.76 |
| NOT_NA_2_n | 110 | 6.47 |
| R_AB_2_n | 108 | 6.35 |
| NA_R_2_n | 108 | 6.35 |
| STENOK_AN | 106 | 6.24 |
| ant_im | 83 | 4.88 |
| inf_im | 80 | 4.71 |
| lat_im | 80 | 4.71 |
| FK_STENOK | 73 | 4.29 |
| post_im | 72 | 4.24 |
| ZSN_A | 54 | 3.18 |
| IBS_POST | 51 | 3.00 |
| nr_11 | 21 | 1.24 |
| nr_01 | 21 | 1.24 |
| nr_08 | 21 | 1.24 |
| nr_07 | 21 | 1.24 |
| nr_04 | 21 | 1.24 |
| nr_03 | 21 | 1.24 |
| nr_02 | 21 | 1.24 |
| np_05 | 18 | 1.06 |
| np_01 | 18 | 1.06 |
| np_04 | 18 | 1.06 |
| np_10 | 18 | 1.06 |
| np_09 | 18 | 1.06 |
| np_08 | 18 | 1.06 |
| np_07 | 18 | 1.06 |
| ASP_S_n | 17 | 1.00 |
| GEPAR_S_n | 17 | 1.00 |
| TRENT_S_n | 16 | 0.94 |
| R_AB_1_n | 16 | 0.94 |

| TIKL_S_n | 16 | 0.94 |
|---|---|---|
| K_SH_POST | 15 | 0.88 |
| MP_TP_POST | 14 | 0.82 |
| ANT_CA_S_n | 13 | 0.76 |
| O_L_POST | 12 | 0.71 |
| SVT_POST | 12 | 0.71 |
| GT_POST | 12 | 0.71 |
| FIB_G_POST | 12 | 0.71 |
| B_BLOK_S_n | 11 | 0.65 |
| endocr_01 | 11 | 0.65 |
| NOT_NA_1_n | 10 | 0.59 |
| LID_S_n | 10 | 0.59 |
| fibr_ter_06 | 10 | 0.59 |
| fibr_ter_08 | 10 | 0.59 |
| fibr_ter_03 | 10 | 0.59 |
| endocr_03 | 10 | 0.59 |
| fibr_ter_01 | 10 | 0.59 |
| fibr_ter_02 | 10 | 0.59 |
| endocr_02 | 10 | 0.59 |
| fibr_ter_05 | 10 | 0.59 |
| fibr_ter_07 | 10 | 0.59 |
| GB | 9 | 0.53 |
| NITR_S | 9 | 0.53 |
| SIM_GIPERT | 8 | 0.47 |
| AGE | 8 | 0.47 |
| zab_leg_01 | 7 | 0.41 |
| zab_leg_02 | 7 | 0.41 |
| zab_leg_03 | 7 | 0.41 |
| zab_leg_04 | 7 | 0.41 |
| zab_leg_06 | 7 | 0.41 |
| NA_R_1_n | 5 | 0.29 |
| INF_ANAM | 4 | 0.24 |
| IM_PG_P | 1 | 0.06 |

Below figure 4.1 and 4.2 provide a visualization of missing values and its missing percentage with respect to all myocardial infarction related features present in the dataset.

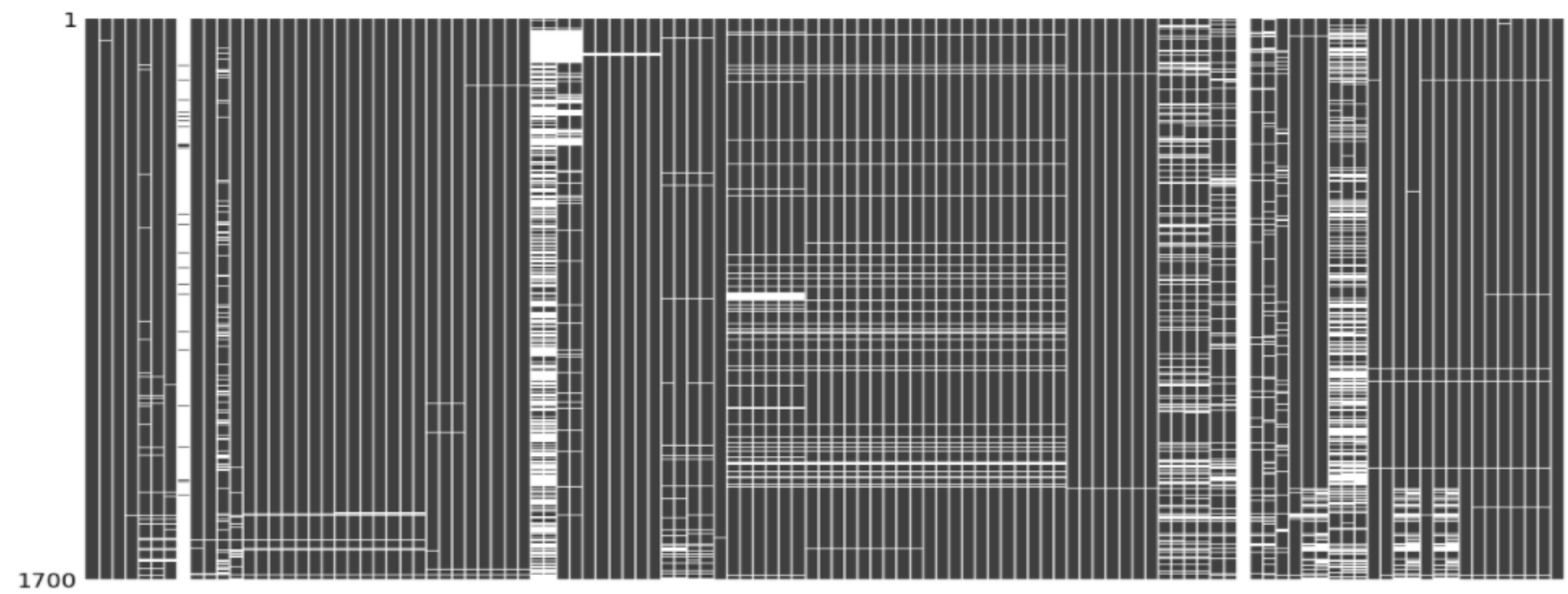

*Figure 4.1Matrix Plot to Visualize the Missing Values in the Presented Dataset*

The missing values in the data frame's columns is indicated by the white space in the above plot.

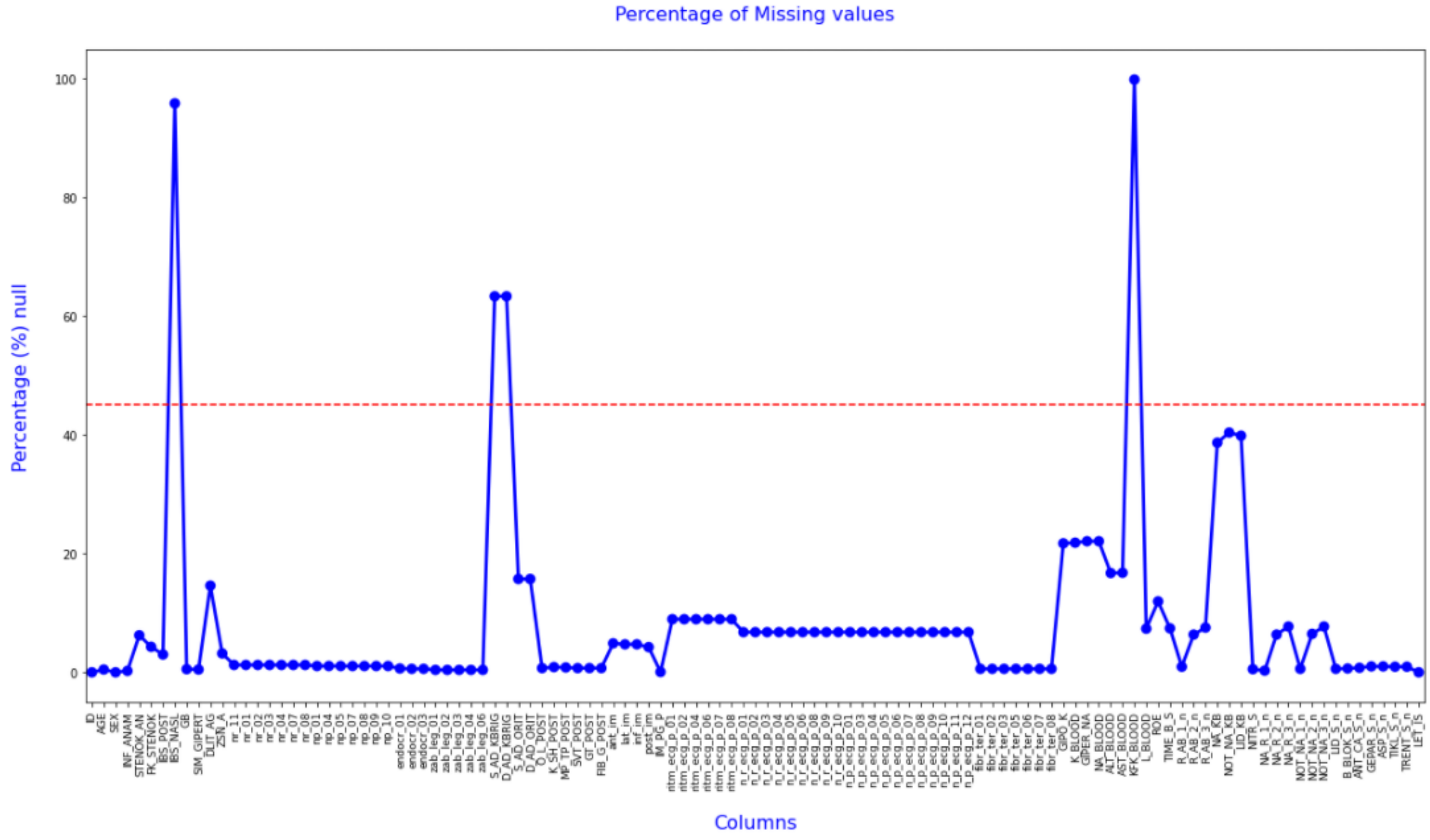

*Figure 4.2 Point Plot to Represent the Missing Values Percentage in the Presented Dataset*

Figure 4.2 shows the percentage of missing values along axis-y in relation to each feature across axis-x. The red line denotes a 45% threshold above which the missing percentage is considered unsatisfactory. The approaches for addressing missing values are described below.

### 4.3.4 Treatment of Missing Values

As part of this research, it has been discovered that roughly 110 related features had missing values from the above missing value analysis, and it is critical to treat the missing value to achieve a fair outcome and prevent bias or errors during further modelling. It has been identified that there are 9 numerical features that consist of missing values from table 4.5 above, and it has been handled using sklearn's iterative imputation module with Python version 3.8.5.
Iterative imputation is a sophisticated method of imputing missing values that entails simulating each feature that predicts each missing value in an attribute as a function of all other features and assessing the function values iteratively. Repeated attempts to forecast missing values using other features result in a more refined estimate, which improves the accuracy.

It has also been identified that there are 97 nominal categorical characteristics that consist of null data from table 4.5 above, and the same has been processed using the modes of the respective columns. Because all the variables are of the nominal category type, the missing values are imputed using modes (Simple imputation with 'most frequent' strategy), which are the values with the highest frequency in the respective columns. To impute missing values for nominal categorical attributes, the panda's package with Python 3.8.5 is employed.

From table 4.5 above, it has also been identified that there are 4 features namely, IBS_NASL, S_AD_KBRIG, D_AD_KBRIG, KFK_BLOOD that consist of more than a 45% null value. Dropping those features would be a better option, as imputing more than 45% of missing values can introduce a bias into the model and can reduce the efficiency of the classification result.

For further investigation, the dataset with imputed missing values was evaluated. There were no missing values in the dataset following this pre-processing phase, and all the observations were identified as legitimate. As a result of this pre-processing stage, the data now contains 167 variables including target variable with 1,700 instances that are linked to myocardial infarction and has been considered for further processing.

Below figure 4.3 provide a visualization of the entire dataset after treating missing value with respect to all myocardial infarction related features present in the dataset.

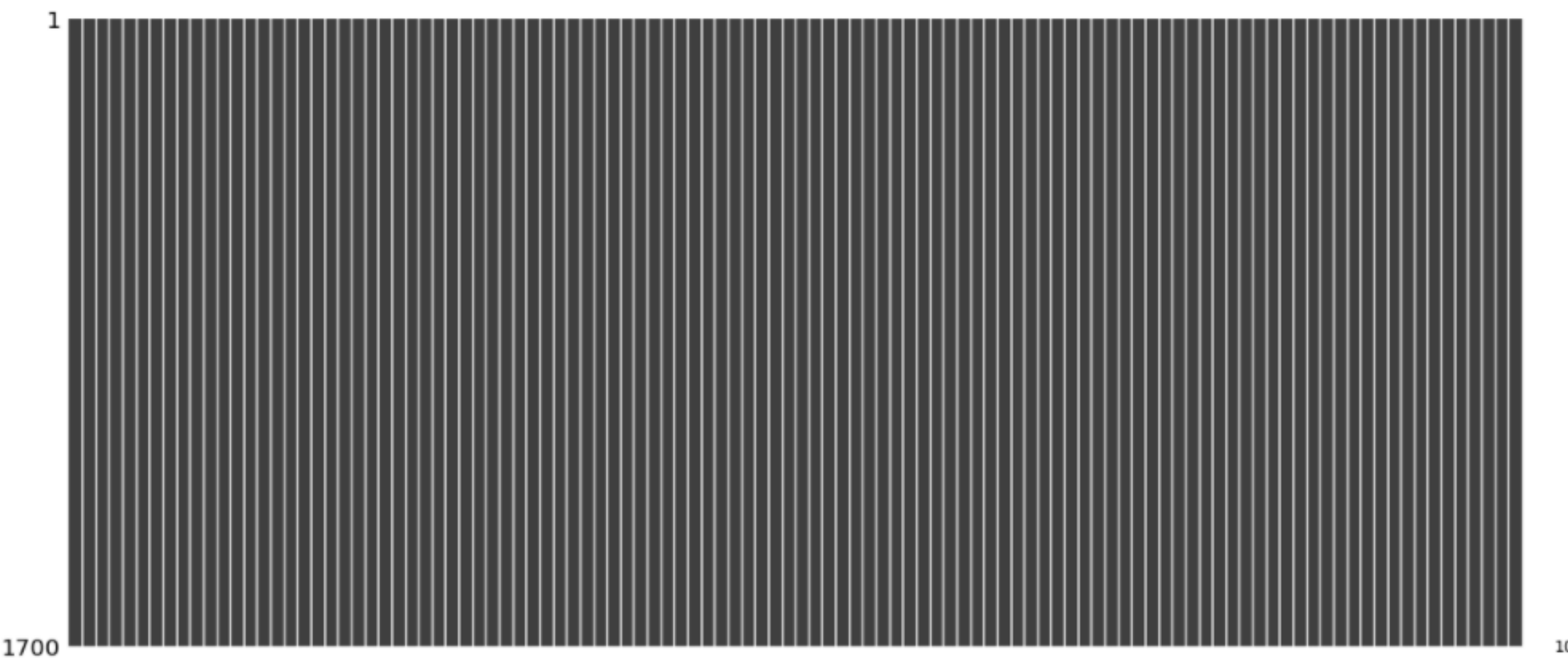

*Figure 4.3 Matrix Plot that indicates there is no Missing Values in the Dataset*

### 4.3.5 Identification of Imbalance Dataset

When the number of examples for one class is greater than the other in a binary or multiclass classification problem, or when the number of observations is not equal for all classes in a multiclass classification problem, there is a serious problem of imbalance. If a predictive machine learning model is built using this imbalance dataset, it is likely to be erroneous and deliver unsatisfactory or misleading results. The constructed model will favor the majority class, while the observations of the minority class will be ignored. As a result, the minority class is more likely to be misclassified than the majority. However, in the medical sector, or in a predictive model created for healthcare, the main focus will be on accurately predicting the minority class. Therefore, imbalance data needs to be treated before passing it of ML pipelines.

There is a high-class imbalance problem in the presented myocardial infarction dataset where the instances of ‘unknown’ class is much higher than rest of the other classes. Below table 4.6 listed down the target class label and number of observations under each class label.

*Table 4.8 Number of Observations in each Class from myocardial infarction dataset*

| Target Class Label | Label Description | Number of observations |
|---|---|---|
| **0** | Unknown | 1429 |
| **1** | Cardiogenic Shock | 110 |
| **2** | Pulmonary Edema | 18 |
| **3** | Myocardial Rupture | 54 |
| **4** | Progress of Congestive Heart Failure | 23 |
| **5** | Thromboembolism | 12 |
| **6** | Asystole | 27 |
| **7** | Ventricular Fibrillation | 27 |

In the presented myocardial infarction dataset, below figure 4.4 depicts a pie chart representation of the percentage of each class's observations, which clearly represented the scenario of class imbalance.

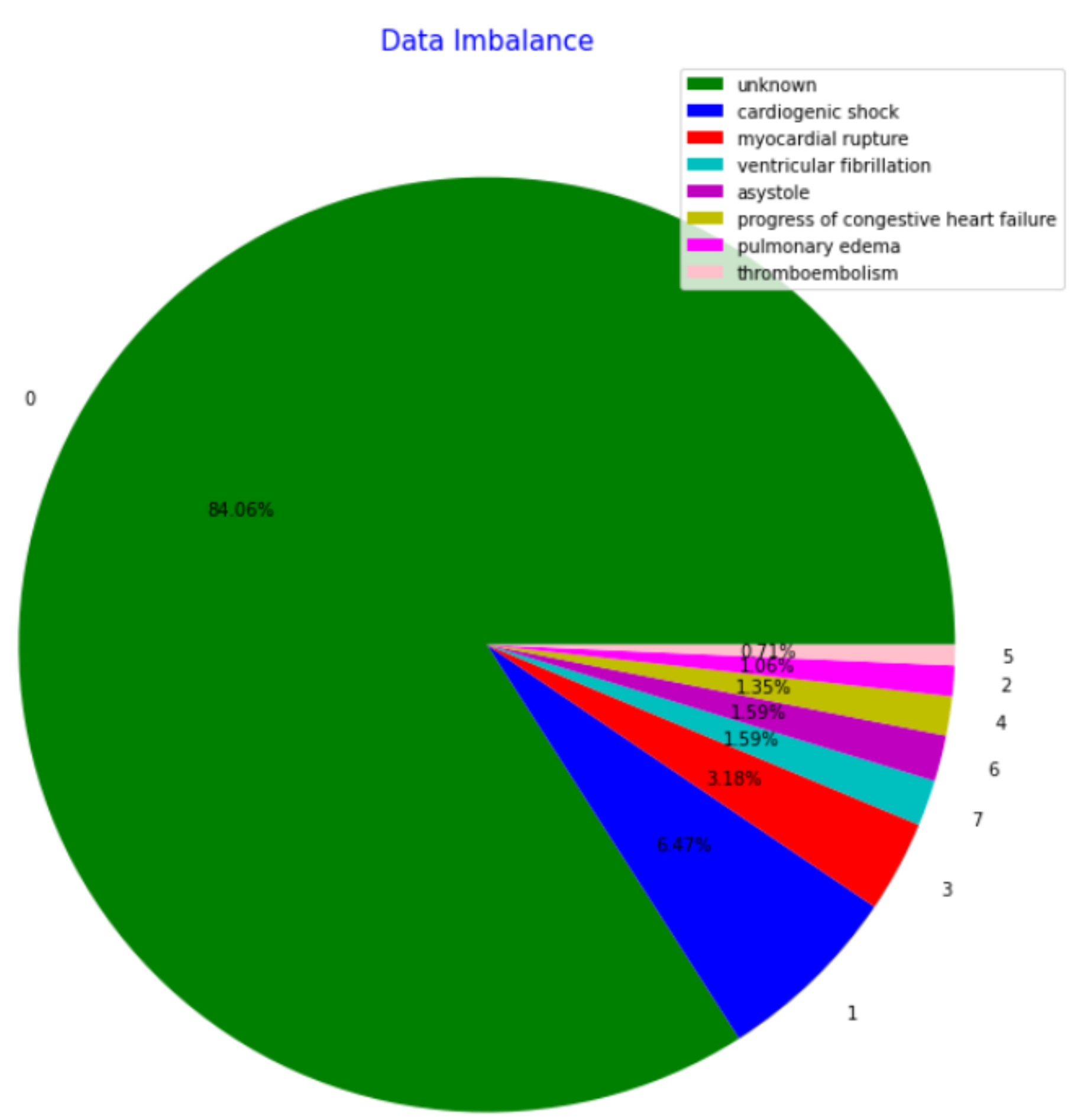

*Figure 4.4 Pie Chart to Visualize the Class Imbalance Problem in the Dataset*

### 4.3.6 Splitting of Original Dataset

The training and testing of the dataset are an important part of the supervised learning model development process. The parameters of the model do not settle on reasonable values without training the algorithm, thus the model is unusable. There is no way to know how good the model is without validating the procedure after training the model parameters on test or validation set, which is as again useless.

Aside from the original dataset with the imputed missing values, two other datasets were constructed from the entire dataset, one including only the training set and the other containing only the test set. The 70-30 split (70% for training set, 30% for test set) was adopted using sklearn's train-test split method with Python version 3.8.5 to allow other researchers to compare their work on the same dataset. In total, three datasets have been created, each of which has all 167 variables with their original categorical/numerical values and no missing values. There are 166 independent factors and one target variable which is the lethal outcome of myocardial infarction.

The three datasets that will be used for further analysis in this study are listed below, along with the total number of observations in each dataset. The training set is used to train the classifiers and develop a predictive model while the test set is used to examine and assess the model's performance.

1) Complete dataset (includes both training and test) - 1,700 records
2) Training dataset – 1,190 records (70%, for model training)
3) Testing dataset – 510 records (30%, for model validation and evaluation)

### 4.3.7 Treating the Imbalance Dataset

Coping with imbalanced datasets necessitates tactics like strengthening classification algorithms or balancing categories in the training data before feeding it to the machine learning model. The main goal of class balancing is to either increase the minority class's occurrence or decrease the dominant class's occurrence. This is done to ensure that each class have roughly the same number of instances. Imbalance treatment is done on train dataset and not in test dataset, hence below table 4.7 listed down the target class label and number of observations under each class label in training dataset before imbalance treatment.

*Table 4.9 Number of Observations in each Class from the train set before imbalance handling*

| Target Class Label | Label Description | Number of observations |
|---|---|---|
| **0** | Unknown | 1003 |
| **1** | Cardiogenic Shock | 76 |
| **2** | Pulmonary Edema | 15 |
| **3** | Myocardial Rupture | 38 |
| **4** | Progress of Congestive Heart Failure | 14 |
| **5** | Thromboembolism | 8 |
| **6** | Asystole | 20 |
| **7** | Ventricular Fibrillation | 16 |

In the presented myocardial infarction dataset, below figure 4.5 depicts a pie chart representation of the percentage of each class's observations in train set before imbalance handling, which clearly represented the scenario of class imbalance.

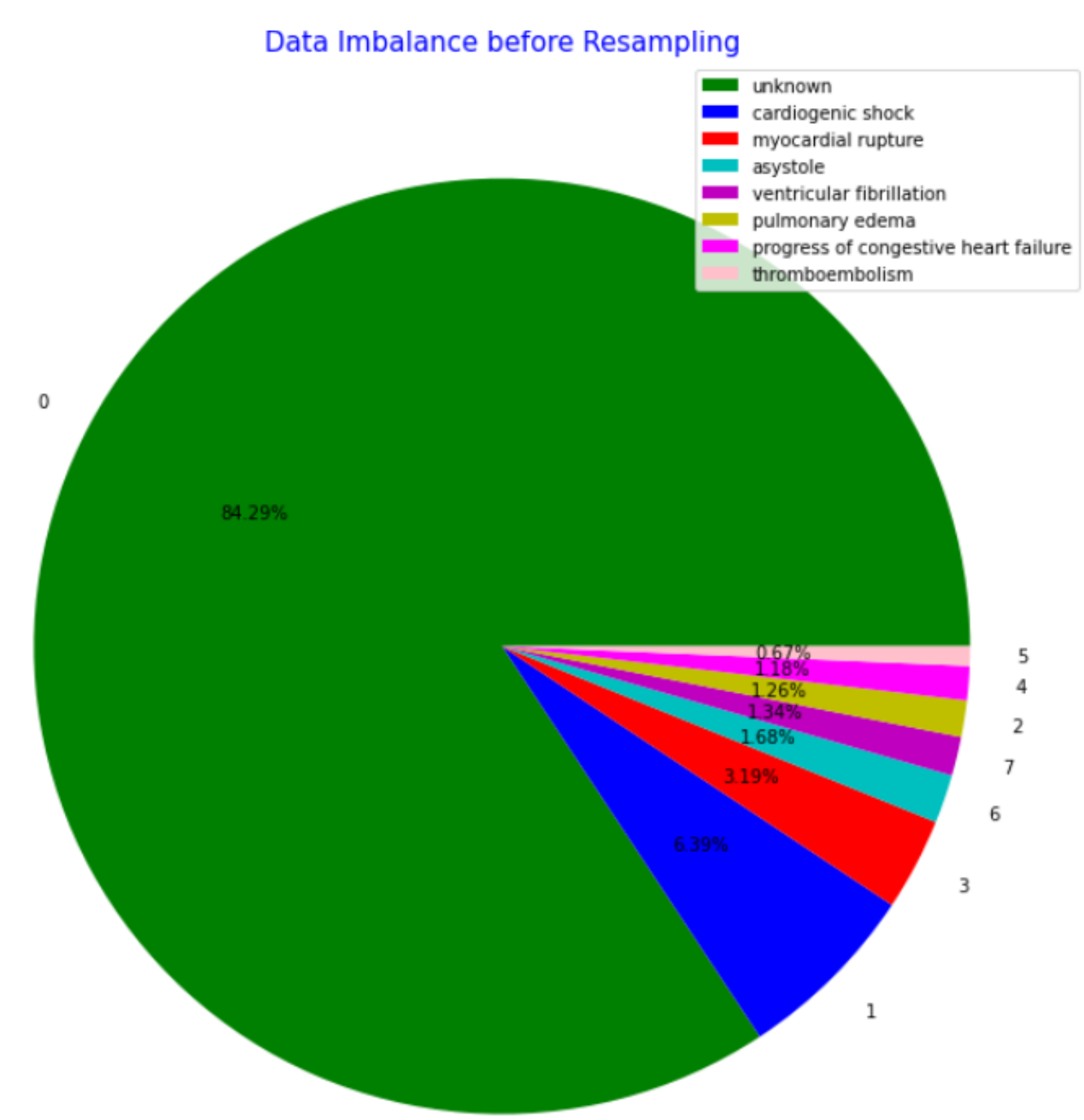

*Figure 4.5 Pie Chart to Visualize the Class Imbalance Problem in the Train Dataset*

As it has been already observed from table 4.7 that this presented myocardial infarction dataset has quite high-class imbalance, hence below sophisticated techniques have been adopted to deal with such scenario.

#### 4.3.7.1 SVM-SMOTE Method of Imbalance Treatment

Support vector points (trained using SVM classifiers) are used to determine the borderline region in SVM-SMOTE. Synthetic data is created by connecting the support vectors of each minority class with the support vectors of several of its closest neighbours. More data is synthesised away from the class overlap zone and more concentrated where the data points are separated in this approach of imbalance therapy. The class imbalance treatment was implemented using the SVMSMOTE method from the imblearn module with Python version 3.8.5.

Below table 4.8 listed down the target class label and number of observations under each class label after imbalance treatment on train set using SVMSMOTE.

*Table 4.10 Number of Observations in each Class from the train set after SVMSMOTE*

| Target Class Label | Label Description | Number of observations |
|---|---|---|
| **0** | Unknown | 1003 |
| **1** | Cardiogenic Shock | 1003 |
| **2** | Pulmonary Edema | 509 |
| **3** | Myocardial Rupture | 576 |
| **4** | Progress of Congestive Heart Failure | 429 |
| **5** | Thromboembolism | 687 |
| **6** | Asystole | 404 |
| **7** | Ventricular Fibrillation | 633 |

In the presented myocardial infarction dataset, below figure 4.6 depicts a pie chart representation of the percentage of each class's observations after imbalance handling using SVMSMOTE.

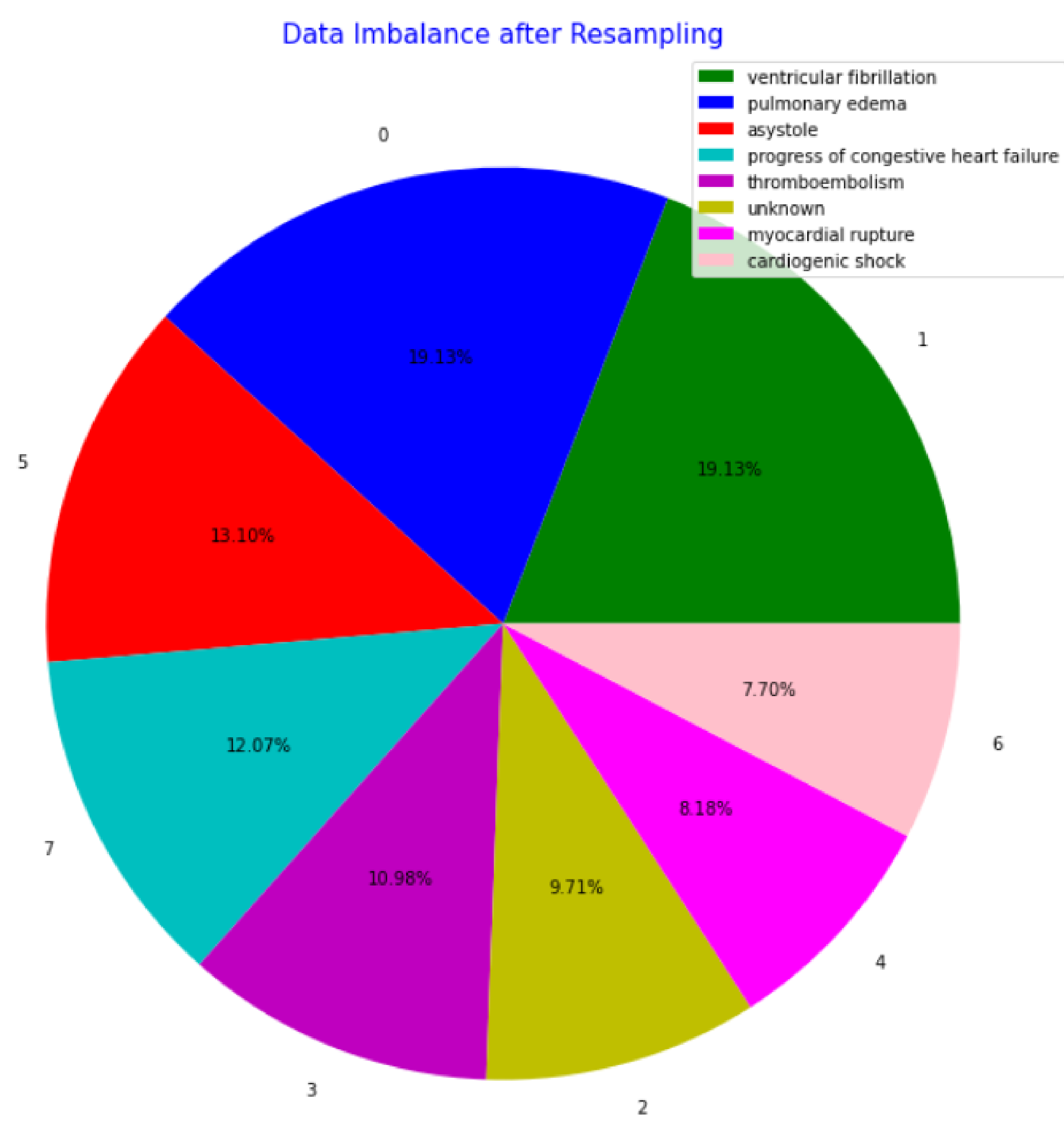

*Figure 4.6 Pie Chart to Visualize each Class observations after applying SVMSMOTE*

The problem of imbalance has been improved after using the SVMSMOTE imbalance handling approach, but it still exists because the percentage of observations in each class is still not equal, as shown in the above visualisation. As a result, to address this issue, the following method of imbalance treatment has been implemented.

#### 4.3.7.2 ADASYN Method of Imbalance Treatment

ADASYN (Adaptive Synthetic Sampling) is quite comparable to SMOTE and is evolved from it, with one major distinction. It will skew the sample space against points that are not in homogeneous neighbourhoods. On points that are not in homogeneous areas, ADASYN use the kind='normal' SMOTE algorithm. As a result, a mix of ordinary SMOTE and borderline SMOTE emerges. This approach inherited SMOTE's main flaw, namely its inability to build inner point outer point bridges. Whether a large focus on anomaly points is a desirable thing or not depends on the application. This class imbalance treatment was implemented using the ADASYN method from the imblearn module with Python version 3.8.5.

Below table 4.9 listed down the target class label and number of observations under each class label after imbalance treatment on train set using ADASYN.

*Table 4.9 Number of Observations in each Class from the train set after ADASYN*

| Target Class Label | Label Description | Number of observations |
|---|---|---|
| **0** | Unknown | 1003 |
| **1** | Cardiogenic Shock | 983 |
| **2** | Pulmonary Edema | 1006 |
| **3** | Myocardial Rupture | 996 |
| **4** | Progress of Congestive Heart Failure | 1005 |
| **5** | Thromboembolism | 1004 |
| **6** | Asystole | 1005 |
| **7** | Ventricular Fibrillation | 1008 |

In the presented myocardial infarction dataset, below figure 4.7 depicts a pie chart representation of the percentage of each class's observations after imbalance handling using ADASYN.

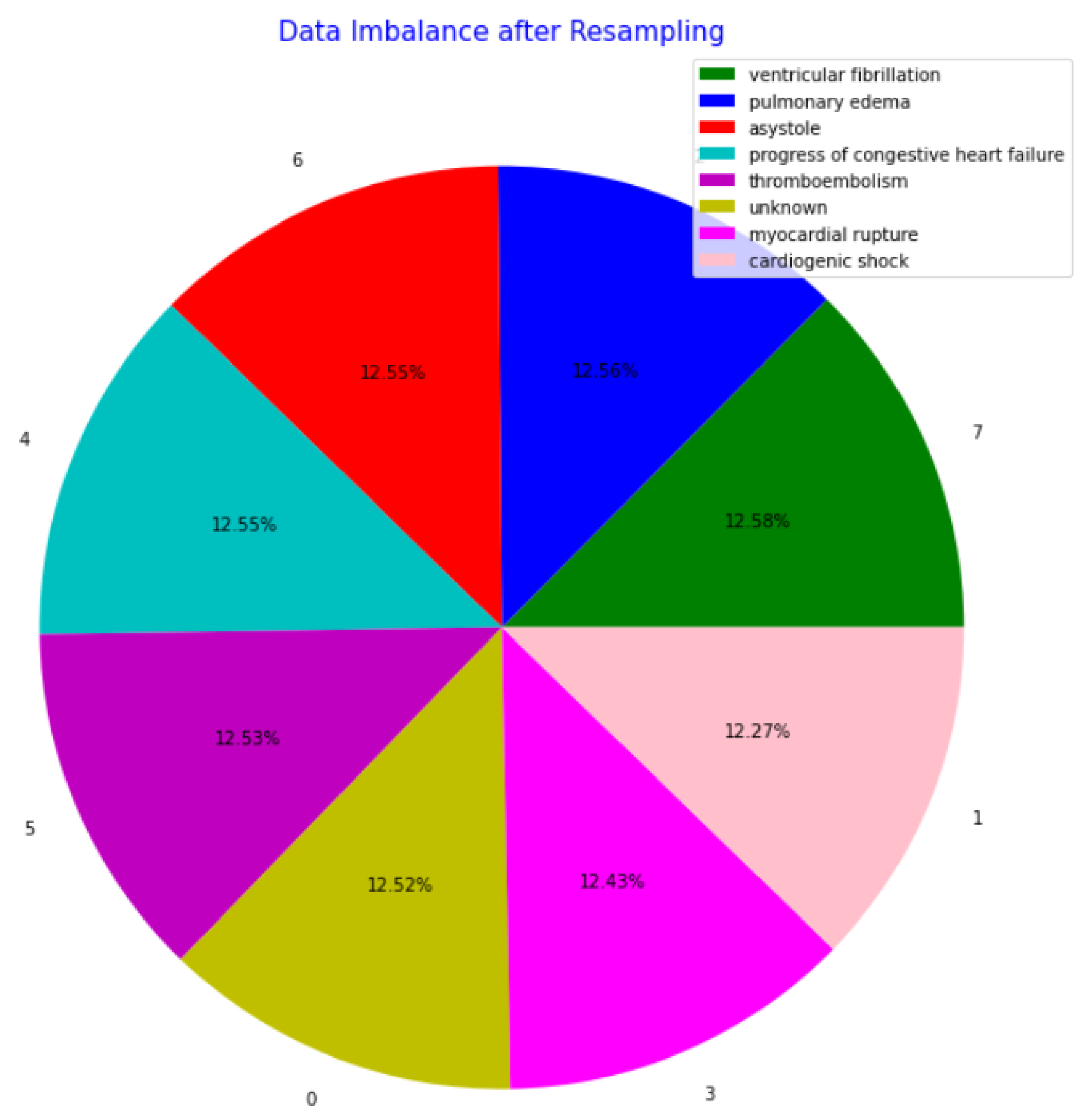

*Figure 4.7 Pie Chart to Visualize each Class observations after applying ADASYN*

The problem of imbalance has been resolved after using the ADASYN imbalance handling approach, and the percentage of observations in each class is now almost equal, as shown in the above visualisation.

The class weighted method of handling imbalance datasets is another method of handling imbalance datasets, but because this is a hyperparameter used in different machine learning models and will not generate any synthetic data, the implementation and the results of this imbalance treatment method will be discussed in the section 4.5 (subsection 4.5.5) and in section 5.3, alongside different classifiers.

Finally, after resampling the dataset using the class imbalance treatment, a new set of training data with 8010 rows and 166 features was created. The classification model will be trained individually using both the training datasets (before and after imbalance handling). To evaluate the classifier's performance, a single test dataset which was sampled earlier will be utilised (without any imbalance treatment).

1) Training dataset before imbalance handling – 1,190 records (for model training)
2) Training dataset after imbalance handling using ADASYN – 8,010 records (for model training)
3) Testing dataset – 510 records (for model validation and evaluation)

As shown in figure 4.6 above, using SVM-SMOTE of imbalance handling, the percentage of observations in each class did not equalizes, there still exists a skewness in the target class, hence rest of modelling procedure has been carried out using ADASYN and class weighted method of imbalance handling.

### 4.3.8 Standardization of Features

One of the most important pre-processing steps in the machine learning algorithms is feature scaling. This ensures that all independent features are scale free, scale independent, or scale equal, allowing the machine learning algorithm to consider each feature equally.
Scaling ensures that all features are updated with the same step size during gradient descent and converges more quickly towards minima for gradient descent-based algorithms like Linear regression, Logistic regression, and neural networks.
Because most of the features in this study are gaussian distributed in nature and some of the features contain true outliers, hence Standardization was chosen over Normalization in this investigation. For feature scaling, the StandardScaler method from the sklearn package with Python 3.8.5 is utilised.

### 4.3.9 Feature Selection Strategies

The number of characteristics or features utilised to train the machine learning model is directly proportional to its prediction power. If irrelevant features are employed to develop ML models, the model's performance will suffer. Feature selection is the process of lowering the number of unnecessary characteristics or dimensionality reduction, either automatically or manually, based on any statistical relationship. This could boost the ML models' performance or predictive potential. Overfitting is reduced, the model's accuracy is improved, and the training time of ML model is reduced. Three feature selection strategies are listed below, along with the findings.

#### 4.3.9.1 Feature Selection using RFE

Recursive feature elimination is an easy-to-configure strategy for selecting relevant features based on feature relevance or coefficient value for predicting the target variable. The above benefits of RFE make it a better strategy to apply in this project's feature section. Logistic regression machine learning algorithm has been used in this study as the heart of this method, which is then wrapped by RFE, which is utilised to identify significant features based on coefficient values. RFE method from sklearn's feature selection library with Python version 3.8.5 was used to implement recursive feature elimination technique.

Because the provided myocardial dataset is imbalanced, RFE was conducted once on the train dataset with 1,190 rows and 166 features before imbalance treatment. Finally, based on coefficient values, RFE selected 100 first-ranked features from a total of 166 independent characteristics.

After addressing the class imbalances in the presented dataset, RFE was run on the train dataset once again with 8010 rows and 166 features. Finally, based on coefficient values, RFE selected 100 first-ranked features from a total of 166 independent characteristics.

Below table 4.10 listed down the 100 first-ranked features selected via RFE before and after imbalance handling.

*Table 4.10 Top 100 Selected features via RFE before and after imbalance handling*

| RFE Selected Features before Imbalance Handling | RFE Selected Features after Imbalance Handling |
|---|---|
| AGE | AGE |
| S_AD_ORIT | S_AD_ORIT |
| D_AD_ORIT | D_AD_ORIT |
| K_BLOOD | K_BLOOD |
| ALT_BLOOD | NA_BLOOD |
| SEX | SEX |
| SIM_GIPERT | SIM_GIPERT |
| nr_11 | nr_04 |
| nr_03 | endocr_01 |

| nr_04 | endocr_02 |
|---|---|
| np_01 | zab_leg_01 |
| np_05 | zab_leg_02 |
| endocr_01 | zab_leg_03 |
| endocr_02 | O_L_POST |
| zab_leg_01 | K_SH_POST |
| zab_leg_02 | MP_TP_POST |
| zab_leg_03 | SVT_POST |
| O_L_POST | IM_PG_P |
| K_SH_POST | ritm_ecg_p_01 |
| MP_TP_POST | ritm_ecg_p_02 |
| SVT_POST | ritm_ecg_p_04 |
| FIB_G_POST | ritm_ecg_p_07 |
| IM_PG_P | ritm_ecg_p_08 |
| ritm_ecg_p_02 | n_r_ecg_p_01 |
| ritm_ecg_p_07 | n_r_ecg_p_03 |
| ritm_ecg_p_08 | n_r_ecg_p_05 |
| n_r_ecg_p_01 | n_p_ecg_p_06 |
| n_r_ecg_p_03 | n_p_ecg_p_07 |
| n_r_ecg_p_04 | n_p_ecg_p_10 |
| n_r_ecg_p_05 | n_p_ecg_p_12 |
| n_p_ecg_p_06 | GIPO_K |
| n_p_ecg_p_08 | NA_KB |
| n_p_ecg_p_12 | NOT_NA_KB |
| GIPO_K | LID_KB |
| NA_KB | NITR_S |
| NOT_NA_KB | LID_S_n |
| LID_KB | B_BLOK_S_n |
| NITR_S | ANT_CA_S_n |
| LID_S_n | GEPAR_S_n |
| B_BLOK_S_n | ASP_S_n |
| ANT_CA_S_n | TRENT_S_n |
| GEPAR_S_n | INF_ANAM_1 |
| ASP_S_n | INF_ANAM_2 |
| TRENT_S_n | INF_ANAM_3 |

| INF_ANAM_1 | STENOK_AN_1 |
|---|---|
| INF_ANAM_2 | STENOK_AN_2 |
| INF_ANAM_3 | STENOK_AN_3 |
| STENOK_AN_1 | STENOK_AN_4 |
| STENOK_AN_2 | STENOK_AN_5 |
| STENOK_AN_3 | STENOK_AN_6 |
| STENOK_AN_4 | FK_STENOK_1 |
| STENOK_AN_5 | FK_STENOK_2 |
| STENOK_AN_6 | FK_STENOK_3 |
| FK_STENOK_1 | IBS_POST_1 |
| FK_STENOK_2 | IBS_POST_2 |
| FK_STENOK_3 | GB_2 |
| IBS_POST_1 | GB_3 |
| IBS_POST_2 | DLIT_AG_1 |
| GB_2 | DLIT_AG_3 |
| GB_3 | DLIT_AG_5 |
| DLIT_AG_2 | DLIT_AG_6 |
| DLIT_AG_3 | DLIT_AG_7 |
| DLIT_AG_5 | ZSN_A_1 |
| DLIT_AG_7 | ZSN_A_2 |
| ZSN_A_1 | ZSN_A_3 |
| ZSN_A_2 | ZSN_A_4 |
| ZSN_A_3 | ant_im_1 |
| ZSN_A_4 | ant_im_4 |
| ant_im_1 | lat_im_1 |
| ant_im_4 | lat_im_2 |
| lat_im_1 | lat_im_3 |
| lat_im_2 | inf_im_1 |
| lat_im_3 | inf_im_2 |
| lat_im_4 | inf_im_3 |
| inf_im_1 | inf_im_4 |
| inf_im_2 | post_im_2 |
| inf_im_3 | post_im_3 |
| inf_im_4 | TIME_B_S_2 |
| TIME_B_S_3 | TIME_B_S_3 |

| TIME_B_S_4 | TIME_B_S_4 |
|---|---|
| TIME_B_S_5 | TIME_B_S_5 |
| TIME_B_S_6 | TIME_B_S_6 |
| TIME_B_S_7 | TIME_B_S_7 |
| TIME_B_S_8 | TIME_B_S_8 |
| TIME_B_S_9 | TIME_B_S_9 |
| R_AB_1_n_1 | R_AB_1_n_1 |
| R_AB_1_n_3 | R_AB_1_n_2 |
| R_AB_2_n_1 | R_AB_1_n_3 |
| R_AB_3_n_1 | R_AB_2_n_1 |
| R_AB_3_n_2 | R_AB_3_n_1 |
| NA_R_1_n_1 | R_AB_3_n_2 |
| NA_R_1_n_2 | NA_R_1_n_1 |
| NA_R_1_n_3 | NA_R_1_n_2 |
| NA_R_1_n_4 | NA_R_1_n_3 |
| NA_R_2_n_1 | NA_R_2_n_1 |
| NA_R_3_n_1 | NA_R_2_n_2 |
| NOT_NA_1_n_1 | NOT_NA_1_n_1 |
| NOT_NA_1_n_2 | NOT_NA_1_n_2 |
| NOT_NA_3_n_1 | NOT_NA_2_n_1 |
| NOT_NA_3_n_2 | NOT_NA_3_n_1 |

#### 4.3.9.2 Feature Selection using Extra Tree Classifier

Extremely Randomized Trees Classifier is an ensemble machine learning technique that uses the resulting of many uncorrelated decision trees collected in a forest to produce classification results (Extra Trees Classifier). The normalised overall reduction in Gini Index is calculated for each characteristic during the creation or construction of the extra tree forest. The Gini Importance of a feature is the name given to this value. Then, using the Gini Index, each characteristic or feature is ranked in descending order, and the most significant features are chosen based on the problem statement.

Extra Tree Classifier for feature selection was implemented using the ExtraTreesClassifier method from sklearn's ensemble library with Python version 3.8.5 and number of trees (n estimators) of 1,000 used as a hyperparameter. The RFE-selected top 100 features are now sent through an additional tree classifier to refine the top 70 features, which are then used for model development.

Because the provided myocardial dataset is imbalanced, feature selection using extra tree was conducted once on the train dataset with 1,190 rows and 100 features (selected via RFE) before imbalance treatment. Finally, based on gini importance, extra tree classifier selected top 70 features from a total of 100 independent characteristics.

After addressing the class imbalances in the presented dataset, feature selection using extra tree was run on the train dataset once again with 8010 rows and 100 features (selected via RFE). Finally, based on gini importance, extra tree classifier selected top 70 features from a total of 100 independent characteristics.

Below figure 4.8 visualize the top 70 features selected via extra tree classifier based on gini importance before class imbalance handling.

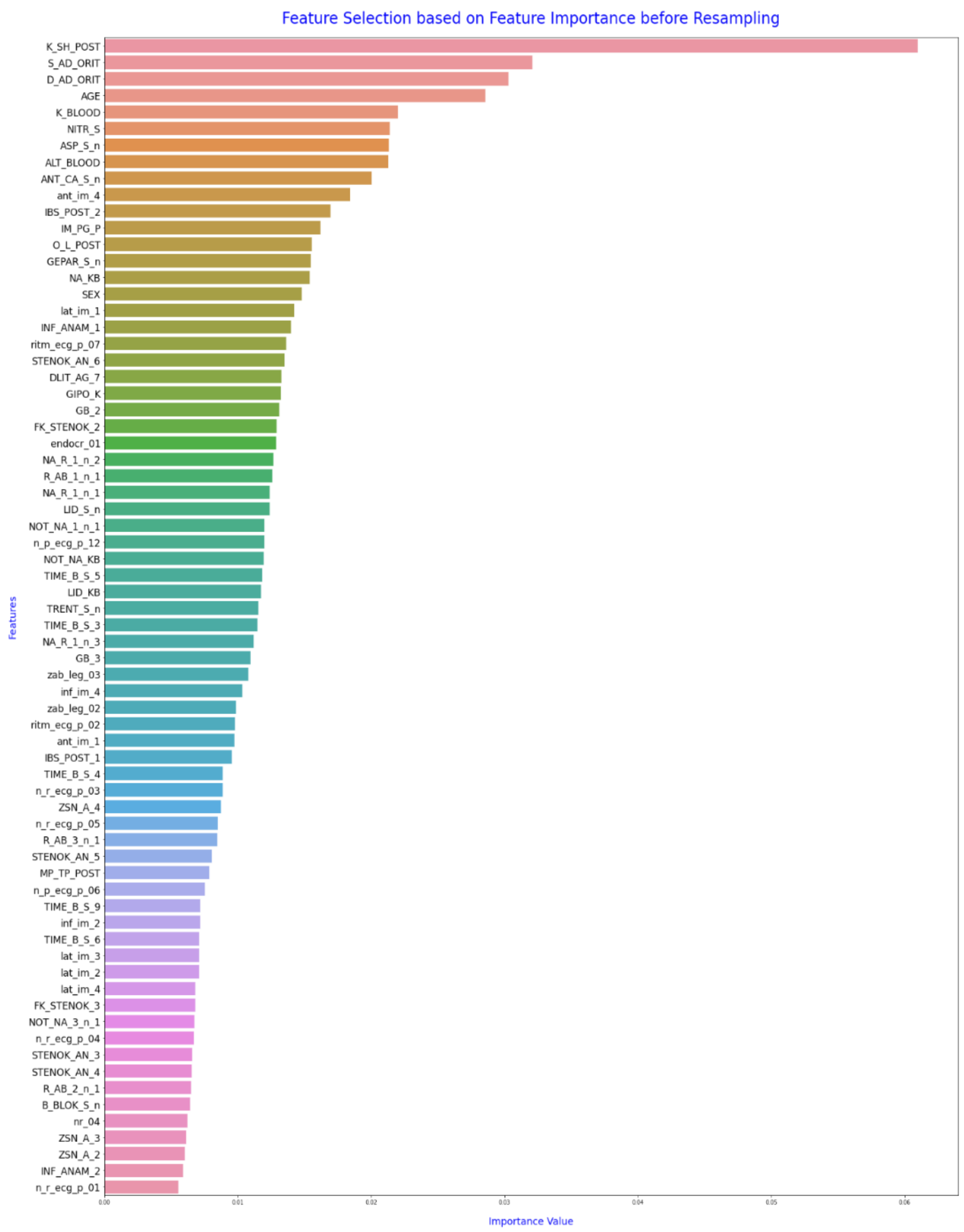

*Figure 4.8 Top 70 Features based on Feature Importance before class imbalance handling*

Below figure 4.9 visualize the top 70 features selected via extra tree classifier based on gini importance after class imbalance handling.

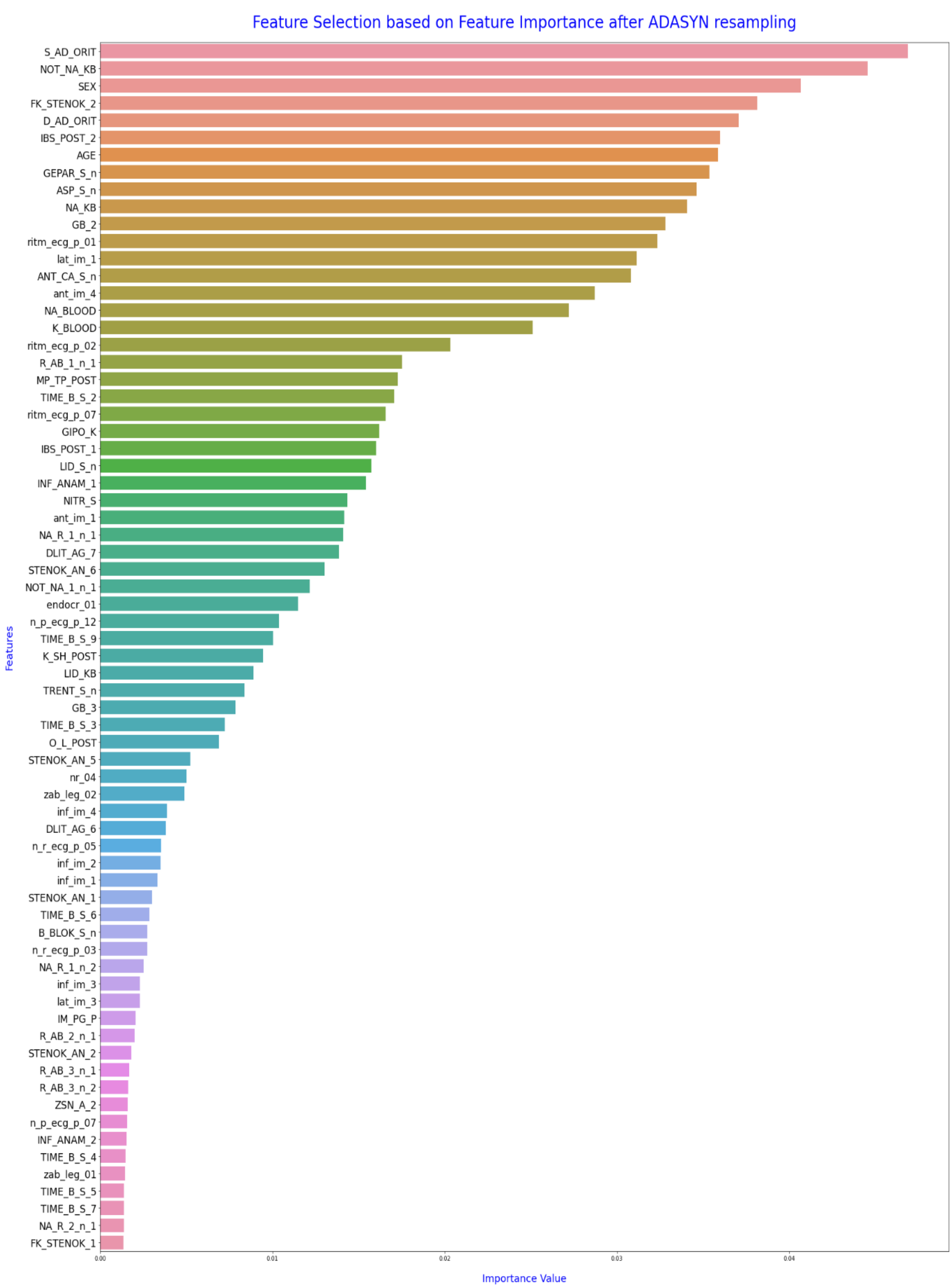

*Figure 4.9 Top 70 Features based on feature importance after class imbalance handling*

#### 4.3.9.3 Feature Selection using Elastic Net

The Elastic Net regularisation approach combines the L1 (LASSO) and L2 (RIDGE) regularisation techniques. Elastic net regularisation includes both the LASSO and RIDGE penalty terms. The elastic net method has addressed the incapability of both L1 and L2 strategies. In elastic net regularisation, the RIDGE regularisation coefficient must first be estimated to minimise the coefficient values, and then LASSO regularisation must be applied to eliminate the irrelevant coefficients. As a result, these variables with a coefficient value of zero can be omitted because they will not contribute to the model's prediction capacity, making this strategy perfect for feature selection mechanisms.

The Logistic regression method from sklearn's linear model library was used to build Elastic Net regularisation for feature selection, with Python version 3.8.5 and a penalty term of 'elastic net' and a solver 'saga' used as a hyperparameter. The top 70 features that were selected through RFE and extra tree classifier were passed into elastic net regularization via logistic regression.

Because the myocardial dataset was imbalanced, feature selection using elastic net regularization was performed once on the train dataset with 1,190 rows and 70 features (chosen by RFE and extra tree) before imbalance treatment. Finally, elastic net regularization shows that there is no common feature among all the 8 classes in the dataset whose coefficients are equal to zero, indicating that all the features are significant in this multiclass classification problem. As a result, for model development, the top 70 features that were previously selected by RFE and an extra tree classifier will be used.

After resolving the class imbalances in the presented dataset, the train dataset with 8010 rows and 70 features (selected via RFE and extra tree) was subjected to feature selection using elastic net regularization. Finally, elastic net regularization shows that there are no common features among all the 8 classes in the dataset whose coefficients are equal to zero, indicating that all the features are significant in this multiclass classification problem. As a result, for model development, the top 70 features that were previously selected by RFE and an extra tree classifier will be used.

Below Venn diagram figure 4.10 visualizes the intersection of all features among all 8 classes with coefficient values calculated as zero before class imbalance handling.

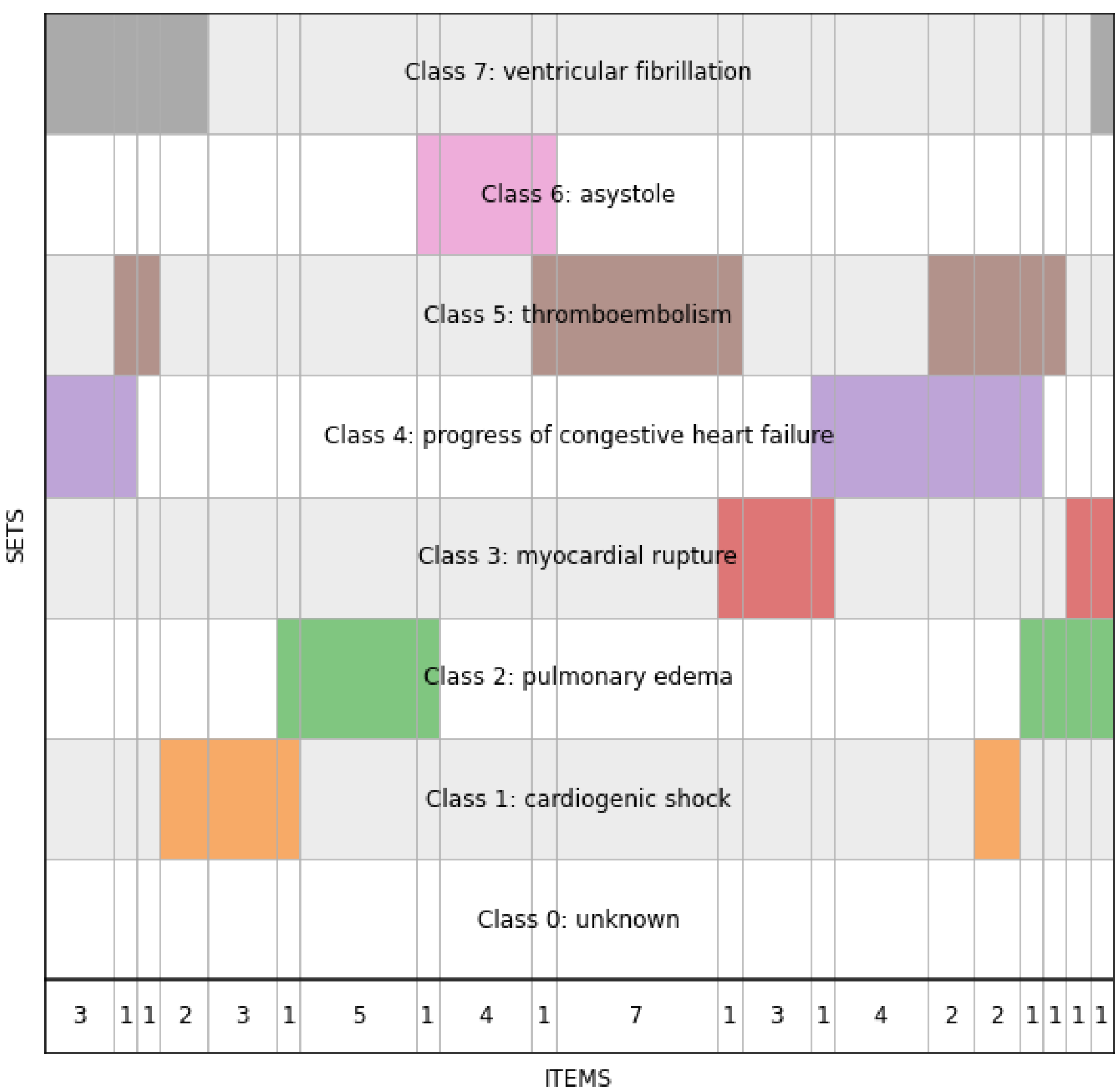

*Figure 4.10 Feature intersection whose coefficients calculated as zero, selected via elastic net regularization among all 8 classes before class imbalance handling*

As shown above, there are many features whose coefficients are calculate as zero individually but there are no features selected whose coefficients are common in all the 8 classes and are calculated as zero, before class imbalance handling. Hence, all the coefficients for selected 70 features are significant in this multiclass classification problem and should be considered for model development.

Below Venn diagram figure 4.11 visualizes the intersection of all features among all 8 classes with coefficient values calculated as zero after class imbalance handling.

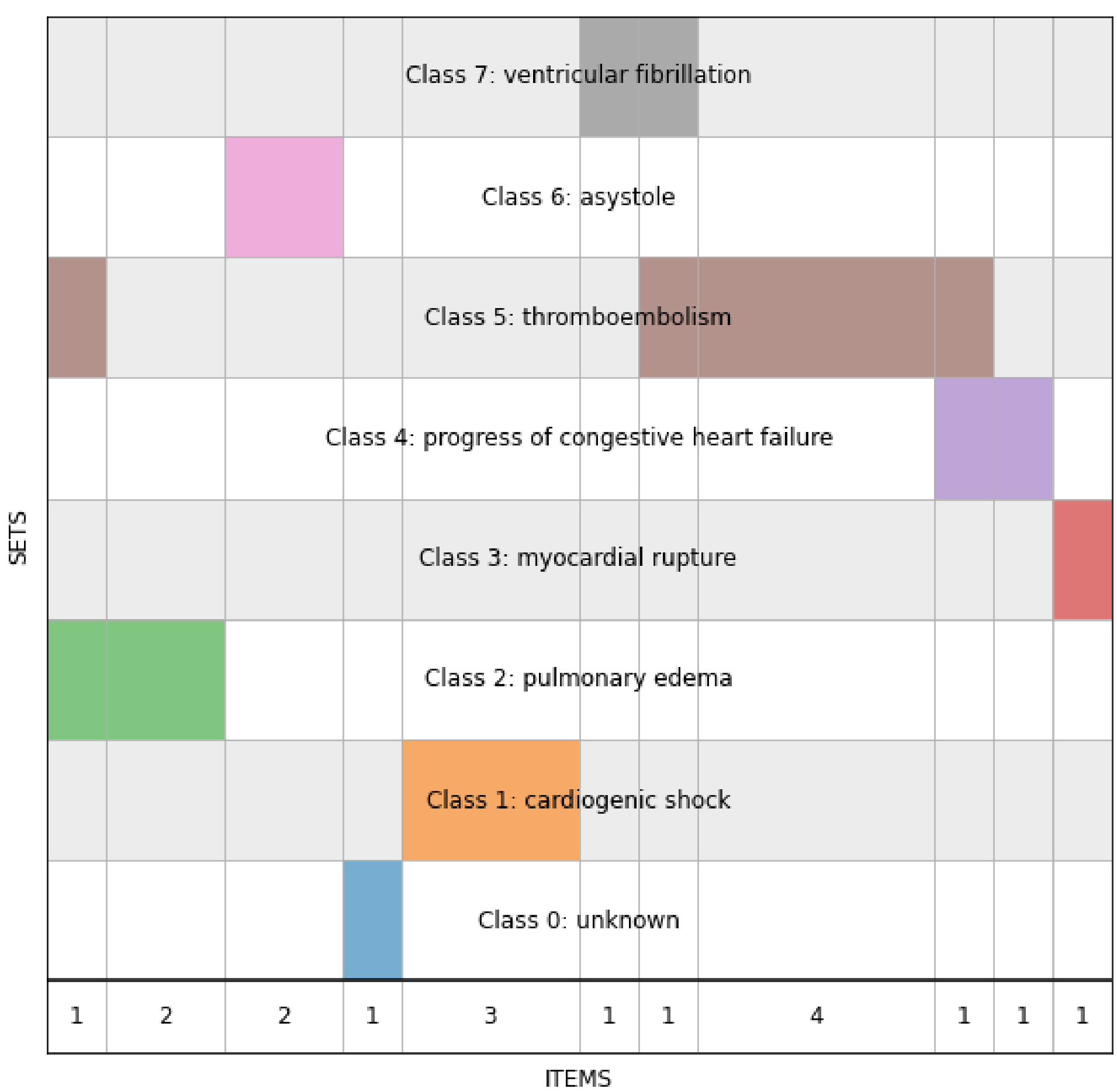

*Figure 4.11 Feature intersection whose coefficients calculated as zero, selected via elastic net regularization among all 8 classes after class imbalance handling*

As shown above, there are many features whose coefficients are calculate as zero individually but there are no features selected whose coefficients are common in all the 8 classes and are calculated as zero, even after class imbalance handling. Hence, all the coefficients for selected 70 features are significant in this multiclass classification problem and should be considered for model development.

## 4.4 Exploratory Data Analysis

Exploratory data analysis (EDA) is the method of using statistical and visualization techniques to analyse data to gain a better grasp of the dataset, recognize various data patterns, and gain a better understanding of the issue statement. Analytical insight in EDA entails obtaining various statistical data such as Mean, Standard Deviation, Median, Max Value, and Min Value, as well as data visualization, which is turning raw data into maps or graphical forms to extract relevant insight. EDA will also aid in the identification of the distribution and relationships among the many features or biomarkers in this dataset. Below subsections will discuss about all the exploratory data analysis that has been carried out in this research.

### 4.4.1 Univariate Analysis

Because 'uni' means one and 'variate' means variable, there is only one trustworthy factor in univariate analysis. The goal of univariate analysis to explore only one variable at one time and to obtain data, characterize and summarize it, and examine any patterns that may exist. It investigates each variable separately in a sample. Central Tendency (average, mode, and median), distribution, frequency/count, frequency percentage, standard deviation, and outliers are some of the patterns that may be easily discovered using univariate analysis. In this study, the univariate analysis has been implemented using two important visualization libraries namely matplotlib and seaborn along with pandas with Python versions 3.8.5.

#### 4.4.1.1 Visualization and Summary Statistics

This section will go over various univariate visualisations, such as the distribution plot, count/frequency plot, percentage plot, and descriptive statistics, for some important independent features that consists of both numerical and categorical variables. Among all the predictive indicators, this section addresses a couple of the most important biomarkers. This analysis focuses at both numerical and categorical characteristics. Statistics such as Central Tendency, Quartiles and distribution plots are reported for numerical variables. On the other side, categorical attributes were examined in frequency, mode value, count plot, and so on. The next sections go over the specifics of each variable.

**a) AGE (Patient's age)**

As it can be observed from below table 4.11, a total of 1,700 observations have been reported for patients with myocardial infarction. The reported average age of patients admitted to the hospital with this cardiovascular illness is 61.85 years old. A patient with this ailment might be as young as 26 years old and as older as 92 years, according to analysis. Also, approximately 25% of the reported patients are under the age of 54, 50% of the reported patients are under the age of 63, and 75% of the reported patients are under the age of 70. This reported analysis is as per the number of observations present in the dataset.

*Table 4.13 Summary Statistics of AGE feature*

| AGE | | | | | | | |
|---|---|---|---|---|---|---|---|
| **Count** | **Mean Value** | **Standard Deviation** | **Minimum Value** | **25%** | **50%** | **75%** | **Maximum Value** |
| 1700 | 61.851765 | 11.234057 | 26.00 | 54.00 | 63.00 | 70.00 | 92.00 |

From the below figure 4.12 shows that the AGE variable almost follows a normal distribution. Hence the reported sample distribution of patient's age admitted to the hospital is a good representation of the population's age those who are suffering from myocardial infarction. Also, it can be observed that most of the patients those who are suffering from this disease ranges between 55 to 75 years of age.

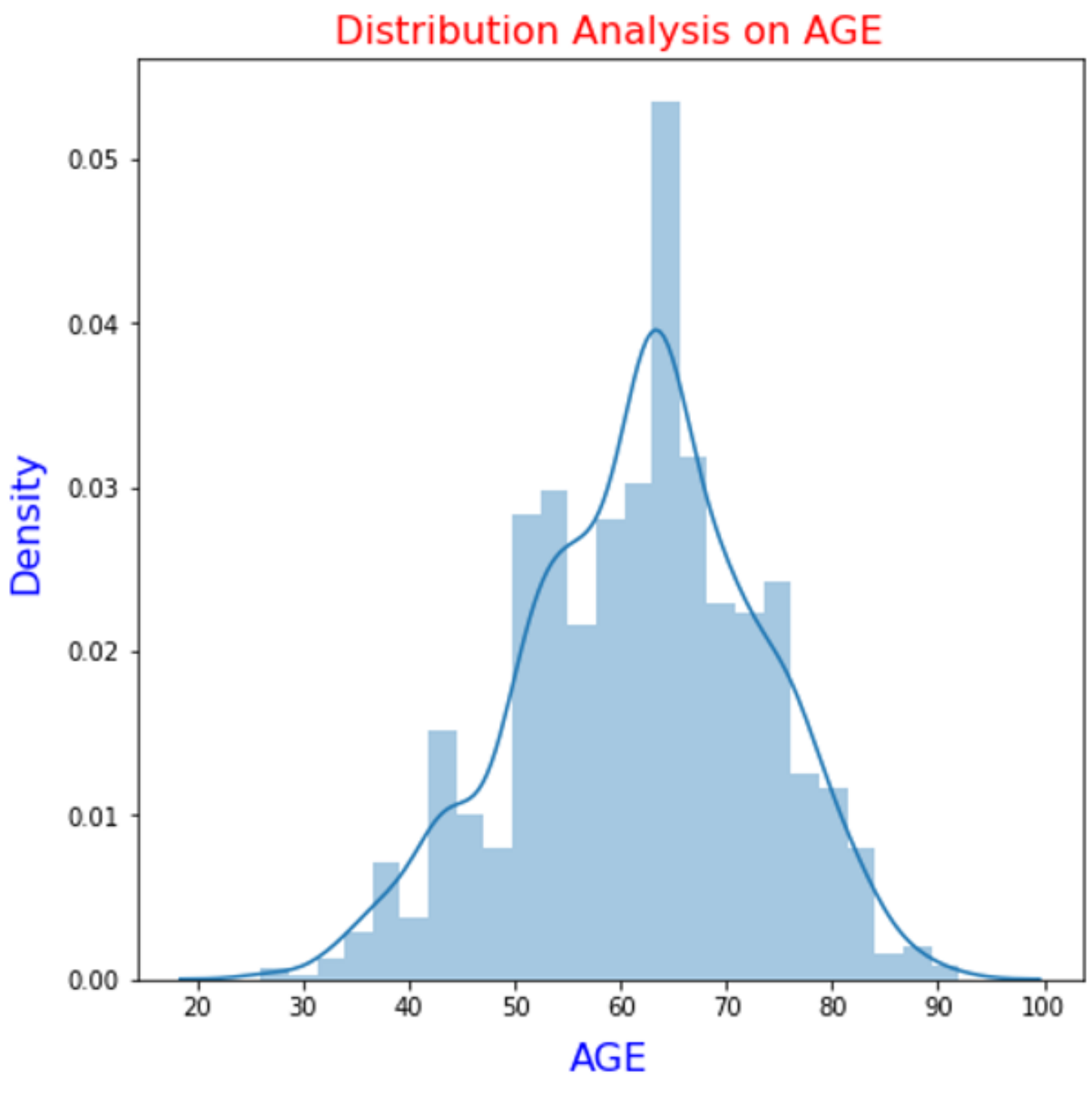

*Figure 4.12 Distribution plot of AGE variable*

**b) S_AD_ORIT (Patient's systolic blood pressure according to ICU)**

As it can be observed from below table 4.12, a total of 1700 observations have been reported for patients with myocardial infarction. The reported average systolic blood pressure of patients admitted to the hospital with this cardiovascular illness is 134.59 mmHg. A patient with this ailment might have 0.00 mmHg minimum systolic blood pressure and maximum of 260.00 mmHg, according to this analysis. Also, approximately 25% of the reported patients having less than 120.00 mmHg systolic pressure, 50% of the reported patients having less than 133.00 mmHg of systolic pressure and 75% of the reported patients having less than 150.00 mmHg of systolic pressure. This reported analysis is as per the number of observations present in the dataset and for the patients admitted to ICU.

*Table 4.12 Summary Statistics of S_AD_ORIT feature*

| S_AD_ORIT | | | | | | | |
|---|---|---|---|---|---|---|---|
| **Count** | **Mean Value** | **Standard Deviation** | **Minimum Value** | **25%** | **50%** | **75%** | **Maximum Value** |
| 1700 | 134.588235 | 28.855282 | 0.00 | 120.00 | 133.00 | 150.00 | 260.00 |

From the below figure 4.13 shows that the S_AD_ORIT variable almost follows a normal distribution. Hence the reported sample distribution of patient's systolic blood pressure admitted to the hospital is a good representation of the population's systolic blood pressure those who are suffering from myocardial infarction. Also, it can be observed that most of the patients those who are suffering from this disease having systolic blood pressure between 120 to 150 mmHg.

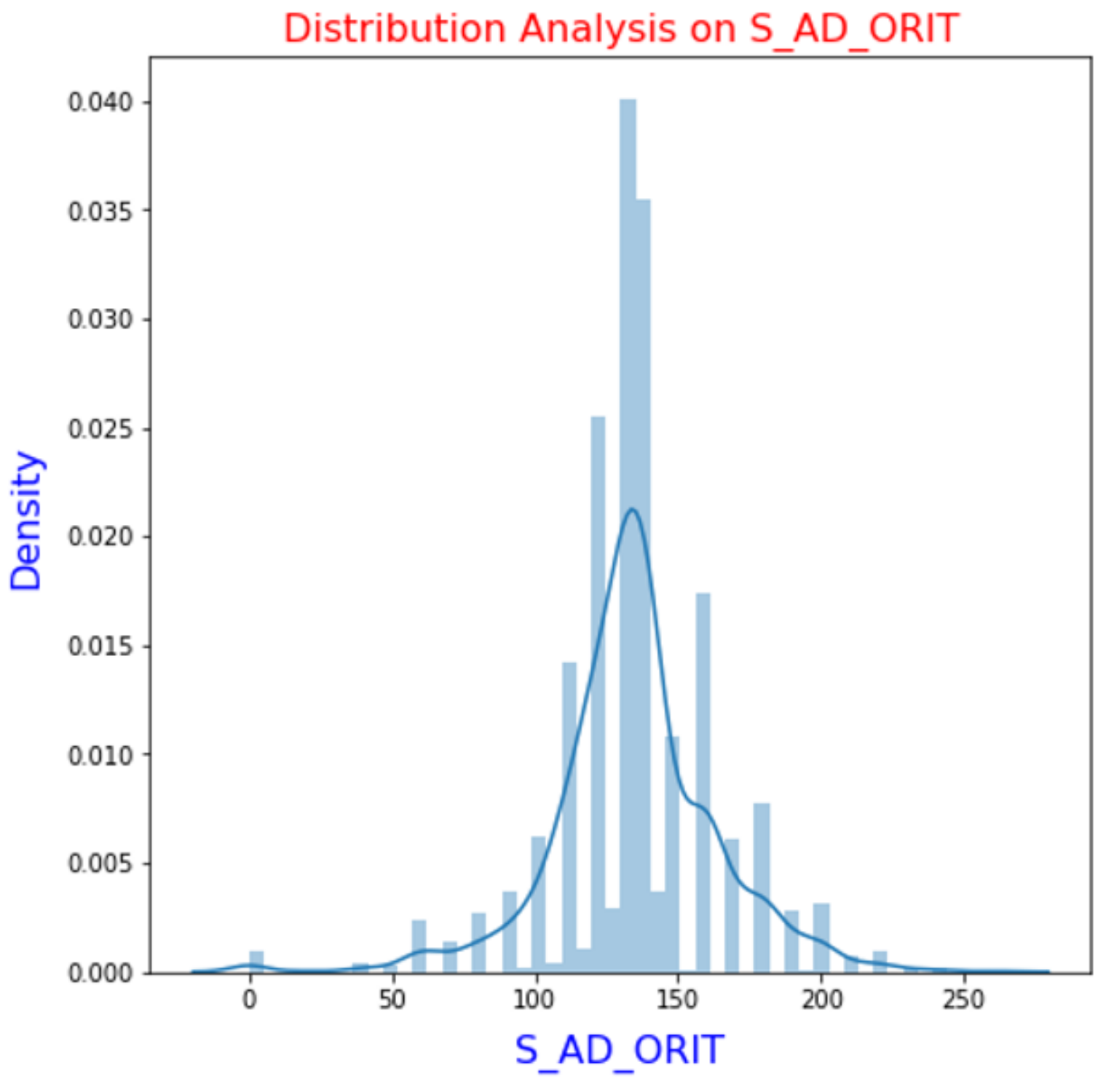

*Figure 4.13 Distribution plot of S_AD_ORIT variable*

**c) D_AD_ORIT (Patient's diastolic blood pressure according to ICU)**

As it can be observed from below table 4.13, a total of 1,700 observations have been reported for patients with myocardial infarction. The reported average diastolic blood pressure of patients admitted to the hospital with this cardiovascular illness is 82.76 mmHg. A patient with this ailment might have 0.00 mmHg minimum diastolic blood pressure and maximum of 190.00 mmHg, according to this analysis. Also, approximately 25% of the reported patients having less than 80.00 mmHg diastolic pressure, 50% of the reported patients are also having less than 80.00 mmHg of diastolic pressure and 75% of the reported patients having less than 90.00 mmHg of diastolic pressure. This reported analysis is as per the number of observations present in the dataset and for the patients admitted to ICU.

*Table 4.15 Summary Statistics of D_AD_ORIT feature*

| D_AD_ORIT | | | | | | | |
|---|---|---|---|---|---|---|---|
| Count | Mean Value | Standard Deviation | Minimum Value | 25% | 50% | 75% | Maximum Value |
| 1700 | 82.766471 | 16.869410 | 0.00 | 80.00 | 80.00 | 90.00 | 190.00 |

From the below figure 4.14 shows that the D_AD_ORIT variable almost follows a normal distribution. Hence the reported sample distribution of patient's diastolic blood pressure admitted to the hospital is a good representation of the population's diastolic blood pressure those who are suffering from myocardial infarction. Also, it can be observed that most of the patients those who are suffering from this disease having diastolic blood pressure between 70 to 90 mmHg.

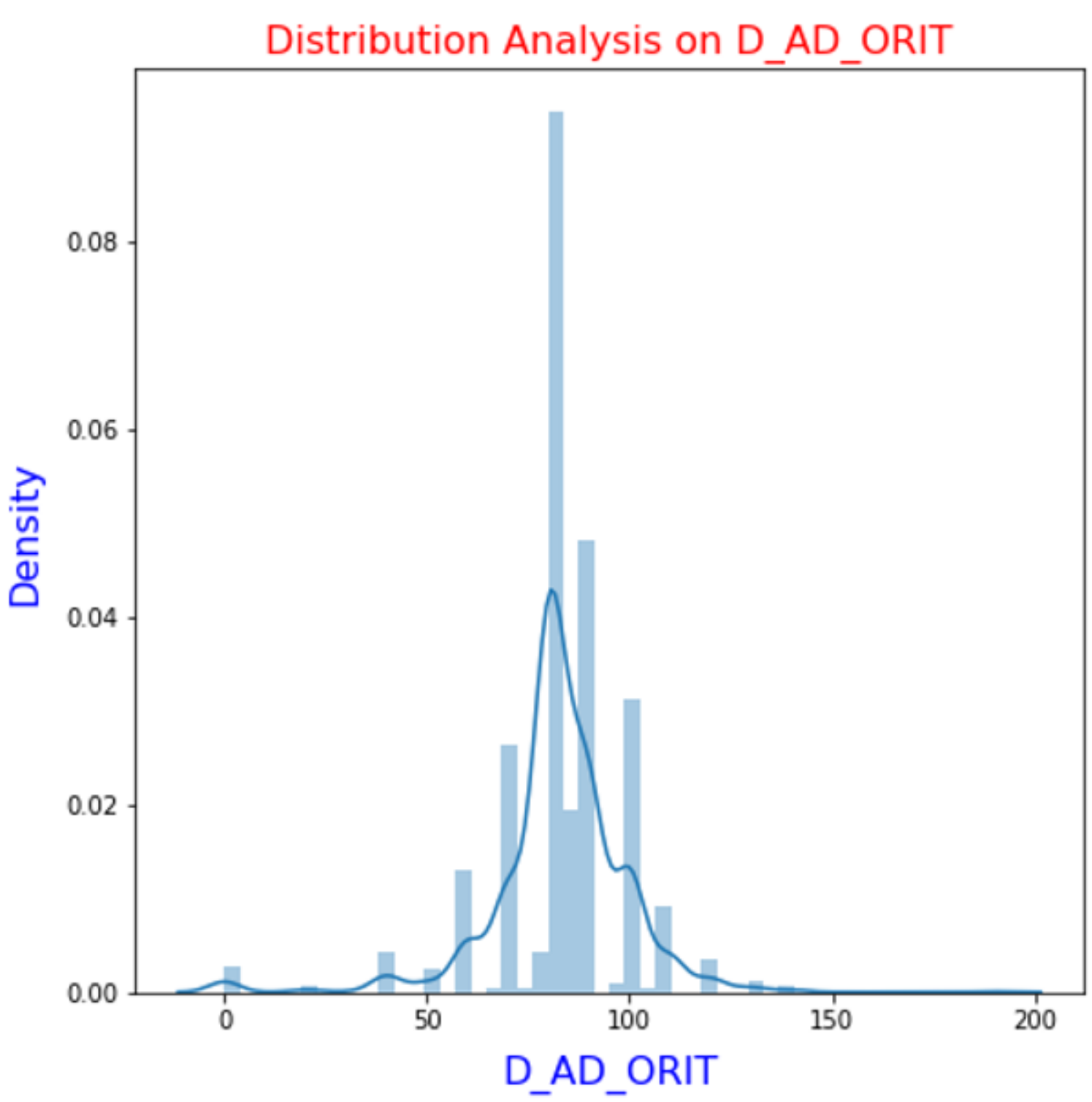

*Figure 4.14 Distribution plot of D_AD_ORIT variable*

**d) K_BLOOD (Patient's serum potassium content)**

As it can be observed from below table 4.14, a total of 1,700 observations have been reported for patients with myocardial infarction. The reported average serum potassium content in patients admitted to the hospital with this cardiovascular illness is 4.19 mmol/L. A patient with this ailment might have 2.300 mmol/L minimum serum potassium content and maximum of 8.200 mmol/L, according to this analysis. Also, approximately 25% of the reported patients having less than 3.800 mmol/L serum potassium content, 50% of the reported patients having less than 4.18 mmol/L serum potassium content and 75% of the reported patients having less than 4.500 mmol/L serum potassium content. This reported analysis is as per the number of observations present in the dataset.

*Table 4.16 Summary Statistics of K_BLOOD feature*

| K_BLOOD | | | | | | | |
|---|---|---|---|---|---|---|---|
| **Count** | **Mean Value** | **Standard Deviation** | **Minimum Value** | **25%** | **50%** | **75%** | **Maximum Value** |
| 1700 | 4.191923 | 0.666948 | 2.300 | 3.800 | 4.182041 | 4.500 | 8.200 |

From the below figure 4.15 shows that the K_BLOOD variable almost follows a right skewed distribution. It can be observed that most of the patients those who are suffering from this disease having serum potassium content between 3.5 to 4.5 mmol/L.

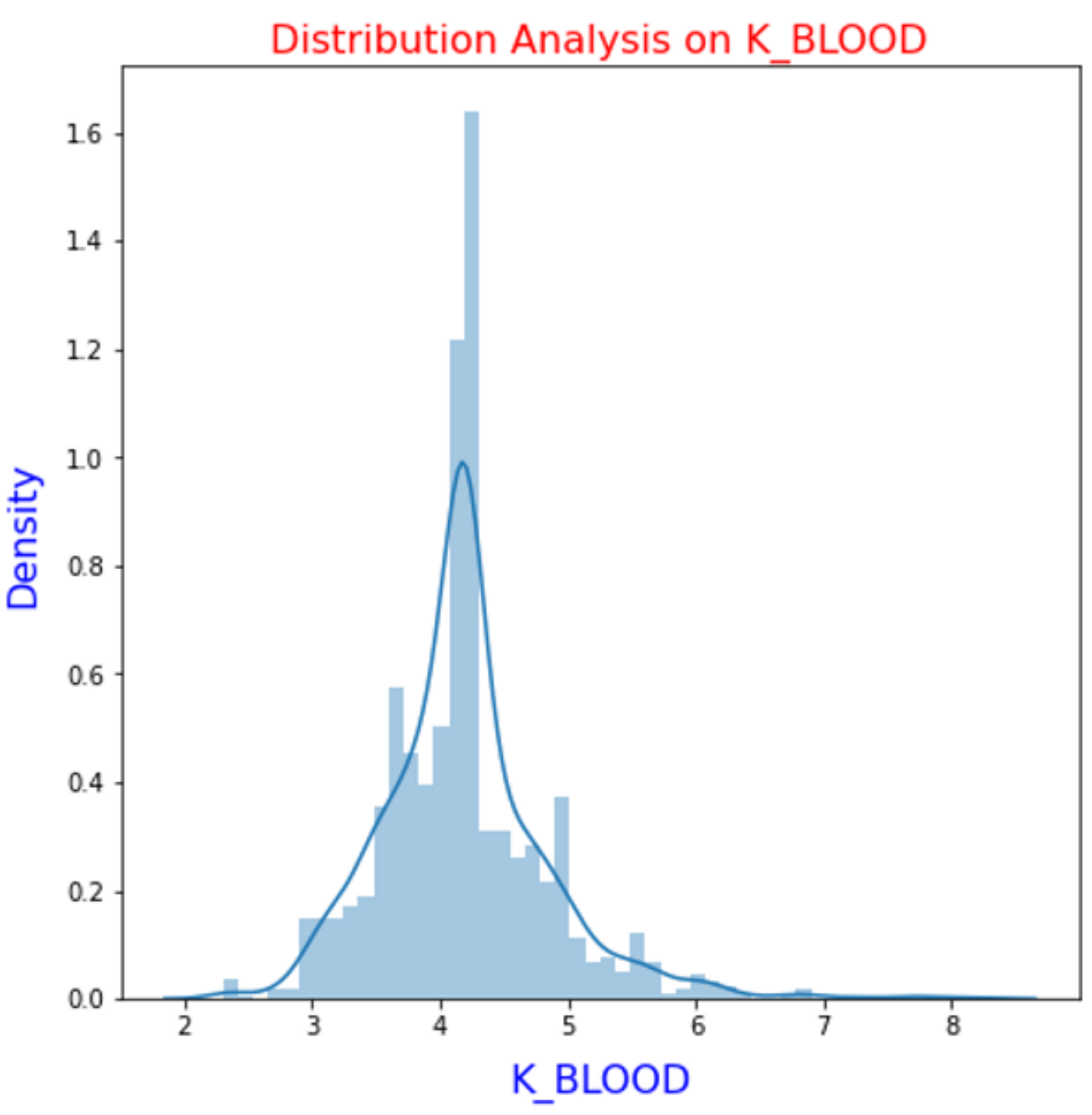

*Figure 4.15 Distribution plot of K_BLOOD variable*

### e) ALT_BLOOD (Patient's Serum AlAT content)

As it can be observed from below table 4.15, a total of 1,700 observations have been reported for patients with myocardial infarction. The reported average serum AlAT content in patients admitted to the hospital with this cardiovascular illness is 0.48 IU/L. A patient with this ailment might have 0.030 IU/L minimum serum AlAT content and maximum of 3.00 IU/L, according to this analysis. Also, approximately 25% of the reported patients having less than 0.230 IU/L serum AlAT content, 50% of the reported patients having less than 0.44 IU/L serum AlAT content and 75% of the reported patients having less than 0.55 IU/L serum AlAT content. This reported analysis is as per the number of observations present in the dataset.

*Table 4.15 Summary Statistics of ALT_BLOOD feature*

| ALT_BLOOD | | | | | | | |
|---|---|---|---|---|---|---|---|
| **Count** | **Mean Value** | **Standard Deviation** | **Minimum Value** | **25%** | **50%** | **75%** | **Maximum Value** |
| 1700 | 0.483044 | 0.354662 | 0.0300 | 0.2300 | 0.442813 | 0.555113 | 3.000000 |

From the below figure 4.16 shows that the ALT_BLOOD variable follows a right skewed distribution. It can be observed that most of the patients those who are suffering from this disease having serum AlAT content between 0.3 to 0.7 IU/L.

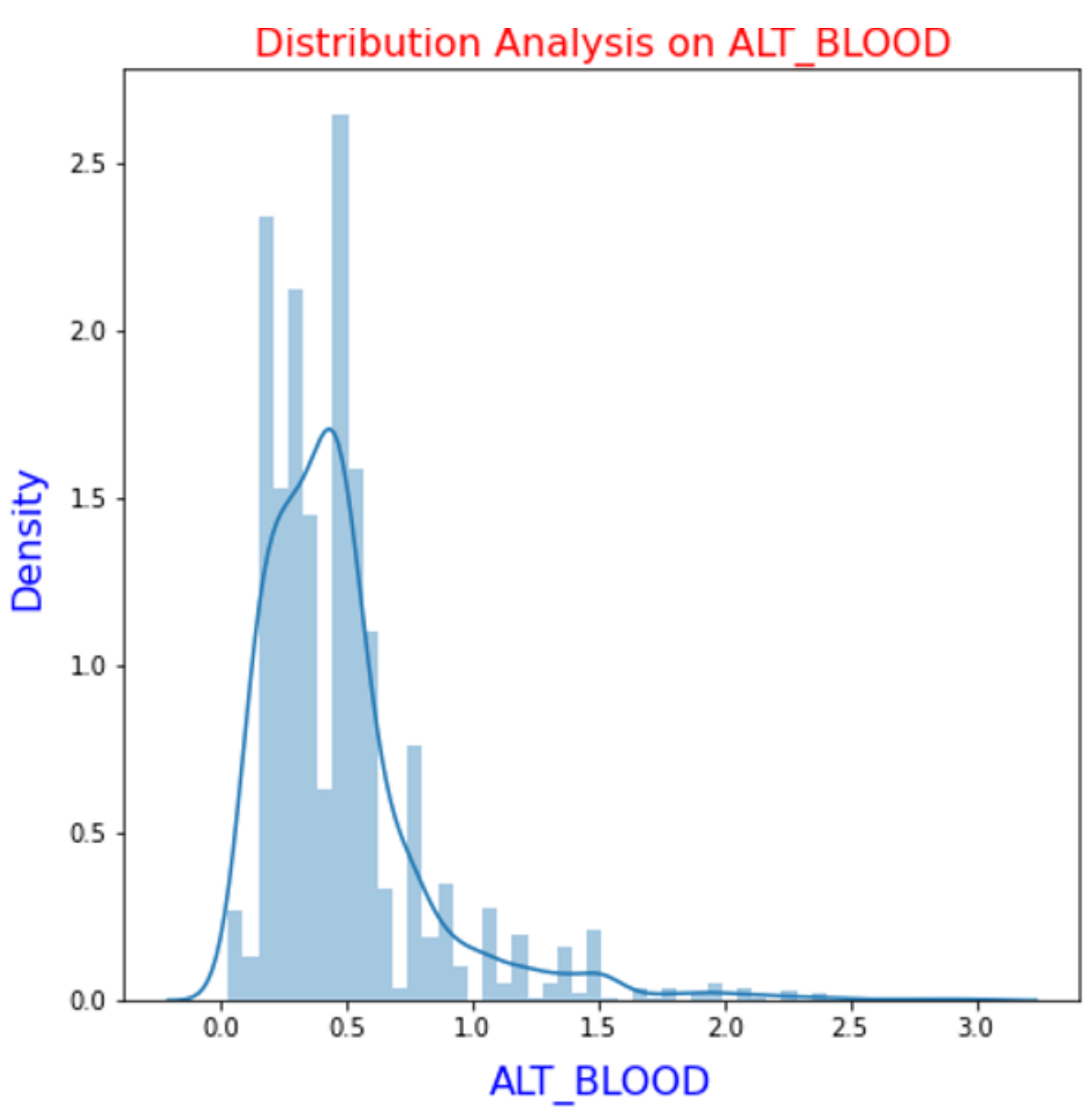

*Figure 4.16 Distribution plot of ALT_BLOOD variable*

#### f) K_SH_POST (Cardiogenic shock at the time of admission to ICU)

As it can be observed from below table 4.16, a total of 1,700 observations have been reported for patients suffering from lethal outcome of acute myocardial infarction. Among 1700 patients, 46 patients or 2.71% of them suffering from lethal outcomes due to cardiogenic shock at the time of admission to the hospital, while the remaining 1654 patients or 97.29% of them still suffering from lethal outcomes of acute myocardial infarction without any cardiogenic shock at the time of admission to the hospital.

*Table 4.18 Frequency and Percentage analysis of K_SH_POST feature*

| K_SH_POST | | | |
|---|---|---|---|
| | **Decoded Value** | **Frequency/Count** | **Percentage** |
| **No** | 0 | 1654 | 97.29 |
| **Yes** | 1 | 46 | 2.71 |
| **Total** | | 1700 | 100.00 |

Below figure 4.17 depicts the percentage of patients admitted to the hospital who had a deadly outcome from an acute myocardial infarction with or without cardiogenic shock at the time of admission to the hospital.

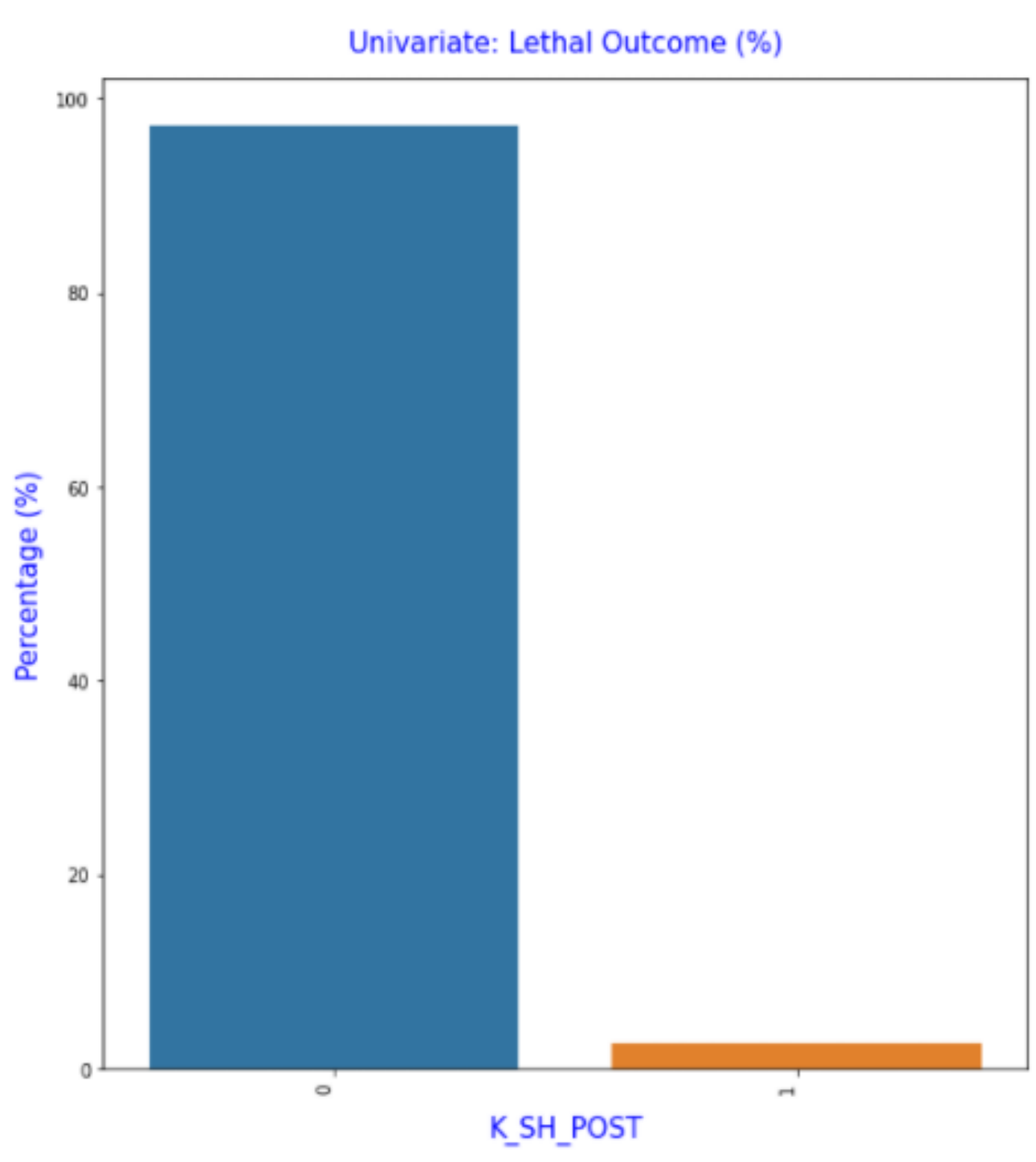

*Figure 4.17 Percentage plot of K_SH_POST feature*

**g) ritm_ecg_p_02 (ECG rhythm at the time of admission to hospital – atrial fibrillation)**

As it can be observed from below table 4.17, a total of 1,700 observations have been reported for patients suffering from lethal outcome of acute myocardial infarction. Among 1700 patients, 95 patients or 5.59% of them suffering from lethal outcomes and having atrial fibrillation (irregular ECG rhythm) at the time of admission to the hospital, while the remaining 1605 patients or 94.41% of them still suffering from lethal outcomes of acute myocardial infarction and without having atrial fibrillation (irregular ECG rhythm) at the time of admission to the hospital.

*Table 4.17*19 *Frequency and Percentage analysis of ritm_ecg_p_02 feature*

| ritm_ecg_p_02 | | | |
|---|---|---|---|
| | **Decoded Value** | **Frequency/Count** | **Percentage** |
| **No** | 0 | 1605 | 94.41 |
| **Yes** | 1 | 95 | 5.59 |
| **Total** | | 1700 | 100.00 |

Below figure 4.18 depicts the percentage of patients admitted to the hospital who had a deadly outcome from an acute myocardial infarction with or without having atrial fibrillation (irregular ECG rhythm) at the time of admission to the hospital.

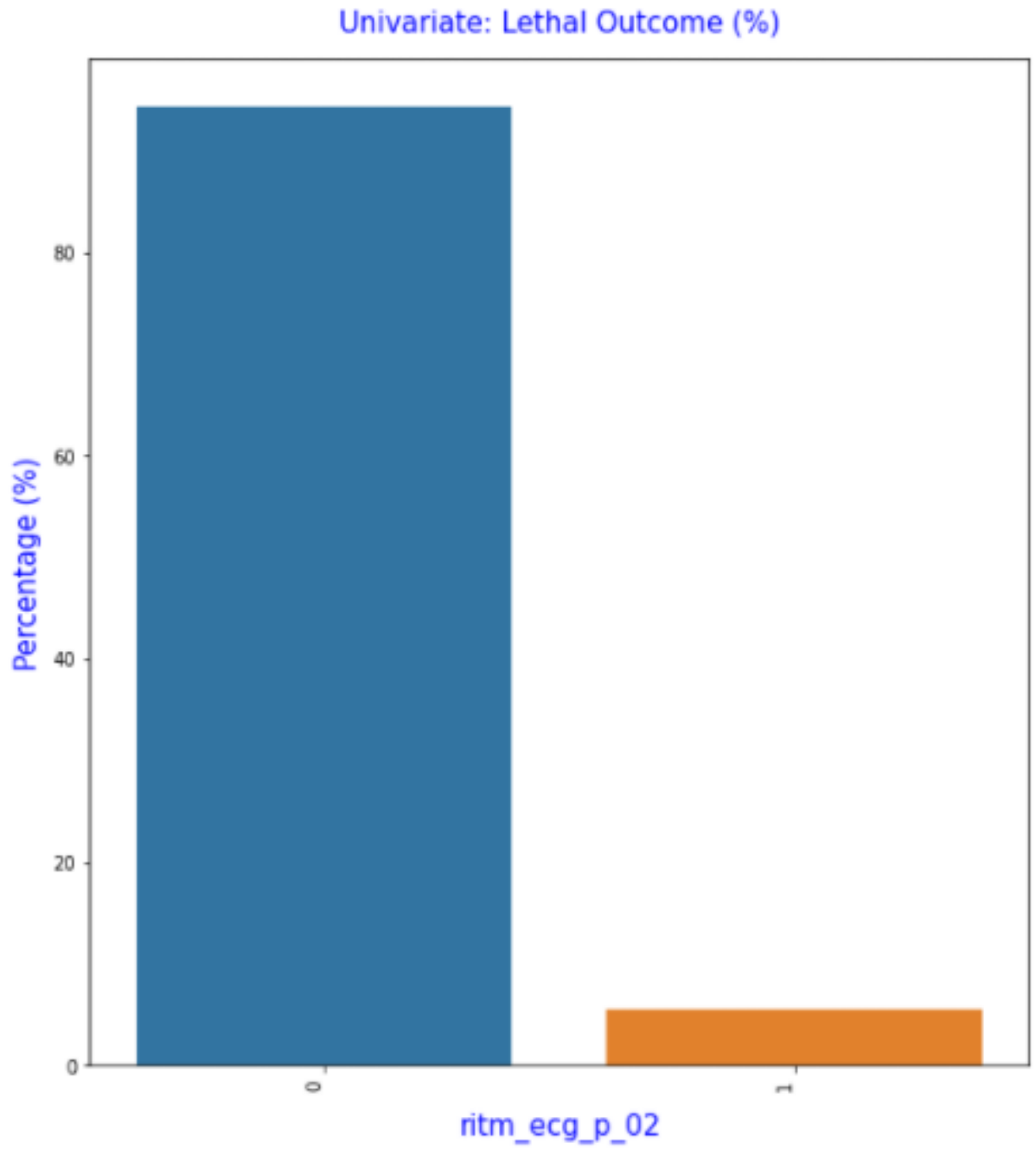

*Figure 4.18* 23*Percentage plot of ritm_ecg_p_02 feature*

**h) MP_TP_POST (Paroxysms of atrial fibrillation at the time of admission to ICU)**

As it can be observed from below table 4.18, a total of 1,700 observations have been reported for patients suffering from lethal outcome of acute myocardial infarction. Among 1700 patients, 114 patients or 6.71% of them suffering from lethal outcomes and having paroxysms of atrial fibrillation (erratic heart rate begins) at the time of admission to the ICU, while the remaining 1586 patients or 93.29% of them still suffering from lethal outcomes of acute myocardial infarction and without having paroxysms of atrial fibrillation (erratic heart rate begins) at the time of admission to the ICU.

*Table 4.20 Frequency and Percentage analysis of MP_TP_POST feature*

| MP_TP_POST | | | |
|---|---|---|---|
| | **Decoded Value** | **Frequency/Count** | **Percentage** |
| **No** | 0 | 1586 | 93.29 |
| **Yes** | 1 | 114 | 6.71 |
| **Total** | | 1700 | 100.00 |

Below figure 4.19 depicts the percentage of patients admitted to the hospital who had a deadly outcome from an acute myocardial infarction with or without having paroxysms of atrial fibrillation (erratic heart rate begins) at the time of admission to the ICU.

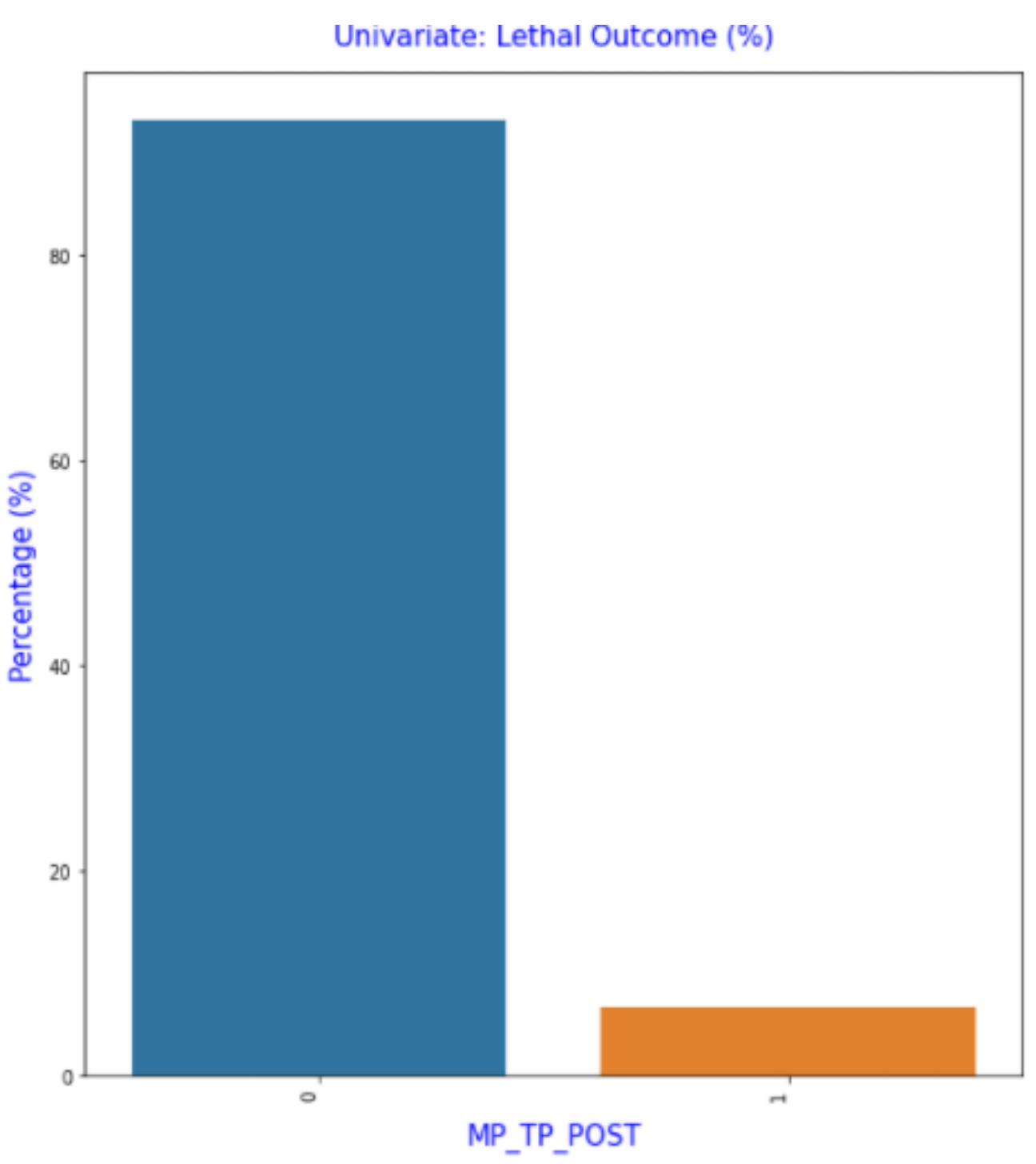

*Figure 4.19 Percentage plot of MP_TP_POST feature*

### i) SEX (Gender)

As it can be observed from below table 4.19, a total of 1,700 observations have been reported for patients suffering from lethal outcome of acute myocardial infarction. Among 1700 patients, 635 patients or 37.35% of them suffering from lethal outcomes of this disease are female, while the remaining 1065 patients or 62.65% of them are male.

*Table 4.21 Frequency and Percentage analysis of SEX feature*

| SEX | | | |
|---|---|---|---|
| | **Decoded Value** | **Frequency/Count** | **Percentage** |
| **Female** | 0 | 635 | 37.35 |
| **Male** | 1 | 1065 | 62.65 |
| **Total** | | 1700 | 100.00 |

Below figure 4.20 depicts the percentage of patients admitted to the hospital who had a deadly outcome from an acute myocardial infarction are male or female.

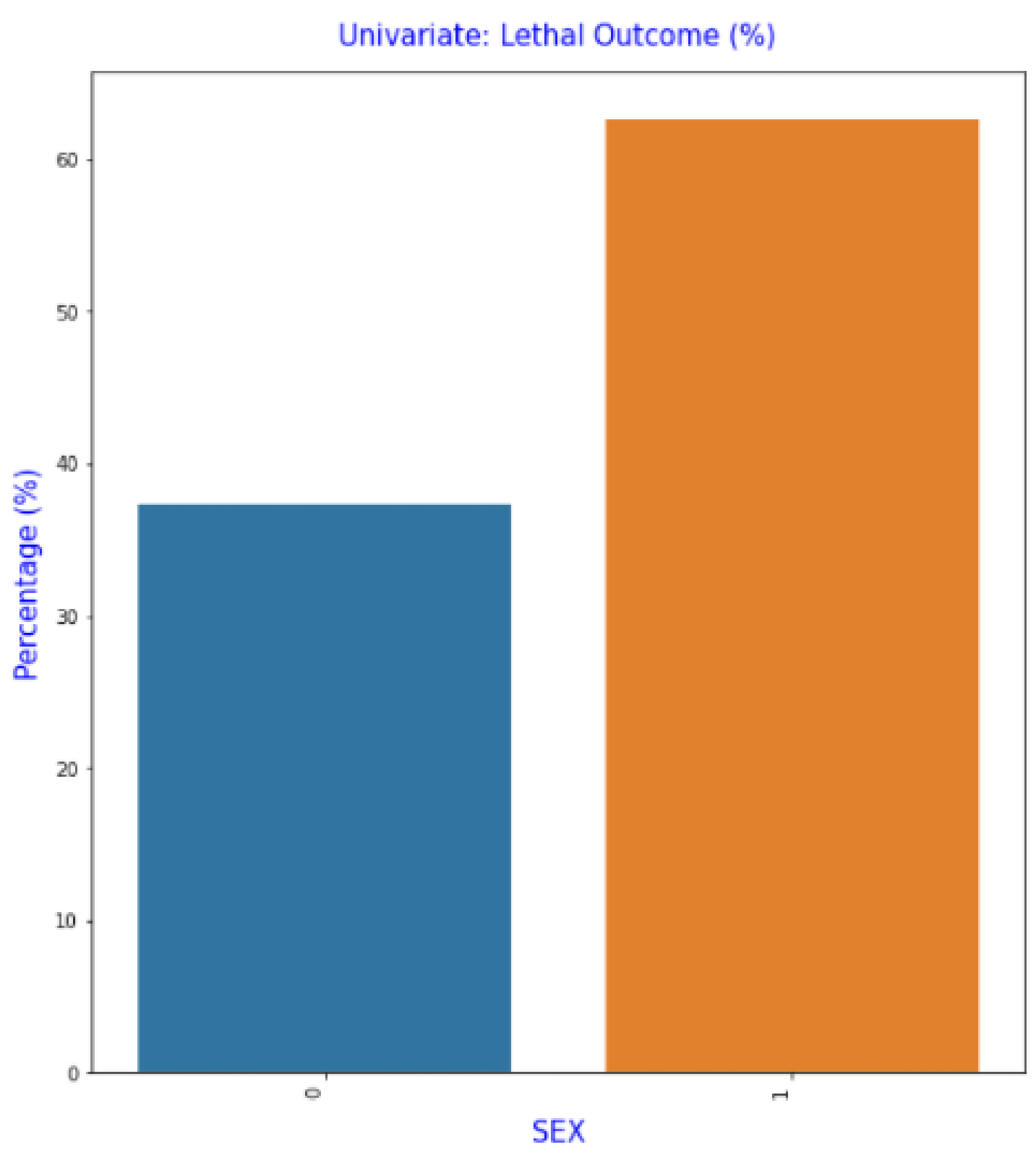

*Figure 4.20 Percentage plot of SEX feature*

### j) NITR_S (Use of liquid nitrates in the ICU)

As it can be observed from below table 4.20, a total of 1,700 observations have been reported for patients suffering from lethal outcome of acute myocardial infarction. Among 1700 patients, 195 patients or 11.47% of them suffering from lethal outcomes of this disease used liquid nitrates in the ICU, while the remaining 1,505 patients or 88.53% of them did not use liquid nitrates in the ICU.

*Table 4.22 Frequency and Percentage analysis of NITR_S feature*

| NITR_S | | | |
|---|---|---|---|
| | **Decoded Value** | **Frequency/Count** | **Percentage** |
| **No** | 0 | 1505 | 88.53 |
| **Yes** | 1 | 195 | 11.47 |
| **Total** | | 1700 | 100.00 |

Below figure 4.21 depicts the percentage of patients admitted to the hospital who had a deadly outcome from an acute myocardial infarction used or not used liquid nitrates in the ICU during their ailments.

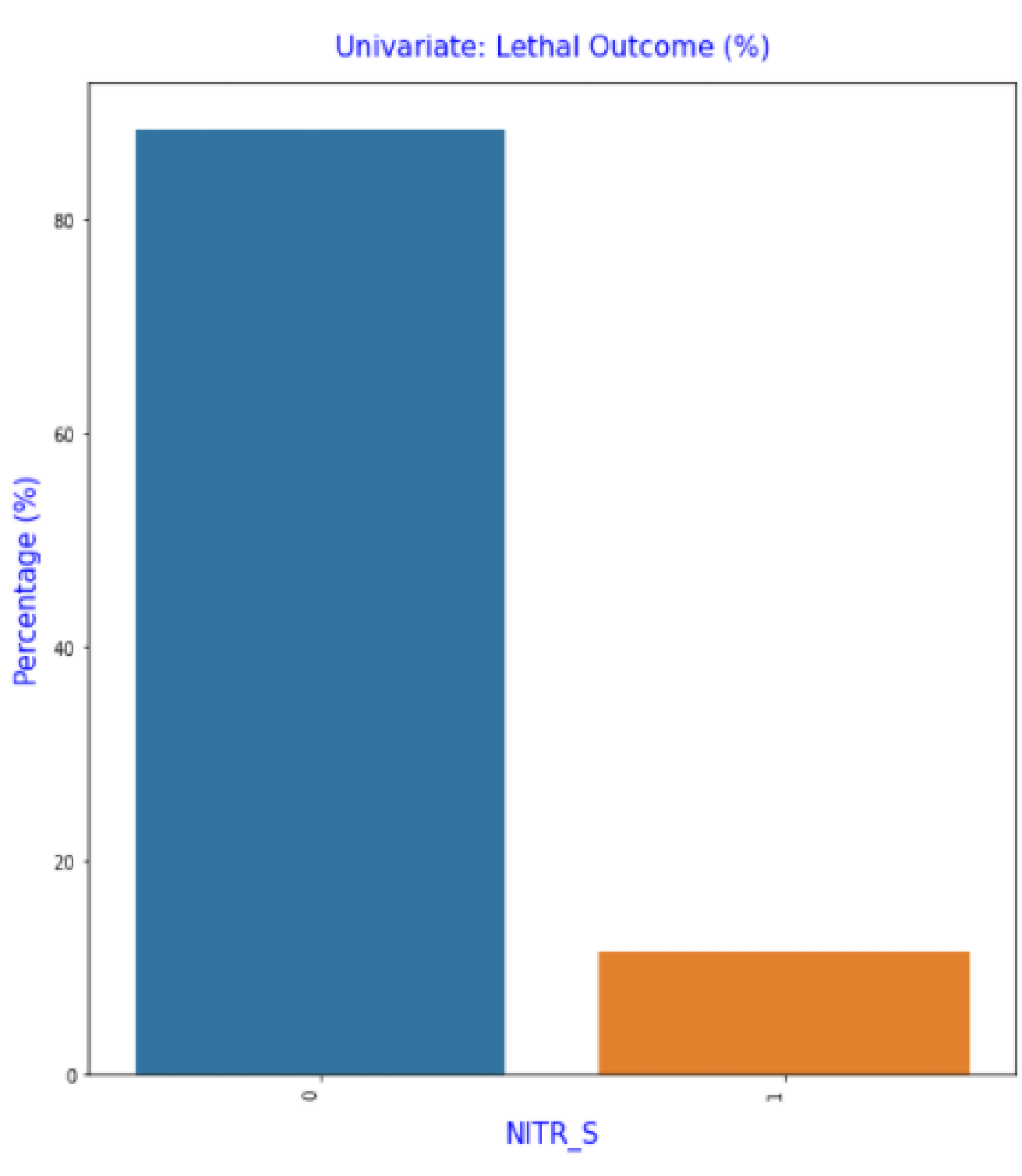

*Figure 4.21 Percentage plot of NITR_S feature*

**k) R_AB_1_n (Relapse of the pain in the first hours of the hospital period)**

As it can be observed from below table 4.21, a total of 1,700 observations have been reported for patients suffering from lethal outcome of acute myocardial infarction. Among 1700 patients, 1298 patients or 76.35% of them suffering from lethal outcomes of this disease with no relapse of heart pain in the first hours after admission to hospital, 298 patients or 17.53% of them suffering from lethal outcomes of this disease with one time relapse of heart pain in the first hours after admission to hospital, 78 patients or 4.59% of them suffering from lethal outcomes of this disease with two time relapse of heart pain in the first hours after admission to hospital and the rest 26 patients or 1.53% of them suffering from lethal outcomes of this disease with three or more times relapse of heart pain in the first hours after admission to hospital.

*Table 4.23 Frequency and Percentage analysis of R_AB_1_n feature*

| R_AB_1_n | | | |
|---|---|---|---|
| | **Decoded Value** | **Frequency/Count** | **Percentage** |
| **there is no relapse** | 0 | 1298 | 76.35 |
| **only once** | 1 | 298 | 17.53 |
| **2 times of relapse** | 2 | 78 | 4.59 |
| **3 or more times of relapse** | 3 | 26 | 1.53 |
| **Total** | | 1700 | 100.00 |

Below figure 4.22 depicts the percentage of patients admitted to the hospital who had a deadly outcome from an acute myocardial infarction with different frequency of relapse in their heart pain during the first hours after admission to hospital.

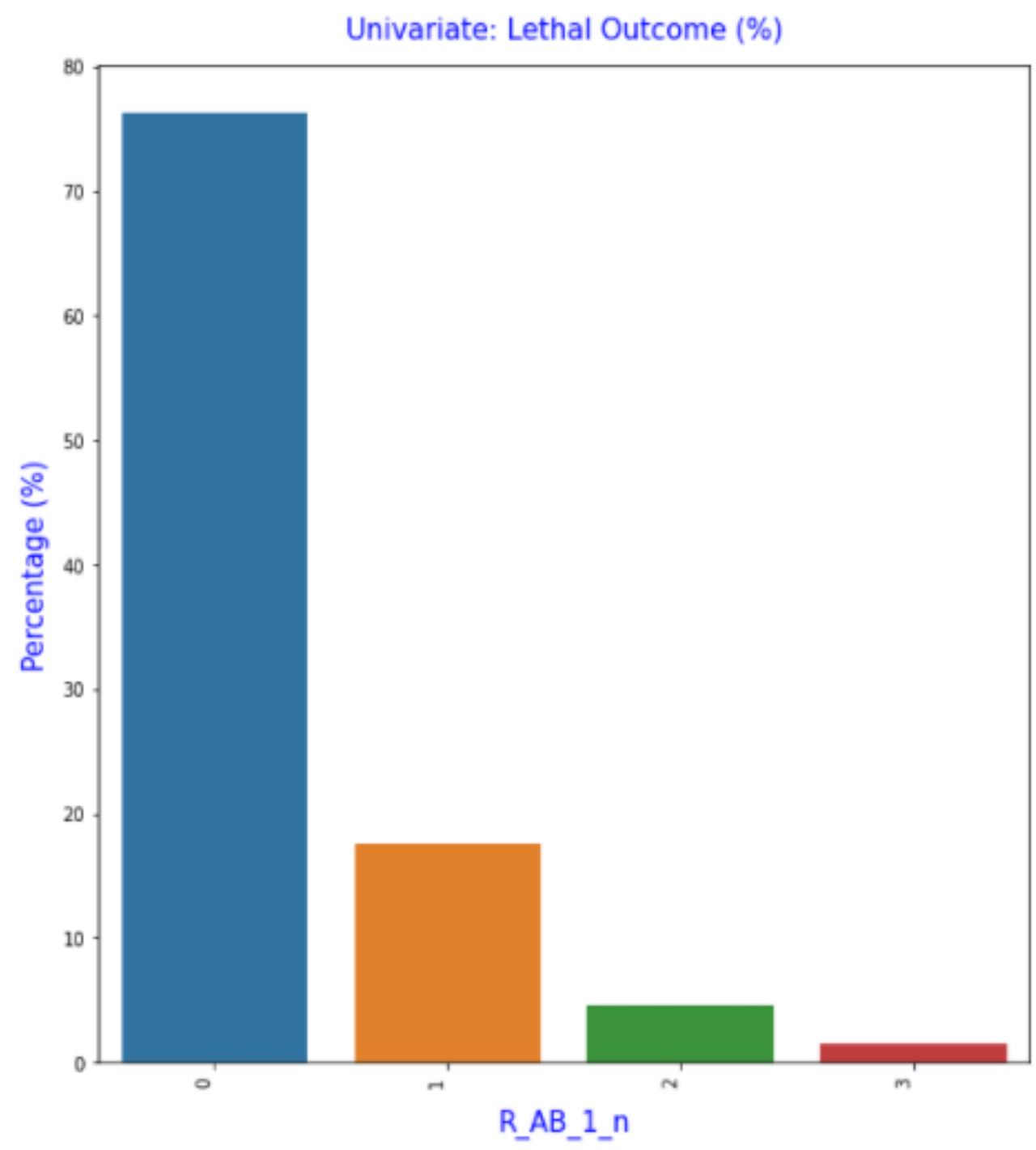

*Figure 4.22 Percentage plot of R_AB_1_n feature*

**l) n_p_ecg_p_12 (Complete RBBB on ECG at the time of admission to hospital)**

As it can be observed from below table 4.22, a total of 1,700 observations have been reported for patients suffering from lethal outcome of acute myocardial infarction. Among 1,700 patients, 78 patients or 4.59% of them suffering from lethal outcomes of this disease having complete right bundle branch block found during ECG at the time of admission to hospital, while the remaining 1622 patients or 95.41% of them did not have complete right bundle branch block found during ECG at the time of admission to hospital.

*Table 4.22 Frequency and Percentage analysis of n_p_ecg_p_12 feature*

| n_p_ecg_p_12 | | | |
|---|---|---|---|
| | **Decoded Value** | **Frequency/Count** | **Percentage** |
| **No** | 0 | 1622 | 95.41 |
| **Yes** | 1 | 78 | 4.59 |
| **Total** | | 1700 | 100.00 |

Below figure 4.23 depicts the percentage of patients admitted to the hospital who had a deadly outcome from an acute myocardial infarction having or not having complete right bundle branch block found during ECG at the time of admission to hospital.

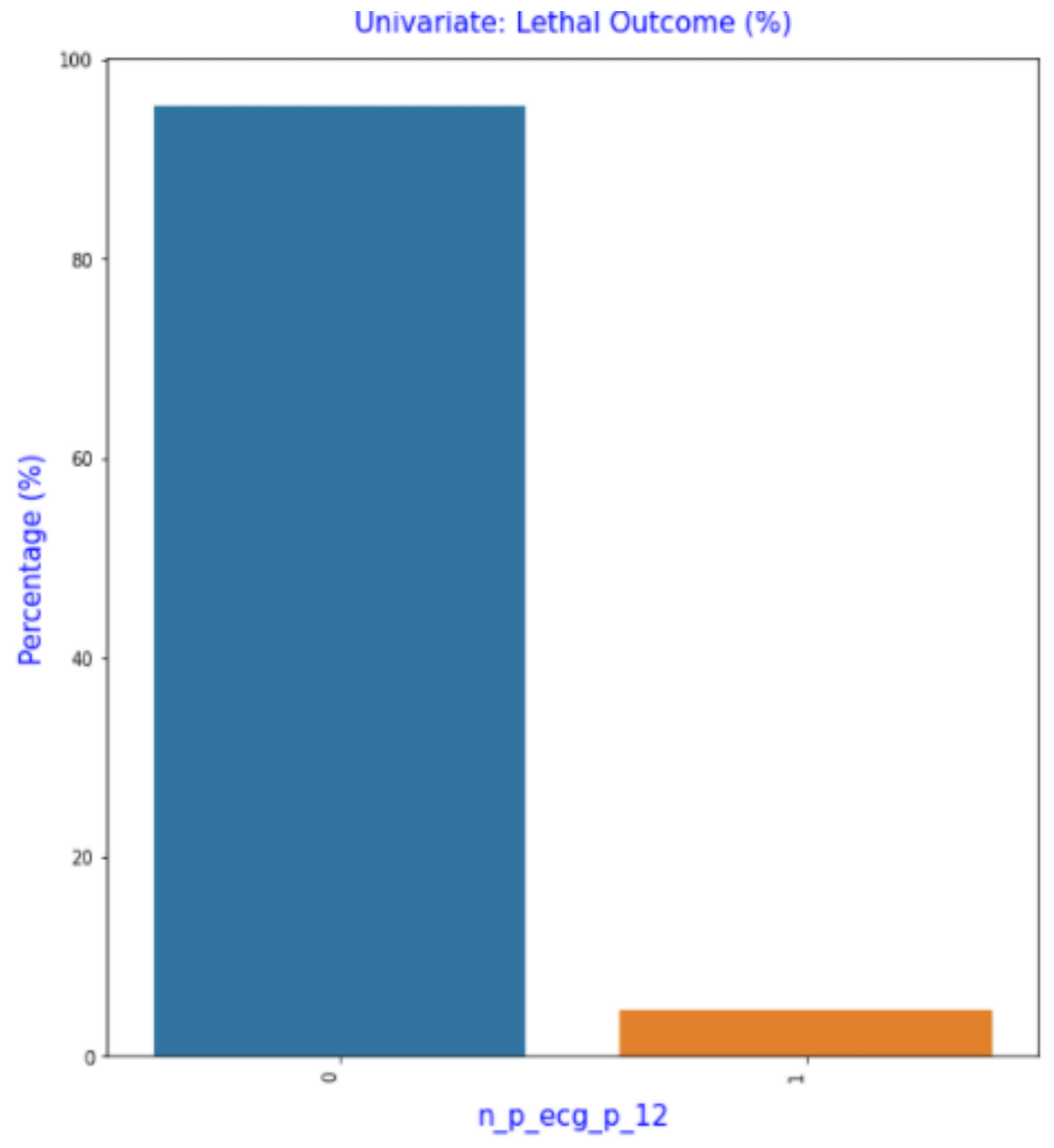

*Figure 4.23 Percentage plot of n_p_ecg_p_12 feature*

**m) NOT_NA_KB (Use of NSAIDs by the emergency cardiology team)**

As it can be observed from below table 4.23, a total of 1,700 observations have been reported for patients suffering from lethal outcome of acute myocardial infarction. Among 1700 patients, 313 patients or 18.41% of them suffering from lethal outcomes of this disease those who were not subjected to any non-steroidal anti-inflammatory drugs by cardiology team, while the remaining 1387 patients or 81.59% of them suffering from lethal outcomes of this disease those who were subjected to certain non-steroidal anti-inflammatory drugs by cardiology team.

*Table 4.25 Frequency and Percentage analysis of NOT_NA_KB feature*

| NOT_NA_KB | | | |
|---|---|---|---|
| | **Decoded Value** | **Frequency/Count** | **Percentage** |
| **No** | 0 | 313 | 18.41 |
| **Yes** | 1 | 1387 | 81.59 |
| **Total** | | 1700 | 100.00 |

Figure 4.24 displays the percentage of patients brought to the hospital who died because of an acute myocardial infarction who were or were not given nonsteroidal anti-inflammatory medicines by the cardiology staff.

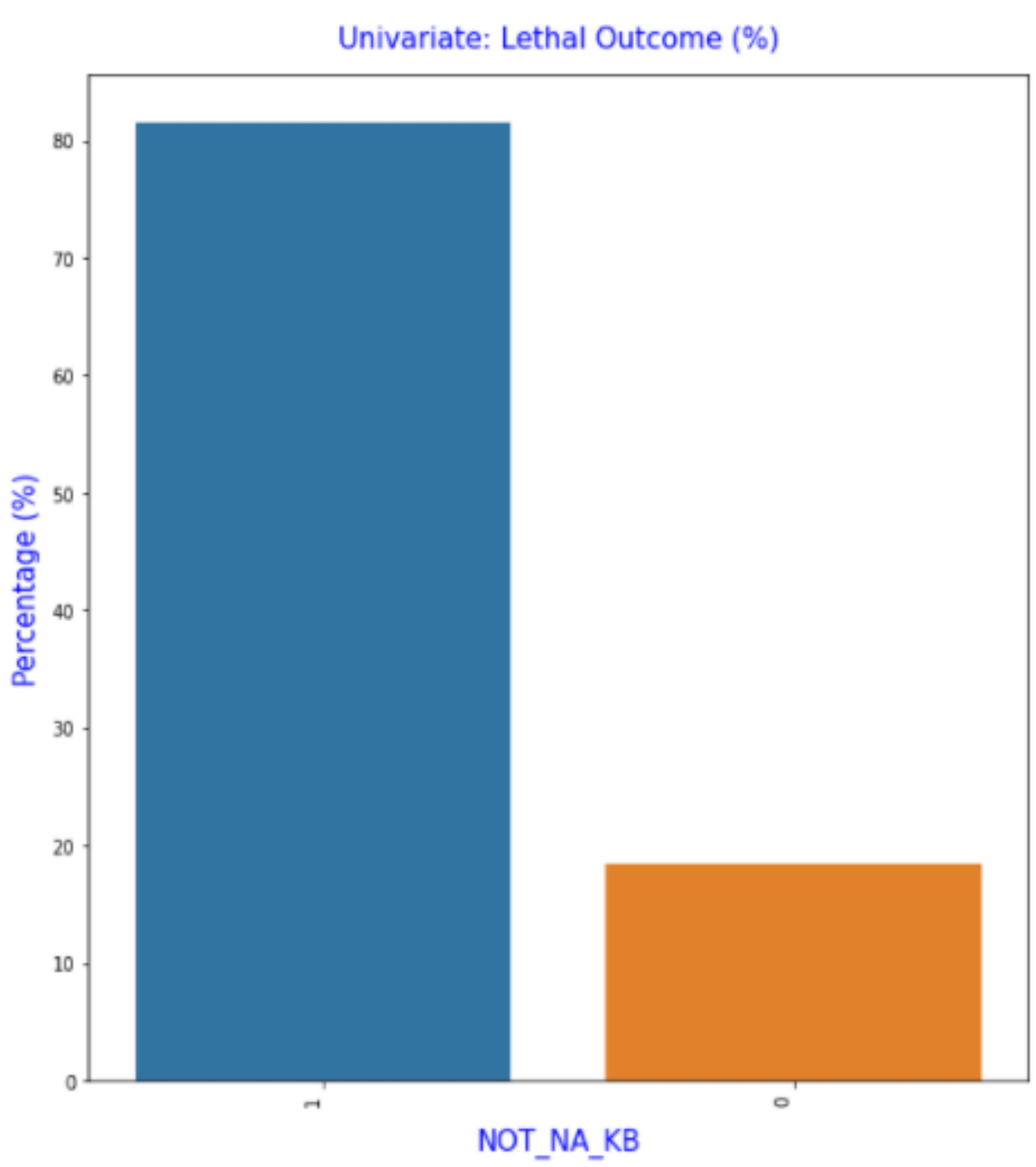

*Figure 4.24 Percentage plot of NOT_NA_KB features*

**n) O_L_POST (Pulmonary edema at the time of admission to ICU)**

As it can be observed from below table 4.24, a total of 1,700 observations have been reported for patients suffering from lethal outcome of acute myocardial infarction. Among 1700 patients, 1590 patients or 93.53% of them suffering from lethal outcomes of this disease not due to pulmonary edema at the time of admission to ICU, while the remaining 110 patients or 6.47% of them suffering from lethal outcomes of this disease due to pulmonary edema at the time of admission to ICU.

*Table 4.26 Frequency and Percentage analysis of O_L_POST feature*

| O_L_POST | | | |
|---|---|---|---|
| | **Decoded Value** | **Frequency/Count** | **Percentage** |
| **No** | 0 | 1590 | 93.53 |
| **Yes** | 1 | 110 | 6.47 |
| **Total** | | 1700 | 100.00 |

Figure 4.25 displays the percentage of patients brought to the hospital who died because of an acute myocardial infarction those who suffered or not suffered from pulmonary edema at the time of admission to ICU.

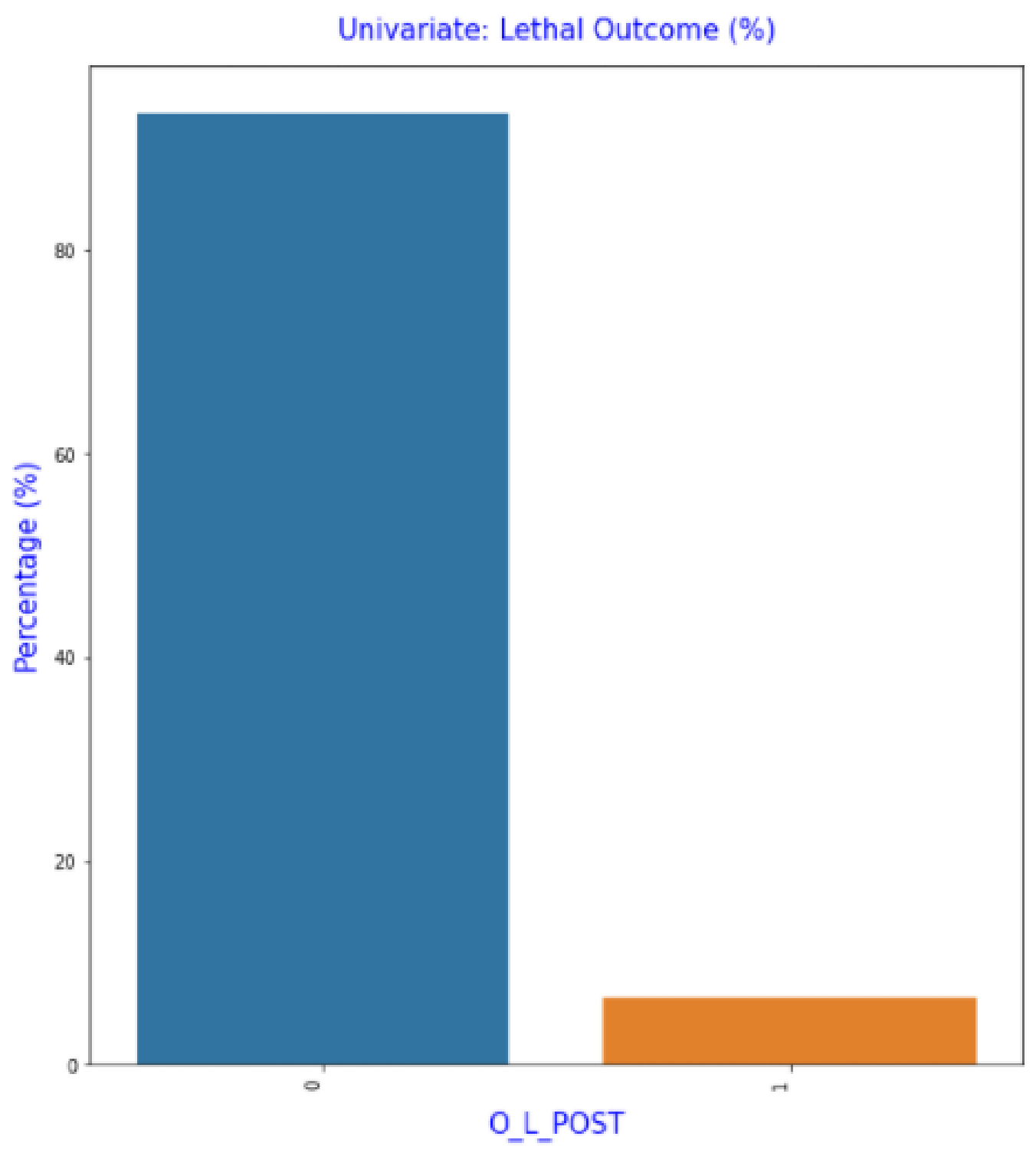

*Figure 4.25 Percentage plot of O_L_POST feature*

**o) GEPAR_S_n (Use of a anticoagulants (heparin) in the ICU)**

As it can be observed from below table 4.25, a total of 1700 observations have been reported for patients suffering from lethal outcome of acute myocardial infarction. Among 1,700 patients, 480 patients or 28.24% of them suffering from lethal outcomes of this disease were not subjected to use of anticoagulants (heparin) in ICU, while the remaining 1,220 patients or 71.76% of them suffering from lethal outcomes of this disease were subjected to use of anticoagulants (heparin) in ICU.

*Table 4.27 Frequency and Percentage analysis of GEPAR_S_n feature*

| GEPAR_S_n | | | |
|---|---|---|---|
| | **Decoded Value** | **Frequency/Count** | **Percentage** |
| **No** | 0 | 480 | 28.24 |
| **Yes** | 1 | 1220 | 71.76 |
| **Total** | | 1700 | 100.00 |

Figure 4.26 displays the percentage of patients brought to the hospital who died because of an acute myocardial infarction those who subjected or not subjected to use of anticoagulants (heparin) in ICU.

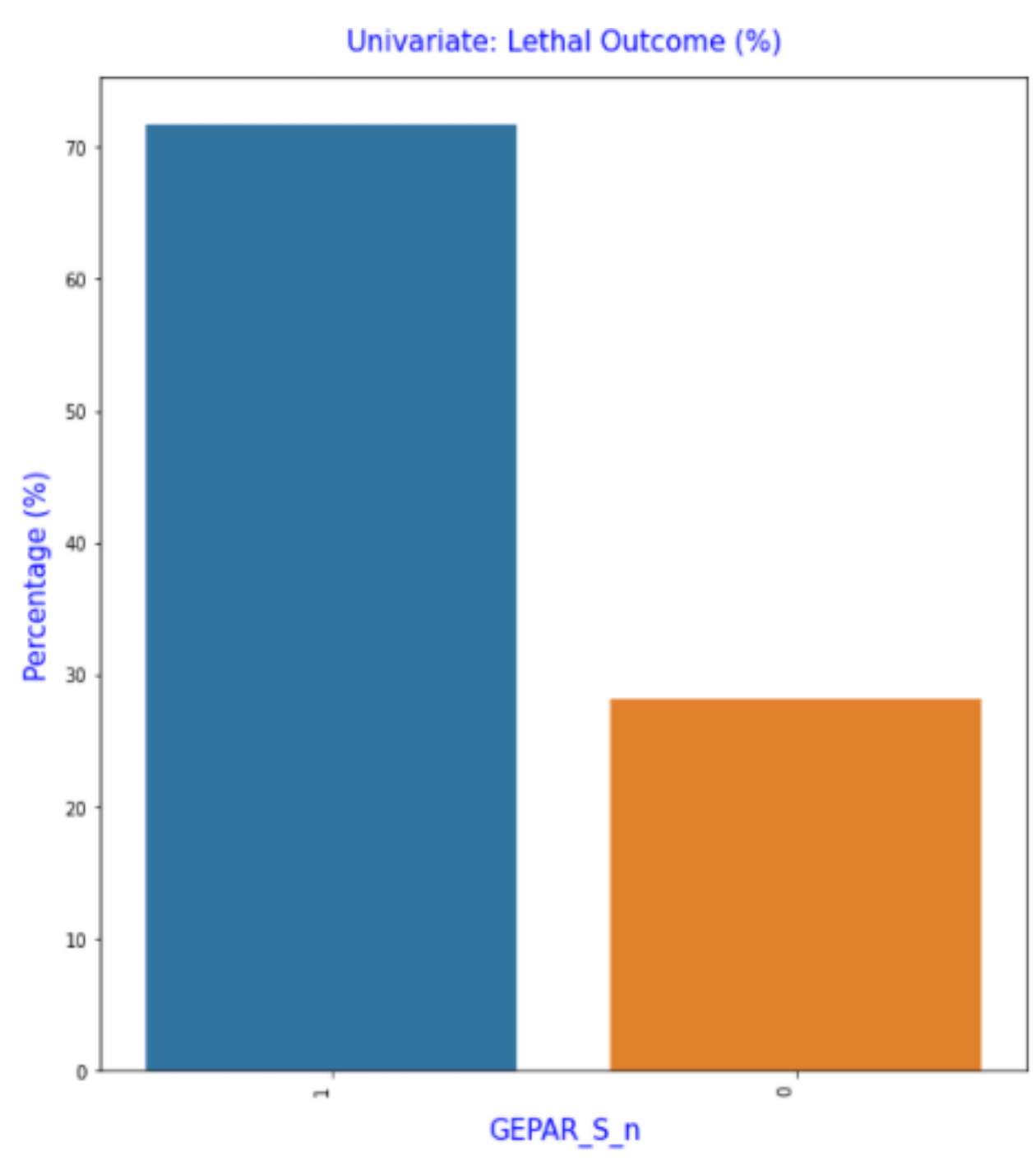

*Figure 4.26 Percentage plot of GEPAR_S_n feature*

### p) ASP_S_n (Use of acetylsalicylic acid in the ICU)

As it can be observed from below table 4.26, a total of 1,700 observations have been reported for patients suffering from lethal outcome of acute myocardial infarction. Among 1,700 patients, 431 patients or 25.35% of them suffering from lethal outcomes of this disease were not subjected to use of acetylsalicylic acid in ICU, while the remaining 1,269 patients or 74.65% of them suffering from lethal outcomes of this disease were subjected to use of acetylsalicylic acid in ICU.

*Table 4.28 Frequency and Percentage analysis of ASP_S_n feature*

| ASP_S_n | | | |
|---|---|---|---|
| | **Decoded Value** | **Frequency/Count** | **Percentage** |
| **No** | 0 | 431 | 25.35 |
| **Yes** | 1 | 1269 | 74.65 |
| **Total** | | 1700 | 100.00 |

Figure 4.27 displays the percentage of patients brought to the hospital who were suffering from fatal outcome because of an acute myocardial infarction those who subjected or not subjected to use of acetylsalicylic acid in ICU.

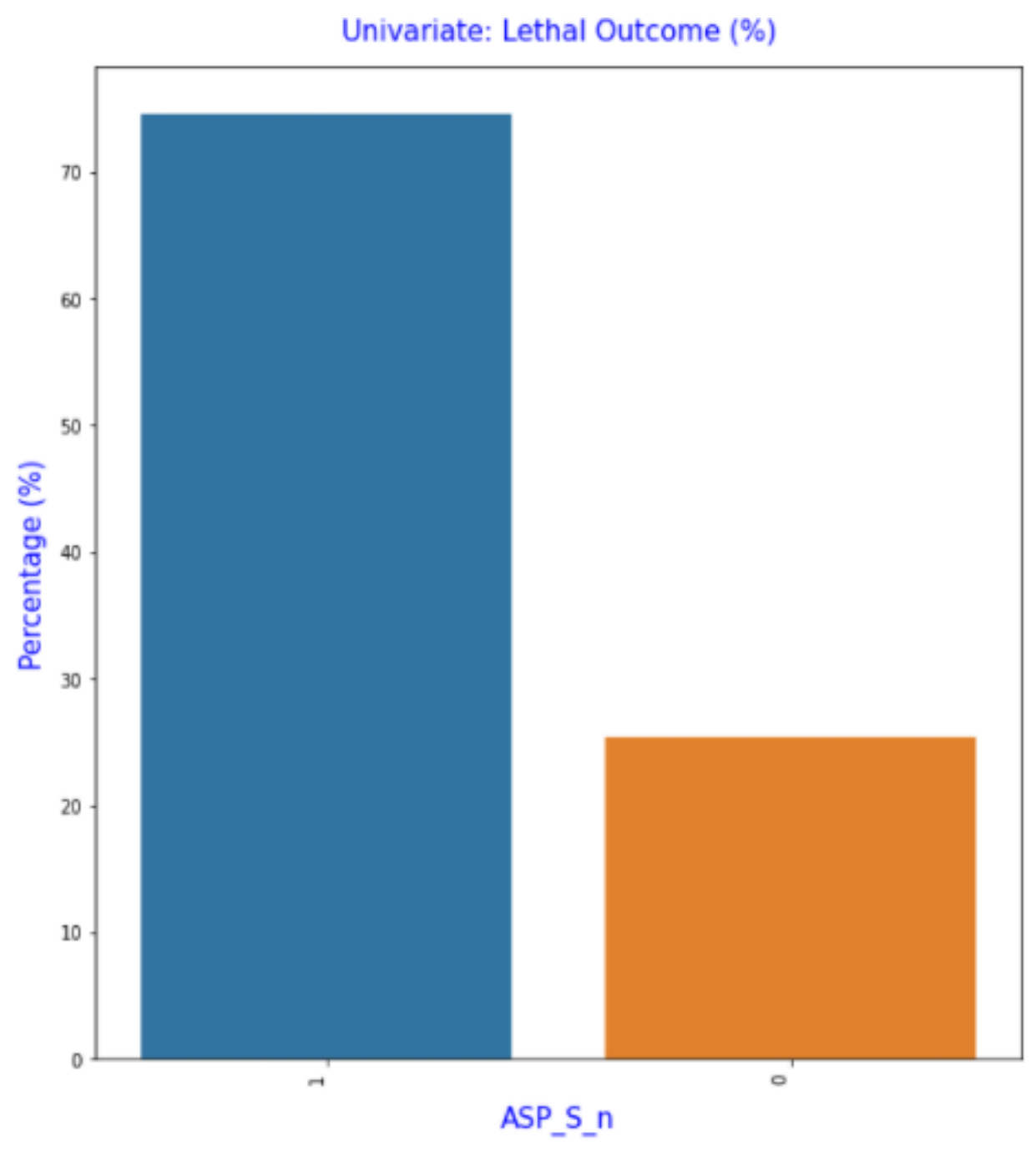

*Figure 4.27 Percentage plot of ASP_S_n feature*

**q) IBS_POST (CHD in recent weeks, days before admission to hospital)**

As it can be observed from below table 4.27, a total of 1,700 observations have been reported for patients suffering from lethal outcome of acute myocardial infarction. Among 1,700 patients, 418 patients or 24.59% of them suffering from lethal outcomes of this disease who have not suffered from coronary heart disease in recent weeks, days before admission to hospital, while the 548 patients or 32.24% of them suffering from lethal outcomes of this disease who have suffered from exertional angina pectoris, a form of coronary heart disease in recent weeks, days before admission to hospital and the rest 734 patients or 43.18% of them suffering from lethal outcomes of this disease who have already suffered from unstable angina pectoris, a form of coronary heart disease in recent weeks, days before admission to hospital.

*Table 4.29 Frequency and Percentage analysis of IBS_POST feature*

| IBS_POST | | | |
|---|---|---|---|
| | **Decoded Value** | **Frequency/Count** | **Percentage** |
| **there was no CHD** | 0 | 418 | 24.59 |
| **exertional angina pectoris** | 1 | 548 | 32.24 |
| **unstable angina pectoris** | 2 | 734 | 43.18 |
| **Total** | | 1700 | 100.00 |

Figure 4.28 displays the percentage of patients with different form of coronary heart disease occurred in recent weeks or days before admission to hospital and were suffering from fatal outcome because of an acute myocardial infarction.

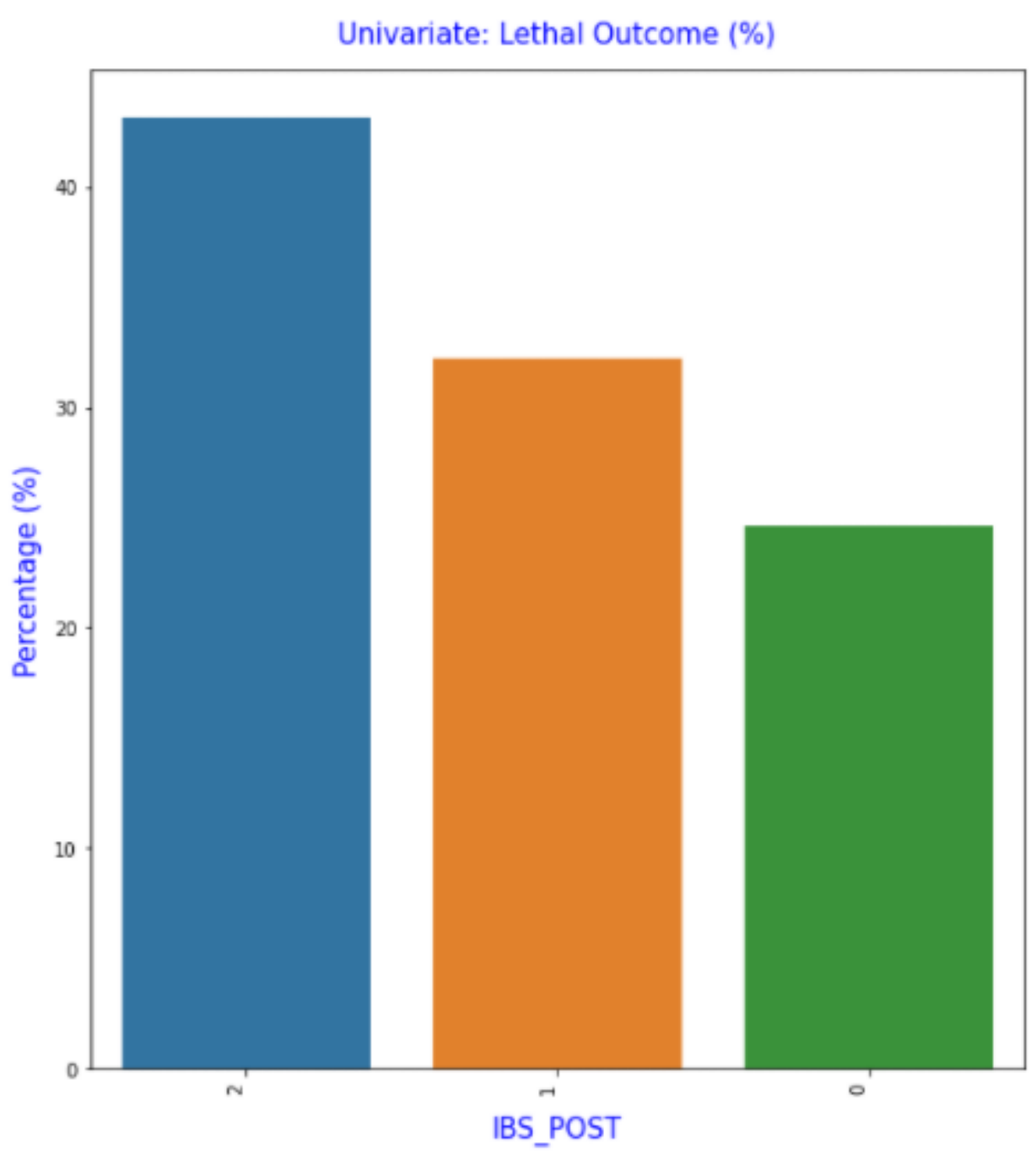

*Figure 4.28 Percentage plot of IBS_POST feature*

**r) FK_STENOK (Functional class (FC) of angina pectoris in the last year)**

As it can be observed from below table 4.28, a total of 1,700 observations have been reported for patients suffering from lethal outcome of acute myocardial infarction. Among 1,700 patients, 661 patients or 38.88% of them having no functional class of angina pectoris in the last year , while 47 patients or 2.76% of them having first functional class of angina pectoris in the last year , 927 patients or 54.53% of them suffering from lethal outcomes of this disease having second functional class of angina pectoris in the last year, another 54 patients or 3.18% of them suffering from lethal outcomes of this disease having third functional class of angina pectoris in the last year  and the rest 11 patients or 0.65% of them suffering from lethal outcomes of this disease having fourth functional class of angina pectoris in the last year.

*Table 4.30 Frequency and Percentage analysis of FK_STENOK feature*

| FK_STENOK | | | |
|---|---|---|---|
| | **Decoded Value** | **Frequency/Count** | **Percentage** |
| **there is no angina pectoris** | 0 | 661 | 38.88 |
| **I FC** | 1 | 47 | 2.76 |
| **II FC** | 2 | 927 | 54.53 |
| **III FC** | 3 | 54 | 3.18 |
| **IV FC** | 4 | 11 | 0.65 |
| **Total** | | 1700 | 100.00 |

Figure 4.29 displays the percentage of patients having different form of functional class of angina pectoris in the last year and were suffering from fatal outcome because of an acute myocardial infarction.

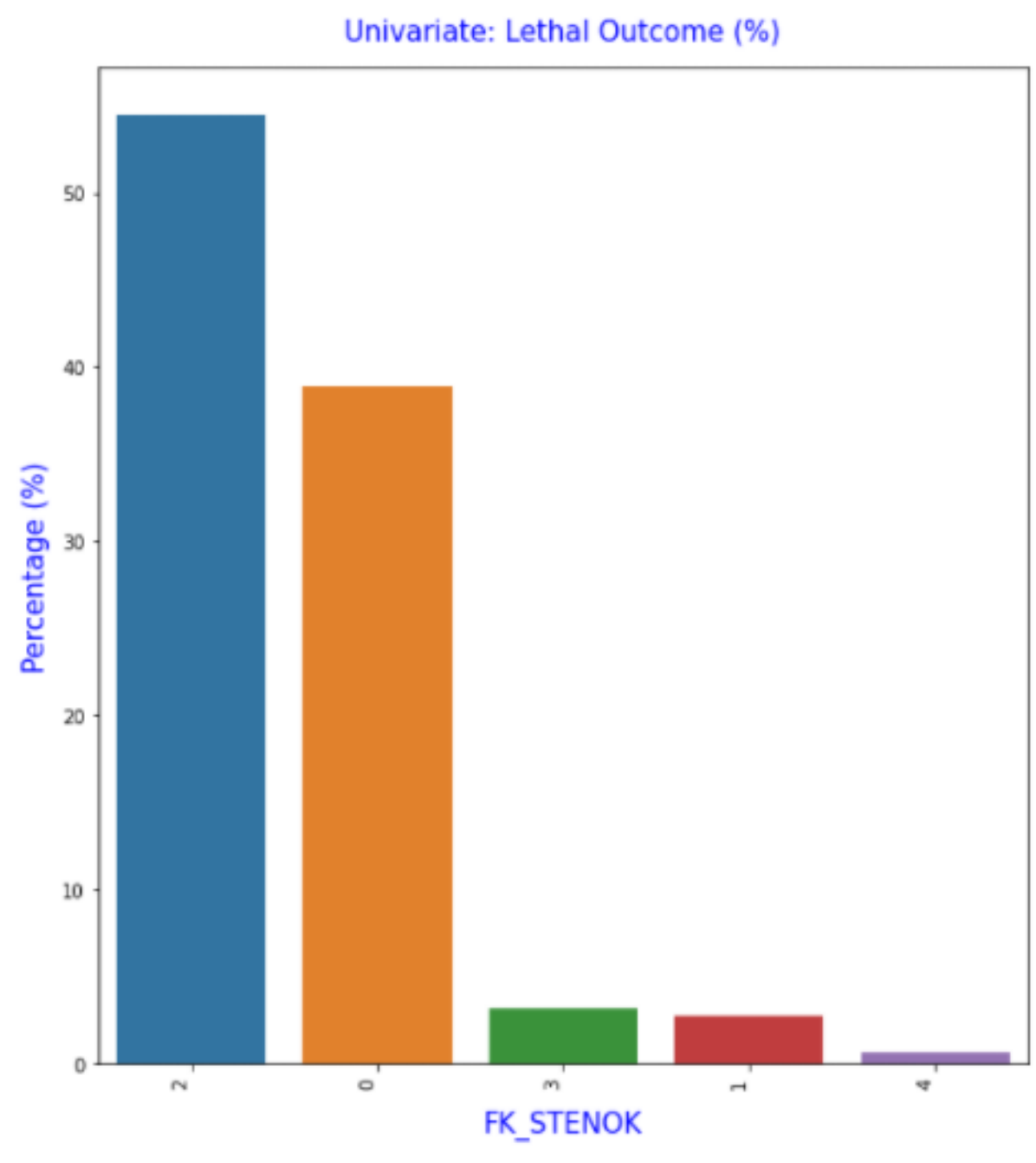

*Figure 4.29  Percentage plot of FK_STENOK feature*

**s) ant_im (Presence of an anterior myocardial infarction (left ventricular))**

As it can be observed from below table 4.29, a total of 1,700 observations have been reported for patients suffering from lethal outcome of acute myocardial infarction. Among 1700 patients, 743 patients or 43.71% of them having no anterior myocardial infarction in left ventricular but still suffering from acute myocardial infarction , while 392 patients or 23.06% of them has QRS no changes, 39 patients or 2.29% of them suffering from lethal outcomes of this disease having QRS is like QR-complex, another 34 patients or 2.00% of them suffering from lethal outcomes of this disease having QRS is like Qr-complex and the rest 492 patients or 28.94% of them suffering from lethal outcomes of this disease having QRS is like QS-complex.

*Table 4.31 Frequency and Percentage analysis of ant_im feature*

| ant_im | | | |
|---|---|---|---|
| | **Decoded Value** | **Frequency/Count** | **Percentage** |
| **there is no infarct in this location** | 0 | 743 | 43.71 |
| **QRS has no changes** | 1 | 392 | 23.06 |
| **QRS is like QR-complex** | 2 | 39 | 2.29 |
| **QRS is like Qr-complex** | 3 | 34 | 2.0 |
| **QRS is like QS-complex** | 4 | 492 | 28.94 |
| **Total** | | 1700 | 100.00 |

Figure 4.30 displays the percentage of patients having their ECG changes in leads V1 – V4 with different anterior myocardial infarction in left ventricular and were suffering from fatal outcome because of an acute myocardial infarction.

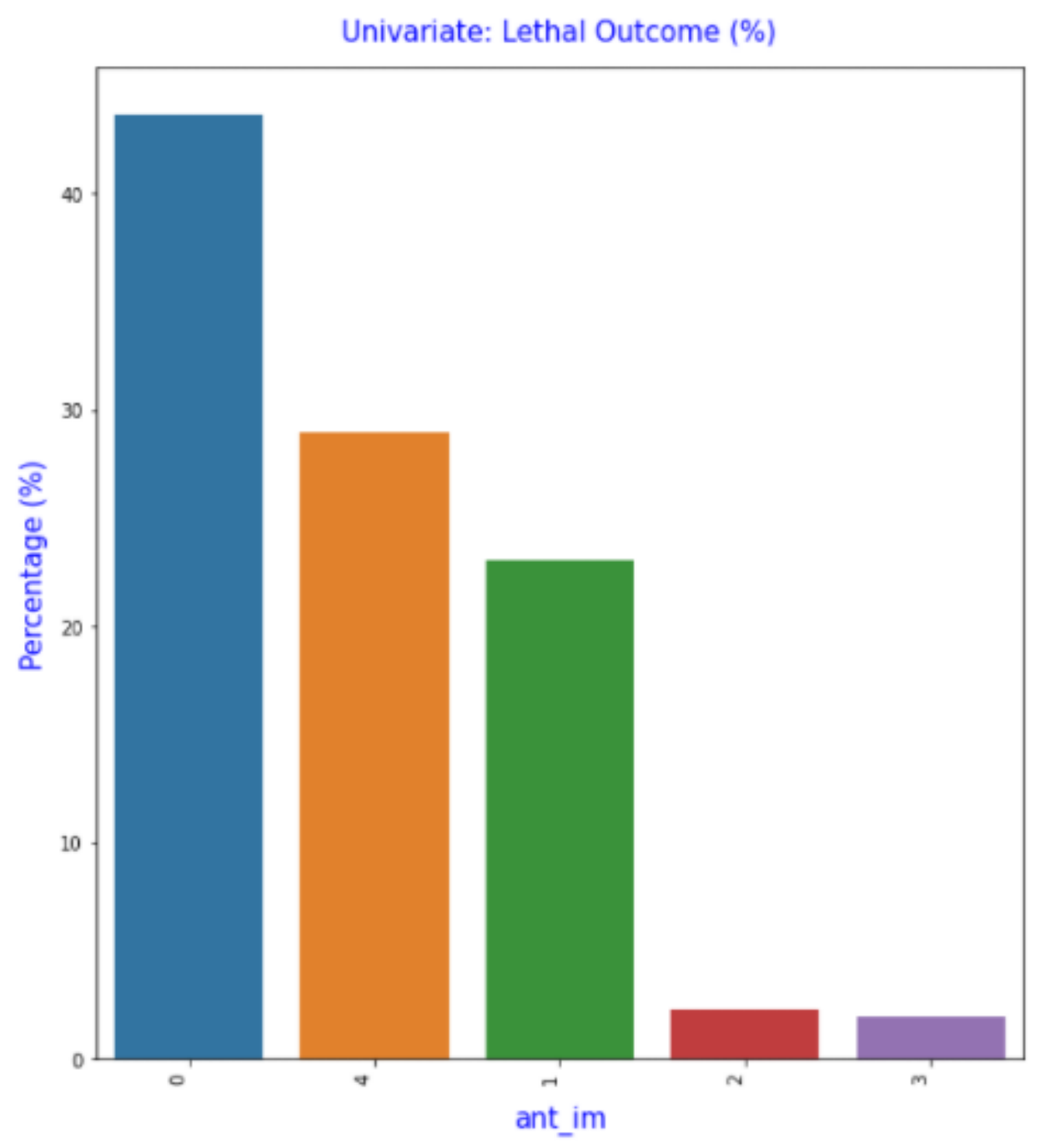

*Figure 4.30 Percentage plot of ant_im feature*

**t) TIME_B_S (Time elapsed from the beginning of the attack of CHD to the hospital)**

As it can be observed from below table 4.30, a total of 1,700 observations have been reported for patients suffering from lethal outcome of acute myocardial infarction. Among 1,700 patients, 198 patients or 11.65% of them having elapsed less than 2 hours from the beginning of the attack of coronary heart disease admitted to the hospital, while 486 patients or 28.59% of them having elapsed between 2-4 hours from the beginning of the attack of coronary heart disease admitted to the hospital, another 175 patients or 10.29% of them having elapsed between 4-6 hours from the beginning of the attack of coronary heart disease admitted to the hospital, 87 patients or 5.12% of them suffering from lethal outcomes of this disease having elapsed between 6-8 hours from the beginning of the attack of coronary heart disease admitted to the hospital, another 92 patients or 5.41% of them suffering from lethal outcomes of this disease having elapsed between 8-12 hours from the beginning of the attack of coronary heart disease admitted to the hospital, 151 patients or 8.88% of them suffering from lethal outcomes of this disease having elapsed between 12-24 hours from the beginning of the attack of coronary heart disease admitted to the hospital, 101 patients or 5.94% of them suffering from lethal outcomes of this disease having elapsed more than 2 days from the beginning of the attack of coronary heart disease admitted to the hospital, 141 patients or 8.29% of them suffering from lethal outcomes of this disease having elapsed more than 1 day from the beginning of the attack of coronary heart disease admitted to the hospital and rest 296 patients or 15.82% of them suffering from lethal outcomes of this disease having elapsed more than 3 days from the beginning of the attack of coronary heart disease admitted to the hospital.

*Table 4.32 Frequency and Percentage analysis of TIME_B_S feature*

| TIME_B_S | | | |
|---|---|---|---|
| | **Decoded Value** | **Frequency/Count** | **Percentage** |
| **less than 2 hours** | 1 | 198 | 11.65 |
| **2-4 hours** | 2 | 486 | 28.59 |
| **4-6 hours** | 3 | 175 | 10.29 |
| **6-8 hours** | 4 | 87 | 5.12 |
| **8-12 hours** | 5 | 92 | 5.41 |
| **12-24 hours** | 6 | 151 | 8.88 |
| **more than 1 days** | 7 | 141 | 8.29 |
| **more than 2 days** | 8 | 101 | 5.94 |
| **more than 3 days** | 9 | 269 | 15.82 |
| **Total** | | 1700 | 100.00 |

Figure 4.31 displays the percentage of patients having time elapsed from the beginning of the attack of CHD admitted to the hospital and were suffering from fatal outcome because of an acute myocardial infarction.

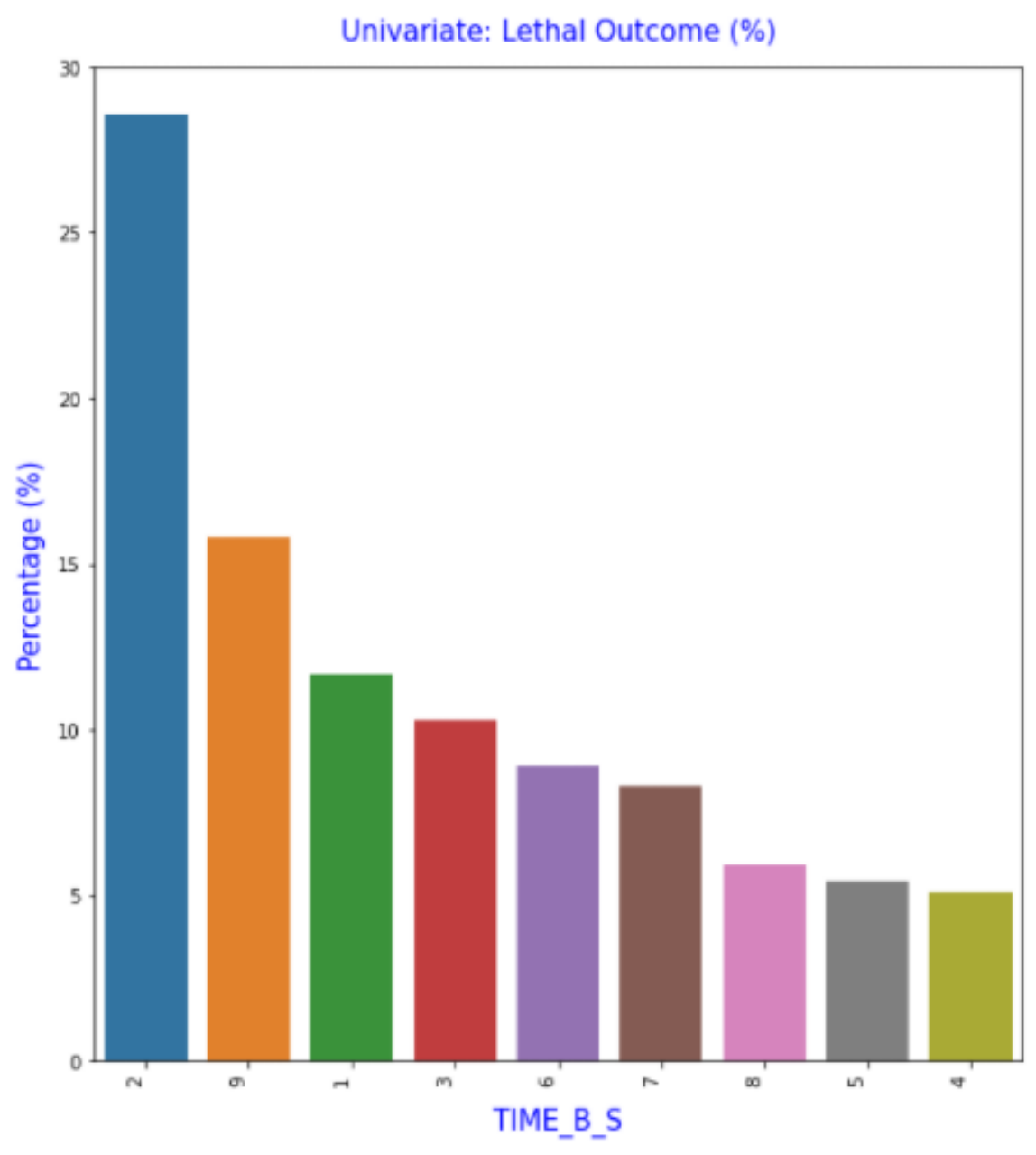

*Figure 4.31 Percentage plot of TIME_B_S feature*

### u) AGE_Interval (Patient's age groups)

As it can be observed from below table 4.31, a total of 1,700 observations have been reported for patients suffering from lethal outcome of acute myocardial infarction. Among 1,700 patients, 0 patients or 0.00% of them having age between 10-20, while 4 patients or 0.24% of them having age between 20-30, another 58 patients or 3.41% of them having age between 30-40, 197 patients or 11.59% of them suffering from lethal outcomes of this disease having age between 40-50, another 446 patients or 26.24% of them suffering from lethal outcomes of this disease having age between 50-60, 622 patients or 36.59% of them suffering from lethal outcomes of this disease having age between 60-70, 305 patients or 17.94% of them suffering from lethal outcomes of this disease having age between 70-80, 66 patients or 3.88% of them suffering from lethal outcomes of this disease having age between 80-90 and rest 2 patients or 0.12% of them suffering from lethal outcomes of this disease having age more than 90.

*Table 4.33 Frequency and Percentage analysis of AGE_Interval feature*

| AGE_Interval | | |
|---|---|---|
| | **Frequency/Count** | **Percentage** |
| **10-20** | 0 | 0.0 |
| **20-30** | 4 | 0.24 |
| **30-40** | 58 | 3.41 |
| **40-50** | 197 | 11.59 |
| **50-60** | 446 | 26.24 |
| **60-70** | 622 | 36.59 |
| **70-80** | 305 | 17.94 |
| **80-90** | 66 | 3.88 |
| **90-above** | 2 | 0.12 |
| **Total** | 1700 | 100.00 |

From the above analysis it can been seen that for age group 60-70 years a greater number of people were suffering from lethal outcomes of acute myocardial infarction, also between age group 20-30 years and above 90 years very less people were suffering from lethal outcomes of acute myocardial infarction.

Figure 4.32 displays the percentage of patient within each age group those were suffering from fatal outcome because of an acute myocardial infarction.

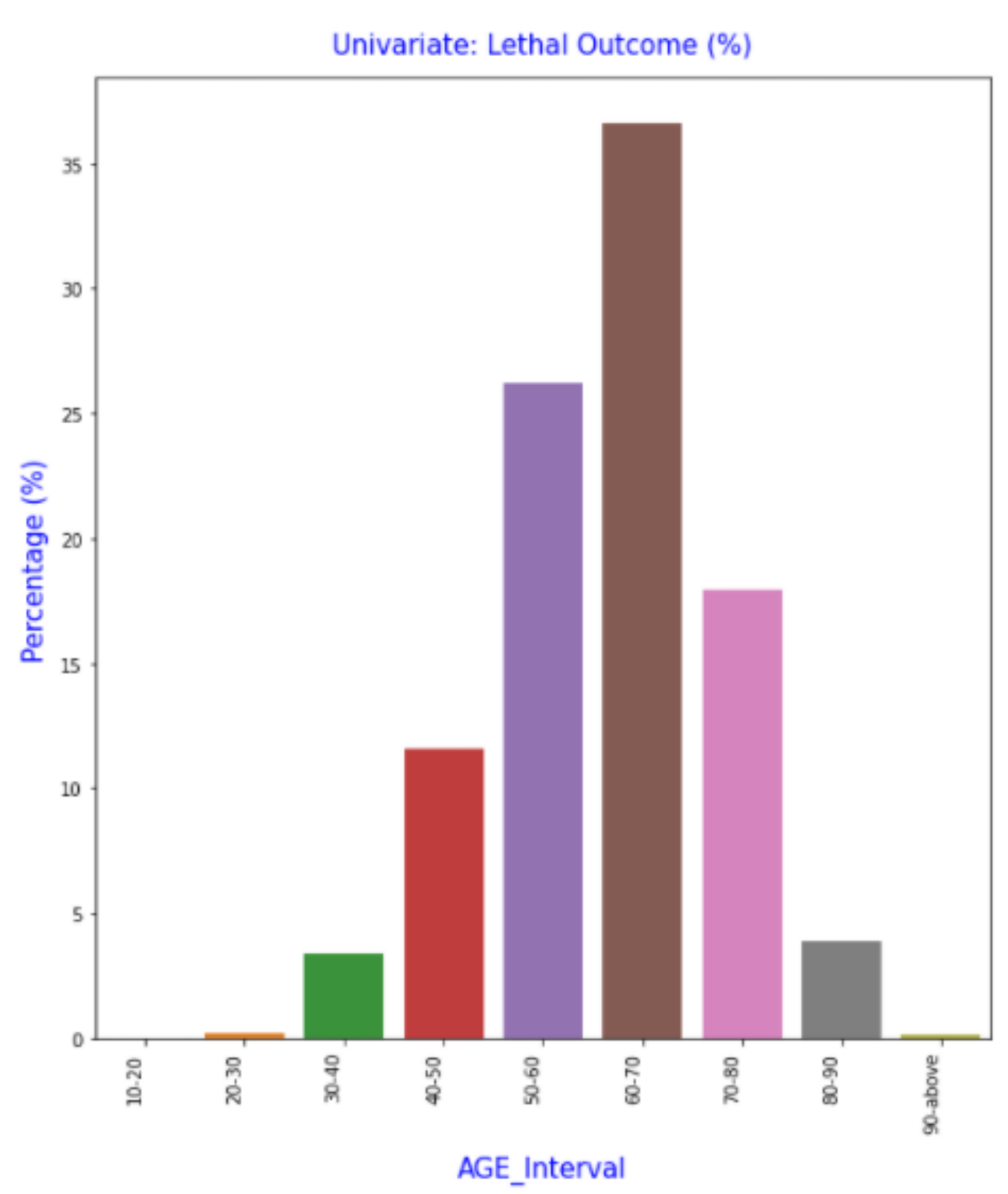

*Figure 4.32 Percentage plot of AGE_Interval feature*

**v) S_AD_ORIT_Interval (Patient's systolic blood pressure groups)**

As it can be observed from below table 4.32, a total of 1,700 observations have been reported for patients suffering from lethal outcome of acute myocardial infarction. Among 1,700 patients, 1 patients or 0.06% of them having systolic blood pressure between 0-30, while 28 patients or 1.65% of them having systolic blood pressure between 30-60, another 69 patients or 4.08% of them having systolic blood pressure between 60-90, 417 patients or 24.65% of them suffering from lethal outcomes of this disease having systolic blood pressure between 90-120, another 827 patients or 48.88% of them suffering from lethal outcomes of this disease having systolic blood pressure between 120-150, 277 patients or 16.37% of them suffering from lethal outcomes of this disease having systolic blood pressure between 150-180, 60 patients or 3.55% of them suffering from lethal outcomes of this disease having systolic blood pressure between 180-210, 12 patients or 0.71% of them suffering from lethal outcomes of this disease having systolic blood pressure between 210-240 and rest 1 patients or 0.06% of them suffering from lethal outcomes of this disease having systolic blood pressure more than 240.

*Table 4.34 Frequency and Percentage analysis of S_AD_ORIT_Interval feature*

| S_AD_ORIT_Interval | | |
|---|---|---|
| | **Frequency/Count** | **Percentage** |
| **0-30** | 1 | 0.06 |
| **30-60** | 28 | 1.65 |
| **60-90** | 69 | 4.08 |
| **90-120** | 417 | 24.65 |
| **120-150** | 827 | 48.88 |
| **150-180** | 277 | 16.37 |
| **180-210** | 60 | 3.55 |
| **210-240** | 12 | 0.71 |
| **240-above** | 1 | 0.06 |
| **Total** | 1700 | 100.00 |

From the above analysis it can been seen that for the systolic blood pressure between 120-150 mmHg a greater number of people is suffering from lethal outcomes of acute myocardial infarction, while for the systolic blood pressure between 0-30 mmHg and above 240 mmHg very less number of people is suffering from lethal outcomes of acute myocardial infarction

Figure 4.33 displays the percentage of patient within each systolic blood pressure group those were suffering from fatal outcome because of an acute myocardial infarction.

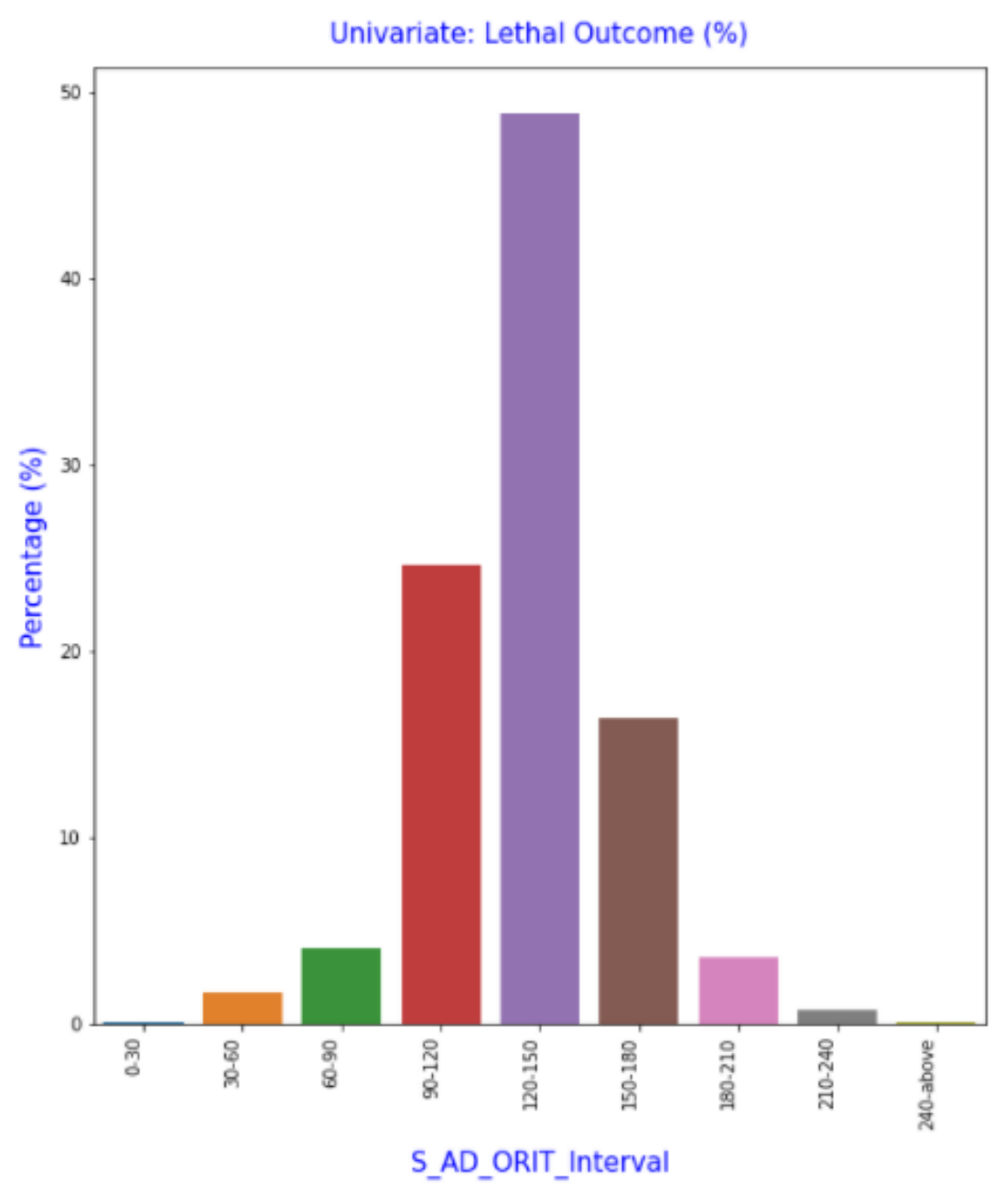

*Figure 4.33 Percentage plot of S_AD_ORIT_Interval feature*

**w) D_AD_ORIT_Interval (Patient's diastolic blood pressure groups)**

As it can be observed from below table 4.33, a total of 1,700 observations have been reported for patients suffering from lethal outcome of acute myocardial infarction. Among 1,700 patients, 4 patients or 0.24% of them having diastolic blood pressure between 0-20, while 29 patients or 1.72% of them having diastolic blood pressure between 20-40, another 100 patients or 5.95% of them having diastolic blood pressure between 40-60, 727 patients or 43.22% of them suffering from lethal outcomes of this disease having diastolic blood pressure between 60-80, another 725 patients or 43.1% of them suffering from lethal outcomes of this disease having diastolic blood pressure between 80-100, 84 patients or 4.99% of them suffering from lethal outcomes of this disease having diastolic blood pressure between 100-120, 12 patients or 0.71% of them suffering from lethal outcomes of this disease having diastolic blood pressure between 120-140, 0 patients or 0.00% of them suffering from lethal outcomes of this disease having diastolic blood pressure between 140-160, 0 patients or 0.00% of them suffering from lethal outcomes of this disease having diastolic blood pressure between 160-180, 1 patients or 0.06% of them suffering from lethal outcomes of this disease having diastolic blood pressure between 180-200 and rest 0 patients or 0.00% of them suffering from lethal outcomes of this disease having diastolic blood pressure more than 200.

*Table 4.35 Frequency and Percentage analysis of D_AD_ORIT_Interval feature*

| D_AD_ORIT_Interval | | |
|---|---|---|
| | **Frequency/Count** | **Percentage** |
| **0-20** | 4 | 0.24 |
| **20-40** | 29 | 1.72 |
| **40-60** | 100 | 5.95 |
| **60-80** | 727 | 43.22 |
| **80-100** | 725 | 43.1 |
| **100-120** | 84 | 4.99 |
| **120-140** | 12 | 0.71 |
| **140-160** | 0 | 0.00 |
| **160-180** | 0 | 0.00 |
| **180-200** | 1 | 0.06 |
| **200-220** | 0 | 0.00 |
| **220-above** | 0 | 0.00 |
| **Total** | 1700 | 100.00 |

From the above analysis it can been seen that for the diastolic blood pressure between 60-80 mmHg a greater number of people is suffering from lethal outcomes of acute myocardial infarction, while for the diastolic blood pressure between 0-20 mmHg and above 140 mmHg very less number of people is suffering from lethal outcomes of acute myocardial infarction.

Figure 4.34 displays the percentage of patient within each systolic blood pressure group those were suffering from fatal outcome because of an acute myocardial infarction.

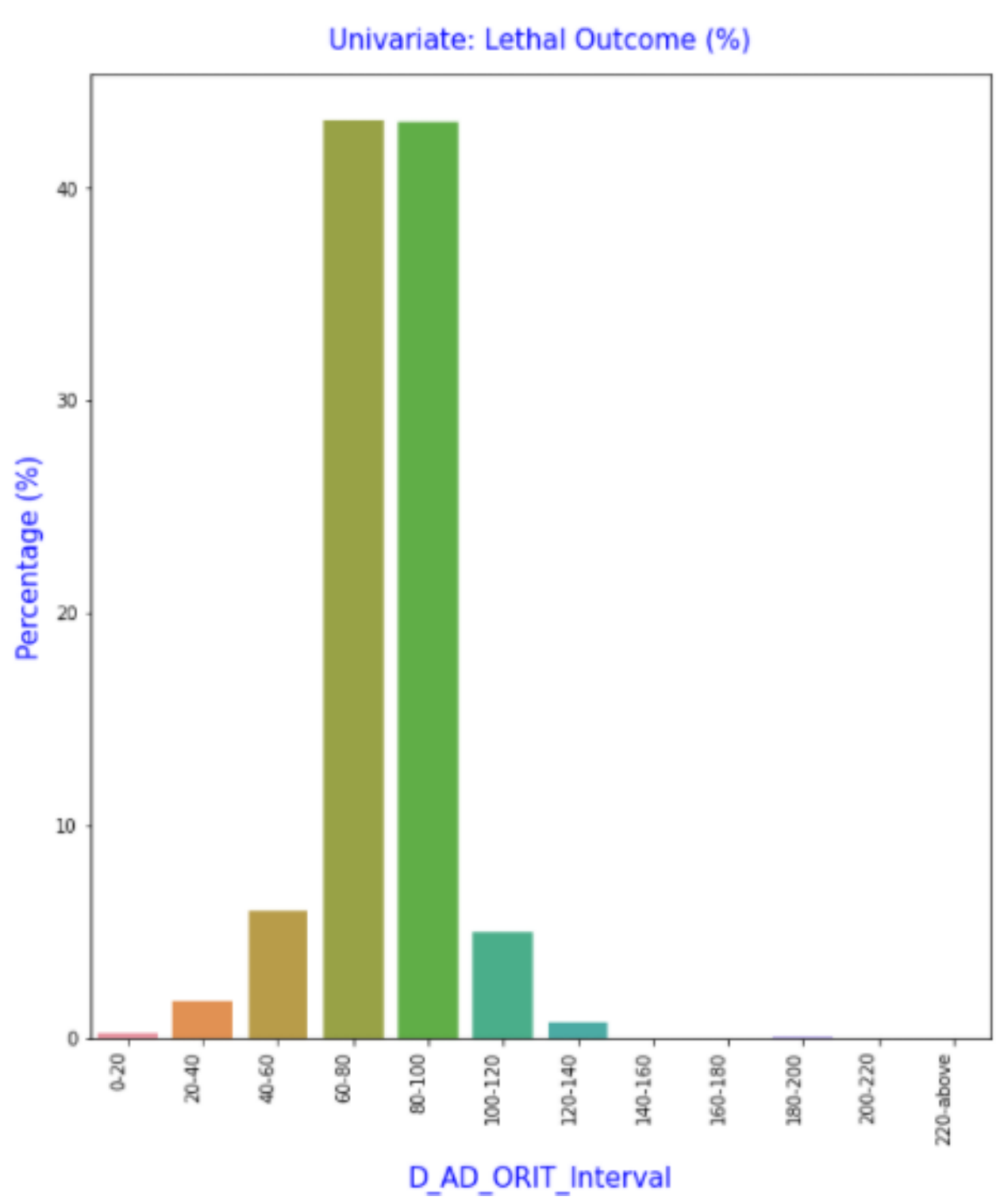

*Figure 4.34 Percentage plot of D_AD_ORIT_Interval feature*

#### 4.4.1.2 Outlier Analysis and Treatment

An outlier is a value in a randomized sampling from a population that is significantly different or noisy from the other values. In this study outlier analysis has been done on some of the important numerical features and based on its significance those outliers have been treated as well. Find the below analysis of the outliers and treatment for some of the important features.

**a) AGE (Patient's Age)**

Figure 4.35 shows the AGE feature with some of the outliers below the lower fence marked in yellow. However, the summary statistic for the AGE feature in table 4.11, shows that the minimum value of age is 26 years, indicating that this outlier for the AGE feature is genuine and may not require further treatment.

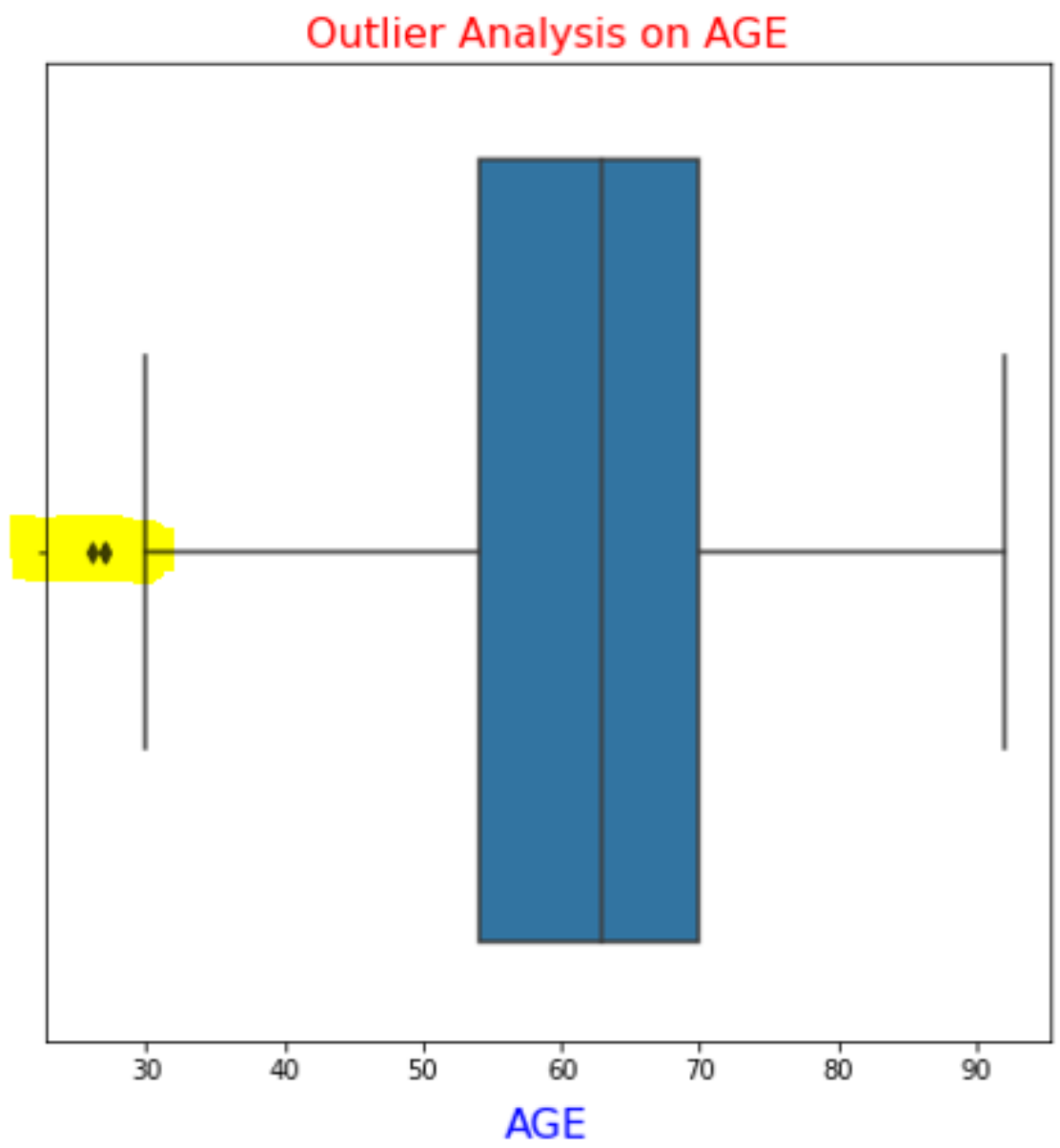

*Figure 4.35 Boxplot for AGE feature*

**b) ALT_BLOOD (Serum AlAT content)**

Figure 4.36 shows the ALT_BLOOD feature with some of the outliers above the upper fence marked in yellow. However, the summary statistic for the ALT_BLOOD feature in table 4.15, shows that the maximum value of serum AlAT content is 3.00 IU/L, indicating that this outlier for the ALT_BLOOD feature is genuine and may not require further treatment.

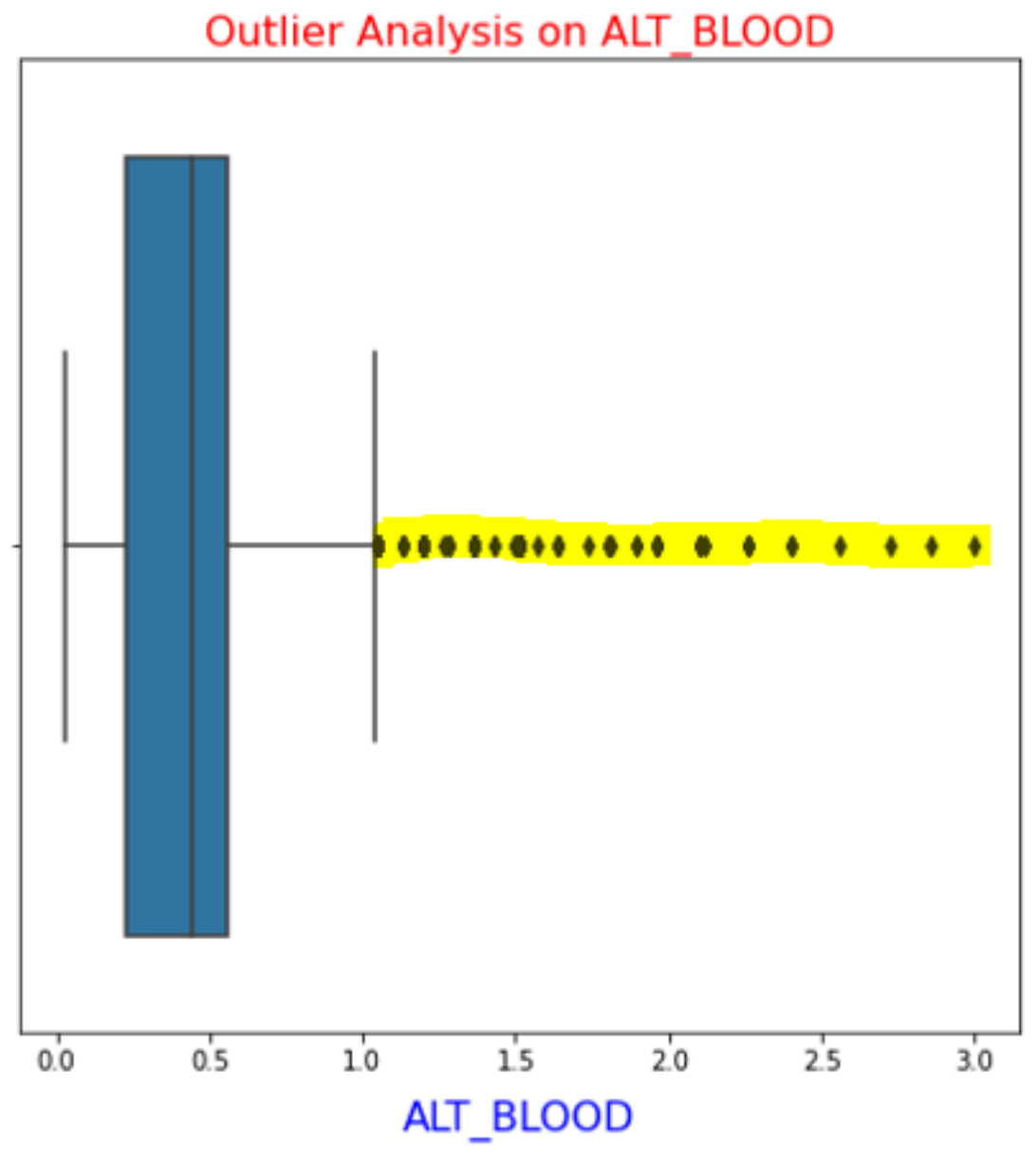

*Figure 4.36 Boxplot for ALT_BLOOD feature*

**c) K_BLOOD (Serum potassium content)**

Figure 4.37 shows the K_BLOOD feature with some of the outliers both above the upper fence and below the lower fence which are marked in yellow. However, the summary statistic for the K_BLOOD feature in table 4.14, shows that the maximum value of serum potassium content is 8.20 mmol/L, and the minimum value of serum potassium content is 2.30 mmol/L, indicating that this outlier for the K_BLOOD feature is genuine and may not require further treatment.

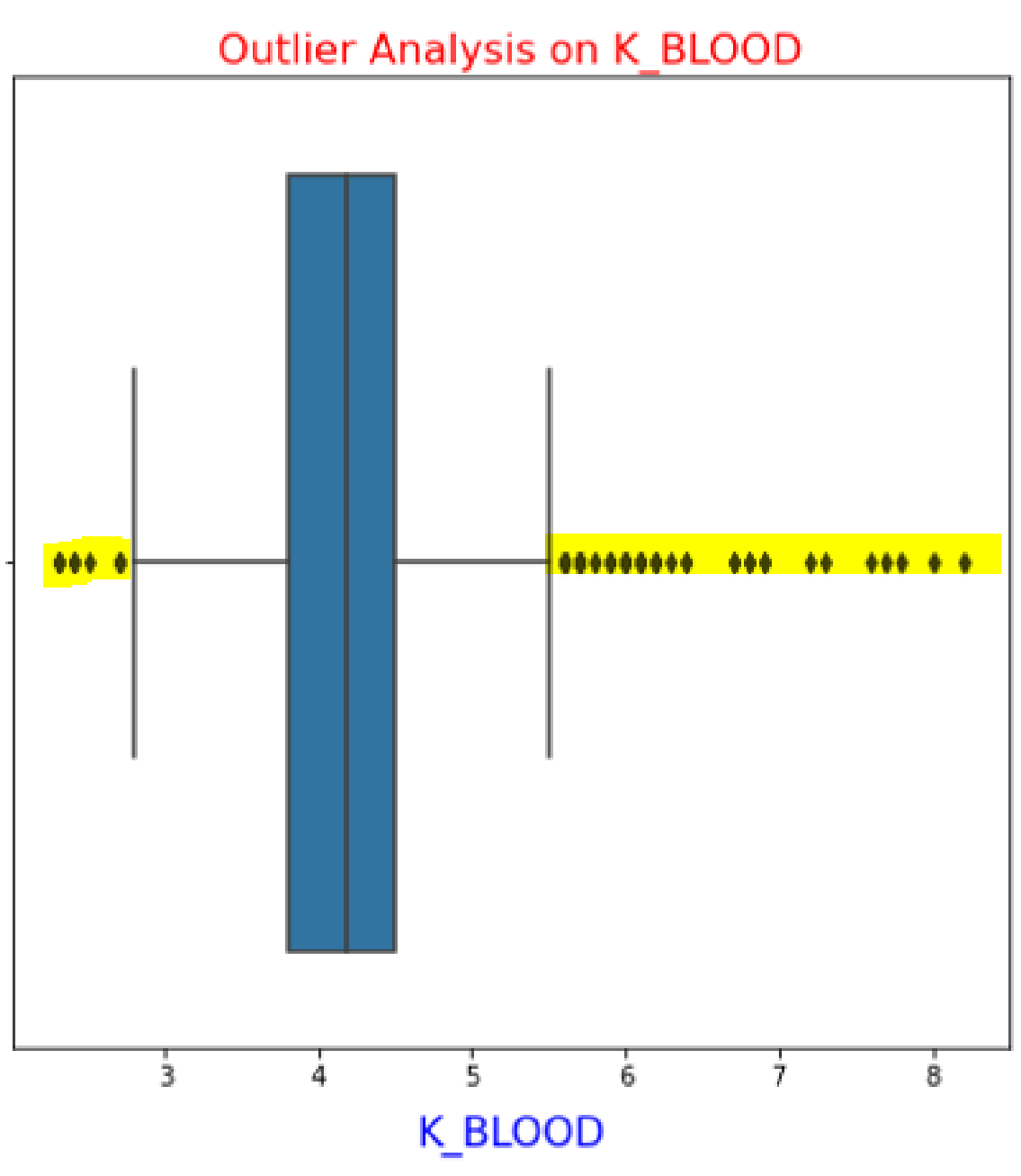

*Figure 4.37 Boxplot for K_BLOOD feature*

**d) S_AD_ORIT (Systolic blood pressure according to ICU)**

Figure 4.38 shows the S_AD_ORIT feature with some of the outliers both above the upper fence and below the lower fence which are marked in yellow. However, the summary statistic for the S_AD_ORIT feature in table 4.12, shows that the maximum value of systolic blood pressure is 260.00 mmHg, and the minimum value of systolic blood pressure is 0.00 mmHg reported as per ICU, however minimum systolic blood pressure for a patient currently under treatment in ICU cannot be 0.00 mmHg, indicating that outliers below the lower fence need treatment for the S_AD_ORIT feature.

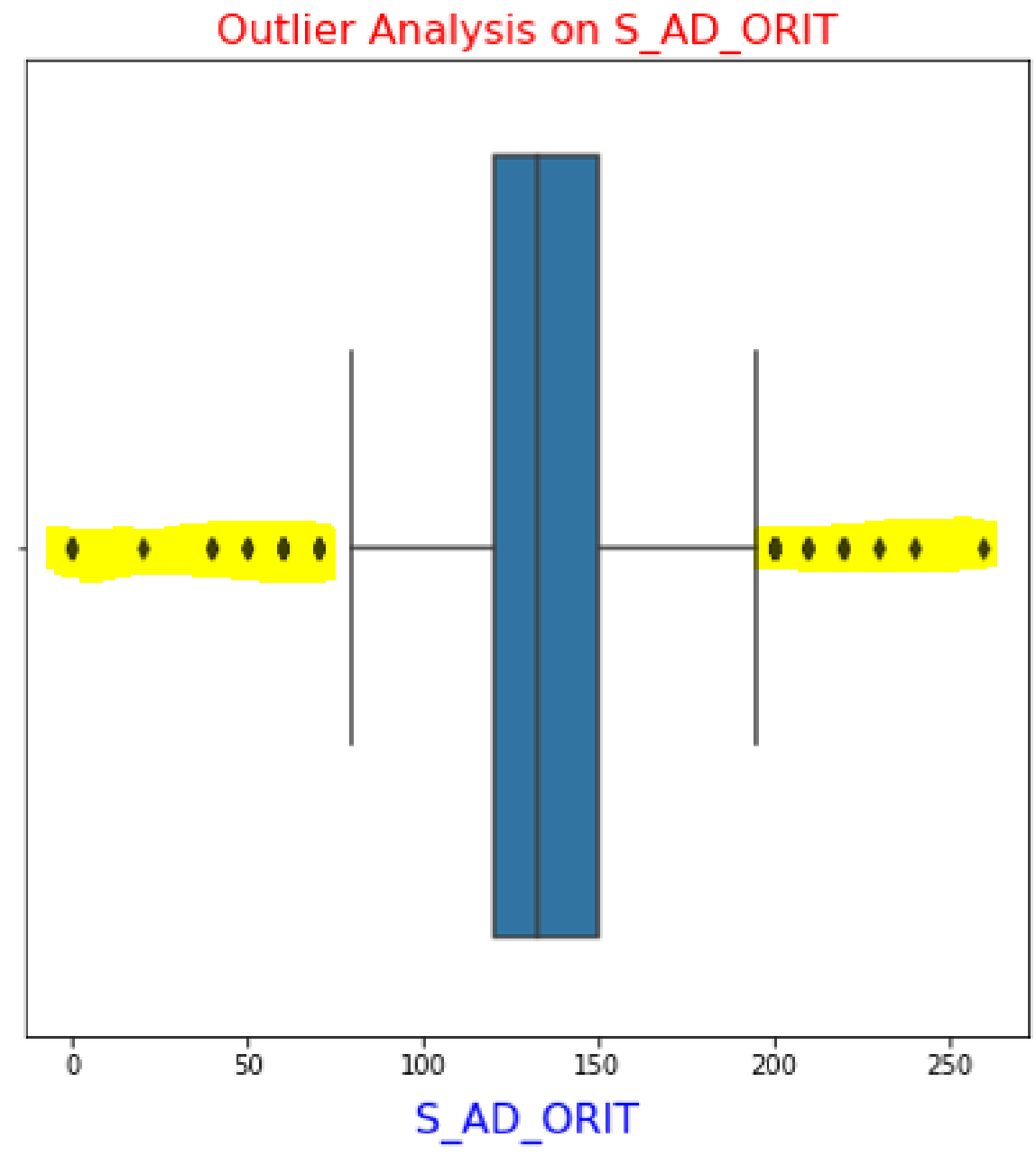

*Figure 4.38 Boxplot for S_AD_ORIT feature before outlier treatment*

For the outliers below the lower fence for feature S_AD_ORIT was treated by hard capping, means all the value below 5 percentiles for this feature was capped with the value of 5 percentile itself. Figure 4.39 shows the boxplot of the feature S_AD_ORIT after the outlier treatment below the lower fence.

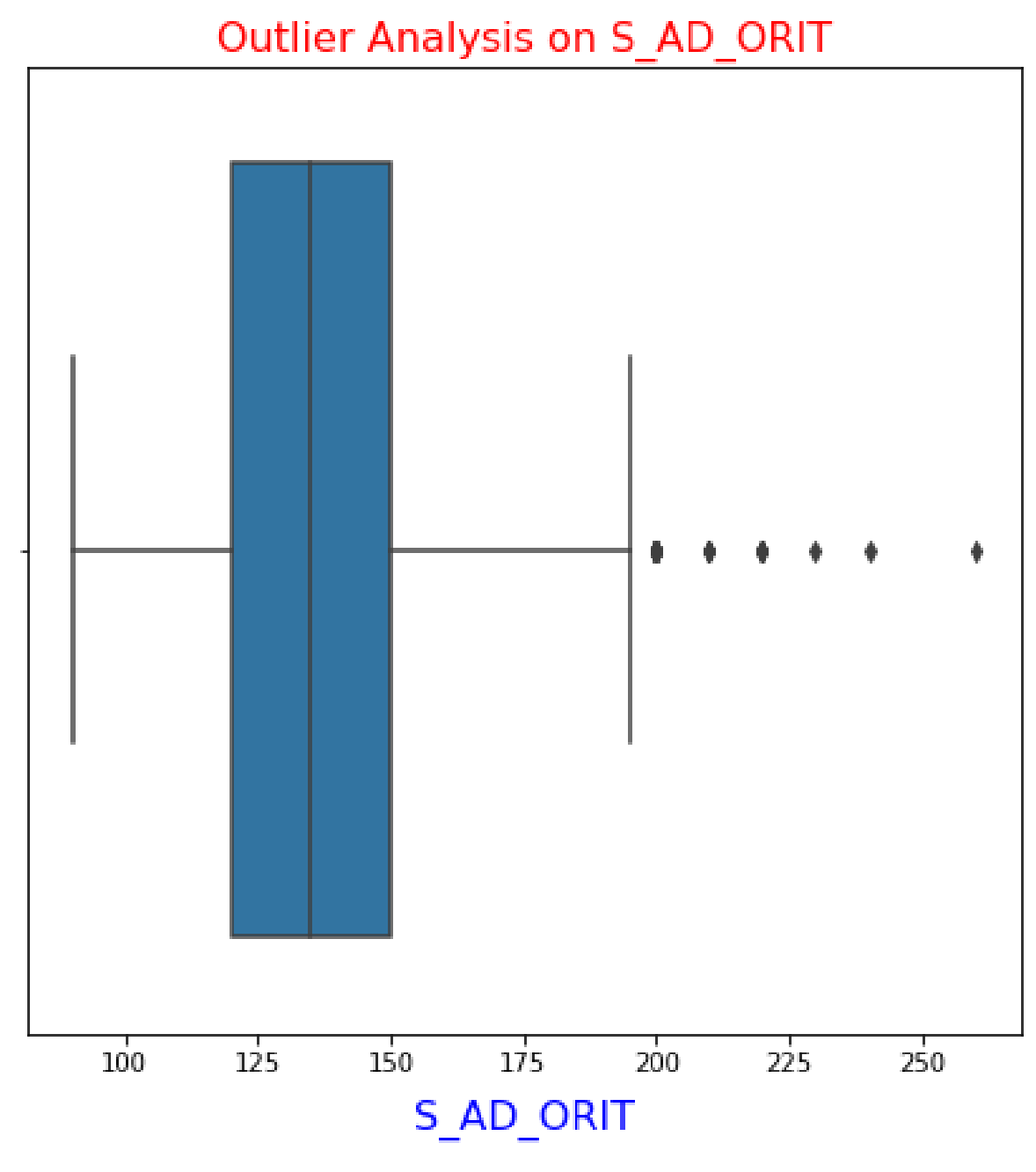

*Figure 4.39 Boxplot for S_AD_ORIT feature after outlier treatment*

**e) D_AD_ORIT (Diastolic blood pressure according to ICU)**

Figure 4.40 shows the D_AD_ORIT feature with some of the outliers both above the upper fence and below the lower fence which are marked in yellow. However, the summary statistic for the D_AD_ORIT feature in table 4.13, shows that the maximum value of diastolic blood pressure is 190.00 mmHg, and the minimum value of diastolic blood pressure is 0.00 mmHg reported as per ICU, however minimum diastolic blood pressure for a patient currently under treatment in ICU cannot be 0.00 mmHg, indicating that outliers below the lower fence need treatment for the D_AD_ORIT feature.

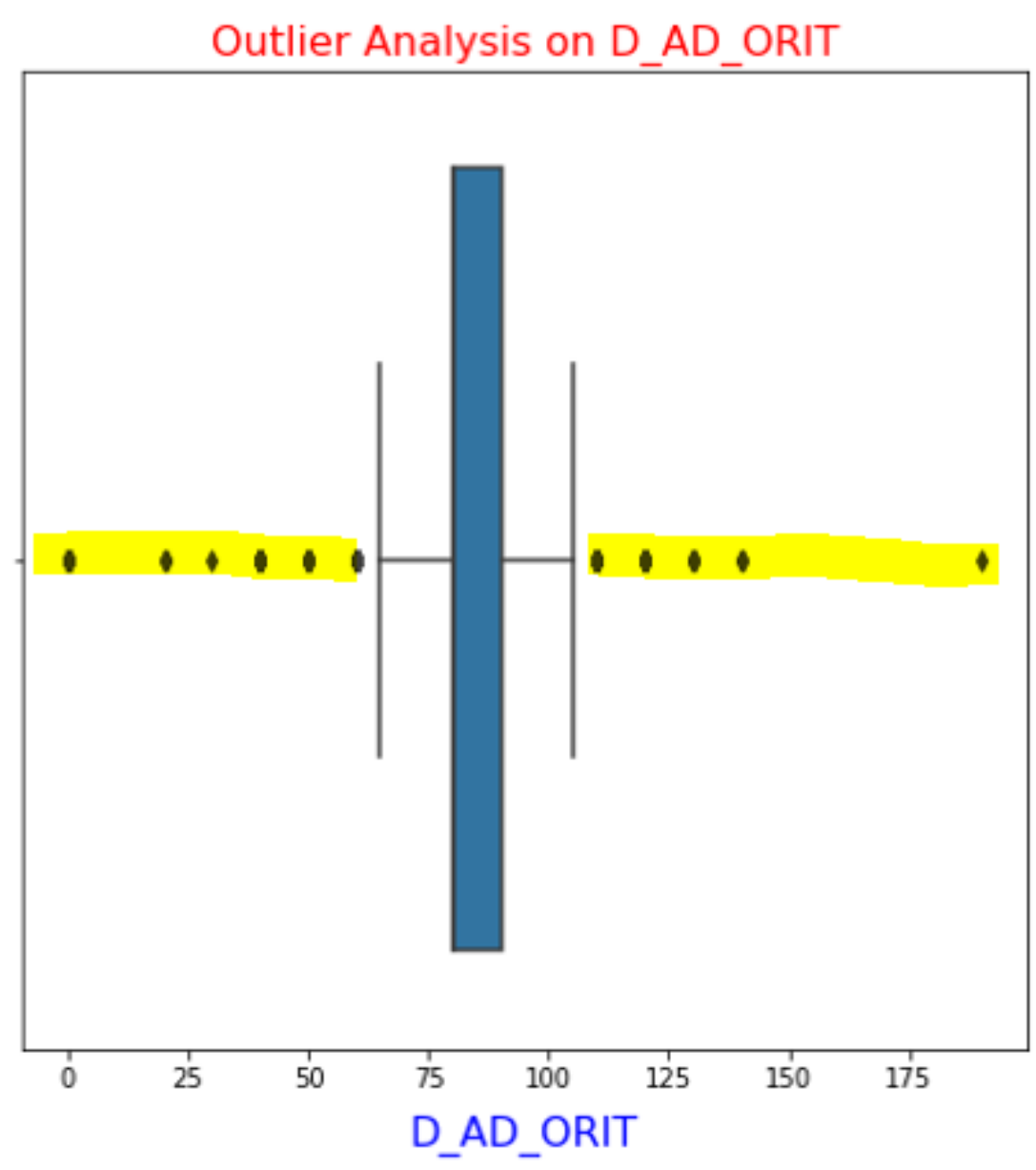

*Figure 4.40 Boxplot for D_AD_ORIT feature before outlier treatment*

For the outliers below the lower fence for feature D_AD_ORIT was treated by hard capping, means all the value below 5 percentiles for this feature was capped with the value of 5 percentile itself. Figure 4.41 shows the boxplot of the feature D_AD_ORIT after the outlier treatment below the lower fence.

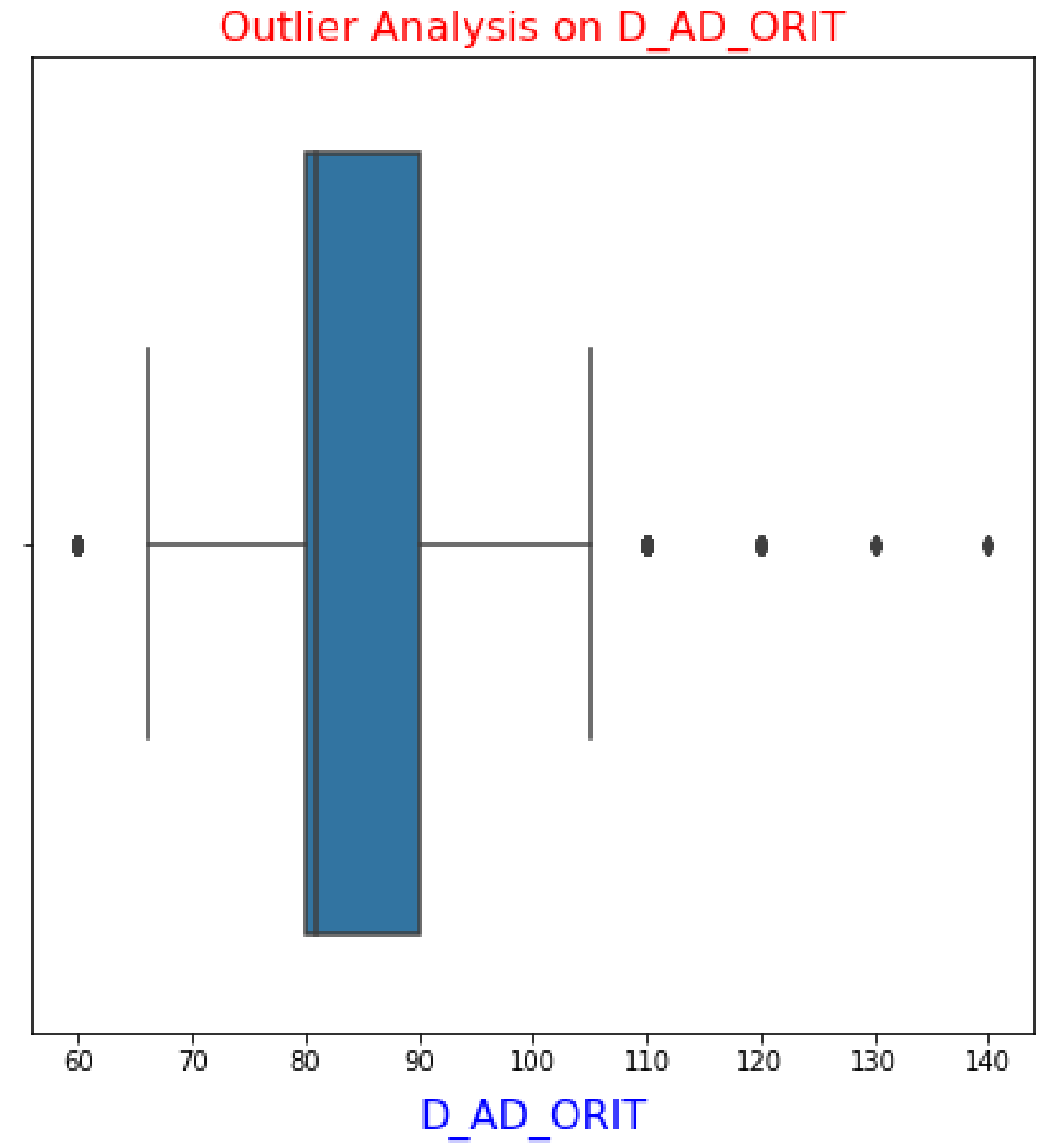

*Figure 4.41 Boxplot for D_AD_ORIT feature after outlier treatment*

### 4.4.2 Bivariate Analysis

Because 'bi' means two and variate means variable, there are two factors involved here. The investigation is focused on the reason and the relationship between the two factors. Bivariate analysis is classified into three categories. Let's discuss each category in detail in below sub-sections.

#### 4.4.2.1 Numerical vs Categorical Analysis

This section will go through different bivariate visualization for some significant independent numerical and categorical characteristics, using bar plot. This section discusses a couple of the most essential biomarkers among all the predictive indicators. Below analysis focuses mostly on the target variable (LET_IS) as a categorical feature versus some essential numerical features.

**a) K_BLOOD (Serum potassium content)**

From the below figure 4.42, the average serum potassium content for the patients having progress of congestive heart failure (Target class 4) is more than 4.30 mmol/L which is greater than any other target class potassium content. Also, the average serum potassium content for the patients having pulmonary edema (Target class 2) is quite less than 4.00 mmol/L which is less compared to any other target class potassium content.

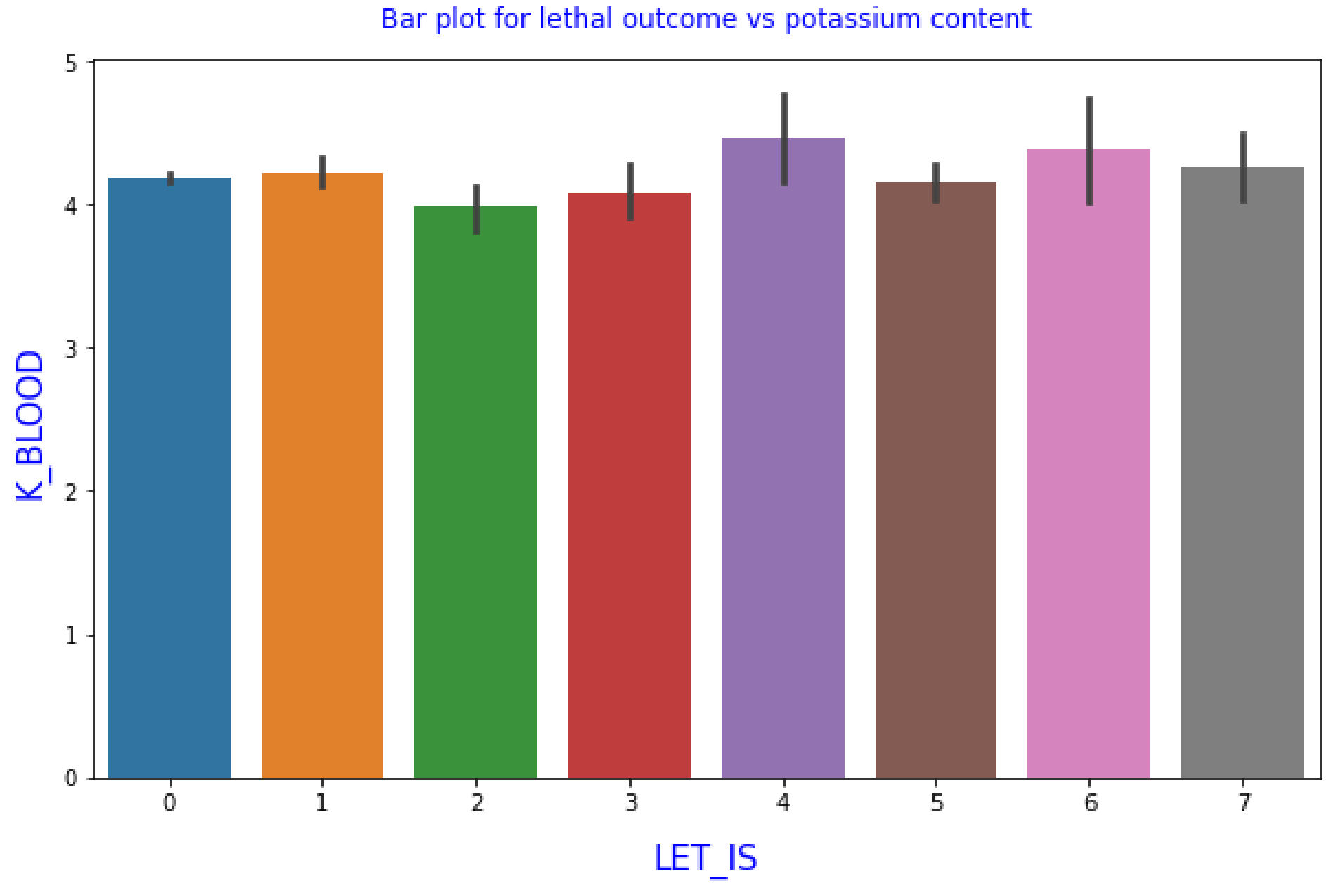

*Figure 4.42 Barplot between K_BLOOD vs LET_IS (target)*

From the below figure 4.43, serum potassium content for all patients suffering from fatal outcomes of acute myocardial infarction is not able to separate each target class distinctively as all the classes are overlapped with each other, hence only serum potassium content cannot be a good predictor for any of the target classes.

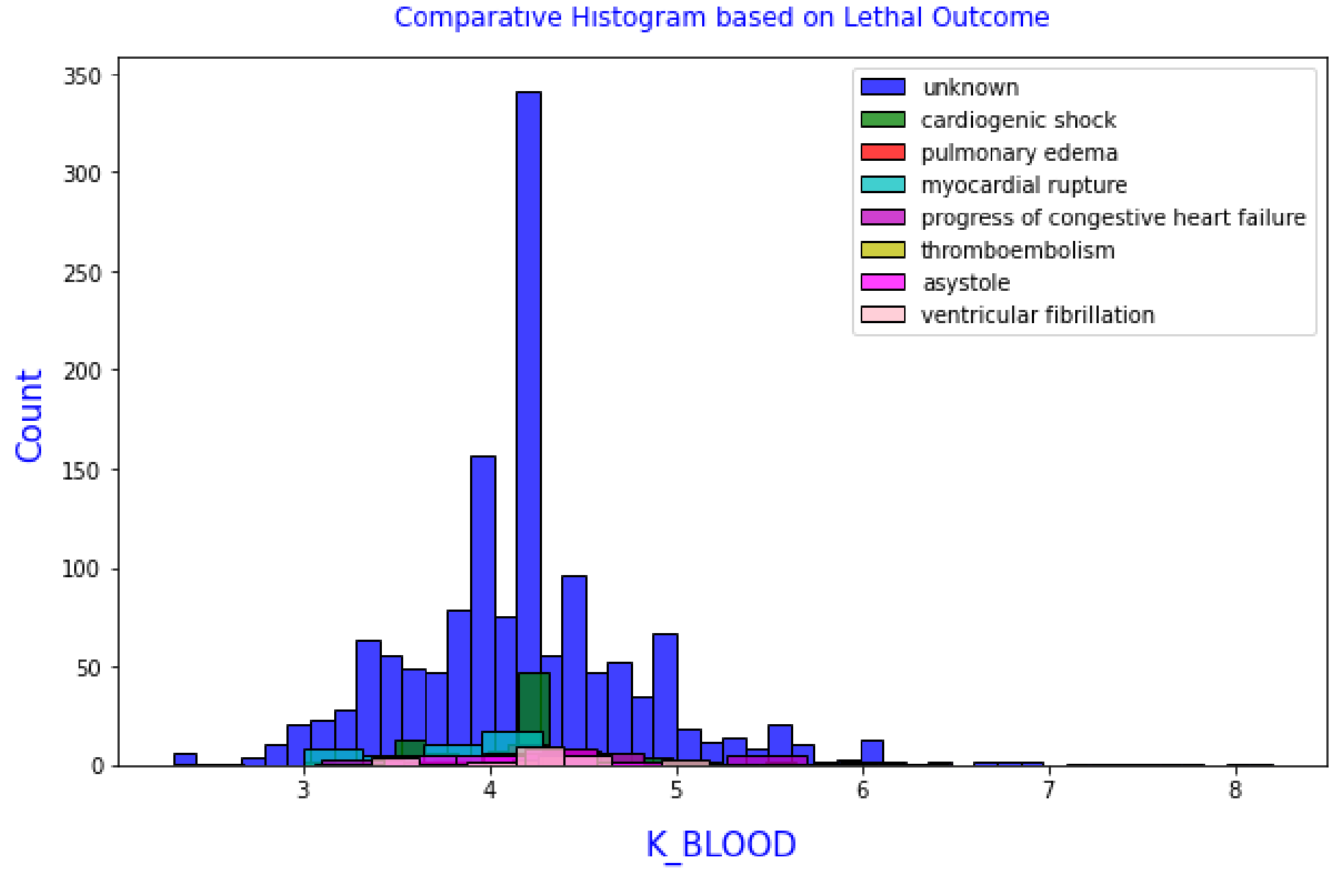

*Figure 4.43 Comparative histogram between K_BLOOD vs LET_IS (all target class)*

**b) ALT_BLOOD (Serum AlAT content)**

From the below figure 4.44, the average serum AlAT content for the patients having progress of congestive heart failure (Target class 4) is more than 0.68 IU/L which is distinctively greater than any other target class A1AT serum content, hence high serum A1AT content might be a good indicator of progression towards congestive heart failure. Also, the average serum AlAT content for the patients having myocardial rupture (Target class 3) is less than 0.4 IU/L which is smallest compared to any other target class A1AT serum content.

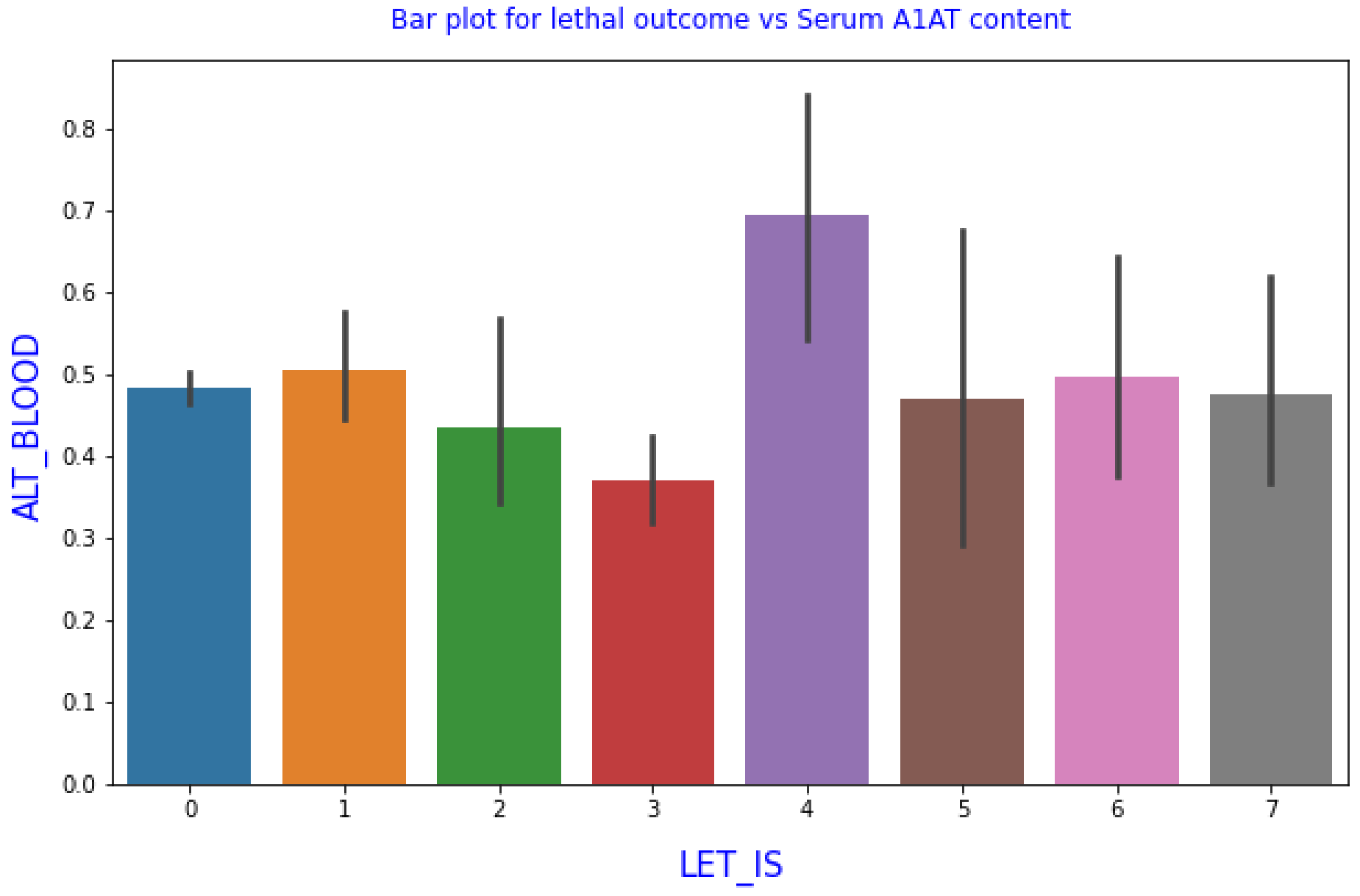

*Figure 4.44 Barplot between ALT_BLOOD vs LET_IS (target)*

Also, from the below figure 4.45, serum AlAT content for all patients suffering from fatal outcomes of acute myocardial infarction is not able to separate each target class distinctively as all the classes are overlapped with each other, hence only serum AlAT content cannot be a good predictor for any of the target classes.

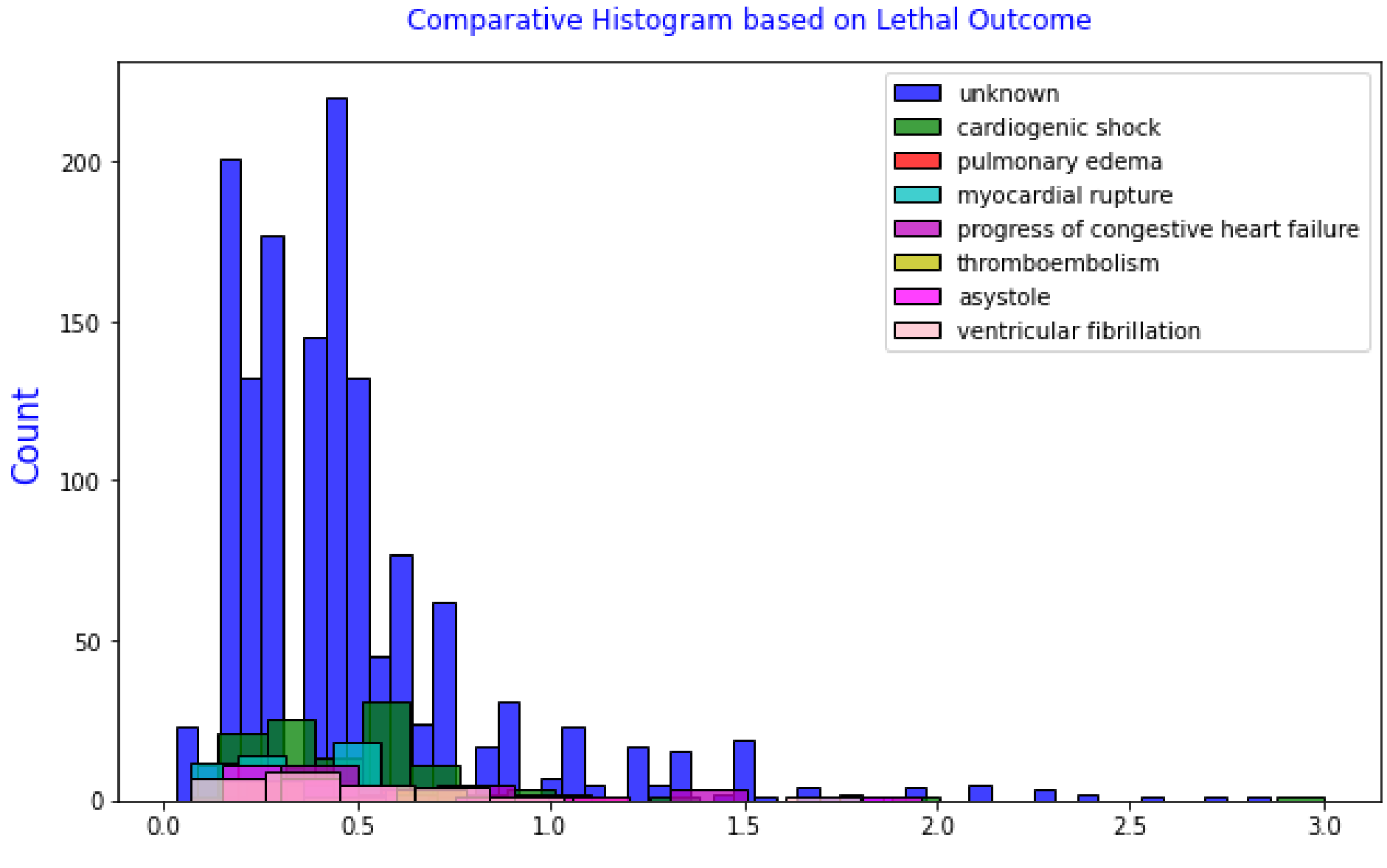

*Figure 4.45 Comparative histogram between ALT_BLOOD vs LET_IS (all target class)*

### c) AGE (Patient's Age)

From the below figure 4.46, the average age for the patients having pulmonary edema (Target class 2), myocardial rupture (Target class 3), progress of congestive heart failure (Target class 4) and asystole (Target class 6) are around 70 years of old which is greater than rest of the target class age. Also, the average age for the patients having ventricular fibrillation (Target Class 7) and unknown cause of acute myocardial infarction (Target class 0) are around 60 years of old which is less compared to any other target class age.

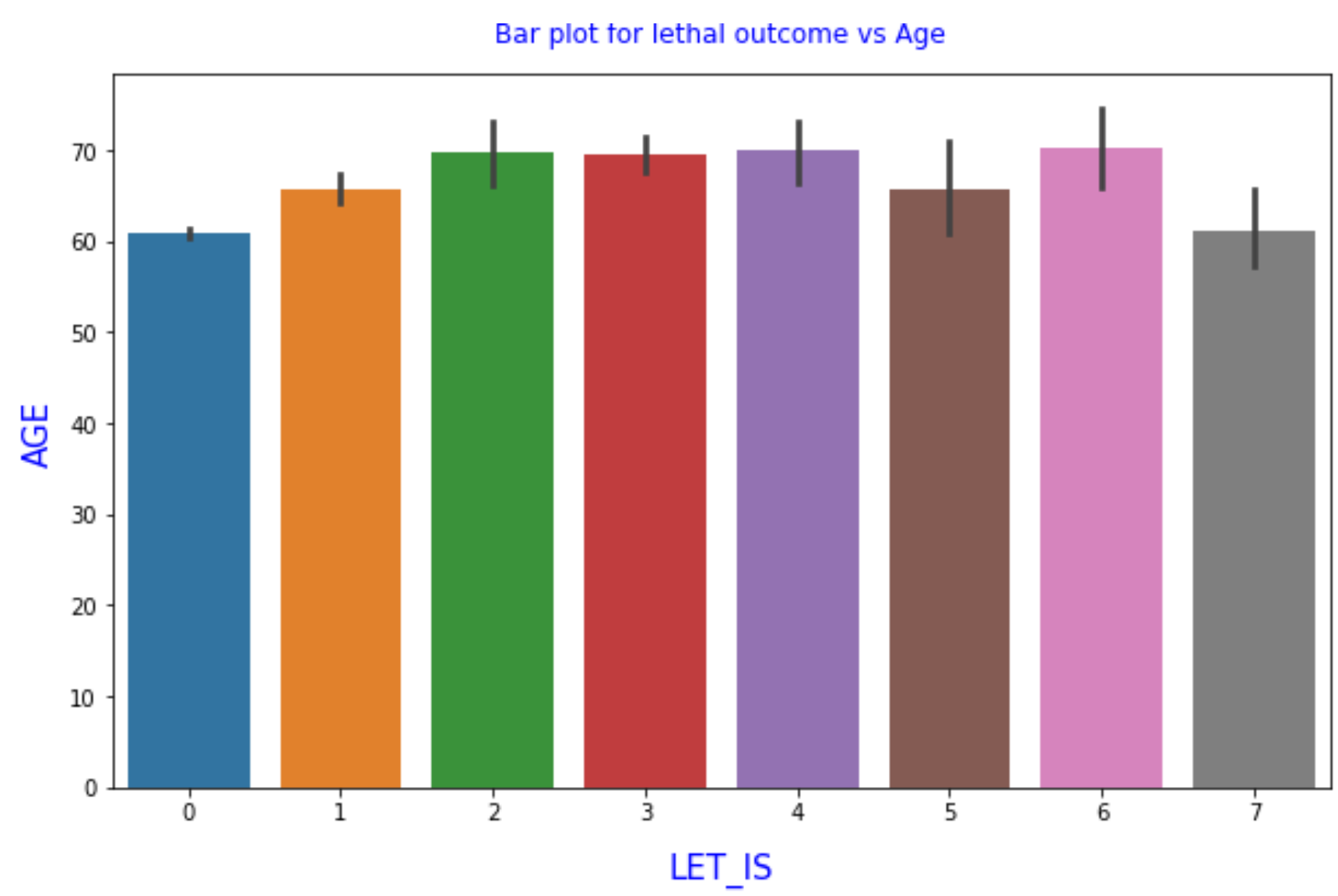

*Figure 4.46 Barplot between AGE vs LET_IS (target)*

Also, from the below figure 4.47, ages for all patients suffering from fatal outcomes of acute myocardial infarction is not able to separate each target class distinctively as all the classes are overlapped with each other, but it can be seen that most of the patients whose ages are from 25 to 42 years of age mostly suffered lethal outcomes of this disease for unknown reasons (Target class 0), hence age cannot be a good predictor for other target class expects unknow class but only from 25 to 42 years of age.

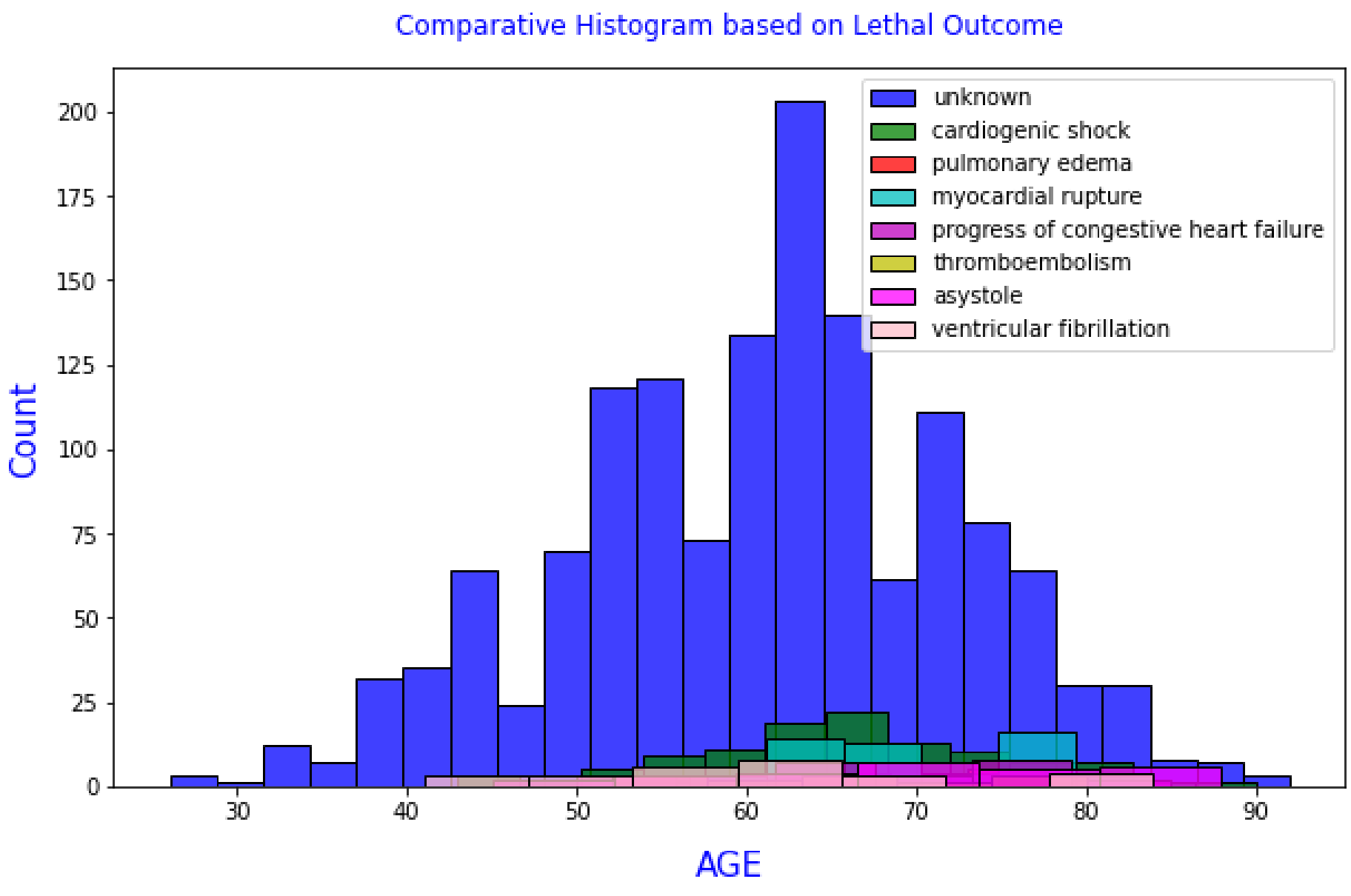

*Figure 4.47 Comparative histogram between AGE vs LET_IS (all target class)*

**d) S_AD_ORIT (Systolic blood pressure according to intensive care unit)**

From the below figure 4.48, the average systolic blood pressure for the patients having pulmonary edema (Target class 2) and myocardial rupture (Target class 3) is more than 140 mmHg which is greater than rest of the target class systolic pressure. Also, the average systolic blood pressure for the patients having cardiogenic shock (Target Class 1) is around 98 mmHg which is distinctively less compared to any other target class systolic pressure, hence low systolic blood pressure can be a good indicator of cardiogenic shock.

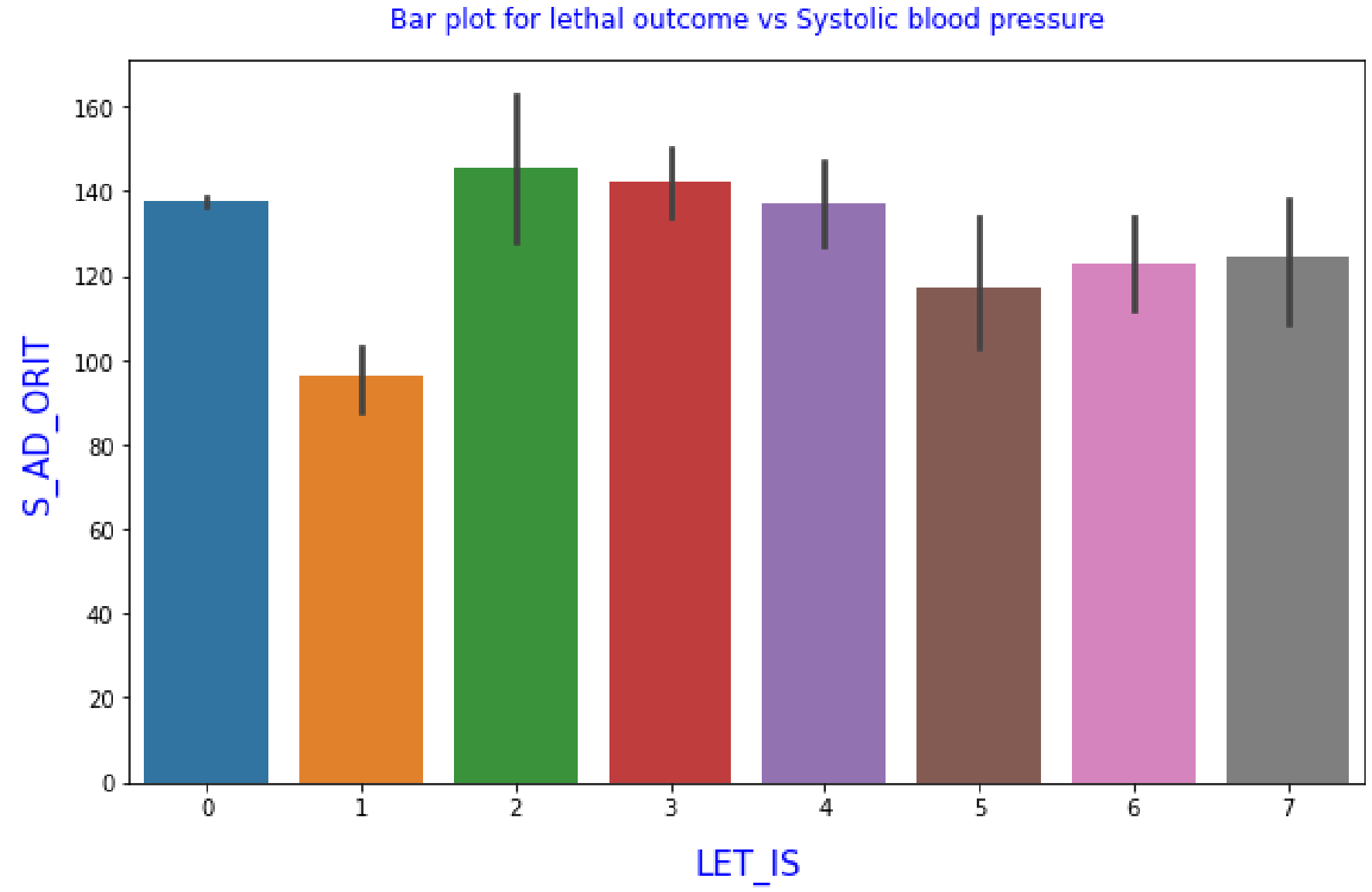

*Figure 4.48 Barplot between S_AD_ORIT vs LET_IS (target)*

From the below figure 4.49, the systolic blood pressure for all patients suffering from fatal outcomes of acute myocardial infarction is not able to separate most of the target class distinctively as all the classes are overlapped with each other, but it can also be seen that low systolic blood pressure can be a good indicator of cardiogenic shock (Target class 1), because for systolic pressure between 0.00 mmHg and 98 mmHg almost all cases belong to cardiogenic shock (Target class 1), hence low systolic blood pressure can be a good predictor of cardiogenic shock.

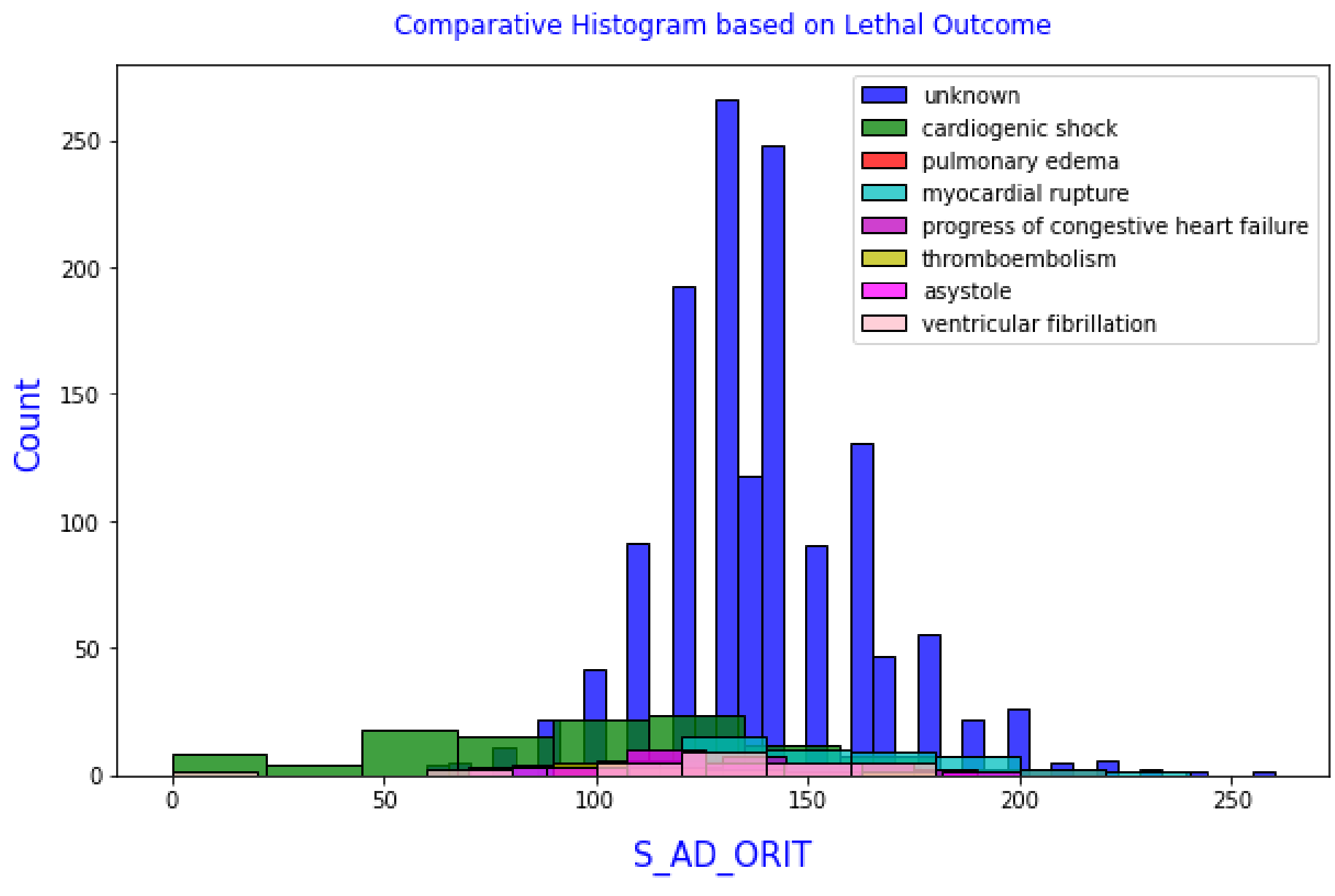

*Figure 4.49 Comparative histogram between S_AD_ORIT vs LET_IS (all target class)*

**e) D_AD_ORIT (Diastolic blood pressure according to intensive care unit)**

From the below figure 4.50, the average diastolic blood pressure for the patients having pulmonary edema (Target class 2) is around 90 mmHg which is greater than rest of the target class diastolic pressure. Also, the average diastolic blood pressure for the patients having cardiogenic shock (Target Class 1) is around 58 mmHg which is distinctively less compared to any other target class diastolic pressure, hence like systolic blood pressure, low diastolic blood pressure can be a good indicator of cardiogenic shock as well.

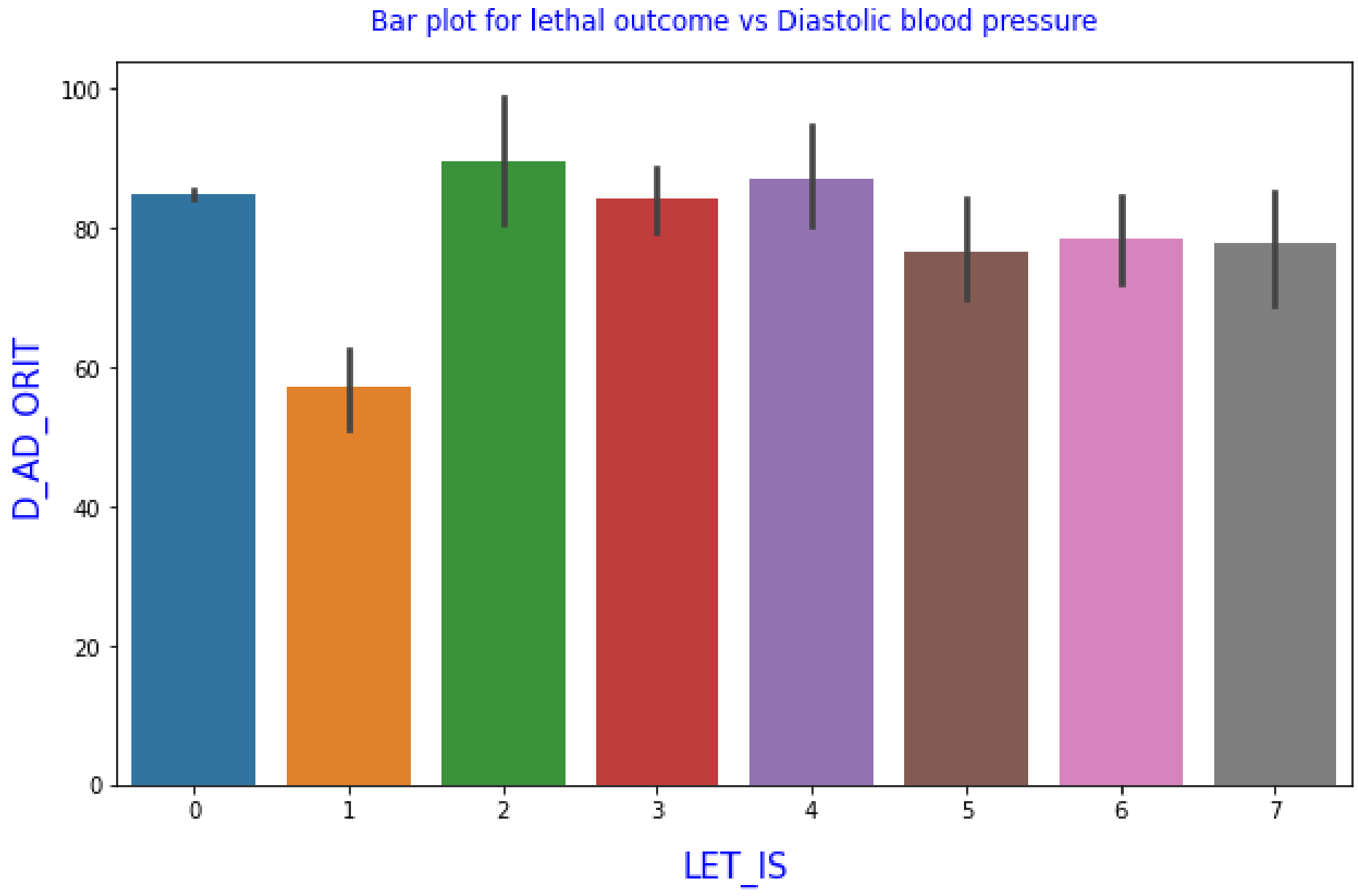

*Figure 4.50 Barplot between D_AD_ORIT vs LET_IS (target)*

From the below figure 4.51, the diastolic blood pressure for all patients suffering from fatal outcomes of acute myocardial infarction is not able to separate most of the target class distinctively as all the classes are overlapped with each other, but it can also be seen that low diastolic blood pressure can be a good indicator of cardiogenic shock (Target class 1) just like systolic blood pressure, because for diastolic pressure between 0.00 mmHg and 58 mmHg almost all cases belong to cardiogenic shock (Target class 1), hence low diastolic blood pressure can be a good predictor of cardiogenic shock.

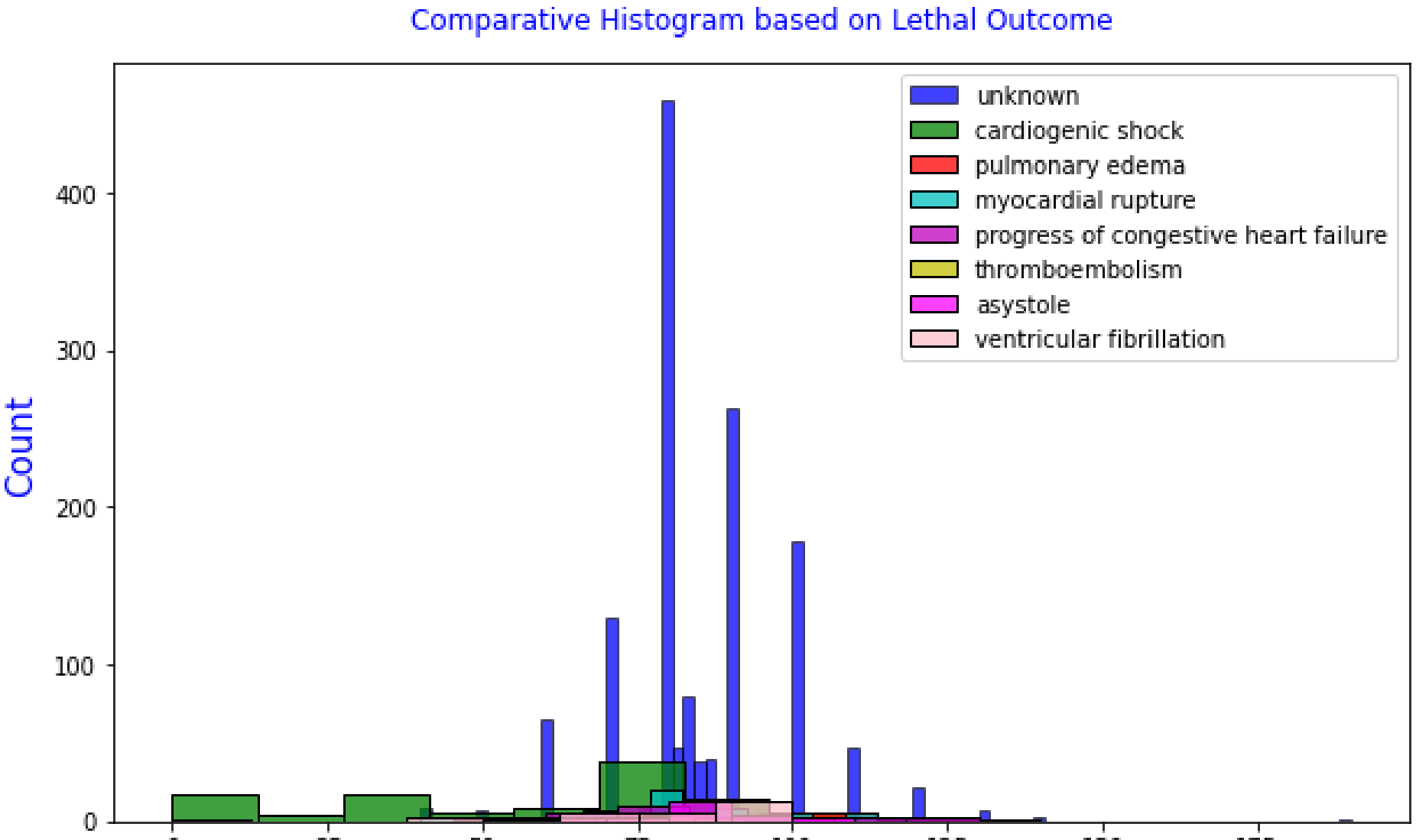

*Figure 4.51 Comparative histogram between D_AD_ORIT vs LET_IS (all target class)*

#### 4.4.2.2 Numerical vs Numerical Analysis

This section will go through different bivariate visualization for some significant independent numerical characteristics, using scatterplot. This section discusses a couple of the most essential biomarkers among all the predictive indicators. Below analysis focuses mostly on the numerical vs numerical variables relationship.

**a) AGE (Patient's age)**

From below scatterplot figure 4.52, age of the patients suffering from lethal outcomes of acute myocardial infarction does not have any type of relationships with any other numerical features like with serum potassium content or with A1AT serum content or with diastolic pressure or with systolic pressure. All the relationships are merely random, and no patterns are observed.

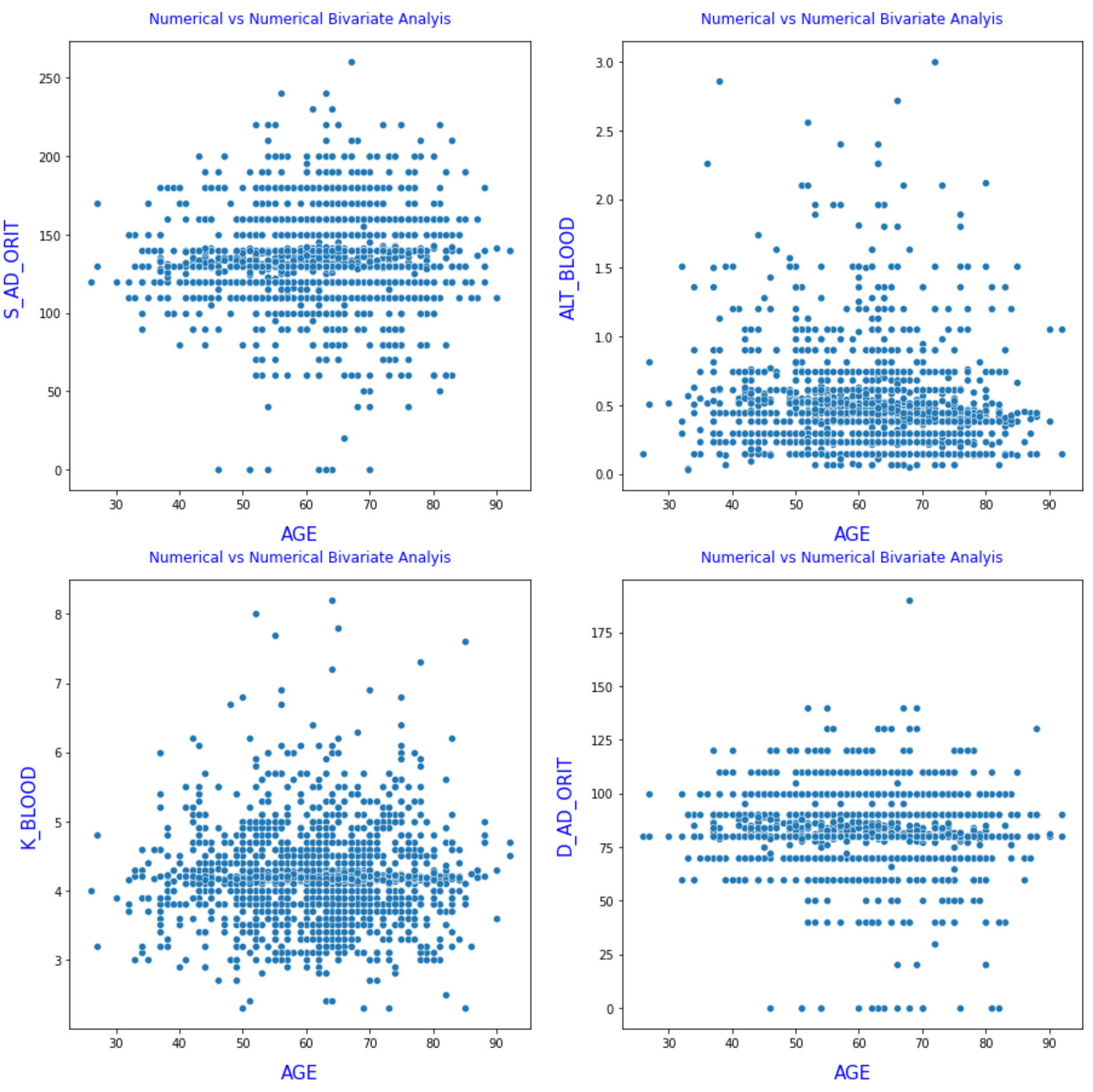

*Figure 4.52 Scatterplot plot between AGE vs D_AD_ORIT, S_AD_ORIT, ALT_BLOOD. K_BLOOD*

**b) S_AD_ORIT (Systolic blood pressure according to ICU)**

From below scatterplot figure 4.53, systolic blood pressure of the patients suffering from lethal outcomes of acute myocardial infarction does not have any type of relationships with any other numerical features like with serum potassium content or with A1AT serum content or with age expect diastolic blood pressure. Systolic blood pressure follows a good positive linear relationship with diastolic blood pressure. Rest other relationships are merely random, and no patterns are observed.

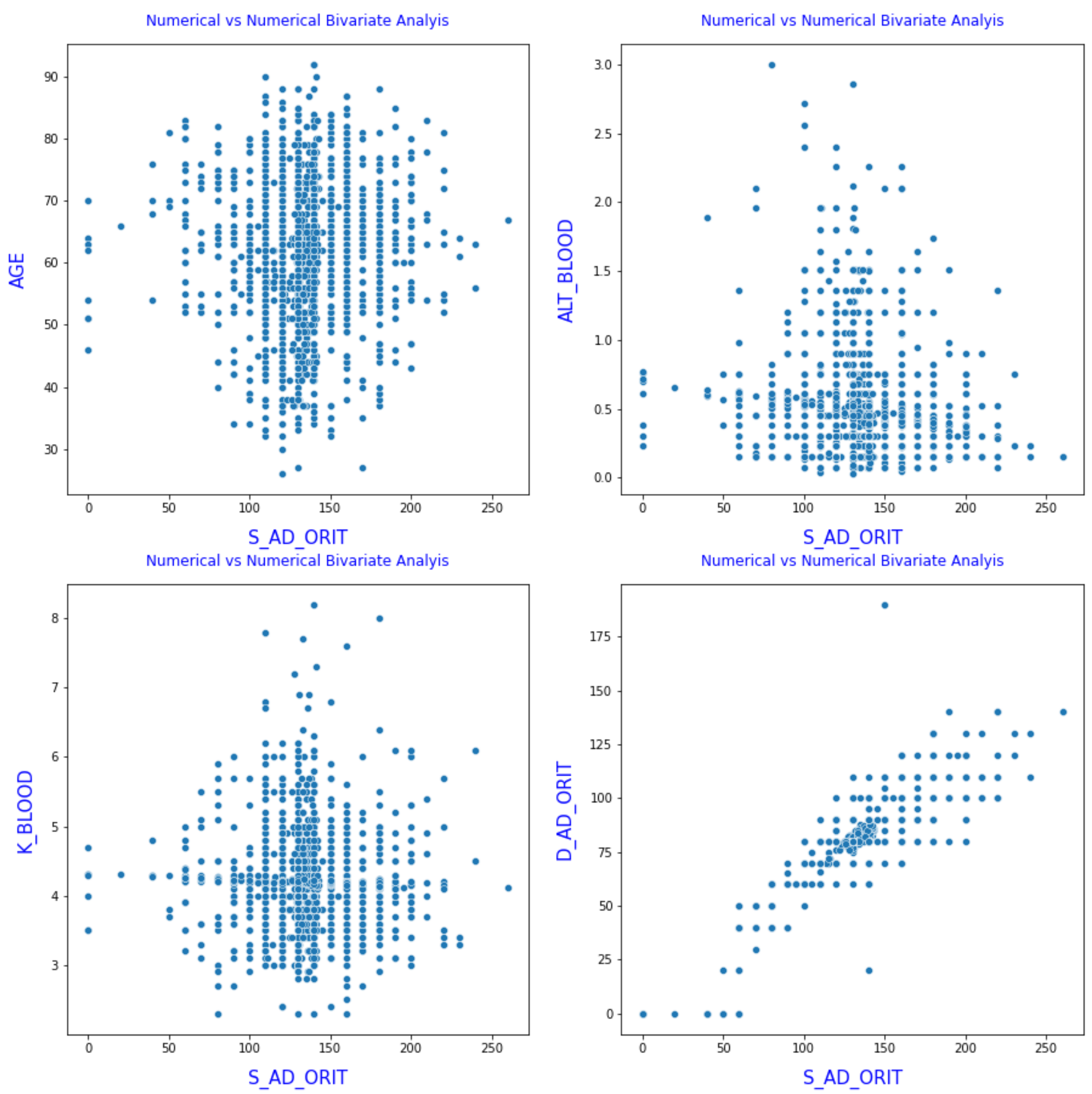

*Figure 4.53 Scatterplot plot between S_AD_ORIT vs AGE, D_AD_ORIT, ALT_BLOOD. K_BLOOD*

**c) D_AD_ORIT (Diastolic blood pressure according to ICU)**

From below scatterplot figure 4.54, diastolic blood pressure of the patients suffering from lethal outcomes of acute myocardial infarction does not have any type of relationships with any other numerical features like with serum potassium content or with A1AT serum content or with age expect systolic blood pressure. Diastolic blood pressure follows a good positive linear relationship with systolic blood pressure. Rest other relationships are merely random, and no patterns are observed.

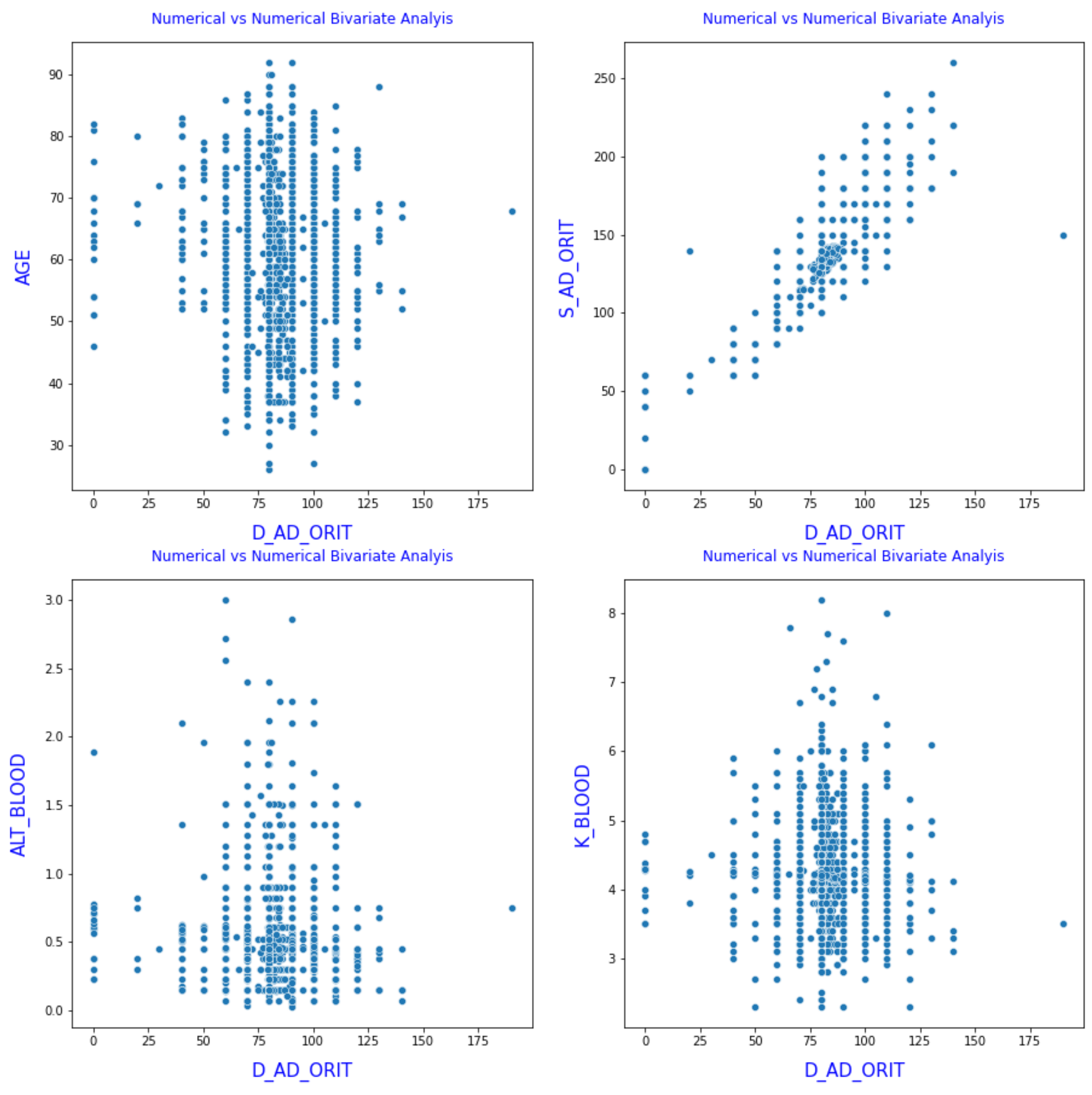

*Figure 4.54 Scatterplot plot between D_AD_ORIT vs AGE, S_AD_ORIT, ALT_BLOOD. K_BLOOD*

### d) K_BLOOD (Serum potassium content)

From below scatterplot figure 4.55, serum potassium content of the patients suffering from lethal outcomes of acute myocardial infarction does not have any type of relationships with any other numerical features like with systolic blood pressure or with A1AT serum content or with age or with diastolic blood pressure. All relationships are merely random, and no patterns are observed.

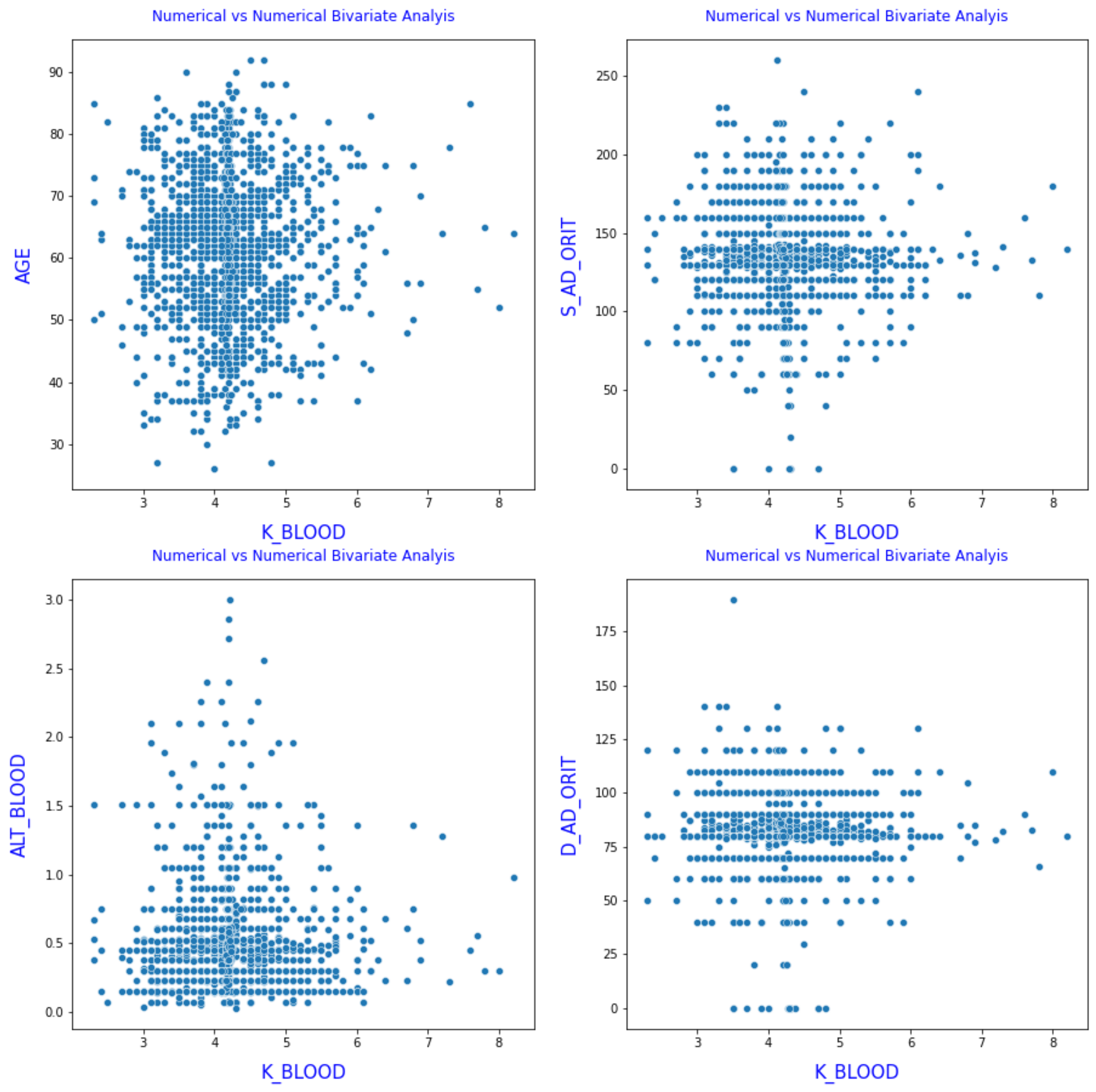

*Figure 4.55 Scatterplot plot between K_BLOOD vs AGE, S_AD_ORIT, D_AD_ORIT, ALT_BLOOD*

**e) ALT_BLOOD (Serum AlAT content)**

From below scatterplot figure 4.56, serum A1AT content of the patients suffering from lethal outcomes of acute myocardial infarction does not have any type of relationships with any other numerical features like with systolic blood pressure or serum potassium content or with age or with diastolic blood pressure. All relationships are merely random, and no patterns are observed.

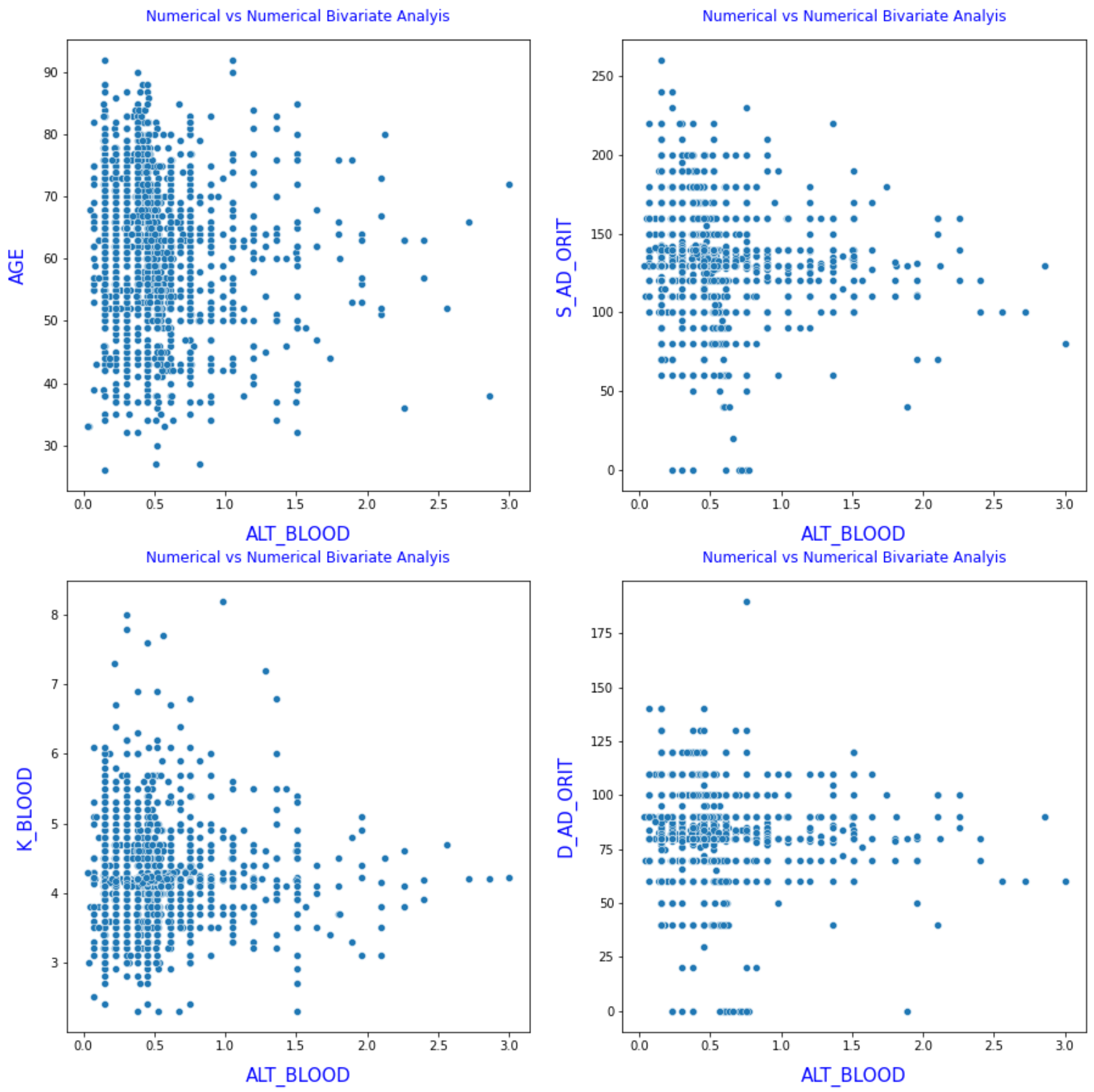

*Figure 4.56 Scatterplot plot between ALT_BLOOD vs AGE, S_AD_ORIT, D_AD_ORIT, K_BLOOD*

#### 4.4.2.3 Categorical vs Categorical Analysis

This section will go through different bivariate visualization for some significant independent categorical characteristics, using count plot and percentage analysis. This section discusses a couple of the most essential biomarkers among all the predictive indicators. Below analysis focuses mostly on the comparative frequency/percentage analysis between categorical vs target (LET_IS) variable.

**a) K_SH_POST (Cardiogenic shock at the time of admission to ICU)**

From below count plot and percentage analysis figure 4.57, 86% of patients who did not have cardiogenic shock (decoded value:0) at the time of admission to ICU suffered fatal outcome of acute myocardial infarction due to unknown reasons and around 0.72% of them suffered fatal outcome of acute myocardial infarction due to thromboembolism.

Again, 89.13% of patients who had cardiogenic shock (decoded value:1) at the time of admission to ICU suffered fatal outcomes of acute myocardial infarction due to cardiogenic shock itself while very less percentage of patients suffered lethal outcome for this disease due to asystole and ventricular fibrillation.

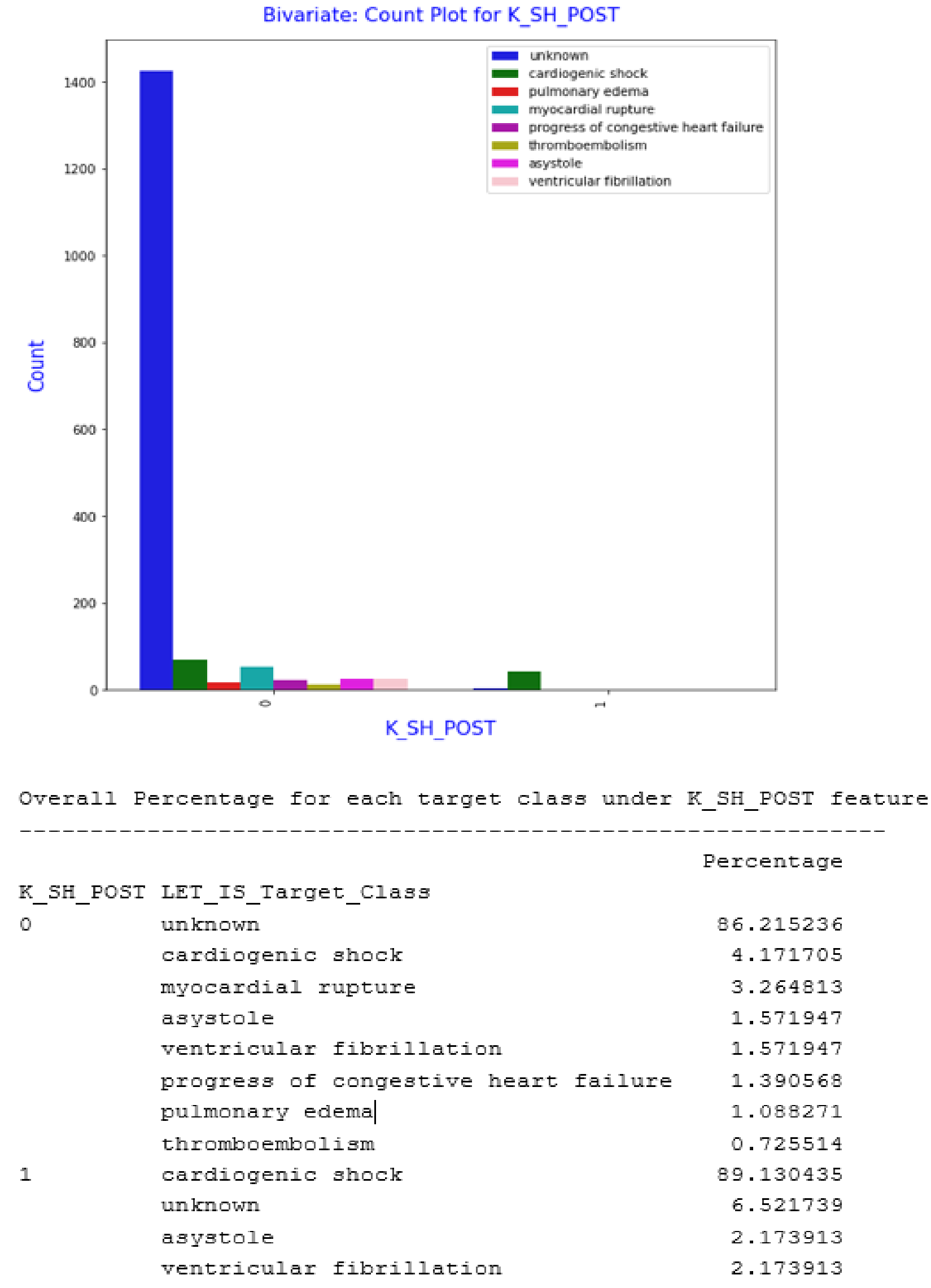

Overall Percentage for each target class under K_SH_POST feature

| K_SH_POST | LET_IS_Target_Class | Percentage |
|---|---|---|
| 0 | unknown | 86.215236 |
| | cardiogenic shock | 4.171705 |
| | myocardial rupture | 3.264813 |
| | asystole | 1.571947 |
| | ventricular fibrillation | 1.571947 |
| | progress of congestive heart failure | 1.390568 |
| | pulmonary edema | 1.088271 |
| | thromboembolism | 0.725514 |
| 1 | cardiogenic shock | 89.130435 |
| | unknown | 6.521739 |
| | asystole | 2.173913 |
| | ventricular fibrillation | 2.173913 |

*Figure 4.57 Countplot and percentage analysis of K_SH_POST feature for all target class*

**b) ritm_ecg_p_02 (ECG rhythm at the time of admission to hospital – atrial fibrillation)**

From below count plot and percentage analysis figure 4.58, 85.23% of patients who did not have arterial fibrillated ECG rhythm (decoded value:0) at the time of admission to hospital suffered fatal outcome of acute myocardial infarction due to unknown reasons while around 6.1% of them suffered fatal outcome of acute myocardial infarction due to cariogenic shock and 0.37% of them suffered fatal outcome of acute myocardial infarction due to thromboembolism.

On the other hand, 64.21% of patients who had arterial fibrillated ECG rhythm (decoded value:1) at the time of admission to hospital suffered fatal outcomes of acute myocardial infarction due to unknown reasons while 12.63% of patients suffered lethal outcome for this disease due to cardiogenic shock and 1.05% of them suffered lethal outcome for this disease due to pulmonary edema.

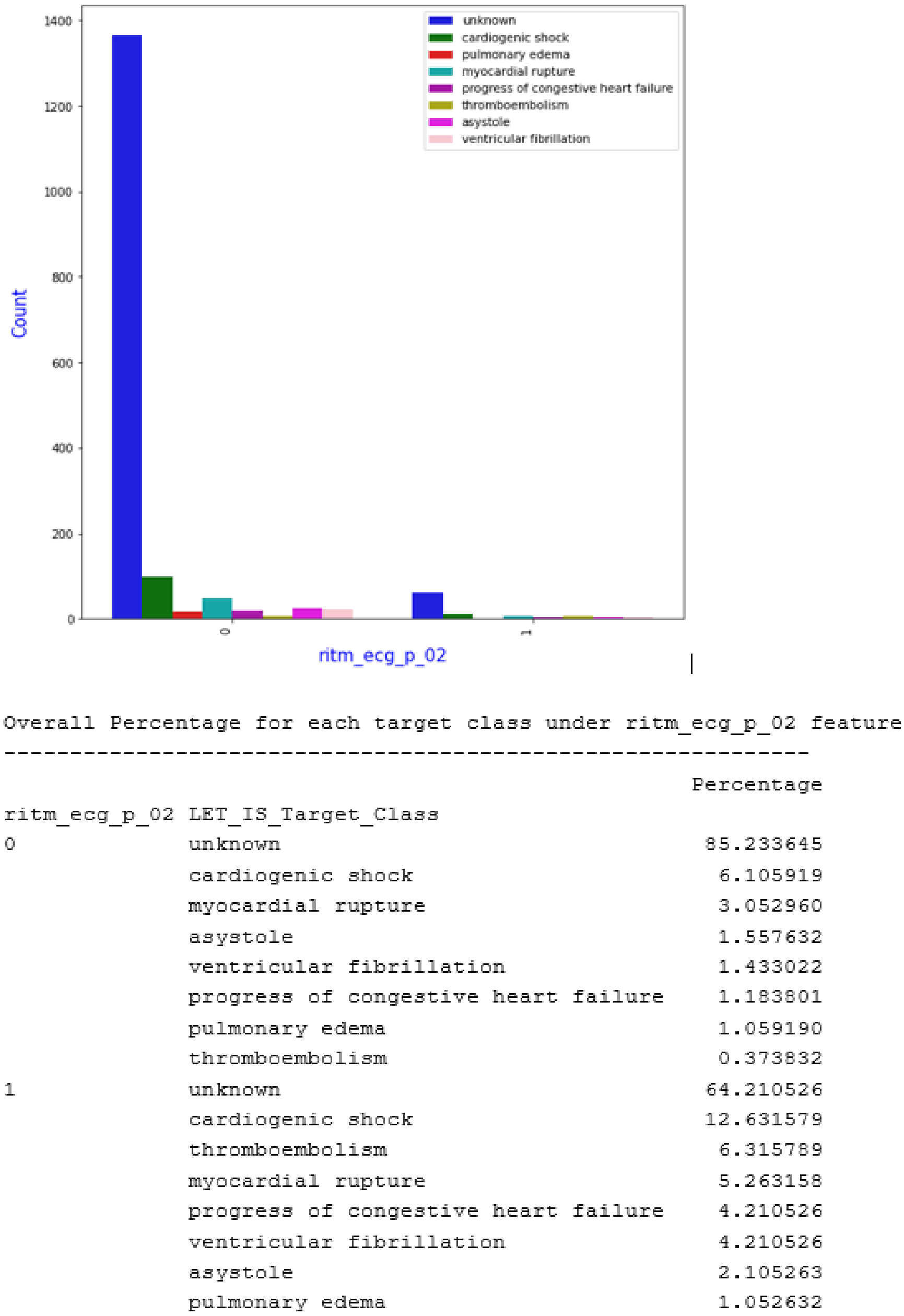

Overall Percentage for each target class under ritm_ecg_p_02 feature

| ritm_ecg_p_02 | LET_IS_Target_Class | Percentage |
|---|---|---|
| 0 | unknown | 85.233645 |
| | cardiogenic shock | 6.105919 |
| | myocardial rupture | 3.052960 |
| | asystole | 1.557632 |
| | ventricular fibrillation | 1.433022 |
| | progress of congestive heart failure | 1.183801 |
| | pulmonary edema | 1.059190 |
| | thromboembolism | 0.373832 |
| 1 | unknown | 64.210526 |
| | cardiogenic shock | 12.631579 |
| | thromboembolism | 6.315789 |
| | myocardial rupture | 5.263158 |
| | progress of congestive heart failure | 4.210526 |
| | ventricular fibrillation | 4.210526 |
| | asystole | 2.105263 |
| | pulmonary edema | 1.052632 |

*Figure 4.58 Countplot and percentage analysis of ritm_ecg_p_02 feature for all target class*

### c) MP_TP_POST (Paroxysms of atrial fibrillation at the time of admission to ICU)

From below count plot and percentage analysis figure 4.59, 85.24% of patients who did not have paroxysms of arterial fibrillated ECG rhythm (decoded value:0) at the time of admission to ICU suffered fatal outcome of acute myocardial infarction due to unknown reasons while around 6.17% of them suffered fatal outcome of acute myocardial infarction due to cariogenic shock and 0.37% of them suffered fatal outcome of acute myocardial infarction due to thromboembolism.

On the other hand, 67.54% of patients who had paroxysms of arterial fibrillated ECG rhythm (decoded value:1) at the time of admission to ICU suffered fatal outcomes of acute myocardial infarction due to unknown reasons while 10.52% of patients suffered lethal outcome for this disease due to cardiogenic shock and 0.87% of them suffered lethal outcome for this disease due to pulmonary edema.

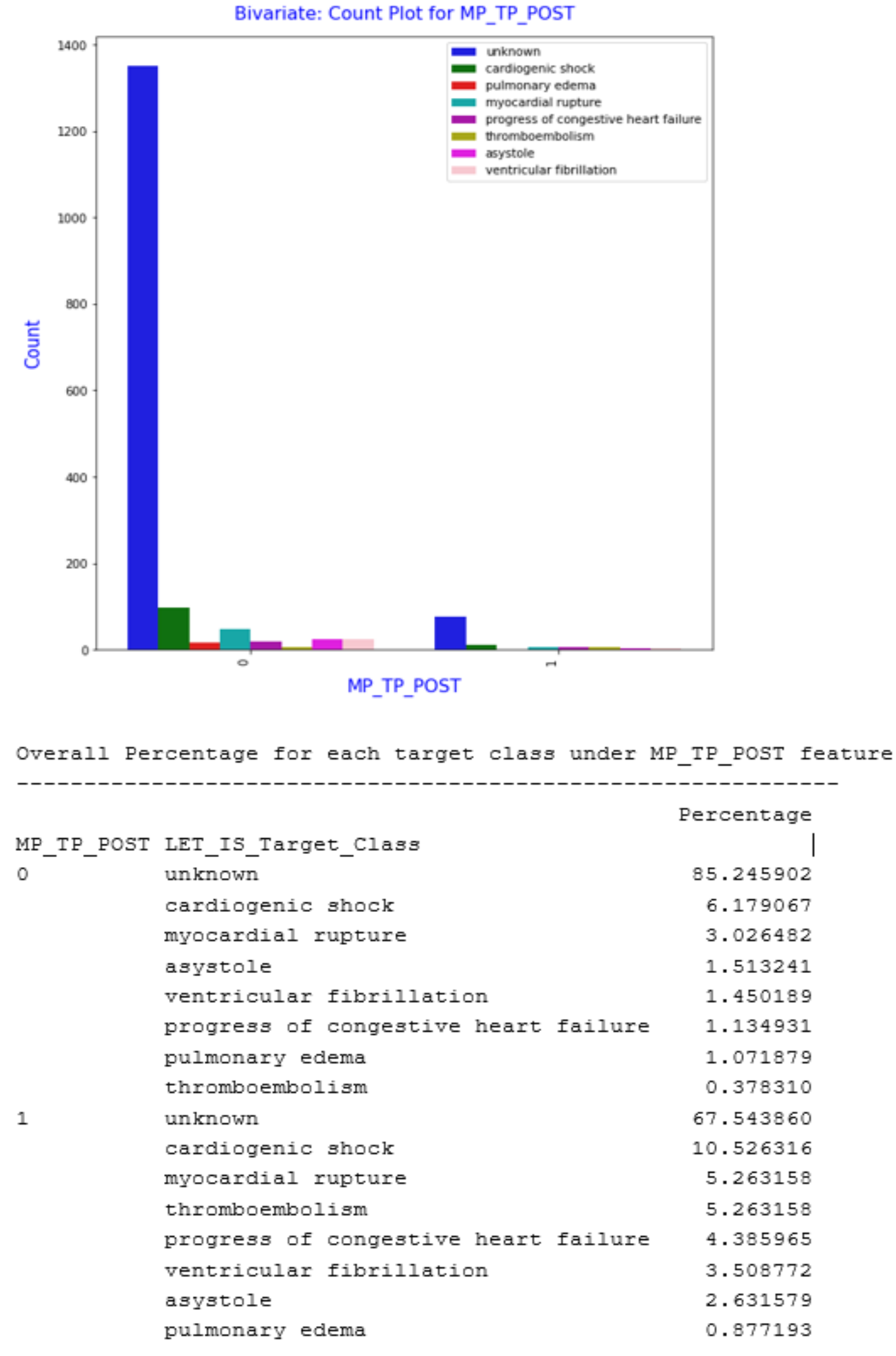

Overall Percentage for each target class under MP_TP_POST feature

| MP_TP_POST | LET_IS_Target_Class | Percentage |
|---|---|---|
| 0 | unknown | 85.245902 |
| | cardiogenic shock | 6.179067 |
| | myocardial rupture | 3.026482 |
| | asystole | 1.513241 |
| | ventricular fibrillation | 1.450189 |
| | progress of congestive heart failure | 1.134931 |
| | pulmonary edema | 1.071879 |
| | thromboembolism | 0.378310 |
| 1 | unknown | 67.543860 |
| | cardiogenic shock | 10.526316 |
| | myocardial rupture | 5.263158 |
| | thromboembolism | 5.263158 |
| | progress of congestive heart failure | 4.385965 |
| | ventricular fibrillation | 3.508772 |
| | asystole | 2.631579 |
| | pulmonary edema | 0.877193 |

*Figure 4.59 Countplot and percentage analysis of MP_TP_POST feature for all target class*

**d) SEX (Patient's gender)**

From below count plot and percentage analysis figure 4.60, 78.42% of patients who were female (decoded value:0) suffered fatal outcome of acute myocardial infarction due to unknown reasons while around 8.50% of them suffered fatal outcome of acute myocardial infarction due to cariogenic shock and 0.94% of them suffered fatal outcome of acute myocardial infarction due to pulmonary edema.

On the other hand, 87.41% of patients who were male (decoded value:1) suffered fatal outcomes of acute myocardial infarction due to unknown reasons while 5.25% of patients suffered lethal outcome for this disease due to cardiogenic shock and 0.28% of them suffered lethal outcome for this disease due to thromboembolism.

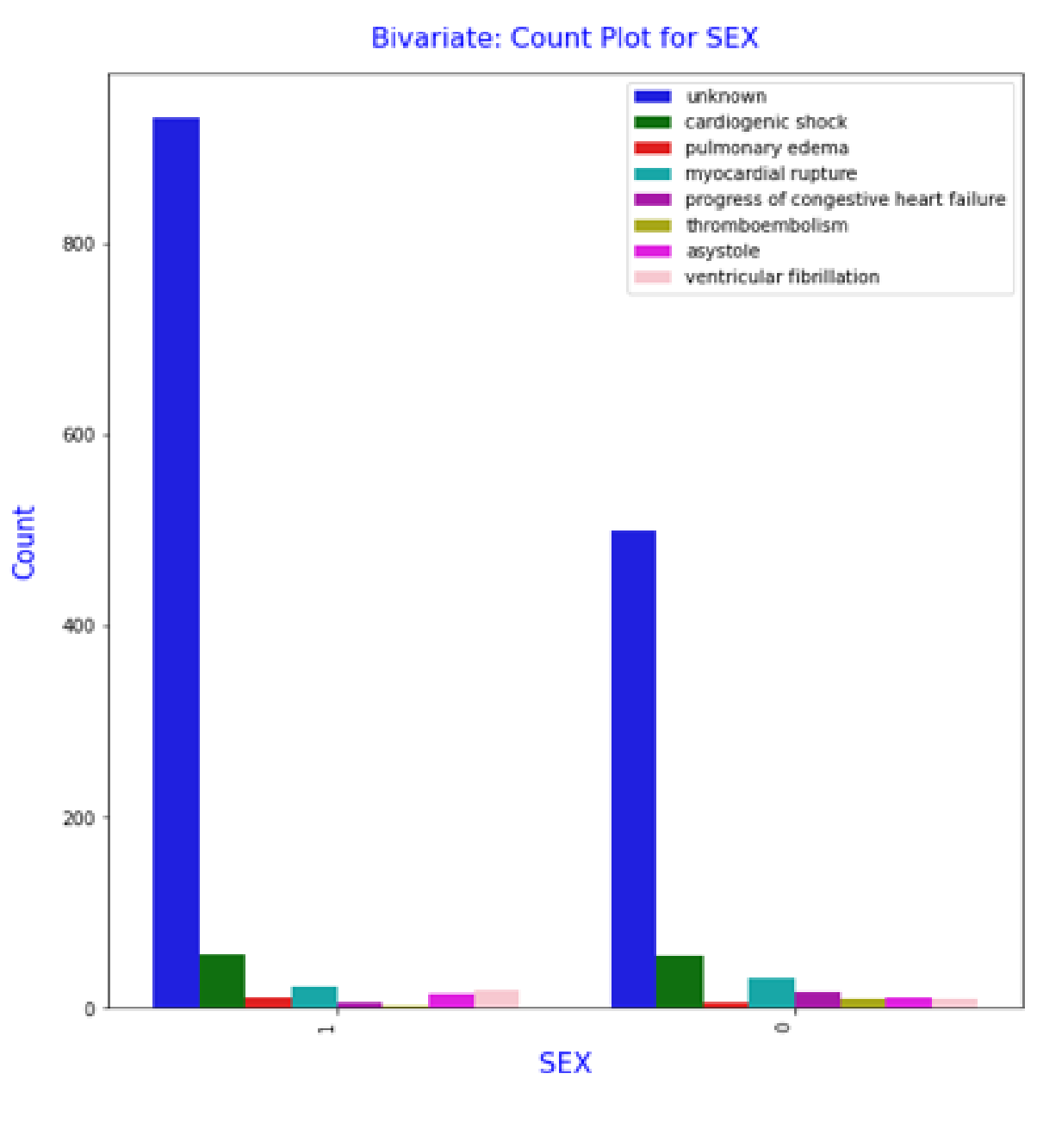

Overall Percentage for each target class under SEX feature

| SEX | LET_IS_Target_Class | Percentage |
|---|---|---|
| 0 | unknown | 78.425197 |
| | cardiogenic shock | 8.503937 |
| | myocardial rupture | 4.881890 |
| | progress of congestive heart failure | 2.677165 |
| | asystole | 1.732283 |
| | thromboembolism | 1.417323 |
| | ventricular fibrillation | 1.417323 |
| | pulmonary edema | 0.944882 |
| 1 | unknown | 87.417840 |
| | cardiogenic shock | 5.258216 |
| | myocardial rupture | 2.159624 |
| | ventricular fibrillation | 1.690141 |
| | asystole | 1.502347 |
| | pulmonary edema | 1.126761 |
| | progress of congestive heart failure | 0.563380 |
| | thromboembolism | 0.281690 |

*Figure 4.60 Countplot and percentage analysis of SEX feature for all target class*

### e) NITR_S (Use of liquid nitrates in the ICU)

From below count plot and percentage analysis figure 4.61, 86.84% of patients who were not subjected to use of liquid nitrates (decoded value:0) in ICU suffered fatal outcome of acute myocardial infarction due to unknown reasons while around 5.18% of them suffered fatal outcome of acute myocardial infarction due to cariogenic shock and 0.66% of them suffered fatal outcome of acute myocardial infarction due to pulmonary edema.

On the other hand, 62.56% of patients who were subjected to use of liquid nitrates (decoded value:1) in ICU suffered fatal outcomes of acute myocardial infarction due to unknown reasons while 16.41% of patients suffered lethal outcome for this disease due to cardiogenic shock and 0.512% of them suffered lethal outcome for this disease due to thromboembolism.

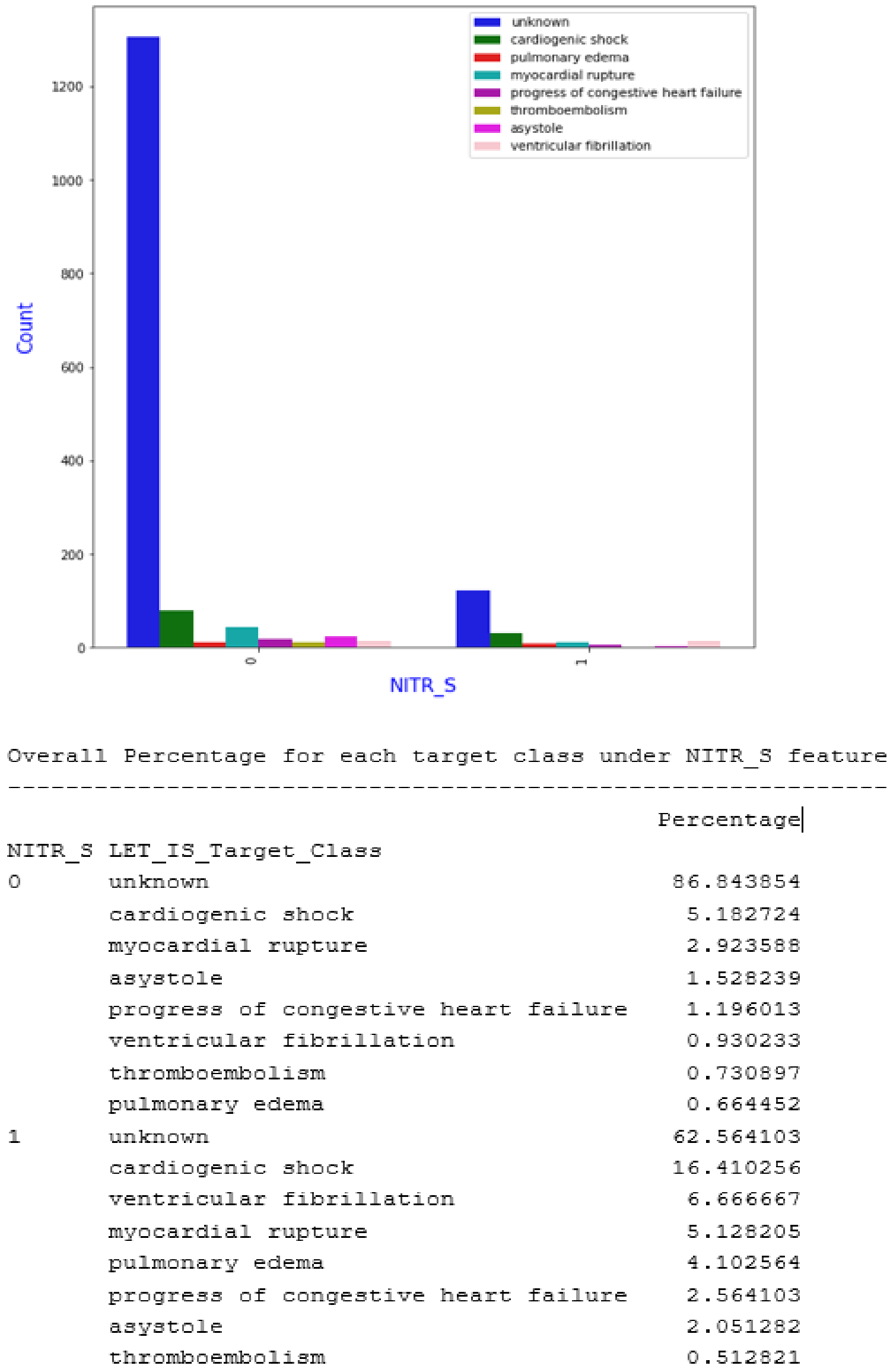

Overall Percentage for each target class under NITR_S feature

| NITR_S | LET_IS_Target_Class | Percentage |
|---|---|---|
| 0 | unknown | 86.843854 |
| | cardiogenic shock | 5.182724 |
| | myocardial rupture | 2.923588 |
| | asystole | 1.528239 |
| | progress of congestive heart failure | 1.196013 |
| | ventricular fibrillation | 0.930233 |
| | thromboembolism | 0.730897 |
| | pulmonary edema | 0.664452 |
| 1 | unknown | 62.564103 |
| | cardiogenic shock | 16.410256 |
| | ventricular fibrillation | 6.666667 |
| | myocardial rupture | 5.128205 |
| | pulmonary edema | 4.102564 |
| | progress of congestive heart failure | 2.564103 |
| | asystole | 2.051282 |
| | thromboembolism | 0.512821 |

*Figure 4.61 Countplot and percentage analysis of NITR_S feature for all target class*

**f) n_p_ecg_p_12 (Complete RBBB on ECG at the time of admission to hospital)**

From below count plot and percentage analysis figure 4.62, 85.38% of patients who were not having complete right bundle branch block (decoded value:0) during hospital admission suffered fatal outcome of acute myocardial infarction due to unknown reasons while around 6.04% of them suffered fatal outcome of acute myocardial infarction due to cariogenic shock and 0.55% of them suffered fatal outcome of acute myocardial infarction due to thromboembolism.

On the other hand, 56.41% of patients who were having complete right bundle branch block (decoded value:1) during hospital admission suffered fatal outcomes of acute myocardial infarction due to unknown reasons while 15.38% of patients suffered lethal outcome for this disease due to cardiogenic shock and 3.84% of them suffered lethal outcome for this disease due to asystole.

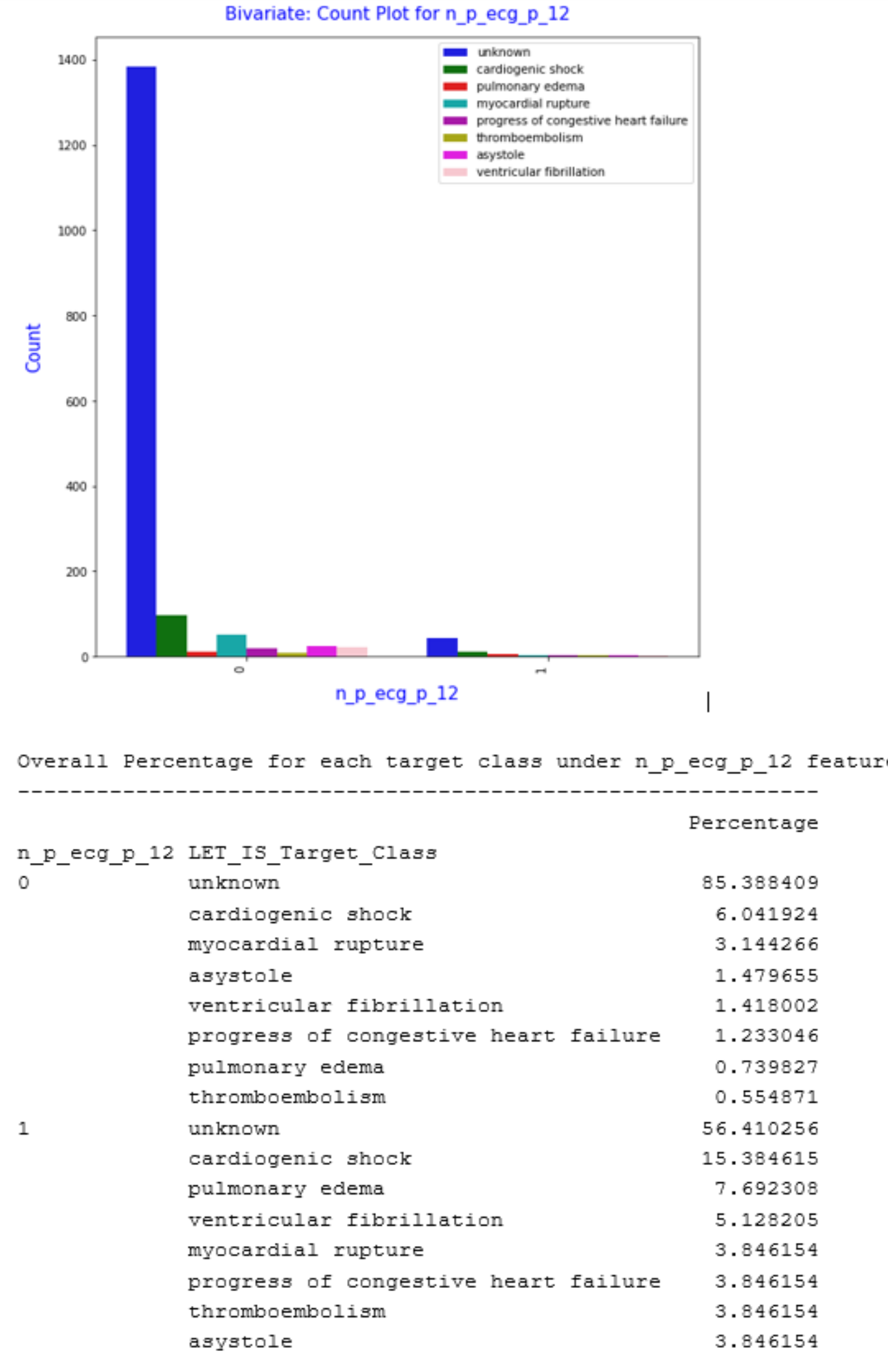

Overall Percentage for each target class under n_p_ecg_p_12 feature

| n_p_ecg_p_12 | LET_IS_Target_Class | Percentage |
|---|---|---|
| 0 | unknown | 85.388409 |
| | cardiogenic shock | 6.041924 |
| | myocardial rupture | 3.144266 |
| | asystole | 1.479655 |
| | ventricular fibrillation | 1.418002 |
| | progress of congestive heart failure | 1.233046 |
| | pulmonary edema | 0.739827 |
| | thromboembolism | 0.554871 |
| 1 | unknown | 56.410256 |
| | cardiogenic shock | 15.384615 |
| | pulmonary edema | 7.692308 |
| | ventricular fibrillation | 5.128205 |
| | myocardial rupture | 3.846154 |
| | progress of congestive heart failure | 3.846154 |
| | thromboembolism | 3.846154 |
| | asystole | 3.846154 |

*Figure 4.62 Countplot and percentage analysis of n_p_ecg_p_12 feature for all target class*

**g) NOT_NA_KB (Use of NSAIDs by the Emergency Cardiology Team)**

From below count plot and percentage analysis figure 4.63, 78.59% of patients who were not subjected to use of NSAID drug (decoded value:0) suffered fatal outcome of acute myocardial infarction due to unknown reasons while around 8.30% of them suffered fatal outcome of acute myocardial infarction due to cariogenic shock and 0.63% of them suffered fatal outcome of acute myocardial infarction due to pulmonary edema.

On the other hand, 85.29% of patients who were subjected to use of NSAID drug (decoded value:1) during hospital admission suffered fatal outcomes of acute myocardial infarction due to unknown reasons while 6.05% of patients suffered lethal outcome for this disease due to cardiogenic shock and 0.64% of them suffered lethal outcome for this disease due to thromboembolism.

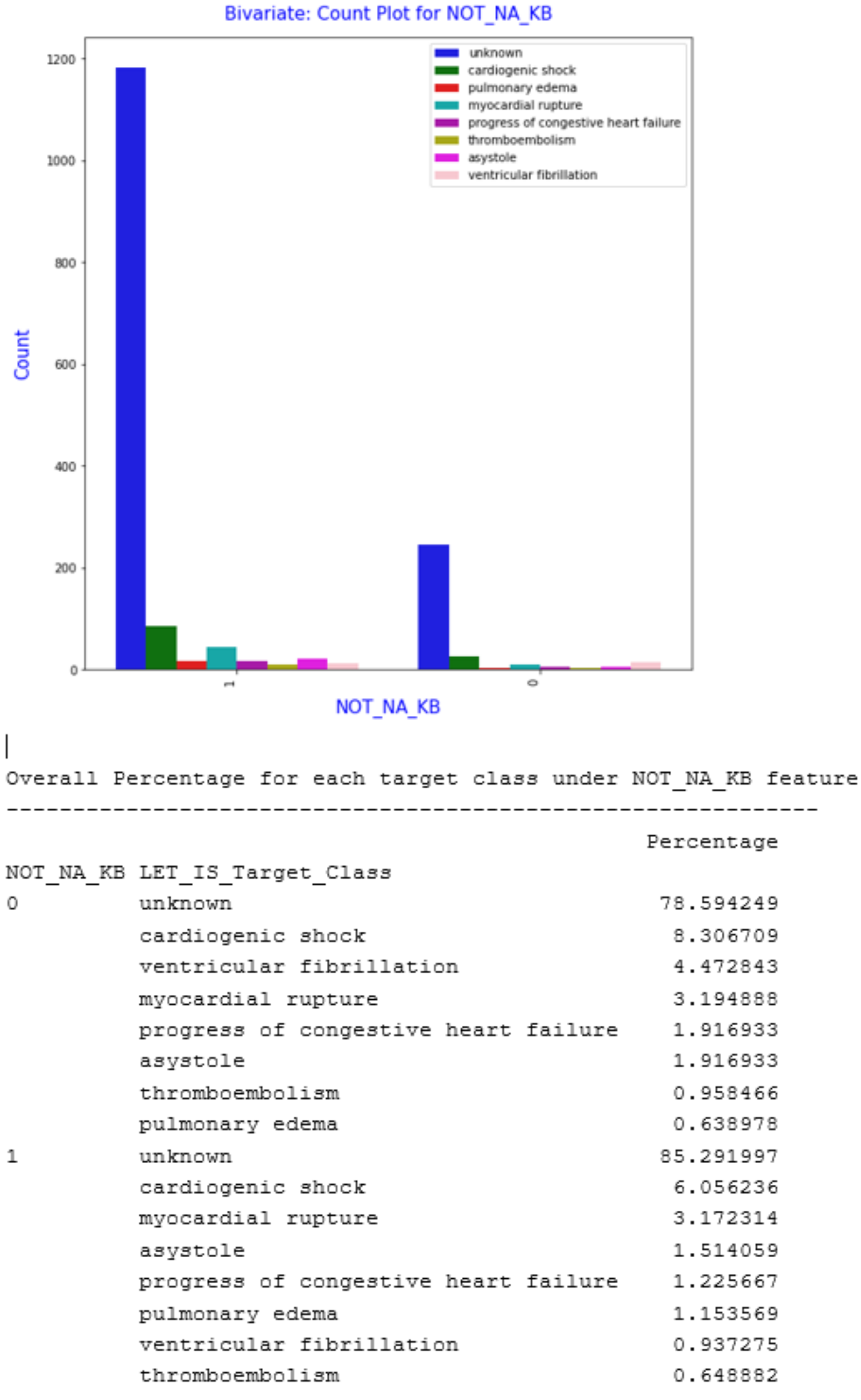

Overall Percentage for each target class under NOT_NA_KB feature

| NOT_NA_KB | LET_IS_Target_Class | Percentage |
|---|---|---|
| 0 | unknown | 78.594249 |
| | cardiogenic shock | 8.306709 |
| | ventricular fibrillation | 4.472843 |
| | myocardial rupture | 3.194888 |
| | progress of congestive heart failure | 1.916933 |
| | asystole | 1.916933 |
| | thromboembolism | 0.958466 |
| | pulmonary edema | 0.638978 |
| 1 | unknown | 85.291997 |
| | cardiogenic shock | 6.056236 |
| | myocardial rupture | 3.172314 |
| | asystole | 1.514059 |
| | progress of congestive heart failure | 1.225667 |
| | pulmonary edema | 1.153569 |
| | ventricular fibrillation | 0.937275 |
| | thromboembolism | 0.648882 |

*Figure 4.63 Countplot and percentage analysis of NOT_NA_KB feature for all target class*

**h) O_L_POST (Pulmonary edema at the time of admission to ICU)**

From below count plot and percentage analysis figure 4.64, 85.66% of patients who were not having pulmonary edema (decoded value:0) at the time of admission to ICU suffered fatal outcome of acute myocardial infarction due to unknown reasons while around 5.53% of them suffered fatal outcome of acute myocardial infarction due to cariogenic shock and 0.75% of them suffered fatal outcome of acute myocardial infarction due to thromboembolism.

On the other hand, 60.90% of patients who were having pulmonary edema (decoded value:1) during ICU admission suffered fatal outcomes of acute myocardial infarction due to unknown reasons while 20.00% of patients suffered lethal outcome for this disease due to cardiogenic shock and 2.72% of them suffered lethal outcome for this disease due to asystole.

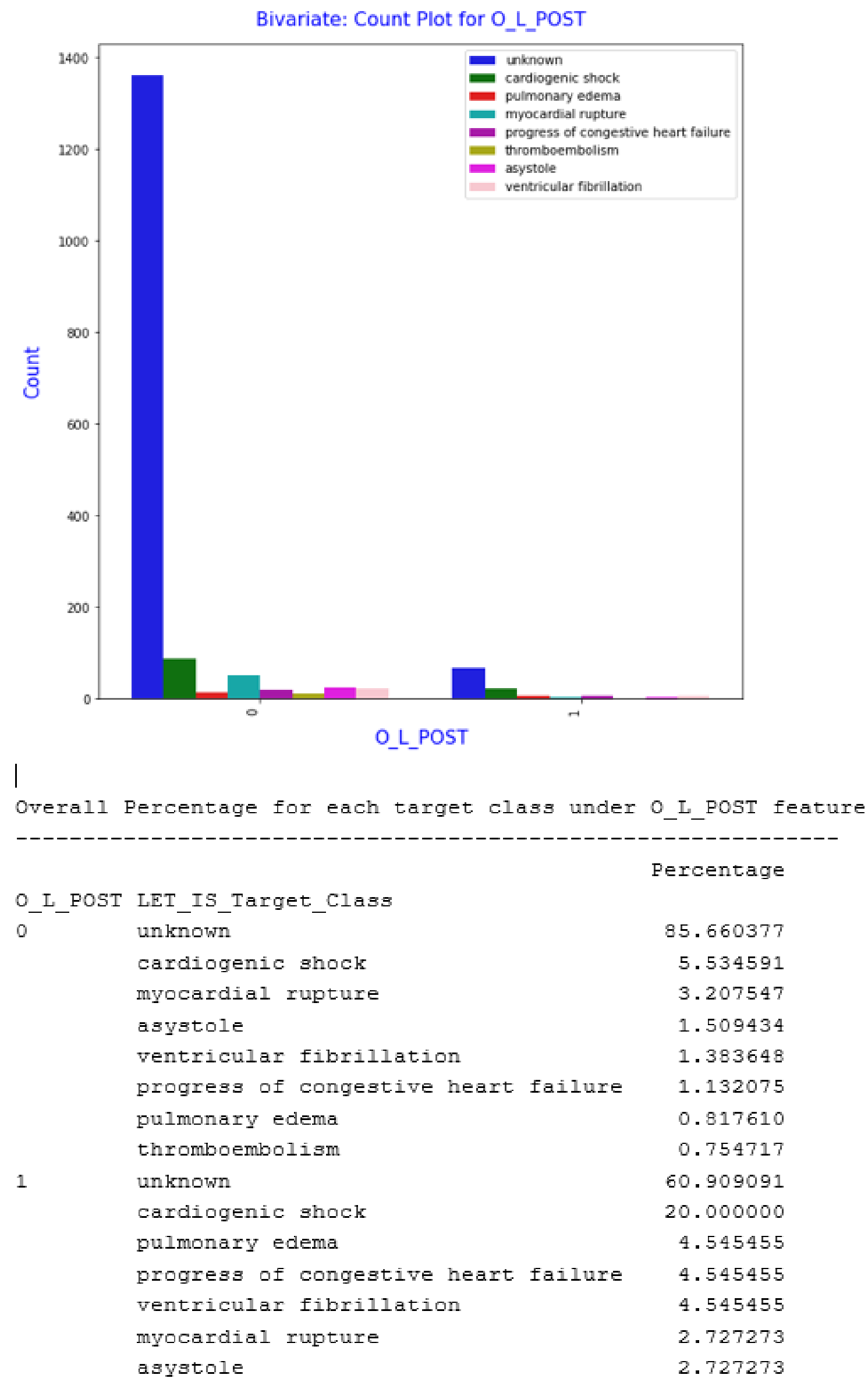

Overall Percentage for each target class under O_L_POST feature

| O_L_POST | LET_IS_Target_Class | Percentage |
|---|---|---|
| 0 | unknown | 85.660377 |
| | cardiogenic shock | 5.534591 |
| | myocardial rupture | 3.207547 |
| | asystole | 1.509434 |
| | ventricular fibrillation | 1.383648 |
| | progress of congestive heart failure | 1.132075 |
| | pulmonary edema | 0.817610 |
| | thromboembolism | 0.754717 |
| 1 | unknown | 60.909091 |
| | cardiogenic shock | 20.000000 |
| | pulmonary edema | 4.545455 |
| | progress of congestive heart failure | 4.545455 |
| | ventricular fibrillation | 4.545455 |
| | myocardial rupture | 2.727273 |
| | asystole | 2.727273 |

*Figure 4.64 Countplot and percentage analysis of O_L_POST feature for all target class*

### i) GEPAR_S_n (Use of a anticoagulants (heparin) in the ICU)

From below count plot and percentage analysis figure 4.65, 81.45% of patients who were not subjected to anticoagulants (heparin) (decoded value:0) in ICU suffered fatal outcome of acute myocardial infarction due to unknown reasons while around 10.20% of them suffered fatal outcome of acute myocardial infarction due to cariogenic shock and 0.62% of them suffered fatal outcome of acute myocardial infarction due to thromboembolism.

On the other hand, 85.08% of patients who were subjected to anticoagulants (heparin) (decoded value:1) in ICU suffered fatal outcomes of acute myocardial infarction due to unknown reasons while 5.00% of patients suffered lethal outcome for this disease due to cardiogenic shock and 0.65% of them suffered lethal outcome for this disease due to pulmonary edema.

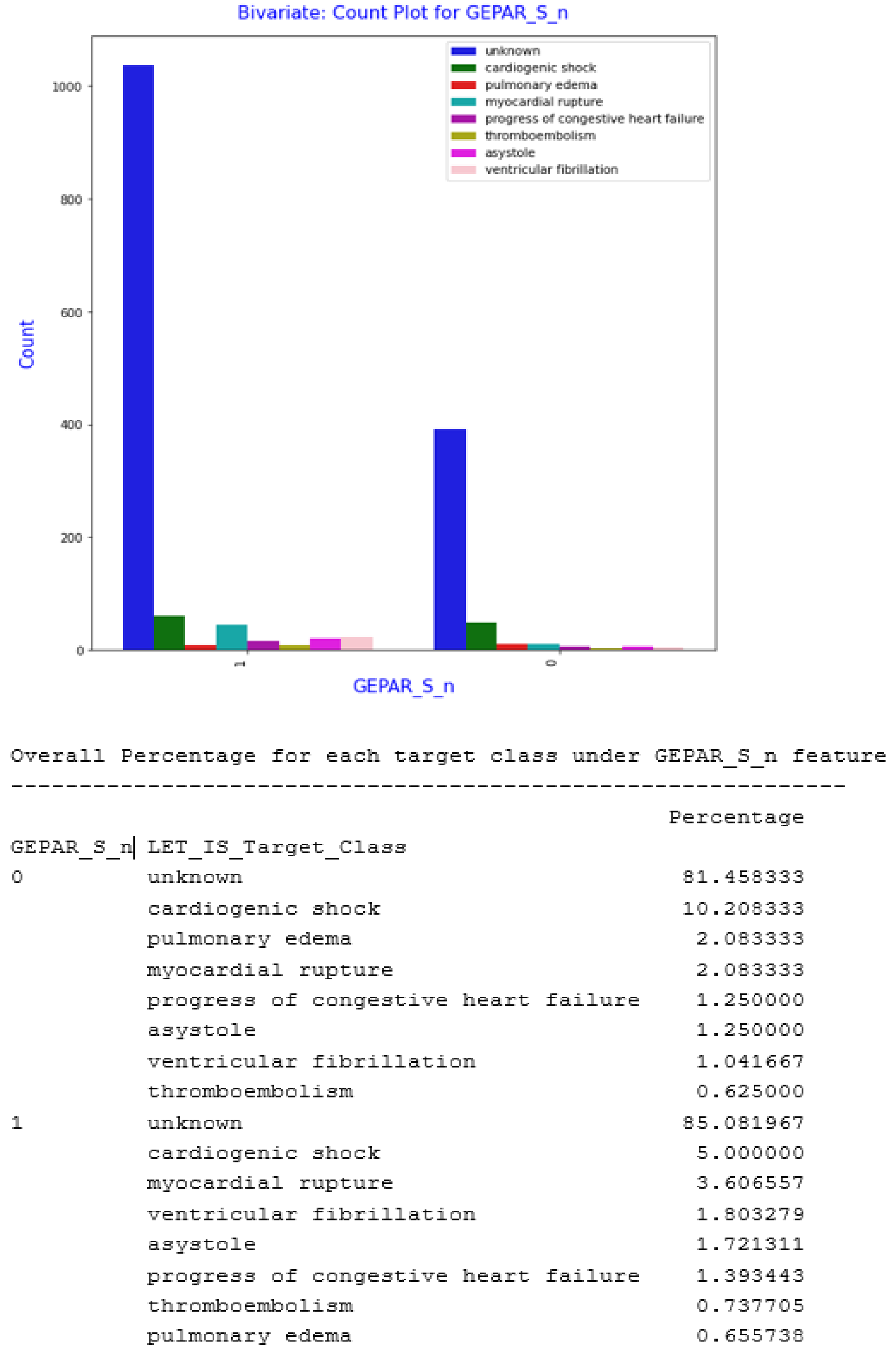

Overall Percentage for each target class under GEPAR_S_n feature

| GEPAR_S_n | LET_IS_Target_Class | Percentage |
|---|---|---|
| 0 | unknown | 81.458333 |
| | cardiogenic shock | 10.208333 |
| | pulmonary edema | 2.083333 |
| | myocardial rupture | 2.083333 |
| | progress of congestive heart failure | 1.250000 |
| | asystole | 1.250000 |
| | ventricular fibrillation | 1.041667 |
| | thromboembolism | 0.625000 |
| 1 | unknown | 85.081967 |
| | cardiogenic shock | 5.000000 |
| | myocardial rupture | 3.606557 |
| | ventricular fibrillation | 1.803279 |
| | asystole | 1.721311 |
| | progress of congestive heart failure | 1.393443 |
| | thromboembolism | 0.737705 |
| | pulmonary edema | 0.655738 |

*Figure 4.65 Countplot and percentage analysis of GEPAR_S_n feature for all target class*

### j) ASP_S_n (Use of acetylsalicylic acid in the ICU)

From below count plot and percentage analysis figure 4.66, 71.69% of patients who were not subjected to acetylsalicylic acid (decoded value:0) in ICU suffered fatal outcome of acute myocardial infarction due to unknown reasons while around 14.61% of them suffered fatal outcome of acute myocardial infarction due to cariogenic shock and 1.39% of them suffered fatal outcome of acute myocardial infarction due to asystole.

On the other hand, 88.25% of patients who were subjected to acetylsalicylic acid (decoded value:1) in ICU suffered fatal outcomes of acute myocardial infarction due to unknown reasons while 3.70% of patients suffered lethal outcome for this disease due to cardiogenic shock and 0.39% of them suffered lethal outcome for this disease due to thromboembolism.

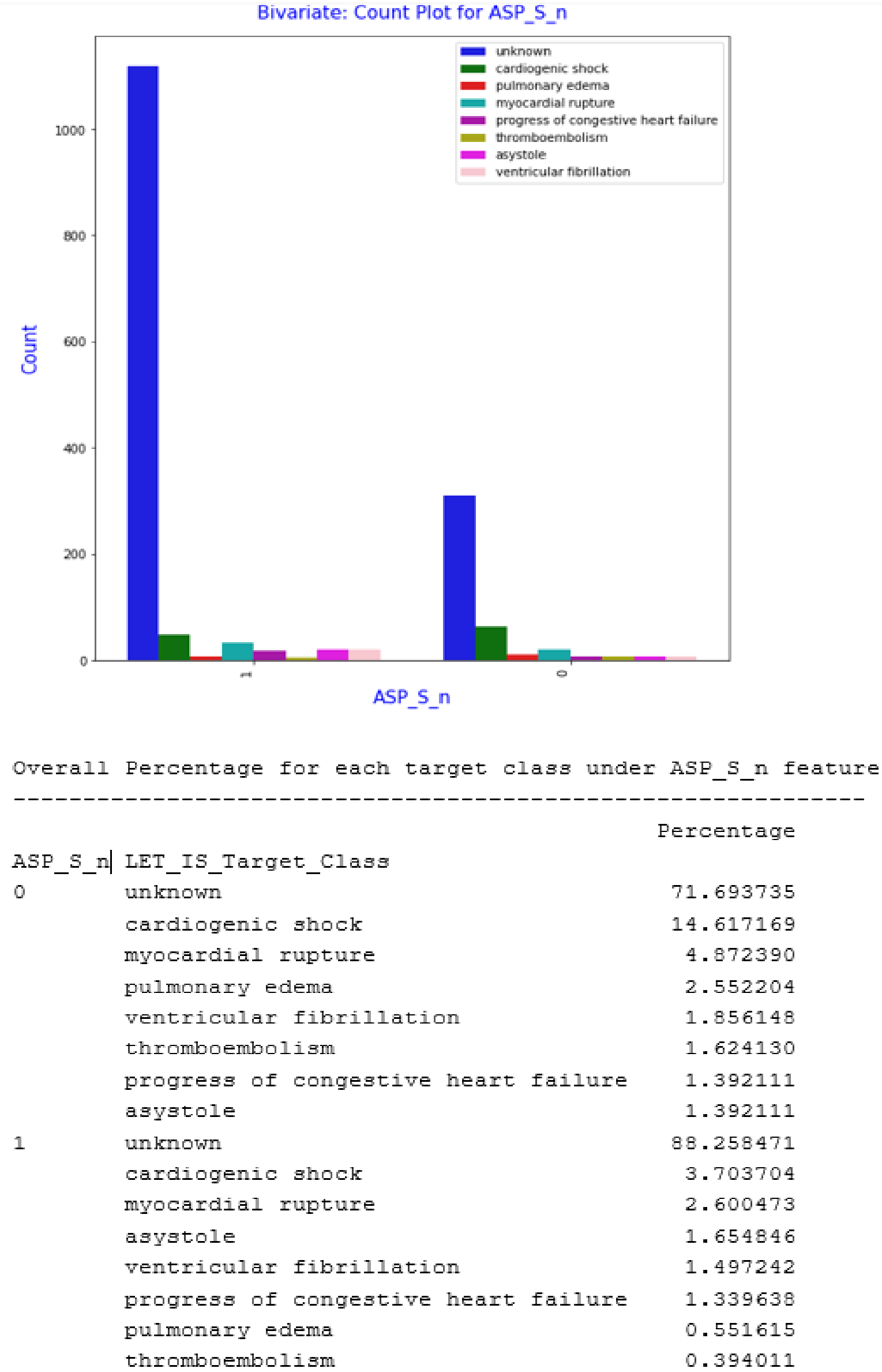

Overall Percentage for each target class under ASP_S_n feature

| ASP_S_n | LET_IS_Target_Class | Percentage |
|---|---|---|
| 0 | unknown | 71.693735 |
| | cardiogenic shock | 14.617169 |
| | myocardial rupture | 4.872390 |
| | pulmonary edema | 2.552204 |
| | ventricular fibrillation | 1.856148 |
| | thromboembolism | 1.624130 |
| | progress of congestive heart failure | 1.392111 |
| | asystole | 1.392111 |
| 1 | unknown | 88.258471 |
| | cardiogenic shock | 3.703704 |
| | myocardial rupture | 2.600473 |
| | asystole | 1.654846 |
| | ventricular fibrillation | 1.497242 |
| | progress of congestive heart failure | 1.339638 |
| | pulmonary edema | 0.551615 |
| | thromboembolism | 0.394011 |

*Figure 4.66 Countplot and percentage analysis of ASP_S_n feature for all target class*

### k) IBS_POST (CHD in recent weeks, days before admission to hospital)

From below count plot and percentage analysis figure 4.67, 91.14% of patients who were not having any coronary heart disease (decoded value:0) in before admission to hospital suffered fatal outcome of acute myocardial infarction due to unknown reasons while around 0.23% of them suffered fatal outcome of acute myocardial infarction due to pulmonary edema.

On the other hand, 87.59% of patients who were having exertional angina pectoris (decoded value:1) before admission to hospital suffered fatal outcomes of acute myocardial infarction due to unknown reasons while 0.72% of patients suffered lethal outcome for this disease due to thromboembolism.

Again, 77.38% of patients who were having unstable angina pectoris (decoded value:2) before admission to hospital suffered fatal outcomes of acute myocardial infarction due to unknown reasons while 0.68% of patients suffered lethal outcome for this disease due to thromboembolism.

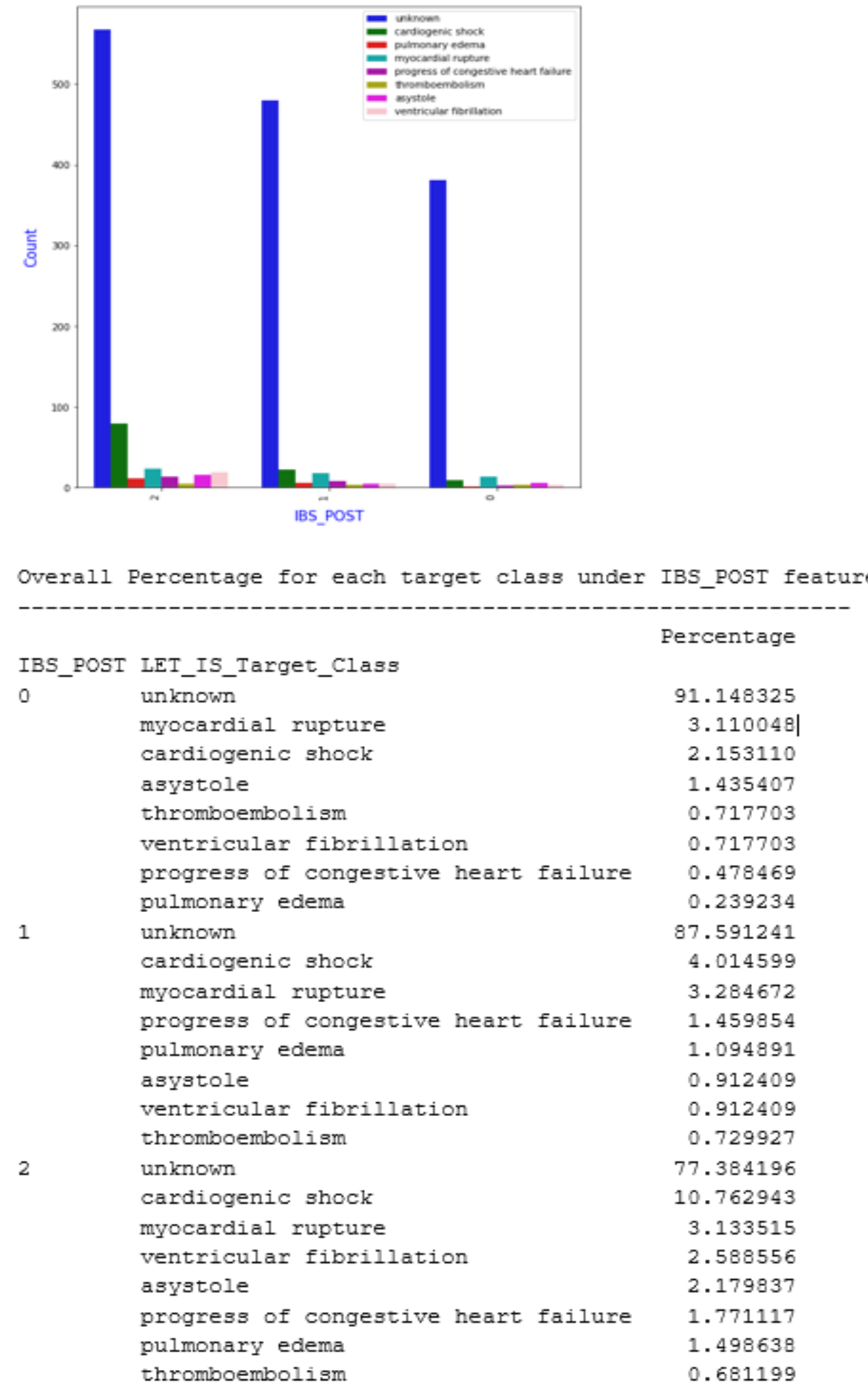

Overall Percentage for each target class under IBS_POST feature

| IBS_POST | LET_IS_Target_Class | Percentage |
|---|---|---|
| 0 | unknown | 91.148325 |
| | myocardial rupture | 3.110048 |
| | cardiogenic shock | 2.153110 |
| | asystole | 1.435407 |
| | thromboembolism | 0.717703 |
| | ventricular fibrillation | 0.717703 |
| | progress of congestive heart failure | 0.478469 |
| | pulmonary edema | 0.239234 |
| 1 | unknown | 87.591241 |
| | cardiogenic shock | 4.014599 |
| | myocardial rupture | 3.284672 |
| | progress of congestive heart failure | 1.459854 |
| | pulmonary edema | 1.094891 |
| | asystole | 0.912409 |
| | ventricular fibrillation | 0.912409 |
| | thromboembolism | 0.729927 |
| 2 | unknown | 77.384196 |
| | cardiogenic shock | 10.762943 |
| | myocardial rupture | 3.133515 |
| | ventricular fibrillation | 2.588556 |
| | asystole | 2.179837 |
| | progress of congestive heart failure | 1.771117 |
| | pulmonary edema | 1.498638 |
| | thromboembolism | 0.681199 |

*Figure 4.67 Countplot and percentage analysis of IBS_POST feature for all target class*

### l) FK_STENOK (Functional class of angina pectoris in the last year)

From below count plot and percentage analysis figure 4.68, 90.62% of patients who were not detected of any functional class of angina pectoris (decoded value:0) last year, suffered fatal outcome of acute myocardial infarction due to unknown reasons while around 0.30% of them suffered fatal outcome of acute myocardial infarction due to pulmonary edema.

On the other hand, 85.10% of patients who were of first functional class of angina pectoris (decoded value:1) last year, suffered fatal outcomes of acute myocardial infarction due to unknown reasons while 2.12% of patients suffered lethal outcome for this disease due to ventricular fibrillation.

Again, 80.15% of patients who were of second functional class of angina pectoris (decoded value:2) last year, suffered fatal outcomes of acute myocardial infarction due to unknown reasons while 0.53% of patients suffered lethal outcome for this disease due to thromboembolism.

Again, 66.66% of patients who were of third functional class of angina pectoris (decoded value:3) last year, suffered fatal outcomes of acute myocardial infarction due to unknown reasons while 1.85% of patients suffered lethal outcome for this disease due to thromboembolism.

Lastly, 100% of patients who were of fourth functional class of angina pectoris (decoded value:4) last year, suffered fatal outcomes of acute myocardial infarction due to unknown reasons.

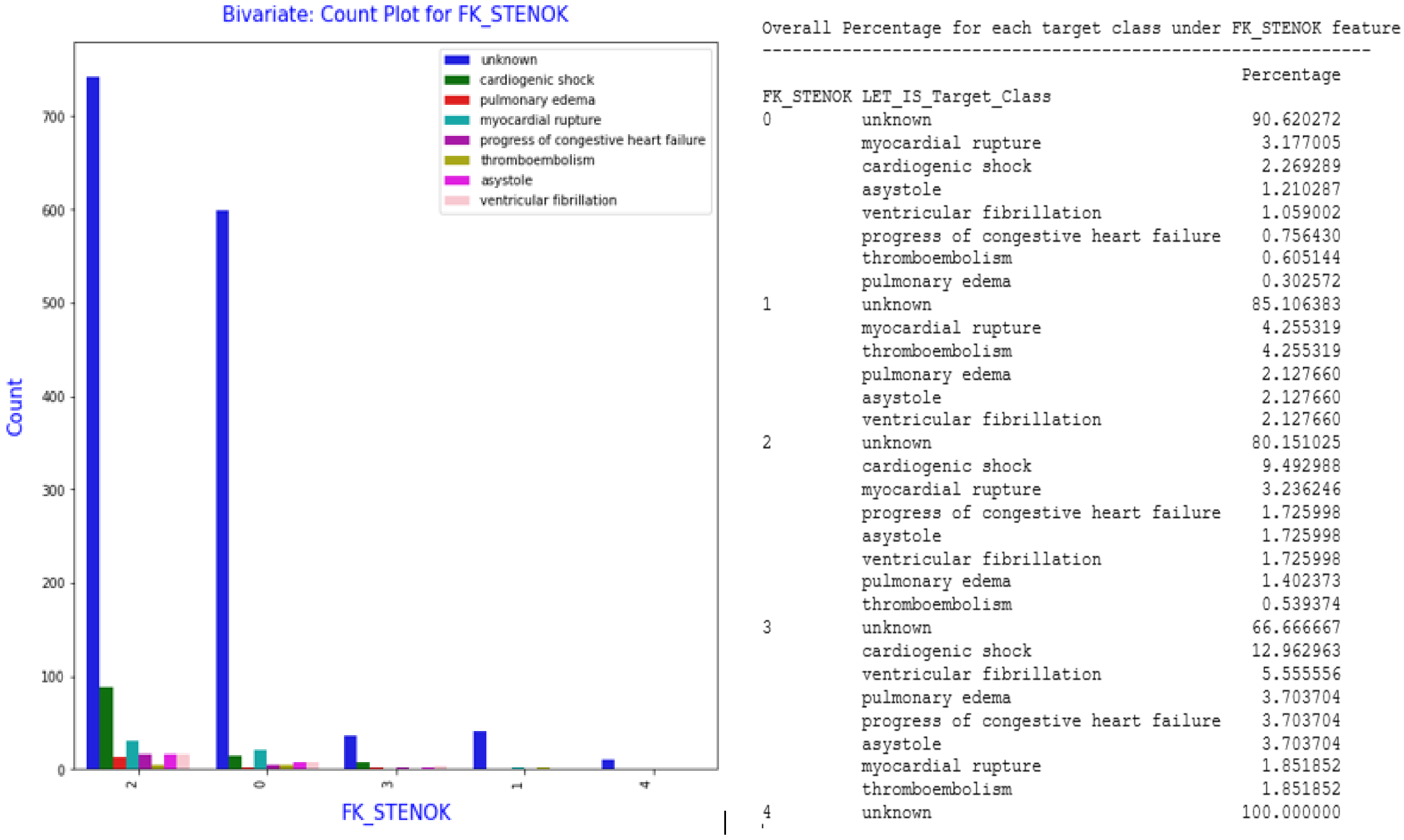

Overall Percentage for each target class under FK_STENOK feature

| FK_STENOK | LET_IS_Target_Class | Percentage |
|---|---|---|
| 0 | unknown | 90.620272 |
| | myocardial rupture | 3.177005 |
| | cardiogenic shock | 2.269289 |
| | asystole | 1.210287 |
| | ventricular fibrillation | 1.059002 |
| | progress of congestive heart failure | 0.756430 |
| | thromboembolism | 0.605144 |
| | pulmonary edema | 0.302572 |
| 1 | unknown | 85.106383 |
| | myocardial rupture | 4.255319 |
| | thromboembolism | 4.255319 |
| | pulmonary edema | 2.127660 |
| | asystole | 2.127660 |
| | ventricular fibrillation | 2.127660 |
| 2 | unknown | 80.151025 |
| | cardiogenic shock | 9.492988 |
| | myocardial rupture | 3.236246 |
| | progress of congestive heart failure | 1.725998 |
| | asystole | 1.725998 |
| | ventricular fibrillation | 1.725998 |
| | pulmonary edema | 1.402373 |
| | thromboembolism | 0.539374 |
| 3 | unknown | 66.666667 |
| | cardiogenic shock | 12.962963 |
| | ventricular fibrillation | 5.555556 |
| | pulmonary edema | 3.703704 |
| | progress of congestive heart failure | 3.703704 |
| | asystole | 3.703704 |
| | myocardial rupture | 1.851852 |
| | thromboembolism | 1.851852 |
| 4 | unknown | 100.000000 |

*Figure 4.68 Countplot and percentage analysis of FK_STENOK feature for all target class*

**m) ant_im (Presence of an anterior myocardial infarction (left ventricular))**

From below count plot and percentage analysis figure 4.69, 82.50% of patients who were not detected of left ventricular myocardial infarction (decoded value:0), suffered fatal outcome of acute myocardial infarction due to unknown reasons while around 0.94% of them suffered fatal outcome of acute myocardial infarction due to thromboembolism.

On the other hand, 95.91% of patients who were having QRS has no changes (decoded value:1), suffered fatal outcome of acute myocardial infarction due to unknown reasons while around 0.25% of them suffered fatal outcome of acute myocardial infarction due to pulmonary edema.

Again, 92.30% of patients who were having QRS is like QR-complex (decoded value:2), suffered fatal outcome of acute myocardial infarction due to unknown reasons while around 2.56% of them suffered fatal outcome of acute myocardial infarction due to myocardial rupture.

Again, 88.23% of patients who having QRS is like Qr-complex (decoded value:3), suffered fatal outcome of acute myocardial infarction due to unknown reasons while around 2.94% of them suffered fatal outcome of acute myocardial infarction due to ventricular fibrillation.

Lastly, 76.01% of patients who having QRS is like QS-complex (decoded value:4), suffered fatal outcome of acute myocardial infarction due to unknown reasons while around 0.40% of them suffered fatal outcome of acute myocardial infarction due to thromboembolism.

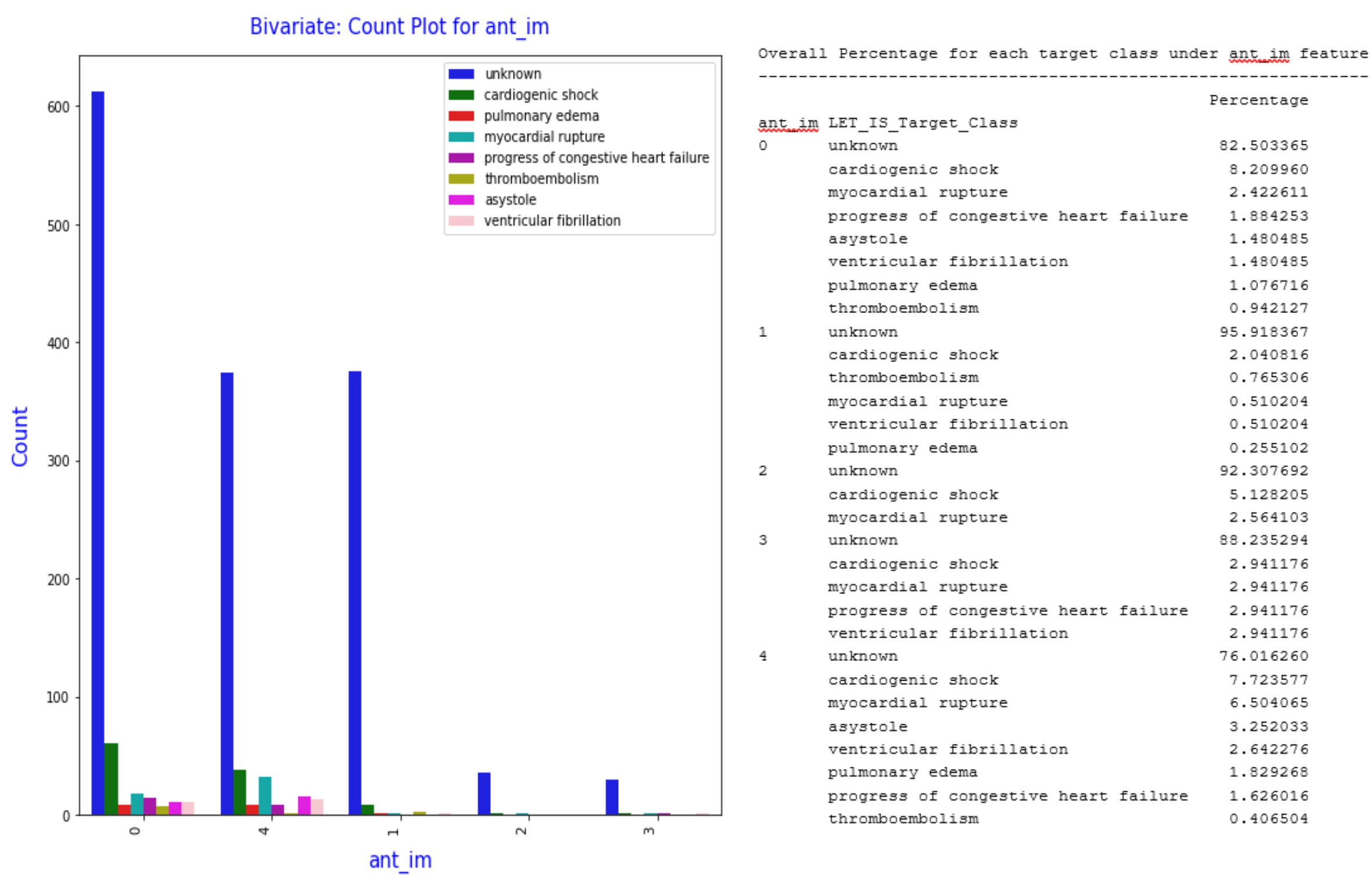

*Figure 4.69 Countplot and percentage analysis of ant_im feature for all target class*

### 4.4.3 Multivariate and Correlation Analysis

Multivariate analysis is looking at various factors (more than two) to see if there is any conceivable relationship between them. As a part of multivariate analysis Pearson correlations between the numerical features has been analysed in this study and top 10 positive and negative correlation has been reported accordingly.

Below table 4.34 shows the top 10 positive correlation between the features

*Table 4.36 Top 10 positive correlations between two features*

| Feature 1 | Feature 2 | Correlation |
|---|---|---|
| **D_AD_ORIT** | S_AD_ORIT | 0.861157 |
| **AST_BLOOD** | ALT_BLOOD | 0.522085 |
| **NA_BLOOD** | K_BLOOD | 0.301366 |
| **ROE** | AGE | 0.224937 |

| **LET_IS** | AGE | 0.157927 |
|---|---|---|
| **LET_IS** | L_BLOOD | 0.118314 |
| **LET_IS** | ROE | 0.090155 |
| **L_BLOOD** | AST_BLOOD | 0.084967 |
| **L_BLOOD** | ALT_BLOOD | 0.048407 |
| **AST_BLOOD** | K_BLOOD | 0.048212 |

From the above analysis, systolic and diastolic blood pressure has very strong positive correlation that means systolic pressure is positively linearly related to diastolic pressure, increase in systolic blood pressure will increase diastolic blood pressure as well. Rest of the features do not have that much stronger correlations between them.

Below table 4.35 shows the top 10 negative correlation between the features.

*Table 4.37 Top 10 negative correlations between two features*

| Feature 1 | Feature 2 | Correlation |
|---|---|---|
| **L_BLOOD** | D_AD_ORIT | -0.184411 |
| **L_BLOOD** | S_AD_ORIT | -0.174162 |
| **ALT_BLOOD** | S_AD_ORIT | -0.126439 |
| **AST_BLOOD** | S_AD_ORIT | -0.116237 |
| **ALT_BLOOD** | AGE | -0.112811 |
| **LET_IS** | S_AD_ORIT | -0.106933 |
| **LET_IS** | D_AD_ORIT | -0.099398 |
| **AST_BLOOD** | D_AD_ORIT | -0.087237 |
| **ALT_BLOOD** | D_AD_ORIT | -0.082208 |
| **AST_BLOOD** | AGE | -0.05646 |

From the above analysis, none of the features have those good negative correlations between them means none of them is negatively linearly related among each other.

Below figure 4.70 visualizes the heatmap which represent the correlations between different numerical features.

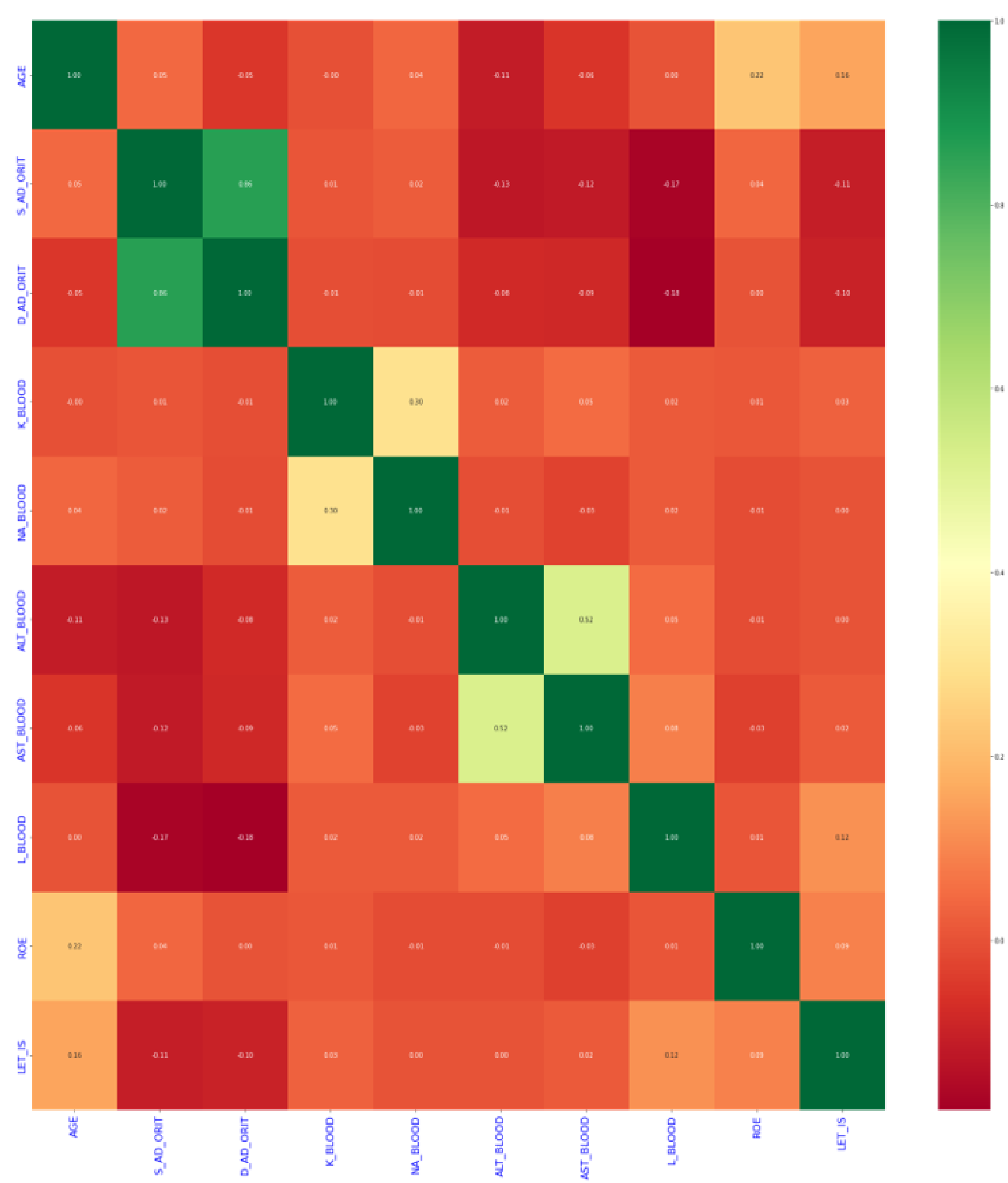

*Figure 4.70 Heatmap representing correlation between numerical features*

The pair plot in below figure 4.71 depicts the relationship between numerical features, and it can also be observed that all the numerical predictor variables are unable to explicitly cluster or identify the target variable namely lethal outcome (LET_IS) due to overlaps between them establishing a nonlinear decision boundary.

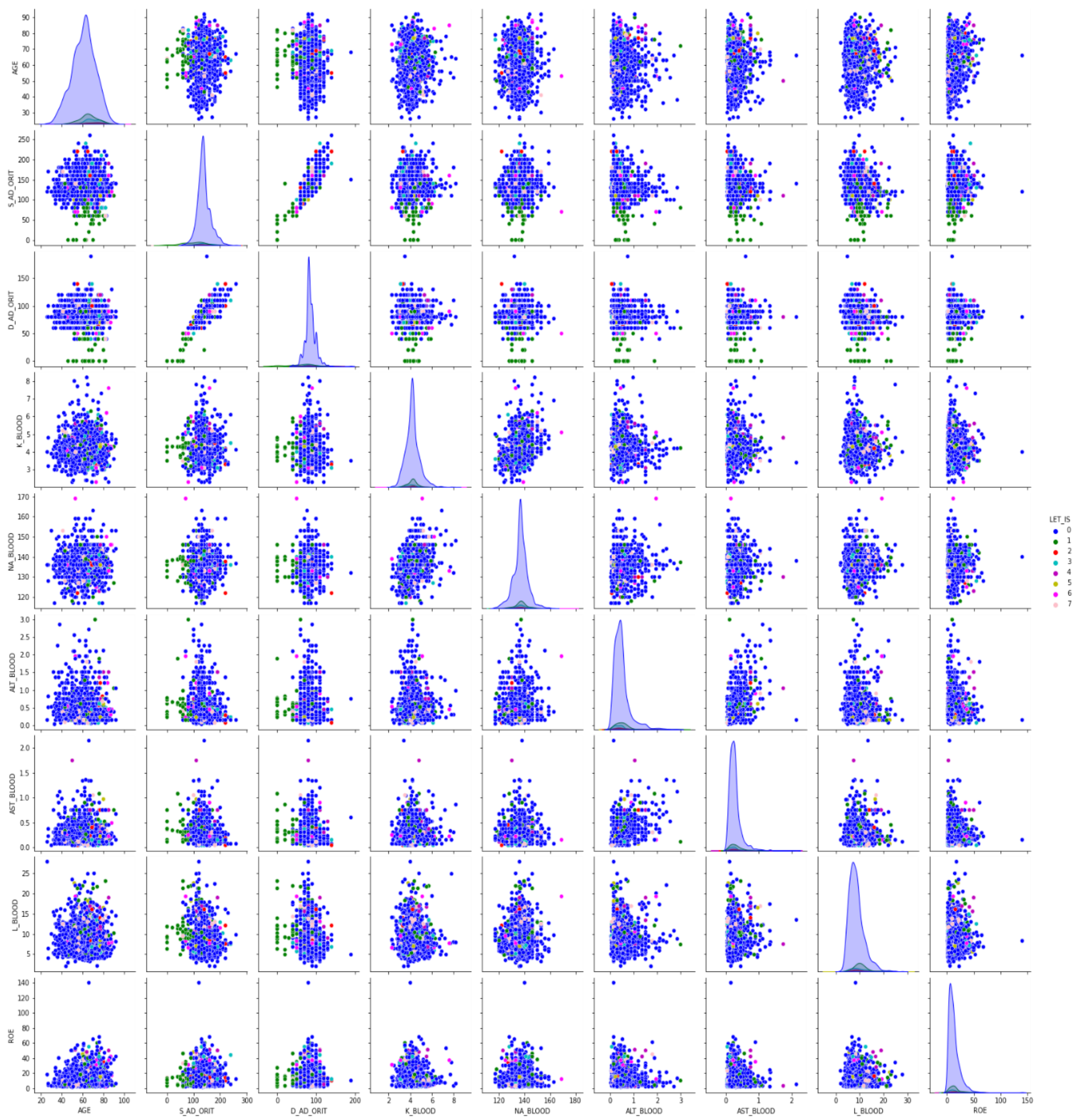

*Figure 4.71 Pairplot representing relation between numerical features*

## 4.5 Model Implementation

The procedures associated with implementing the classification algorithms to construct a prediction model for lethal outcomes of acute myocardial infarction for the dataset used in this study are depicted in below figure 4.72. Logistic regression, Light GBM, Random Forest, Bagging SVM, Stacking Blending, and Artificial neural network were among the six techniques used. The process flow of these classifiers is shown in the diagram below.

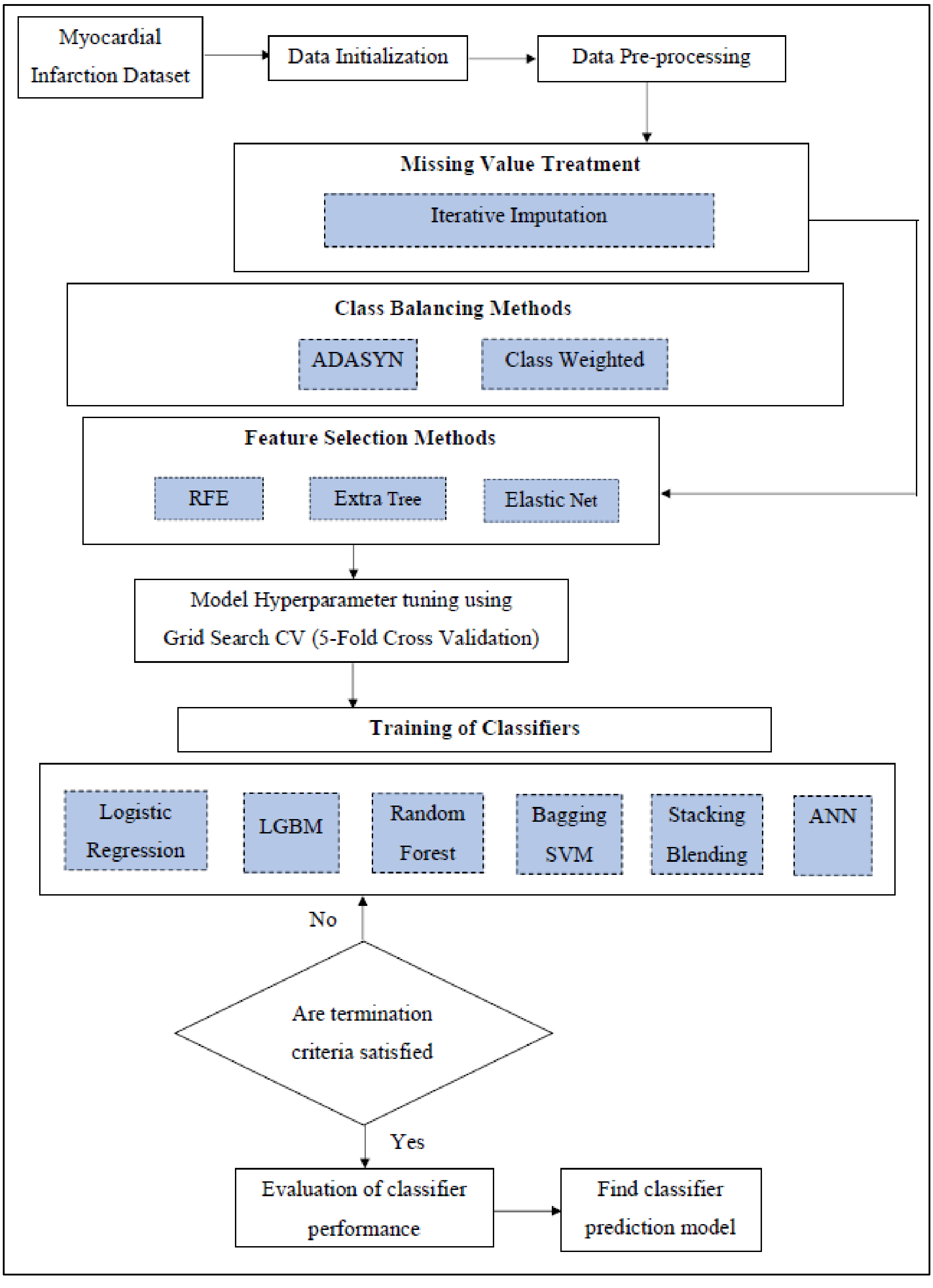

*Figure 4.72 Process flow diagram for classifiers*

### 4.5.1 Initialization of Myocardial Infarction Dataset

The whole myocardial infarction dataset in.csv format was imported and loaded into jupyter notebook using Python 3.8.5. There were 113 attributes and 1,700 instances when the data was loaded. Each independent attribute's type was shown, and all the variables in this dataset were a mix of category and numerical. As the target class variable, the variable LET_IS (lethal outcome) was used.

### 4.5.2 Data Pre-processing

The missing values, marked by '?' (converted to NaN or null), were examined and replaced. Except for the variables SEX, ID, and LET_IS (target variable), all variables have missing values, which were imputed using the iterative imputation technique (with hyperparameter max_iter= 50, random_state= 42, and rest are default) for all numerical attributes and simple imputation (using mode or most frequent hyperparameter) for categorical characteristics. The missing values were imputed after the entire dataset was imported as a.csv file into the local jupyter notebook. Also, one of the irrelevant features like 'ID' and features with more than 45% missing value (IBS_NASL, S_AD_KBRIG, D_AD_KBRIG, KFK_BLOOD) were dropped. Using the default split percentage, the entire dataset was then partitioned into training and test datasets.

### 4.5.3 Class Imbalance Handling

The class rebalancing algorithms were employed on a training dataset of 1,190 instances. These balancing procedures were used to address the target variable's class imbalance, LET_IS (lethal outcome). ADASYN (adaptive synthetic sampling) and the class weighted technique were used for class imbalance handling, as SVM-SMOTE has already been discarded from this study due to its inability to equally balance the target class in the presented myocardial infarction dataset, as shown in table 4.8 and figure 4.6. For ADASYN, the default hyperparameters were applied along with random state= 42 for imbalance management, and the class weighted imbalance handling hyperparameter (class_weight= 'balanced') was employed during machine learning model implementation. The use of these approaches modifies the number of instances in the training dataset. The model is also built using the imbalanced dataset as well for comparative analysis. The following lists the number of instances or observations in the training dataset for each of these techniques.

1) Training dataset before imbalance handling – 1,190 records (for model training)
2) Training dataset after imbalance handling using ADASYN – 8,010 records (for model training)
3) Training dataset for applying class weighted method - 1,190 records (for model training)

#### 4.5.4 Feature Selection Methods

The machine learning model's prediction power is directly proportional to the number of attributes or features utilised to train the model. When irrelevant features are employed to develop ML models, the model performance suffers. Feature Selection is the process of automatically or manually minimising the number of unnecessary features or dimensionality based on any statistical relationship. This may boost the ML models' performance or predictive potential. Three feature selection techniques along with their hyperparameters (refer to below table 4.36, table 4.37 and table 4.38) were used once before class imbalance handling and once after class imbalance treatment in this study for comparative analysis.

Below table 4.36 refers to hyperparameters that was used for recursive feature elimination technique.

*Table 4.38 RFE hyperparameter's name and values*

| Hyperparameter Name | Hyperparameter Value |
|---|---|
| estimator | LogisticRegression |
| n_features_to_select | 100 |

Below table 4.37 refers to hyperparameters that was used in extra tree classifier.

*Table 4.39 Extra tree hyperparameter's name and values*

| Hyperparameter Name | Hyperparameter Value |
|---|---|
| random_state | 42 |
| n_estimators | 1000 |

Below table 4.38 refers to hyperparameters that was used for elastic net regularization (via logistic regression).

*Table 4.40 Elastic net hyperparameter's name and values*

| Hyperparameter Name | Hyperparameter Value |
|---|---|
| random_state | 42 |
| multi_class | multinomial |
| penalty | elasticnet |
| solver | saga |
| l1_ratio | 0.2 |

Finally, as shown in figure 4.8 and 4.9, the top 70 independent characteristics were chosen from a total of 166 independent features from each dataset (before imbalance handling and after imbalance handling) for further processing and classifiers training/modelling.

### 4.5.5 Classifier's and Model Training

The classification model was trained using the techniques of logistic regression, light GBM, random forest, SVM bagging classifier, stacking blending, and artificial neural network on the training dataset, one including and one without different class balancing techniques implemented. The model was calibrated using Grid search CV and 5-fold cross validation, with one-fold used for testing and the remaining four for training. This technique was performed five times until each fold had been used as a test set once. Below subsections will provide a detail about the configurations and hyperparameters that was selected for the classifiers after a proper hyperparameter tunning which was then used for model training with and without class imbalance treatment.

#### 4.5.5.1 Logistic Regression Classifier Model Training

A simple multiclass logistic regression was used for implementing a classification model, this classifier was trained once without imbalance handling and once after class imbalance treatment (with ADASYN and class weighted method).

##### 4.5.5.1.1 Training without Class Imbalance Handling

Training dataset before imbalance handling consists of 1,190 records, which was used to train the model. Below table 4.39 refers to hyperparameters that was used for logistic regression model training without class imbalance handling, and rest of the hyperparameters were used as default.

*Table 4.41 Logistic regression's hyperparameters name and values without class imbalance handling*

| Hyperparameter Name | Hyperparameter Value |
|---|---|
| random_state | 42 |
| multi_class | multinomial |
| penalty | l2 |
| solver | saga |
| C | 1.0 |

#### 4.5.5.1.2 Training after Class Imbalance Handling (with ADASYN method)

Training dataset after imbalance handling consists of 8,010 records, which was used to train the model. Below table 4.40 refers to hyperparameters that was used for logistic regression model training after class imbalance handling with ADASYN imbalance treatment method, and rest of the hyperparameters were used as default.

*Table 4.42 Logistic regression's hyperparameters name and values after class imbalance handling*

| Hyperparameter Name | Hyperparameter Value |
|---|---|
| random_state | 42 |
| multi_class | multinomial |
| penalty | none |
| solver | sag |
| C | 0.6 |

#### 4.5.5.1.3 Training with Class Weighted Method on Imbalance Dataset

Training dataset before imbalance handling consists of 1,190 records, which was used to train the model. Below table 4.41 refers to hyperparameters that was used for logistic regression model training with class weighted method on imbalance dataset, and rest of the hyperparameters were used as default.

*Table 4.43 Logistic regression's hyperparameters name and values for class weighted method*

| Hyperparameter Name | Hyperparameter Value |
|---|---|
| random_state | 42 |

| multi_class | multinomial |
|---|---|
| penalty | none |
| solver | sag |
| C | 0.6 |
| class_weight | balanced |

#### 4.5.5.2 Light Gradient Boosting Machine Classifier Model Training

A light gradient boosting machine was used for implementing a classification model, this classifier was trained once without imbalance handling and once after class imbalance treatment (with ADASYN and class weighted method).

##### 4.5.5.2.1 Training without Class Imbalance Handling

Training dataset before imbalance handling consists of 1,190 records, which was used to train the model. Below table 4.42 refers to hyperparameters that was used for light gradient boosting machine model training without class imbalance handling, and rest of the hyperparameters were used as default.

*Table 4.44 LGBM hyperparameters name and values without class imbalance handling*

| Hyperparameter Name | Hyperparameter Value |
|---|---|
| random_state | 42 |
| objective | multiclass |
| colsample_bytree | 1 |
| max_depth | 20 |
| n_estimators | 600 |
| num_leaves | 50 |
| subsample | 0.1 |
| min_child_samples | 20 |
| learning_rate | 0.1 |

##### 4.5.5.2.2 Training after Class Imbalance Handling (with ADASYN method)

Training dataset after imbalance handling consists of 8,010 records, which was used to train the model. Below table 4.43 refers to hyperparameters that was used for light gradient boosting machine model training after class imbalance handling with ADASYN imbalance treatment method, and rest of the hyperparameters were used as default.

*Table 4.45 LGBM hyperparameters name and values after class imbalance handling*

| Hyperparameter Name | Hyperparameter Value |
|---|---|
| random_state | 42 |
| objective | multiclass |
| colsample_bytree | 1 |
| n_estimators | 10000 |
| reg_alpha | 50 |
| num_leaves | 50 |
| subsample | 0.1 |
| learning_rate | 0.2 |
| min_child_samples | 20 |

#### 4.5.5.2.3 Training with Class Weighted Method on Imbalance Dataset

Training dataset before imbalance handling consists of 1,190 records, which was used to train the model. Below table 4.44 refers to hyperparameters that was used for light gradient boosting machine model training with class weighted method on imbalance dataset, and rest hyperparameters were used as default.

*Table 4.46 LGBM hyperparameters name and values for class weighted method*

| Hyperparameter Name | Hyperparameter Value |
|---|---|
| random_state | 42 |
| objective | multiclass |
| colsample_bytree | 1 |
| max_depth | 50 |
| n_estimators | 1000 |
| class_weight | balanced |
| num_leaves | 50 |
| subsample | 0.1 |
| min_child_samples | 20 |
| learning_rate | 0.1 |

#### 4.5.5.3 Random Forest Classifier Model Training

A random forest algorithm was used for implementing a classification model, this classifier was trained once without imbalance handling and once after class imbalance treatment (with ADASYN and class weighted method).

##### 4.5.5.3.1 Training without Class Imbalance Handling

Training dataset before imbalance handling consists of 1,190 records, which was used to train the model. Below table 4.45 refers to hyperparameters that was used for random forest model training without class imbalance handling, and rest of the hyperparameters were used as default.

*Table 4.47 Random Forest hyperparameters name and values without class imbalance handling*

| Hyperparameter Name | Hyperparameter Value |
|---|---|
| random_state | 42 |
| max_features | 18 |
| max_depth | 6 |
| n_estimators | 50 |
| min_samples_leaf | 3 |
| min_samples_split | 70 |

##### 4.5.5.3.2 Training after Class Imbalance Handling (with ADASYN method)

Training dataset after imbalance handling consists of 8,010 records, which was used to train the model. Below table 4.46 refers to hyperparameters that was used for random forest model training after class imbalance handling with ADASYN imbalance treatment method, and rest of the hyperparameters were used as default.

*Table 4.48 Random Forest hyperparameters name and values after class imbalance handling*

| Hyperparameter Name | Hyperparameter Value |
|---|---|
| random_state | 42 |
| max_features | 12 |
| max_depth | 10 |
| n_estimators | 100 |
| min_samples_leaf | 3 |
| min_samples_split | 50 |

#### 4.5.5.3.3 Training with Class Weighted Method on Imbalance Dataset

Training dataset before imbalance handling consists of 1,190 records, which was used to train the model. Below table 4.47 refers to hyperparameters that was used for random forest model training with class weighted method on imbalance dataset, and rest hyperparameters were used as default.

*Table 4.49 Random Forest hyperparameters name and values for class weighted method*

| Hyperparameter Name | Hyperparameter Value |
|---|---|
| random_state | 42 |
| max_features | 17 |
| max_depth | 30 |
| n_estimators | 2000 |
| min_samples_leaf | 3 |
| min_samples_split | 70 |
| class_weight | balanced |

### 4.5.5.4 Bagging SVM Classifier Model Training

A SVM bagging classifier algorithm was used for implementing a classification model, this classifier was trained once without imbalance handling and once after class imbalance treatment (with ADASYN and class weighted method).

#### 4.5.5.4.1 Training without Class Imbalance Handling

Training dataset before imbalance handling consists of 1,190 records, which was used to train the model. Below table 4.48 refers to hyperparameters that was used SVM bagging classifier model training without class imbalance handling, and rest of the hyperparameters were used as default.

*Table 4.50 Bagging SVM hyperparameters name and values without class imbalance handling*

| Classifiers | Hyperparameter Name | Hyperparameter Value |
|---|---|---|
| SVC | random_state | 42 |
| | C | 1000 |
| | gamma | 0.001 |
| | degree | 3 |
| | decision_function_shape | ovr |

| | kernel | rbf |
|---|---|---|
| BaggingClassifier | base_estimator | SVC |
| | random_state | 42 |
| | n_estimators | 20 |

#### 4.5.5.4.2 Training after Class Imbalance Handling (with ADASYN method)

Training dataset after imbalance handling consists of 8,010 records, which was used to train the model. Below table 4.49 refers to hyperparameters that was used for SVM bagging classifier model training after class imbalance handling with ADASYN imbalance treatment method, and rest of the hyperparameters were used as default.

*Table 4.51 Bagging SVM hyperparameters name and values after class imbalance handling*

| Classifiers | Hyperparameter Name | Hyperparameter Value |
|---|---|---|
| SVC | random_state | 42 |
| | C | 1000 |
| | gamma | 0.001 |
| | degree | 3 |
| | decision_function_shape | ovr |
| | kernel | rbf |
| BaggingClassifier | base_estimator | SVC |
| | random_state | 42 |
| | n_estimators | 20 |

#### 4.5.5.4.3 Training with Class Weighted Method on Imbalance Dataset

Training dataset before imbalance handling consists of 1,190 records, which was used to train the model. Below table 4.50 refers to hyperparameters that was used for SVM bagging classifier model training with class weighted method on imbalance dataset, and rest hyperparameters were used as default.

*Table 4.52 Bagging SVM hyperparameters name and values for class weighted method*

| Classifiers | Hyperparameter Name | Hyperparameter Value |
|---|---|---|
| SVC | random_state | 42 |
| | C | 1000 |
| | gamma | 0.001 |
| | degree | 3 |
| | class_weight | balanced |
| | decision_function_shape | ovr |
| | kernel | rbf |
| BaggingClassifier | base_estimator | SVC |
| | random_state | 42 |
| | n_estimators | 1000 |

#### 4.5.5.5 Stacking and Blending Classifier Model Training

A stacking and blending classifier algorithm were used for implementing a classification model, this classifier was trained once without imbalance handling and once after class imbalance treatment (with ADASYN and class weighted method).

##### 4.5.5.5.1 Training without Class Imbalance Handling

Training dataset before imbalance handling consists of 1,190 records, which was used to train the model. Below table 4.51 refers to hyperparameters that was used for stacking blending classifier model training without class imbalance handling, and rest of the hyperparameters were used as default.

*Table 4.53 Stacking blending's hyperparameters name and values without class imbalance handling*

| Classifiers | Hyperparameter Name | Hyperparameter Value |
|---|---|---|
| StackingClassifier | final_estimator | LogisticRegression(multi_class='multinomial' ) |
| | estimators | LogisticRegression |
| | | LGBMClassifier |
| | | RandomForestClassifier |
| | | BaggingClassifier |
| | cv | StratifiedKFold(n_splits=5) |
| SVC | random_state | 42 |
| | C | 1000 |
| | gamma | 0.001 |
| | degree | 3 |

| | decision_function_shape | ovr |
|---|---|---|
| | kernel | rbf |
| BaggingClassifier | base_estimator | SVC |
| | random_state | 42 |
| | n_estimators | 20 |
| RandomForestClassifier | random_state | 42 |
| | max_features | 18 |
| | max_depth | 6 |
| | n_estimators | 50 |
| | min_samples_leaf | 3 |
| | min_samples_split | 70 |
| LGBMClassifier | random_state | 42 |
| | objective | multiclass |
| | colsample_bytree | 1 |
| | max_depth | 20 |
| | n_estimators | 600 |
| | num_leaves | 50 |
| | subsample | 0.1 |
| | min_child_samples | 20 |
| | learning_rate | 0.1 |
| LogisticRegression | random_state | 42 |
| | multi_class | multinomial |
| | penalty | l2 |
| | solver | saga |
| | C | 1.0 |

#### 4.5.5.5.2 Training after Class Imbalance Handling (with ADASYN method)

Training dataset after imbalance handling consists of 8,010 records, which was used to train the model. Below table 4.52 refers to hyperparameters that was used for stacking blending classifier model training after class imbalance handling with ADASYN imbalance treatment method, and rest of the hyperparameters were used as default.

*Table 4.54 Stacking blending's hyperparameters name and values after class imbalance handling*

| Classifiers | Hyperparameter Name | Hyperparameter Value |
|---|---|---|
| StackingClassifier | final_estimator | LogisticRegression(multi_class='multinomial') |
| | estimators | LogisticRegression |
| | | LGBMClassifier |
| | | RandomForestClassifier |
| | | BaggingClassifier |
| | cv | StratifiedKFold(n_splits=5) |
| SVC | random_state | 42 |
| | C | 1000 |
| | gamma | 0.001 |
| | degree | 3 |
| | decision_function_shape | ovr |
| | kernel | rbf |
| BaggingClassifier | base_estimator | SVC |
| | random_state | 42 |
| | n_estimators | 20 |
| RandomForestClassifier | random_state | 42 |
| | max_features | 12 |
| | max_depth | 10 |
| | n_estimators | 100 |
| | min_samples_leaf | 3 |
| | min_samples_split | 50 |
| LGBMClassifier | random_state | 42 |
| | objective | multiclass |
| | colsample_bytree | 1 |
| | n_estimators | 10000 |
| | reg_alpha | 50 |
| | num_leaves | 50 |
| | subsample | 0.1 |
| | learning_rate | 0.2 |
| | min_child_samples | 20 |
| LogisticRegression | random_state | 42 |
| | multi_class | multinomial |
| | penalty | none |
| | solver | sag |
| | C | 0.6 |

#### 4.5.5.5.3 Training with Class Weighted Method on Imbalance Dataset

Training dataset before imbalance handling consists of 1,190 records, which was used to train the model. Below table 4.53 refers to hyperparameters that was used for stacking blending classifier model training with class weighted method on imbalance dataset, and rest hyperparameters were used as default.

*Table 4.55 Bagging SVM hyperparameters name and values for class weighted method*

| Classifiers | Hyperparameter Name | Hyperparameter Value |
|---|---|---|
| StackingClassifier | final_estimator | LogisticRegression(multi_class='multinomial’ ) |
| | estimators | LogisticRegression |
| | | LGBMClassifier |
| | | RandomForestClassifier |
| | | BaggingClassifier |
| | cv | StratifiedKFold(n_splits=5) |
| SVC | random_state | 42 |
| | C | 1000 |
| | gamma | 0.001 |
| | degree | 3 |
| | decision_function_shape | ovr |
| | class_weight | balanced |
| | kernel | rbf |
| BaggingClassifier | base_estimator | SVC |
| | random_state | 42 |
| | n_estimators | 1000 |
| RandomForestClassifier | random_state | 42 |
| | max_features | 17 |
| | max_depth | 30 |
| | n_estimators | 2000 |
| | min_samples_leaf | 3 |
| | min_samples_split | 70 |
| LGBMClassifier | random_state | 42 |
| | objective | multiclass |
| | colsample_bytree | 1 |
| | max_depth | 50 |
| | n_estimators | 1000 |
| | num_leaves | 50 |
| | subsample | 0.1 |
| | min_child_samples | 20 |
| | class_weight | balanced |
| LogisticRegression | random_state | 42 |
| | multi_class | multinomial |
| | penalty | none |
| | solver | sag |
| | class_weight | balanced |
| | C | 0.6 |

#### 4.5.5.6 Artificial Neural Network Model Training

A deep artificial neural network algorithm was used for implementing a classification model, this classifier was trained once without imbalance handling and once after class imbalance treatment (with ADASYN and class weighted method).

##### 4.5.5.6.1 Training without Class Imbalance Handling

Training dataset before imbalance handling consists of 1,190 records, which was used to train the model. Below table 4.54 refers to hyperparameters that was used for artificial neural network model training without class imbalance handling, and rest of the hyperparameters were used as default.

*Table 4.56 ANN hyperparameters name and values without class imbalance handling*

| Layers | Hyperparameter Name | Hyperparameter Value |
|---|---|---|
| Sequential () | layers | none |
| | name | none |
| Dense | units | 128 |
| | input_dim | 70 |
| | activation | relu |
| Dense | units | 256 |
| | activation | relu |
| Dropout | rate | 0.2 |
| Dense | units | 8 |
| | activation | softmax |

#### 4.5.5.6.2 Training after Class Imbalance Handling (with ADASYN method)

Training dataset after imbalance handling consists of 8,010 records, which was used to train the model. Below table 4.55 refers to hyperparameters that was used for artificial neural network model training after class imbalance handling with ADASYN imbalance treatment method, and rest of the hyperparameters were used as default.

*Table 4.57 ANN hyperparameters name and values after class imbalance handling*

| Layers | Hyperparameter Name | Hyperparameter Value |
|---|---|---|
| Sequential () | layers | none |
| | name | none |
| Dense | units | 64 |
| | input_dim | 70 |
| | activation | relu |
| Dropout | rate | 0.4 |
| Dense | units | 64 |
| | activation | relu |
| Dropout | rate | 0.2 |
| Dense | units | 8 |
| | activation | softmax |

#### 4.5.5.6.3 Training with Class Weighted Method on Imbalance Dataset

Training dataset before imbalance handling consists of 1,190 records, which was used to train the model. Below table 4.56 refers to hyperparameters that was used for artificial neural network model training with class weighted method on imbalance dataset, and rest hyperparameters were used as default.

*Table 4.58 ANN hyperparameters name and values for class weighted method*

| Layers | Hyperparameter Name | Hyperparameter Value |
|---|---|---|
| Sequential () | layers | none |
| | name | none |
| Dense | units | 128 |
| | input_dim | 70 |
| | activation | relu |
| Dense | units | 256 |
| | activation | relu |
| Dropout | rate | 0.5 |
| Dense | units | 512 |
| | activation | relu |
| Dropout | rate | 0.5 |
| Dense | units | 8 |
| | activation | softmax |

### 4.5.6 Evaluation of Classifier's Performance

The performance evaluation criteria were used to evaluate the classifiers that were built using the training datasets. The best balancing strategy across all classifiers that were trained using a training dataset and evaluated using a test dataset will be selected based on performance measures. Next, the performance measurements will be used to evaluate all the classifiers developed from the chosen class-balanced dataset to determine the best predictive model. For multiclass class labels or multiclass classification, the confusion matrix for each classifier was utilized to extract various measures such as accuracy, weighted average F1 score, weighted average precision, weighted average recall, and the ROCAUC curve.

If class label No is considered as the negative case and class label Yes is considered as the positive case, the TP, FP, FN, and TN are presented in the sequence stated in below table 4.57 confusion matrix.

*Table 4.59 Generalize confusion matrix for binary classification*

| Confusion Matrix | | Predicted | |
|---|---|---|---|
| | | **No** | **Yes** |
| **Actual** | **No** | TN | FP |
| | **Yes** | FN | TP |

True-positive (TP)= the number of cases correctly defined as positive case

True-negative (TN)= the number of cases correctly defined as negative case

False-positive (FP)= the number of cases incorrectly defined as positive case

False-negative (FN)= the number of cases incorrectly defined as negative case

The classifiers were contrasted after evaluating all the performance parameters produced by the confusion matrix. The classifier with the overall best effectiveness was chosen as the final classification or prediction model for myocardial infarction's deadly result.

## 4.6 Summary

Essentially, this chapter describes the dataset used in this research as well as how the data was used to do analysis. The data on myocardial infarction in this study came from the online open-source UCI repository database, which contained 112 potential risk factors for myocardial infarction. The target variable, LET_IS (lethal outcome), indicates whether a patient died from one of the eight lethal outcomes of acute myocardial infarction. In this collection, there were 1,700 records available. To make the data easier to analyse, unrelated variables were removed, and categorical variables were turned into numerical variables using dummification to make the analysis and modelling process easier. Missing data were detected and imputed using the iterative/mode imputation technique.

The exploratory data analysis was carried out on the myocardial dataset with imputed missing values. A univariate analysis was used further to understand the distribution of the data in each variable. The overview of statistical analysis shown in frequency tables, as well as the count plot or percentage plot for a few key variables, were analysed in depth. The percentage analysis for each target class, as well as the variable relationships, were checked and presented as part of the bivariate and correlation analysis. Each visualisation report demonstrated if the predictor variable has any impact on causing any of the deadly outcomes of acute myocardial infarction. The report also shows the connection between the numerical features that determine if the combined relationship has any effect on the target variable, LET_IS (lethal outcome).

On the presented dataset, the wrapping and embedding methods of feature selection, namely RFE (recursive feature elimination) and extra tree classifier, were used to choose the most appropriate and relevant feature for classification model building before and class imbalance handling. The top 70 features were chosen as the most important elements in predicting the target class, and they were later used to develop a classification model.

The original dataset with imputed missing values was then divided into two sets, with the training set accounting for 70% and the test set for 30% using the default split. On the training dataset, a class balancing method, namely ADASYN, and a class weighted method, were used. The Grid search cross validation technique with five folds was used to tune the six classifiers. These classifiers were trained on both the imbalanced and class balanced training datasets, and they were then validated/evaluated against the test dataset. At the end of the project, 18 prediction models were created. Finally, the confusion matrix was shown for each of the eight classes to derive the model's performance measures. Finally, the algorithm that built the best overall predictive model on the class-balanced data was chosen based on the comparison of performances.

# CHAPTER 5

# RESULTS AND DISCUSSIONS

## 5.1 Introduction

This chapter will include all the findings from the analysis and output results from the implementation using Python (version 3.8.5) libraries which was used in this study. The results of the assessment of several modelling strategies with balanced and unbalanced data will be discussed in detail. The confusion matrices obtained from the development of the six classifiers (Logistic regression, LGBM, Bagging SVM, Random Forest, Stacking Blending, and ANN) using the two training datasets and evaluated with a test dataset utilising different class balancing methods will be showcased and inferred in more detail. Thus, each classification model will be made up of three confusion matrices derived from evaluating the models on a test set that was previously trained using the following datasets: a) before class imbalance treatment, b) after class imbalance treatment with the ADASYN method, and c) before class imbalance treatment with the class weighted method of handling imbalance.

This sub-chapter will interpret 18 confusion matrices, which will be generated by six classifiers. A table comparing the balancing techniques used on the training set and assessed on the test set will be displayed with the classifiers' performance measures for each classification algorithm. Because this is a multiclass classification problem, accuracy, weighted recall, weighted precision, weighted F-measure, and ROC curve for the eighth-class labels for target class LET_IS (lethal outcome) are all used. The class balancing strategy that results in the best classification model will be picked across all six classifiers based on this.

The final predictive model for predicting the lethal outcome (cause) of acute myocardial infarction will be chosen as the classifier with the best overall performance. Based on the balancing method and classification model, this study will highlight the best approach for predicting the lethal outcome (cause) of acute myocardial infarction.

### 5.2 Significant Biomarkers from Visualizations and Feature Selection

In Chapter 4, the various types of visualisations were interpreted to show the interplay of demographic and clinical factors with the target variable, LET_IS (lethal outcome). The connections between demographic and clinical characteristics were also plotted to see whether there were any patterns or trends between these characteristics and myocardial infarction. The interactions and visualisations which provide a critical and deeper insight into the myocardial infraction dataset were evaluated, which gives a clear picture of all the variables in the dataset as well as a summary of their relationships. These key biomarkers or variables will provide doctors with a quick overview of the essential factors associated with the fatal consequences of acute myocardial infraction and will act as a guideline for cardiac patients in understanding the impact of certain risk variables when predicting the cause of death due to acute myocardial infarction.

The important biomarkers or variables that were identified via feature selection (through RFE and extra tree) were then shown, providing a critical and deeper insight into the myocardial infraction dataset as well as a synopsis of their relationships, where a total of 1700 observations for individuals with myocardial infarction have been reported. During the visualization it has been observed that patients admitted to the hospital with this cardiovascular ailment have an average age of 61.85 years. According to research, a patient with this disease could be as young as 26 years old or as old as 92 years old. Also, it was discovered that a bigger number of people in the age group 60-70 years were suffering from lethal consequences of acute myocardial infarction, while very few people in the age groups 20-30 years and above 90 years were suffering from lethal outcomes of this disease. On the other hand, it has been observed that most patients aged 25 to 42 years old died because of this condition due to 'unknown' reasons (Target class 0). Furthermore, there is no significant positive or negative connection between patient age and any of the other features in this dataset.

A patient with this ailment might have 0.00 mmHg minimum systolic blood pressure and maximum of 260.00 mmHg, according to this analysis. Also, it can be observed that most of the patients those who are suffering from this disease having systolic blood pressure between 120 to 150 mmHg. Also, it has been observed that for the systolic blood pressure between 120-150 mmHg a greater number of people is suffering from lethal outcomes of acute myocardial infarction, while for the systolic blood pressure between 0-30 mmHg and above 240 mmHg very less number of people is suffering from lethal outcomes of acute myocardial infarction.

Also, low systolic blood pressure can be a good indicator of cardiogenic shock because for systolic pressure between 0.00 mmHg and 98 mmHg almost all cases belong to cardiogenic shock (Target class 1). In terms of relationships, systolic blood pressure follows a good positive linear relationship with diastolic blood pressure.

According to this analysis, a patient with this condition may have a minimum diastolic blood pressure of 0.00 mmHg and a maximum diastolic blood pressure of 190.00 mmHg. It has been discovered that most individuals with this condition had a diastolic blood pressure between 70 and 90 mmHg. It can also be shown that persons with diastolic blood pressure between 60 and 80 mmHg have a higher risk of dying from acute myocardial infarction, but those with diastolic blood pressure between 0-20 mmHg and beyond 140 mmHg have a much lower risk. Low diastolic blood pressure, just like systolic blood pressure, can be a good indicator of cardiogenic shock, because diastolic pressure between 0.00 mmHg and 58 mmHg almost always indicates cardiogenic shock (Target class 1). Diastolic blood pressure has a good positive linear relationship with systolic blood pressure in terms of correlations.

635 patients, or 37.35%, of the 1,700 patients who have died because of this condition are female, while the remaining 1065 cases, or 62.65%, are male. 8.50% of females died of acute myocardial infarction because of cariogenic shock, while 0.94% died of acute myocardial infarction because of pulmonary edema. On the other hand, 5.25% of male patients died because of cardiogenic shock, while 0.28% of male patients died because of thromboembolism.

Another, 95 patients, or 5.59%, had lethal outcomes and had atrial fibrillation (irregular ECG rhythm) at the time of admission to the hospital, while the remaining 1605 patients, or 94.41%, had lethal outcomes of acute myocardial infarction but did not have atrial fibrillation (irregular ECG rhythm) at the time of admission to the hospital. 85.23% of patients who did not have an arterial fibrillated ECG rhythm at the time of admission to the hospital died of acute myocardial infarction for unknown reasons, while 6.1% died of acute myocardial infarction due to cariogenic shock and 0.37% died of thromboembolism. On the other hand, 64.21% of patients who exhibited an arterial fibrillated ECG rhythm at the time of admission to the hospital died of acute myocardial infarction for unknown reasons, 12.63% died of cardiogenic shock, and 1.05% died of pulmonary edema.

Patients hospitalised to the hospital with this cardiovascular ailment had an average serum potassium value of 4.19 mmol/L, according to reports. According to this analysis, a patient with this condition may have a minimum serum potassium concentration of 2.300 mmol/L and a maximum of 8.200 mmol/L. It has been discovered that most individuals with this condition had serum potassium levels between 3.5 and 4.5 mmol/L. Patients with advanced congestive heart failure (Target class 4) have a serum potassium content of more than 4.30 mmol/L, which is higher than any other target class potassium content. Furthermore, the average serum potassium content for patients with pulmonary edema (Target class 2) is less than 4.00 mmol/L, which is lower than the potassium content of any other target class.

Patients hospitalised to the hospital with this cardiovascular ailment have an average serum AlAT concentration of 0.48 IU/L, according to reports. According to this analysis, a patient with this disease may have a minimum serum AlAT concentration of 0.030 IU/L and a maximum of 3.00 IU/L. It has been discovered that most individuals with this condition had serum AlAT levels ranging from 0.3 to 0.7 IU/L. The average serum AlAT content for patients with congestive heart failure (Target class 4) is more than 0.68 IU/L, which is significantly higher than any other target class A1AT serum content, suggesting that a high serum A1AT content could be a good indicator of congestive heart failure progression. Furthermore, the average serum AlAT level for myocardial rupture patients (Target class 3) is less than 0.4 IU/L, which is the lowest of any target class A1AT serum content.

Among 1,700 patients, 418 patients (or 24.59%) had lethal outcomes of this disease but had not had coronary heart disease in the previous weeks, days before admission to hospital, while 548 patients (or 32.24%) had lethal outcomes of this disease but had had exertional angina pectoris, a type of coronary heart disease in the previous weeks, days before admission to hospital, and the remaining 734 patients (or 43.18% ) had lethal outcomes of this disease who have already suffered from unstable angina pectoris in the previous weeks, days before admission to hospital.

Among 1,700 patients, 431 patients, or 25.35%, were not given acetylsalicylic acid in the ICU, while 1,269 patients, or 74.65%, were given acetylsalicylic acid in the ICU. In addition, it was discovered that patients who were not subjected to acetylsalicylic acid, 14.61% of them died from an acute myocardial infarction caused by cariogenic shock, and 1.39% died from an acute myocardial infarction caused by asystole. On the other side, patients who were subjected to acetylsalicylic acid, 3.70% of them died from this disease because of cardiogenic shock, and 0.39% died from thromboembolism.

A total of 1,700 observations for individuals who died because of an acute myocardial infarction have been documented. Among the 1,700 patients, 480 (28.24%) did not get anticoagulants (heparin) in the ICU, while 1,220 (71.76%) did. Around 10.20% of patients who were not given anticoagulants (heparin) in the ICU died of acute myocardial infarction because of cariogenic shock, while 0.62% died of acute myocardial infarction because of thromboembolism. Patients who were given anticoagulants (heparin) in the ICU, on the other hand, had a fatal outcome of acute myocardial infarction due to cardiogenic shock 5.00% of the time, and 0.65% of the time due to pulmonary edema.

Apart from the above key variables there are some more important features like functional class of angina pectoris in the last year, presence of an anterior myocardial infarction (left ventricular), pulmonary edema at the time of admission to ICU, Use of NSAIDs by the emergency cardiology team, complete RBBB on ECG at the time of admission to hospital, use of liquid nitrates in the ICU, and cardiogenic shock at the time of admission to ICU based on which doctors or physician can get a quick overview of the factors related with the fatal consequences of acute myocardial infraction and these features can act as a guideline for cardiac patients in understanding the impact of certain risk variables when predicting the cause of death for this disease.

## 5.3 Modelling Evaluations and Results on Test Dataset

The effectiveness of each classifier was assessed using the balancing approaches, and the classifier with the best overall performance was chosen as the best prediction model for acute myocardial infarction lethal results. The performance of the classifiers was evaluated using many assessment metrics, including the properly classified instances percentage or accuracy, weighted average F1 score, weighted average precision, weighted average recall, and the ROCAUC curve for these multiclass classification tasks. Because the myocardial infarction dataset used in this study contains health information, specific evaluation criteria like accuracy, sensitivity or recall, precision, and AUCROC area are critical in assessing the prediction model constructed using a classification algorithm.

The greater the true positive rate for an ideal disease prediction model, the better the forecast of the diseased patient. Any illness prediction model's goal is to do this. Higher true negative rates are sought as well, however this metric is less relevant than true positive rates. It is critical that the predictive model accurately detects the existence of the disease and avoids misdiagnosis. The harmful repercussions of false negative and false positive rates can be fatal to the patient, as acute myocardial infarction is a lethal disease, and the earlier the diagnosis is made, the better the patient's chances of survival. The lower the false positive and false negative rates, and the higher the true positive and true negative rates, the better the classification model's performance.

The main goal of this study is to correctly predict the lethal outcomes or causes of acute myocardial infarction in cardiac patients, and the fact that there are eight target classes or lethal outcomes that the model must correctly predict makes it a multiclass classification problem, therefore the important assessment metrics for this multiclass classification task would be weighted average F1 score, weighted average precision, weighted average recall, and the ROCAUC curve. Also, the individual target class's recall or precision or from each modelling approach, as well as the results of each balancing method, will be compared.

The number of correctly and wrongly classified cases described in the confusion matrix usually signifies by the four elements True-positive (TP), False-positive (FP), True-negative (TN), and False-negative (FN) values. Because this study is dealing with a multi-class classification problem, the TP, TN, FP, and FN values are not retrieved directly, as they would be in a binary classification task. In a multi-class problem, these TP, TN, FP, and FN values must be determined for each class. Below figure 5.1 represents a sample confusion matrix for multiclass classification from a myocardial infraction predictive model, which shows that the True positive (TP) value is the point at which the actual and predicted values are the same. The True Negative (TN) value for a class is the sum of all column and row values except the values of the class being computed. Excluding the TP value, the False-negative (FN) value for a class is the summation of the corresponding rows. Excluding the TP value, the False-positive (FP) value for a class is the total of the values in the corresponding columns. Also, below figure 5.1, depicts the TP, TN, FP, and FN in relation to the 'unknown' class (Class label 0). Similarly, the confusion matrix can be used to calculate these components for each class. In this multiclass classification research, this will be used to describe the confusion matrix.

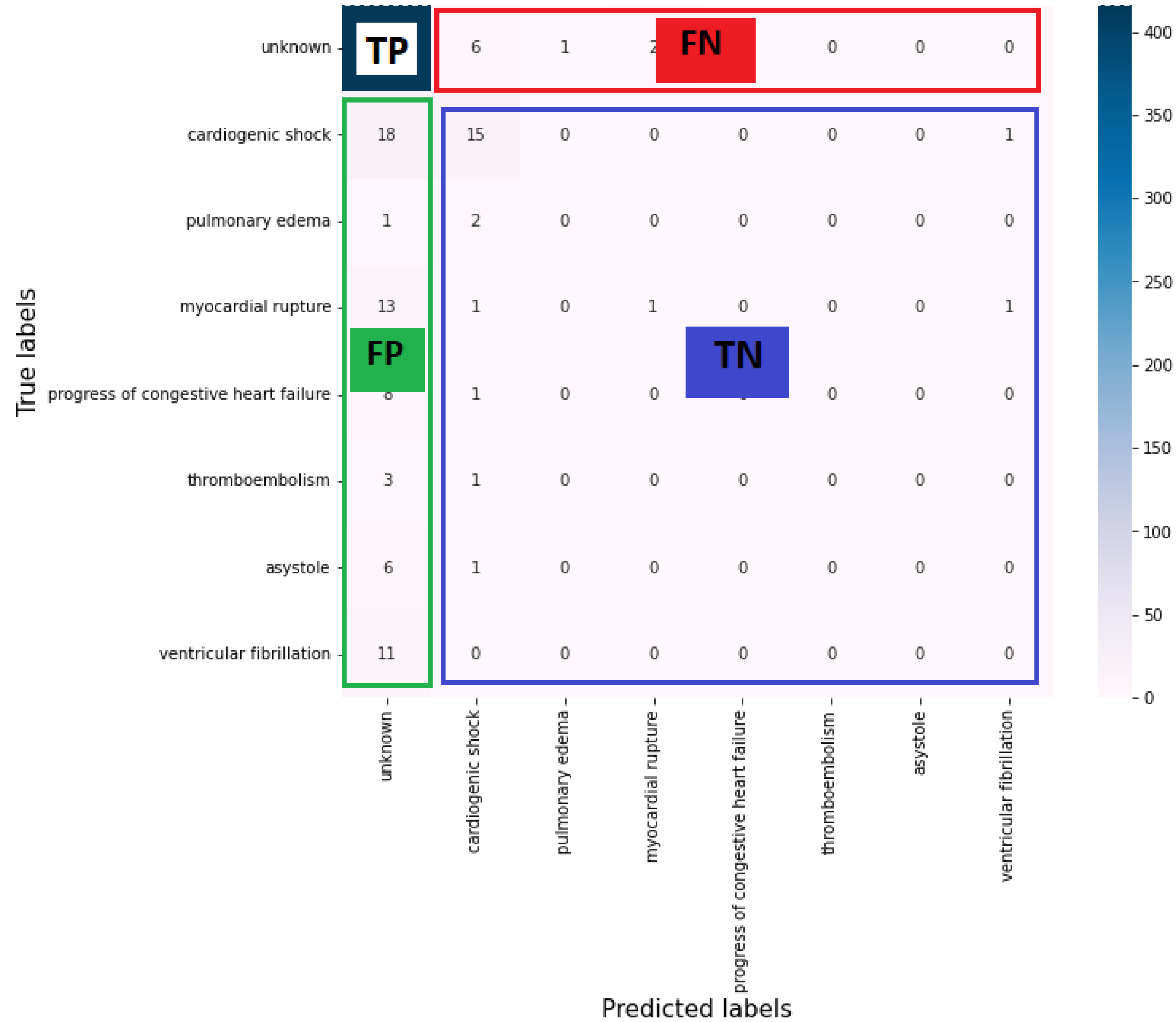

*Figure 5.1 A sample confusion matrix for multiclass classification from myocardial dataset*

The performance measures and confusion matrix for almost all classification models on training dataset after imbalance handling has given a good result. Training dataset after imbalance handling with ADASYN method consists of 8,010 records, which was used to train the model. Below figure 5.2 shows the confusion matrix from training dataset for one of the best classification models namely, random forest classification model on train dataset.

As part of the true positive (TP) counts, it can be shown that 927 cases out of 1003 were correctly predicted for the 'unknow' class label. For the 'cardiogenic shock' class label, 845 cases out of 983 were correctly predicted. For the 'pulmonary edema' class label, 992 cases were anticipated correctly out of 1006 cases. For the 'myocardial rupture' class label, 929 instances out of 996 were correctly predicted. For the 'progress of congestive heart failure' class label, 995 cases were correctly predicted out of 1005 cases. For the 'thromboembolism' class label, 984 cases were anticipated correctly out of 1004 cases. For the 'asystole' class label, 993 cases were anticipated correctly out of 1005 cases. Furthermore, 971 cases were predicted correctly for the 'ventricular fibrillation' class label out of 1008 cases.

As part of the false negative (FN) counts, 76 examples that truly correspond to the 'unknown' class label were forecasted as other class labels. 138 cases were projected as different class labels, despite the fact that they all belong to the 'cardiogenic shock' category. 14 cases, all of which correspond to the 'pulmonary edema' class label, were projected to have different class labels. 67 cases with the class label "myocardial rupture" were also predicted as other class labels. 10 instances were projected as different class labels, despite the fact that they all belong to the 'progress of congestive heart failure' class label. 20 examples, all of which fall under the 'thromboembolism' class label, were predicted to have different class labels. 12 instances with the class label 'asystole' were projected as different class labels, while 37 cases with the class label 'ventricular fibrillation' was also predicted as different class labels.

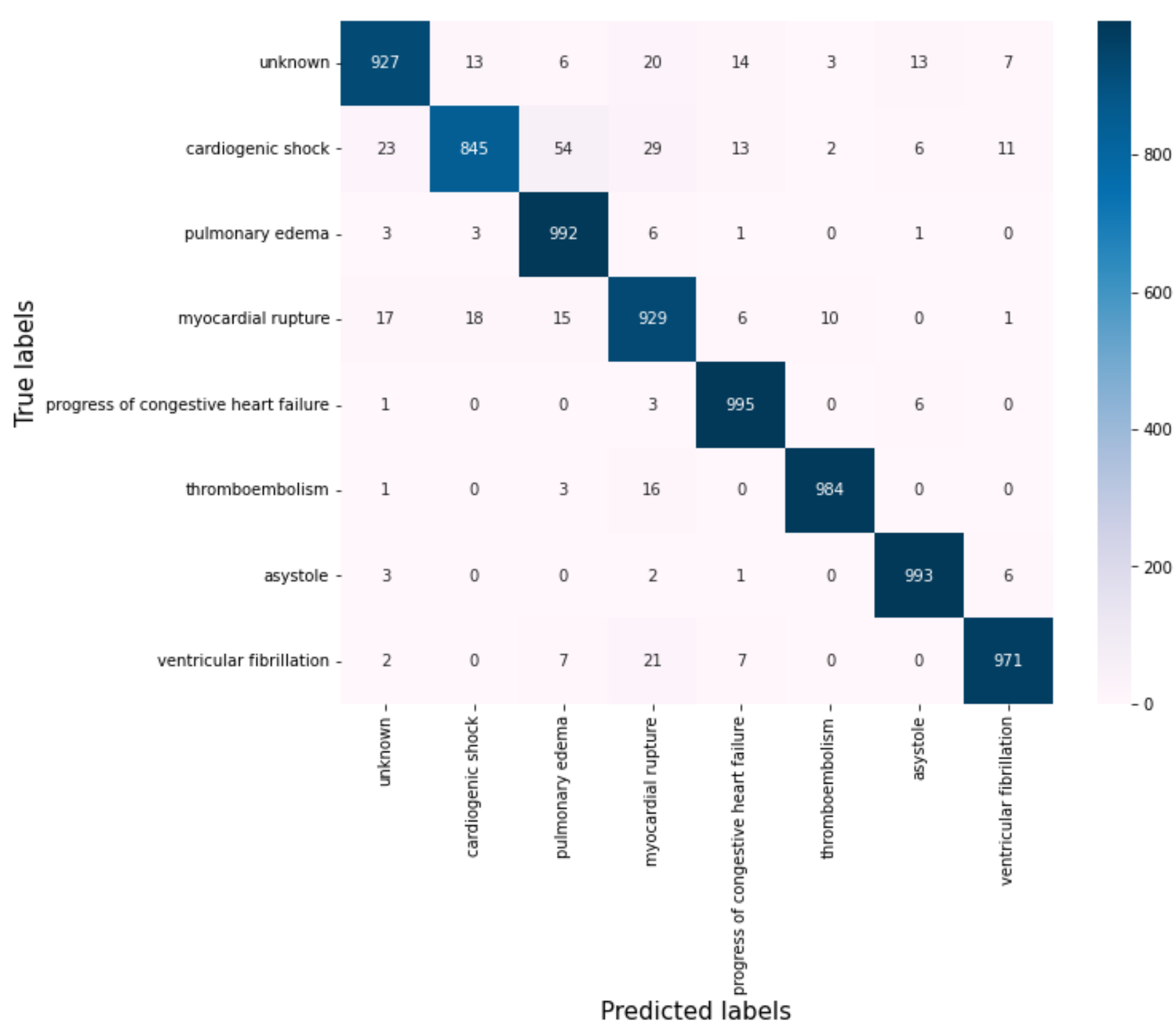

*Figure 5.2 A confusion matrix from train dataset for predicting lethal outcomes using random forest classifier after handling class imbalance*

### 5.3.1 Logistic Regression Classifier's Evaluation and Results

For building a classification model, a simple multiclass logistic regression was utilised; this classifier was trained twice, once without imbalance handling and once after class imbalance treatment (with ADASYN and class weighted method). The evaluations/results of the classifier on the test set, as well as performance metrics such as the confusion matrix, weighted average F1 score, weighted average precision, weighted average recall, and the AUCROC curve, will be shown in the subsections below. As mentioned earlier, the true positive (TP) counts and false negative (FN) counts are important to predict appropriately from confusion matrix in any of the medical diagnosis or prediction model. Hence the true positive counts and false negative counts will be the main area of focus along with the other performance metrices.

#### 5.3.1.1 Evaluation without Class Imbalance Handling

Training dataset before imbalance handling consists of 1,190 records, which was used to train the model. Test dataset consists of 510 records which was used to evaluate the model performance. Below figure 5.3 refers to the confusion matrix where rows are representing actual class labels and columns are representing predicted class labels.

As part of the true positive (TP) counts, it can be shown that 416 cases out of 426 were correctly predicted for the 'unknow' class label. For the 'cardiogenic shock' class label, 15 cases out of 34 were correctly predicted. For the 'pulmonary edema' class label, nothing (0 cases) was anticipated out of 3 cases. For the 'myocardial rupture' class label, 1 instance out of 16 was correctly predicted. For the 'progress of congestive heart failure' class label, nothing (0 cases) was predicted out of 9 cases. For the 'thromboembolism' class label, nothing (0 cases) was anticipated out of 4 cases. For the 'asystole' class label, nothing (0 cases) was anticipated out of 7 cases. Furthermore, nothing (0 cases) was predicted for the 'ventricular fibrillation' class label out of 11 cases.

As part of the false negative (FN) counts, 10 examples that truly correspond to the 'unknown' class label were forecasted as other class labels. 19 cases were projected as different class labels, despite the fact that they all belong to the 'cardiogenic shock' category. 3 cases, all of which correspond to the 'pulmonary edema' class label, were projected to have different class labels. 15 cases with the class label "myocardial rupture" were also predicted as other class labels. 9 instances were projected as different class labels, despite the fact that they all belong to the 'progress of congestive heart failure' class label. 4 examples, all of which fall under the 'thromboembolism' class label, were predicted to have different class labels. 7 instances with the class label 'asystole' were projected as different class labels, while 11 cases with the class label 'ventricular fibrillation' was also predicted as different class labels.

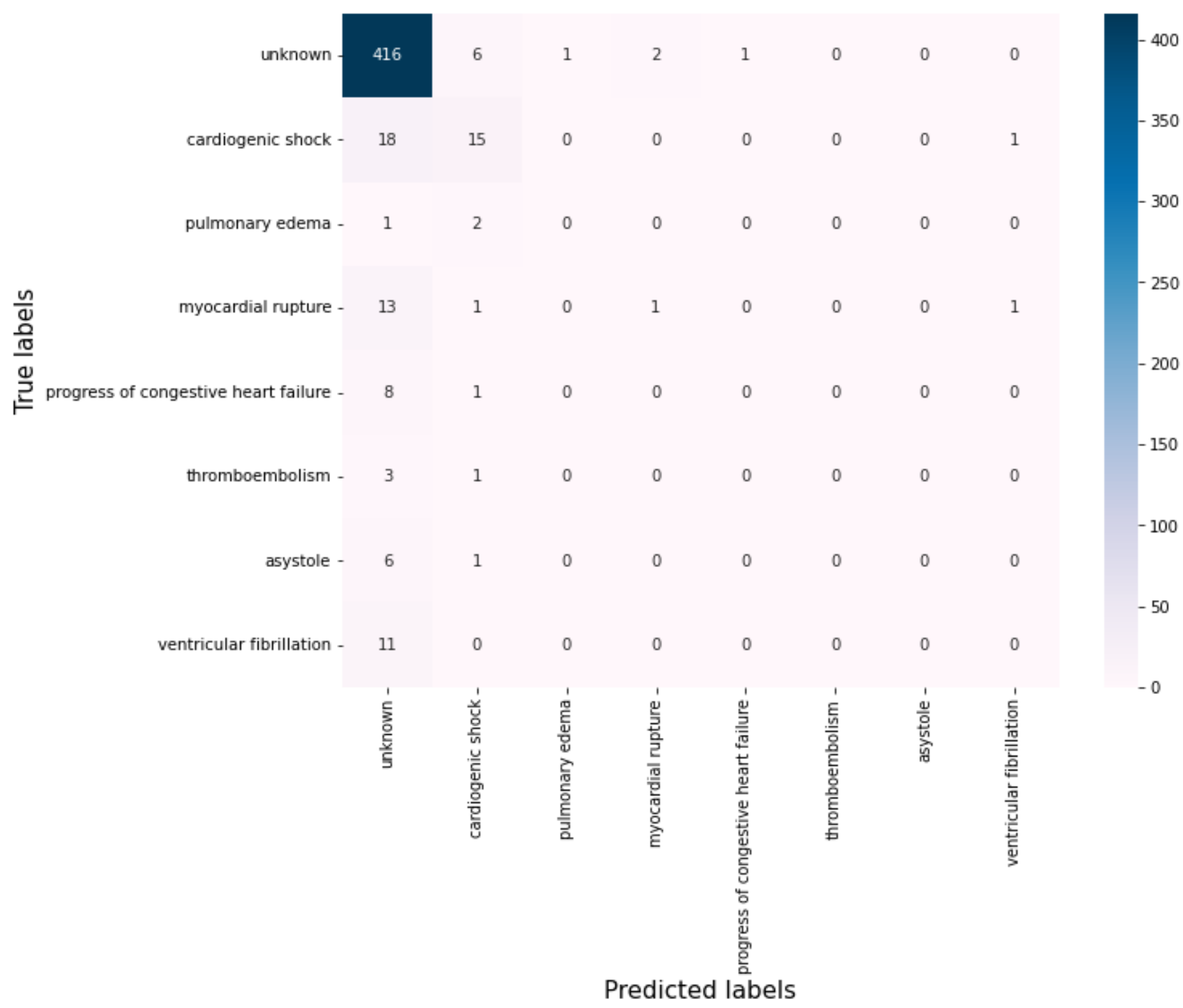

*Figure 5.3 A confusion matrix from logistic regression without class imbalance handling*

The classification report, shown in table 5.1, illustrates the various performance metrics for this multiclass classification task before class imbalance handling, as well as the precision, recall, and f1-score for each class. Additionally, the weighted average precision, weighted average recall, weighted average f1-score, and accuracy for the measuring overall model performance are 78%, 85%, 81%, and 85%, respectively.

*Table 5.1 The classification report from logistic regression without class imbalance handling*

<table>
<tr><th>Target class label</th><th>precision</th><th>recall</th><th>f1-score</th><th>Weighted avg precision</th><th>Weighted avg recall</th><th>Weighted avg f1-score</th><th>accuracy</th></tr>
<tr><td>unknown</td><td>0.87</td><td>0.98</td><td>0.92</td><td rowspan="4">0.78</td><td rowspan="4">0.85</td><td rowspan="4">0.81</td><td rowspan="4">0.85</td></tr>
<tr><td>cardiogenic shock</td><td>0.56</td><td>0.44</td><td>0.49</td></tr>
<tr><td>pulmonary edema</td><td>0</td><td>0</td><td>0</td></tr>
<tr><td>myocardial rupture</td><td>0.33</td><td>0.06</td><td>0.11</td></tr>
</table>

<table>
<tr><td>progress of congestive heart failure</td><td>0</td><td>0</td><td>0</td><td rowspan="4">0.78</td><td rowspan="4">0.85</td><td rowspan="4">0.81</td><td rowspan="4">0.85</td></tr>
<tr><td>thromboembolism</td><td>0</td><td>0</td><td>0</td></tr>
<tr><td>asystole</td><td>0</td><td>0</td><td>0</td></tr>
<tr><td>ventricular fibrillation</td><td>0</td><td>0</td><td>0</td></tr>
</table>

The ROC curve shown below in figure 5.4, illustrate how good the model is performing for predicting each class label before class imbalance handling. Also, the weighted average AUCROC score for the overall model is 85.29%.

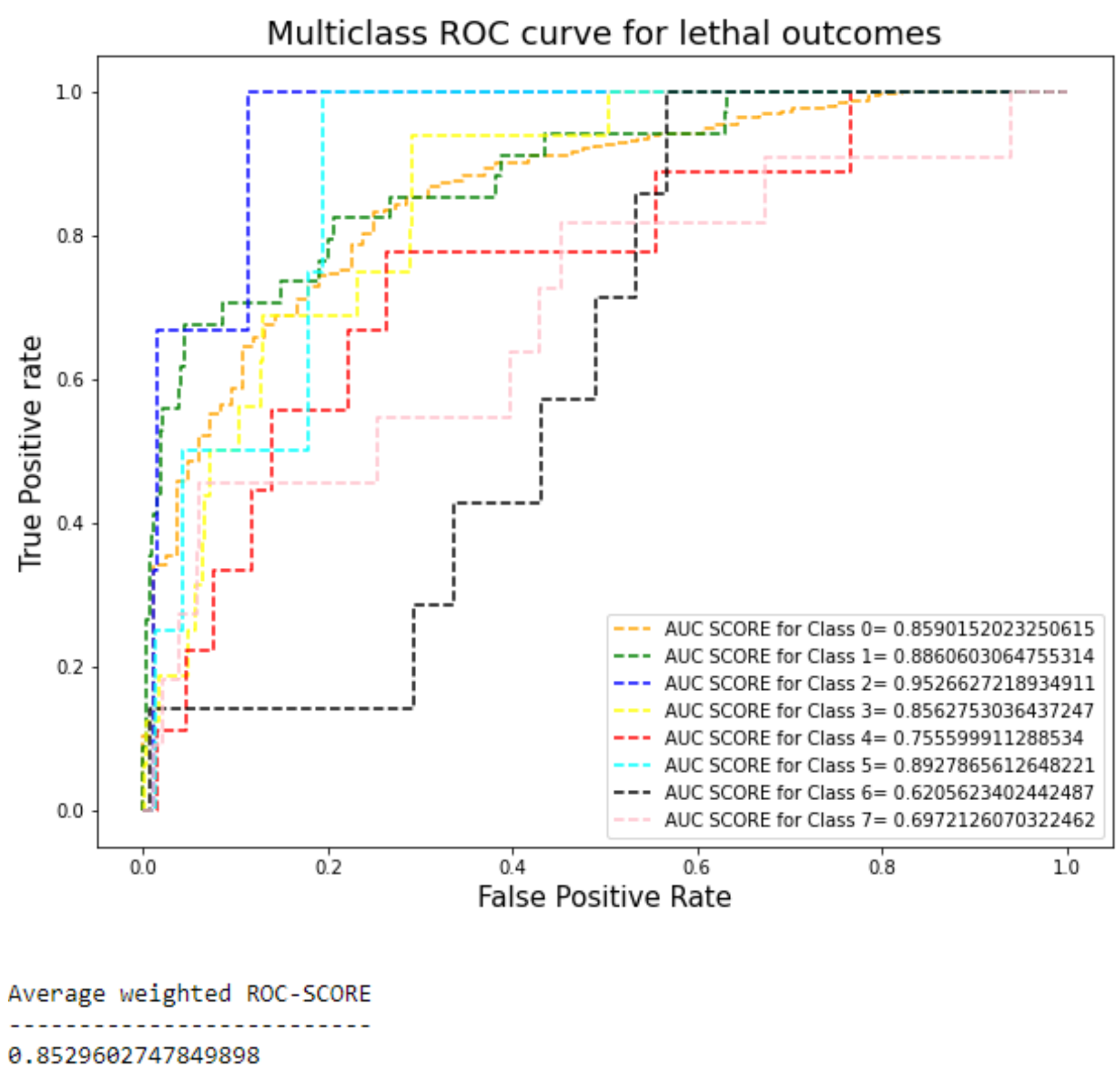

*Figure 5.4 AUCROC curve from logistic regression without class imbalance handling*

#### 5.3.1.2 Evaluation after Class Imbalance Handling (with ADASYN method)

Training dataset after imbalance handling with ADASYN method consists of 8,010 records, which was used to train the model. Test dataset consists of 510 records which was used to evaluate the model performance. Below figure 5.5 refers to the confusion matrix where rows are representing actual class labels and columns are representing predicted class labels.

As part of the true positive (TP) counts, it can be shown that 365 cases out of 426 were correctly predicted for the 'unknow' class label. For the 'cardiogenic shock' class label, 12 cases out of 34 were correctly predicted. For the 'pulmonary edema' class label, nothing (0 cases) was anticipated out of 3 cases. For the 'myocardial rupture' class label, nothing (0 cases) was anticipated out of 16 cases. For the 'progress of congestive heart failure' class label, 2 cases were predicted out of 9 cases. For the 'thromboembolism' class label, nothing (0 cases) was anticipated out of 4 cases. For the 'asystole' class label, 1 case was correctly predicted out of 7 cases. Furthermore, 1 case was predicted for the 'ventricular fibrillation' class label out of 11 cases.

As part of the false negative (FN) counts, 61 examples that truly correspond to the 'unknown' class label were forecasted as other class labels. 22 cases were projected as different class labels, despite the fact that they all belong to the 'cardiogenic shock' category. 3 cases, all of which correspond to the 'pulmonary edema' class label, were projected to have different class labels. 16 cases with the class label "myocardial rupture" were also predicted as other class labels. 7 instances were projected as different class labels, despite the fact that they all belong to the 'progress of congestive heart failure' class label. 4 examples, all of which fall under the 'thromboembolism' class label, were predicted to have different class labels. 6 instances with the class label 'asystole' were projected as different class labels, while 10 cases with the class label 'ventricular fibrillation' was also predicted as different class labels.

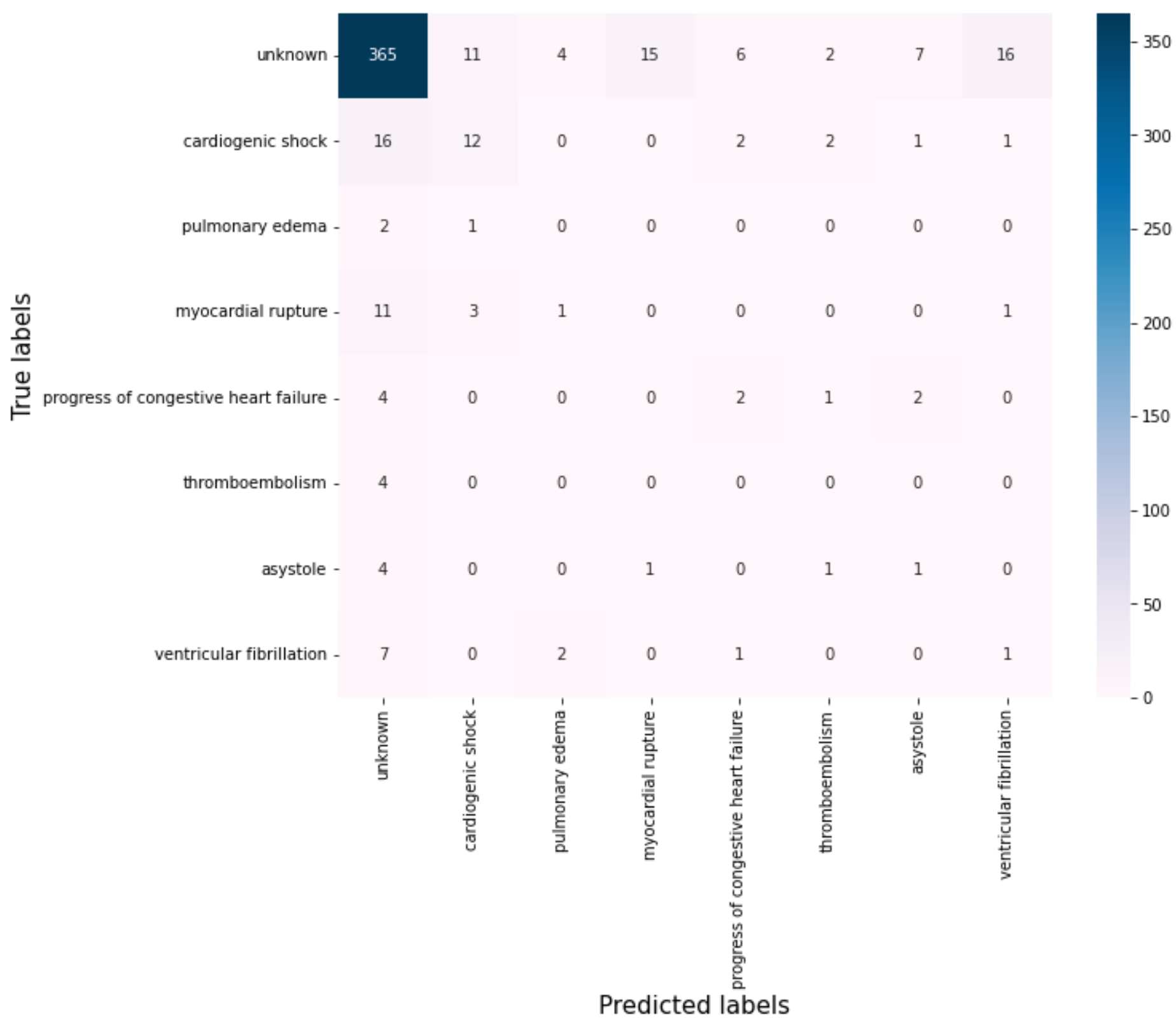

*Figure 5.5 A confusion matrix from logistic regression after class imbalance handling (ADASYN)*

The classification report, shown in table 5.2, illustrates the various performance metrics for this multiclass classification task after class imbalance treatment, as well as the precision, recall, and f1-score for each class. Additionally, the weighted average precision, weighted average recall, weighted average f1-score, and accuracy for the measuring overall model performance are 77%, 75%, 76%, and 75%, respectively.

*Table 5.2 The classification report from logistic regression after class imbalance handling (ADASYN)*

| **Target class label** | **precision** | **recall** | **f1-score** | **Weighted avg precision** | **Weighted avg recall** | **Weighted avg f1-score** | **accuracy** |
|---|---|---|---|---|---|---|---|
| unknown | 0.88 | 0.86 | 0.87 | **0.77** | **0.75** | **0.76** | **0.75** |
| cardiogenic shock | 0.44 | 0.35 | 0.39 | | | | |
| pulmonary edema | 0.00 | 0.00 | 0.00 | | | | |
| myocardial rupture | 0.00 | 0.00 | 0.00 | | | | |

| progress of congestive heart failure | 0.18 | 0.22 | 0.20 | **0.77** | **0.75** | **0.76** | **0.75** |
|---|---|---|---|---|---|---|---|
| thromboembolism | 0.00 | 0.00 | 0.00 | | | | |
| asystole | 0.09 | 0.14 | 0.11 | | | | |
| ventricular fibrillation | 0.05 | 0.09 | 0.07 | | | | |

The ROC curve shown below in figure 5.6, illustrate how good the model is performing for predicting each class label after class imbalance treatment. Also, the weighted average AUCROC score for the overall model is 73.75%.

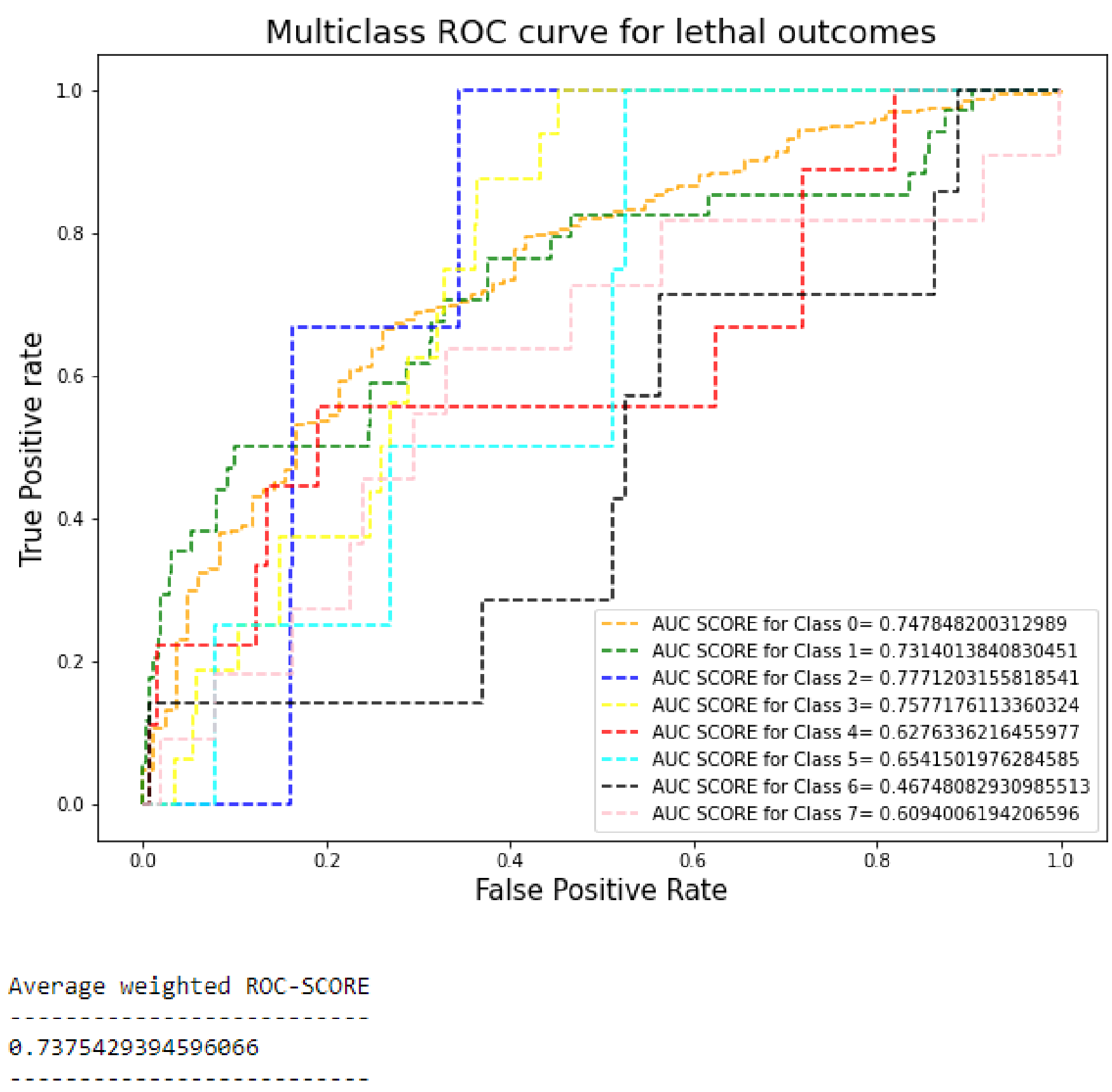

*Figure 5.6 AUCROC curve from logistic regression after class imbalance handling (ADASYN)*

#### 5.3.1.3 Evaluation after applying Class Weighted Method

Training dataset before imbalance handling consists of 1,190 records, which was used to train the model with class weighted hyperparameter (class_weight=’balanced’) to handle the imbalance class. Test dataset consists of 510 records which was used to evaluate the model performance. Below figure 5.7 refers to the confusion matrix where rows are representing actual class labels and columns are representing predicted class labels.

As part of the true positive (TP) counts, it can be shown that 312 cases out of 426 were correctly predicted for the 'unknow' class label. For the 'cardiogenic shock' class label, 15 cases out of 34 were correctly predicted. For the 'pulmonary edema' class label, nothing (0 cases) was anticipated out of 3 cases. For the 'myocardial rupture' class label, 4 cases were predicted correctly out of 16 cases. For the 'progress of congestive heart failure' class label, nothing (0 cases) was anticipated out of 9 cases. For the 'thromboembolism' class label, nothing (0 cases) was anticipated out of 4 cases. For the 'asystole' class label, 1 case was predicted correctly out of 7 cases. Furthermore, 1 case was predicted for the 'ventricular fibrillation' class label out of 11 cases.

As part of the false negative (FN) counts, 114 examples that truly correspond to the 'unknown' class label were forecasted as other class labels. 19 cases were projected as different class labels, despite the fact that they all belong to the 'cardiogenic shock' category. 3 cases, all of which correspond to the 'pulmonary edema' class label, were projected to have different class labels. 12 cases with the class label "myocardial rupture" were also predicted as other class labels. 9 instances were projected as different class labels, despite the fact that they all belong to the 'progress of congestive heart failure' class label. 4 examples, all of which fall under the 'thromboembolism' class label, were predicted to have different class labels. 6 instances with the class label 'asystole' were projected as different class labels, while 10 cases with the class label 'ventricular fibrillation' was also predicted as different class labels.

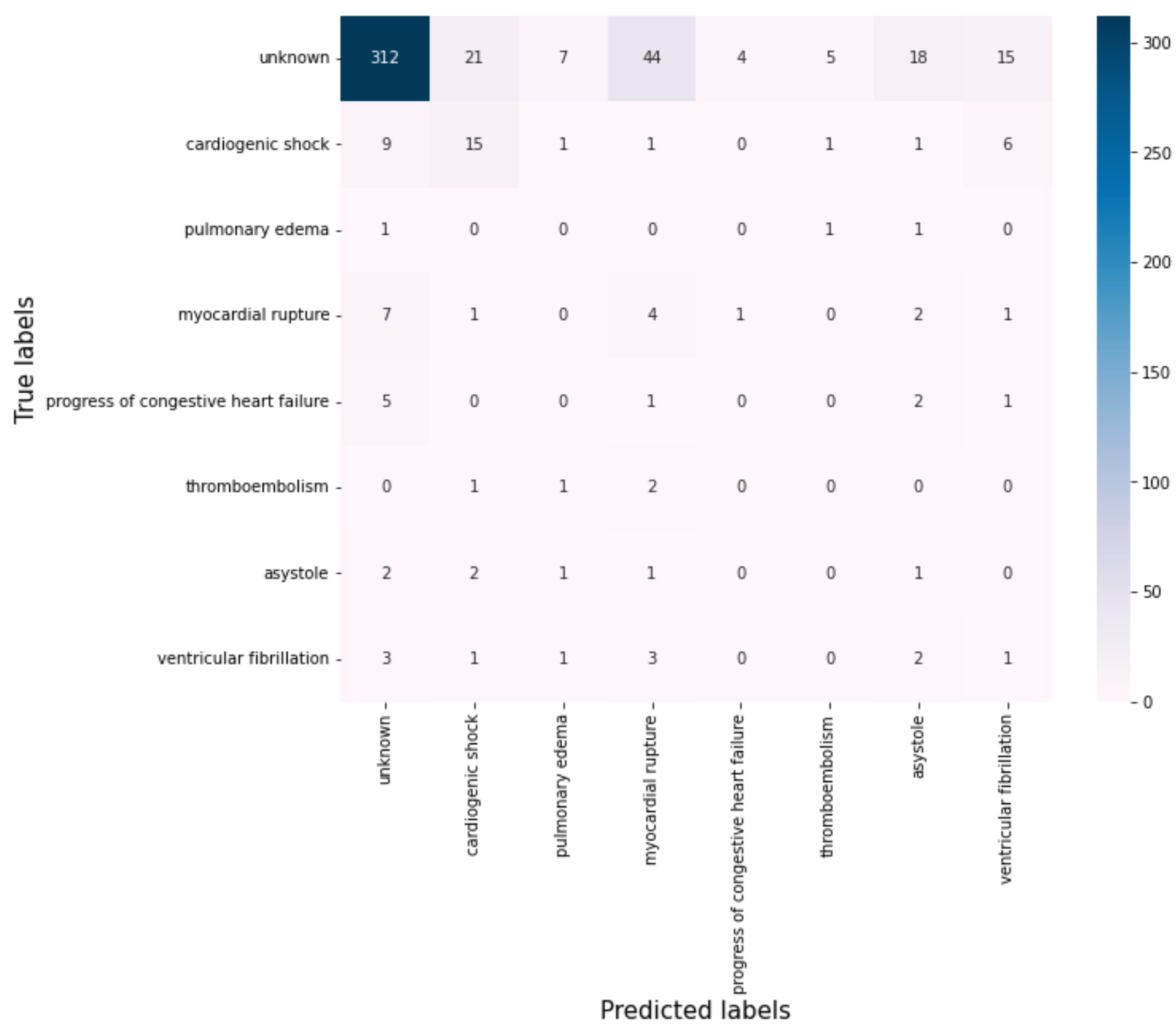

*Figure 5.7 A confusion matrix from logistic regression using class weight method*

The classification report, shown in table 5.3, illustrates the various performance metrics for this multiclass classification task with class weighted method of imbalance handling, as well as the precision, recall, and f1-score for each class. Additionally, the weighted average precision, weighted average recall, weighted average f1-score, and accuracy for the measuring overall model performance are 80%, 65%, 71%, and 65%, respectively.

*Table 5.3 The classification report from logistic regression with class weighted method*

| **Target class label** | **precision** | **recall** | **f1-score** | **Weighted avg precision** | **Weighted avg recall** | **Weighted avg f1-score** | **accuracy** |
|---|---|---|---|---|---|---|---|
| unknown | 0.92 | 0.73 | 0.82 | **0.80** | **0.65** | **0.71** | **0.65** |
| cardiogenic shock | 0.37 | 0.44 | 0.40 | | | | |
| pulmonary edema | 0.00 | 0.00 | 0.00 | | | | |
| myocardial rupture | 0.07 | 0.25 | 0.11 | | | | |

<table>
<tr><td>progress of congestive heart failure</td><td>0.00</td><td>0.00</td><td>0.00</td><td rowspan="4">0.80</td><td rowspan="4">0.65</td><td rowspan="4">0.71</td><td rowspan="4">0.65</td></tr>
<tr><td>thromboembolism</td><td>0.00</td><td>0.00</td><td>0.00</td></tr>
<tr><td>asystole</td><td>0.04</td><td>0.14</td><td>0.06</td></tr>
<tr><td>ventricular fibrillation</td><td>0.04</td><td>0.09</td><td>0.06</td></tr>
</table>

The ROC curve shown below in figure 5.8, illustrate how good the model is performing for predicting each class label with class weighted method of imbalance handling. Also, the weighted average AUCROC score for the overall model is 78.94%.

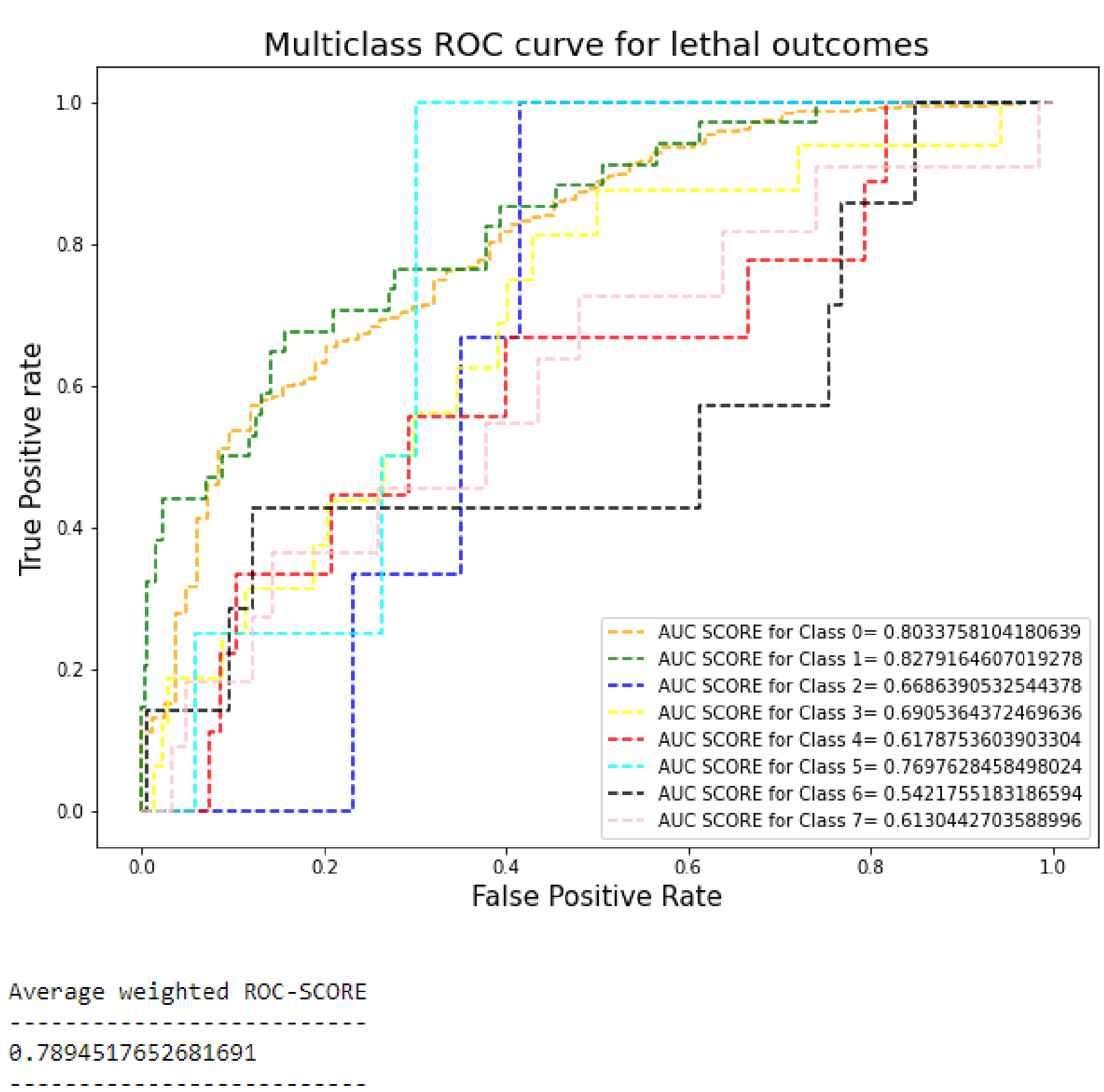

*Figure 5.8 AUCROC curve from logistic regression with class weighted method*

### 5.3.2 Light Gradient Boosting Machine Classifier's Evaluation and Results

For building a classification model, a multiclass LGBM model was utilised; this classifier was trained twice, once without imbalance handling and once after class imbalance treatment (with ADASYN and class weighted method). The evaluations/results of the classifier on the test set, as well as performance metrics such as the confusion matrix, weighted average F1 score, weighted average precision, weighted average recall, and the AUCROC curve, will be shown in the subsections below. As mentioned earlier, the true positive (TP) counts and false negative (FN) counts are important to predict appropriately from confusion matrix in any of the medical diagnosis or prediction model. Hence the true positive counts and false negative counts will be the main area of focus along with the other performance metrices.

#### 5.3.2.1 Evaluation without Class Imbalance Handling

Training dataset before imbalance handling consists of 1,190 records, which was used to train the model. Test dataset consists of 510 records which was used to evaluate the model performance. Below figure 5.9 refers to the confusion matrix where rows are representing actual class labels and columns are representing predicted class labels.

As part of the true positive (TP) counts, it can be shown that 424 cases out of 426 were correctly predicted for the 'unknow' class label. For the 'cardiogenic shock' class label, 12 cases out of 34 were correctly predicted. For the 'pulmonary edema' class label, 1 case was correctly predicted out of 3 cases. For the 'myocardial rupture' class label, 1 instance out of 16 was correctly predicted. For the 'progress of congestive heart failure' class label, nothing (0 cases) was predicted out of 9 cases. For the 'thromboembolism' class label, nothing (0 cases) was anticipated out of 4 cases. For the 'asystole' class label, nothing (0 cases) was anticipated out of 7 cases. Furthermore, nothing (0 cases) was predicted for the 'ventricular fibrillation' class label out of 11 cases.

As part of the false negative (FN) counts, 2 examples that truly correspond to the 'unknown' class label were forecasted as other class labels. 22 cases were projected as different class labels, despite the fact that they all belong to the 'cardiogenic shock' category. 2 cases, all of which correspond to the 'pulmonary edema' class label, were projected to have different class labels. 15 cases with the class label "myocardial rupture" were also predicted as other class labels. 9 instances were projected as different class labels, despite the fact that they all belong to the 'progress of congestive heart failure' class label. 4 examples, all of which fall under the 'thromboembolism' class label, were predicted to have different class labels. 7 instances with the class label 'asystole' were projected as different class labels, while 11 cases with the class label 'ventricular fibrillation' was also predicted as different class labels.

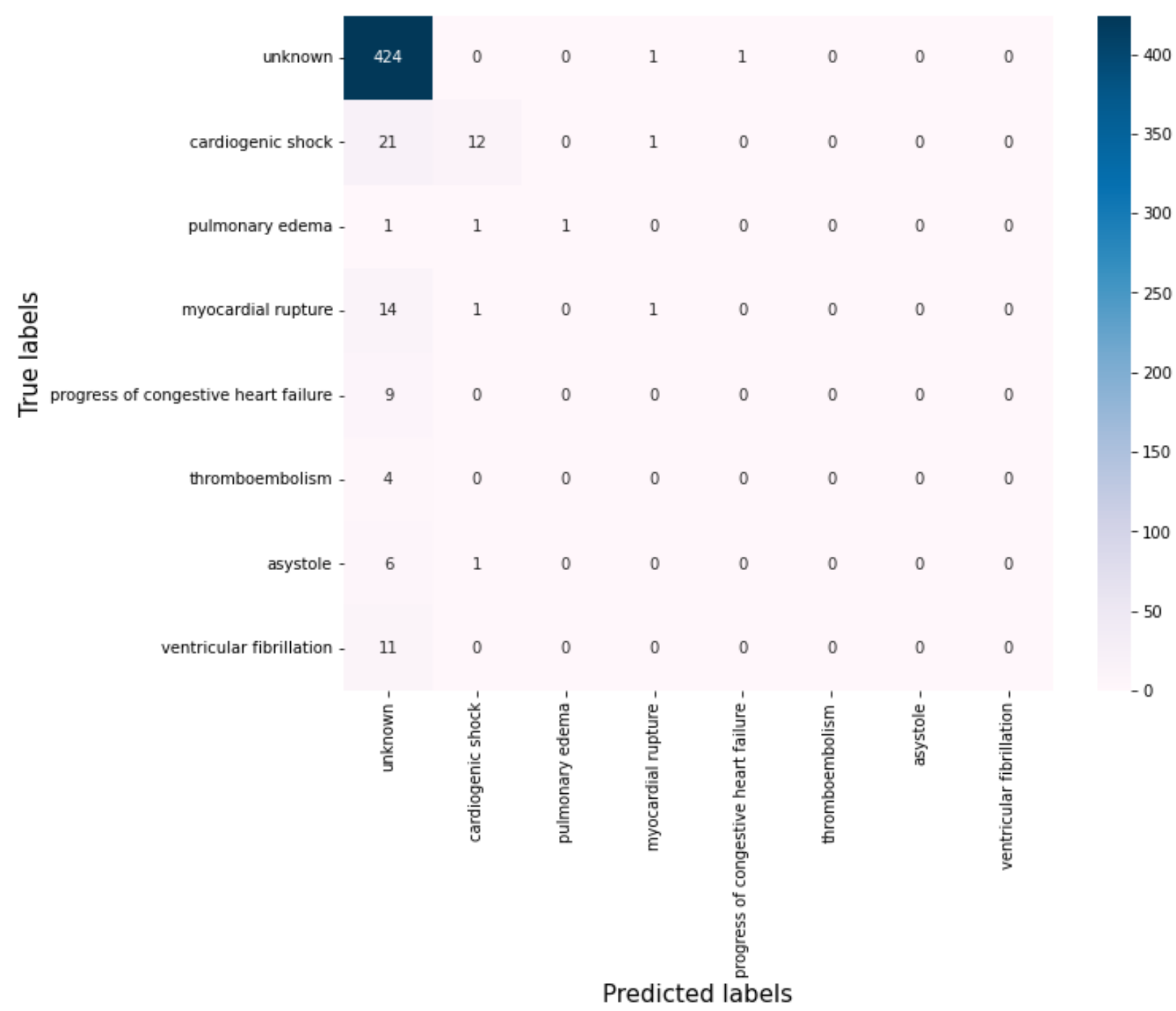

*Figure 5.9 A confusion matrix from LGBM modelling without class imbalance handling*

The classification report, shown in table 5.4, illustrates the various performance metrics for this multiclass classification task before class imbalance handling, as well as the precision, recall, and f1-score for each class. Additionally, the weighted average precision, weighted average recall, weighted average f1-score, and accuracy for the measuring overall model performance are 79%, 86%, 81%, and 86%, respectively.

*Table 5.4 The classification report from LGBM modelling without class imbalance handling*

<table>
<tr><th>Target class label</th><th>precision</th><th>recall</th><th>f1-score</th><th>Weighted avg precision</th><th>Weighted avg recall</th><th>Weighted avg f1-score</th><th>accuracy</th></tr>
<tr><td>unknown</td><td>0.87</td><td>1.00</td><td>0.93</td><td rowspan="4">0.79</td><td rowspan="4">0.86</td><td rowspan="4">0.81</td><td rowspan="4">0.86</td></tr>
<tr><td>cardiogenic shock</td><td>0.80</td><td>0.35</td><td>0.49</td></tr>
<tr><td>pulmonary edema</td><td>1.00</td><td>0.33</td><td>0.50</td></tr>
<tr><td>myocardial rupture</td><td>0.33</td><td>0.06</td><td>0.11</td></tr>
</table>

| progress of congestive heart failure | 0.00 | 0.00 | 0.00 | **0.79** | **0.86** | **0.81** | **0.86** |
|---|---|---|---|---|---|---|---|
| thromboembolism | 0.00 | 0.00 | 0.00 | | | | |
| asystole | 0.00 | 0.00 | 0.00 | | | | |
| ventricular fibrillation | 0.00 | 0.00 | 0.00 | | | | |

The ROC curve shown below in figure 5.10, illustrate how good the model is performing for predicting each class label before class imbalance handling. Also, the weighted average AUCROC score for the overall model is 85.62%.

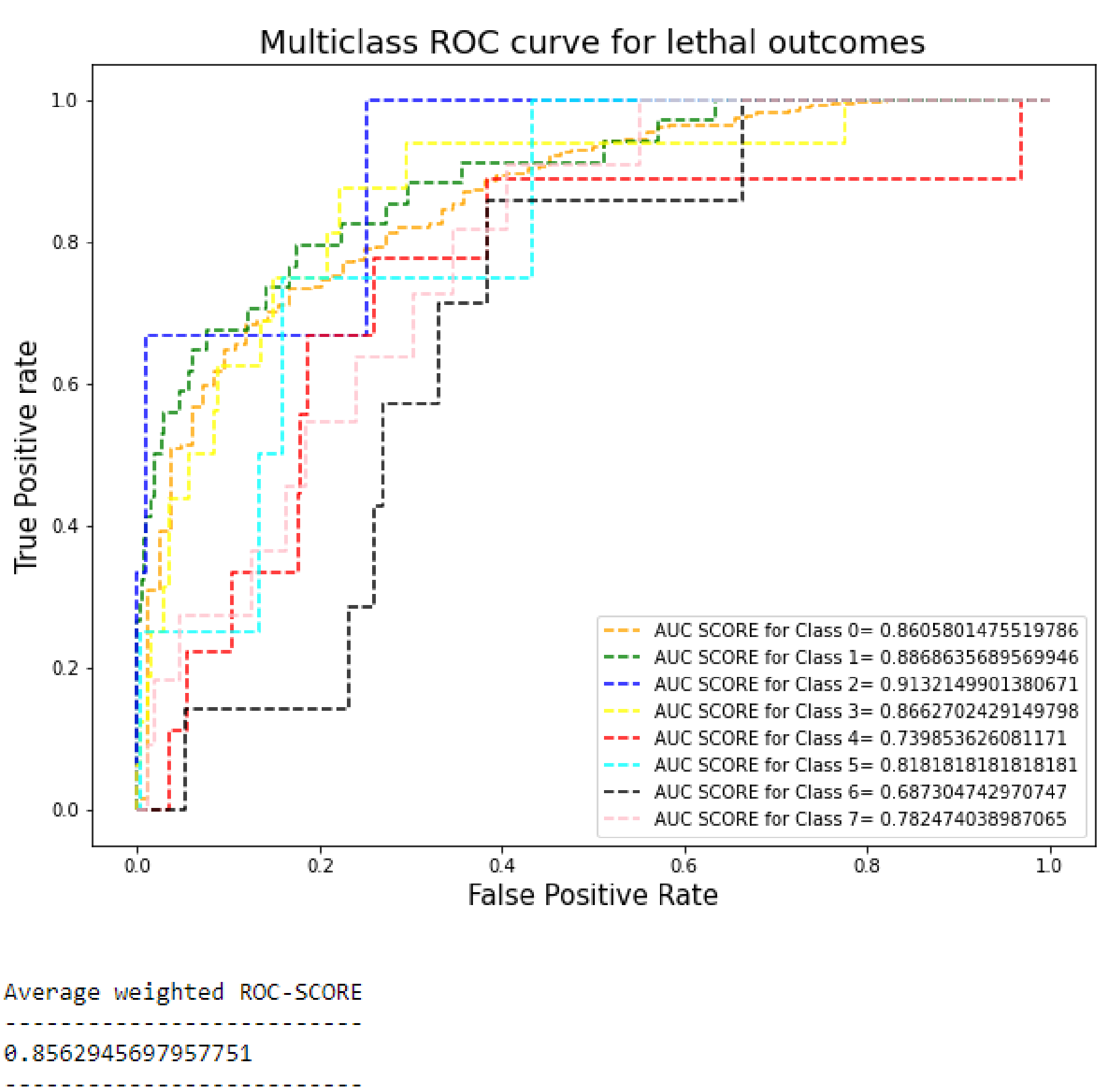

*Figure 5.10 AUCROC curve from LGBM modelling without class imbalance handling*

#### 5.3.2.2 Evaluation after Class Imbalance Handling (with ADASYN method)

Training dataset after imbalance handling with ADASYN method consists of 8,010 records, which was used to train the model. Test dataset consists of 510 records which was used to evaluate the model performance. Below figure 5.11 refers to the confusion matrix where rows are representing actual class labels and columns are representing predicted class labels.

As part of the true positive (TP) counts, it can be shown that 375 cases out of 426 were correctly predicted for the 'unknow' class label. For the 'cardiogenic shock' class label, 15 cases out of 34 were correctly predicted. For the 'pulmonary edema' class label, nothing (0 cases) was anticipated out of 3 cases. For the 'myocardial rupture' class label, 2 cases were correctly predicted out of 16 cases. For the 'progress of congestive heart failure' class label, nothing (0 cases) was anticipated out of 9 cases. For the 'thromboembolism' class label, nothing (0 cases) was anticipated out of 4 cases. For the 'asystole' class label, nothing (0 cases) was anticipated out of 7 cases. Furthermore, nothing (0 cases) was predicted for the 'ventricular fibrillation' class label out of 11 cases.

As part of the false negative (FN) counts, 51 examples that truly correspond to the 'unknown' class label were forecasted as other class labels. 19 cases were projected as different class labels, despite the fact that they all belong to the 'cardiogenic shock' category. 3 cases, all of which correspond to the 'pulmonary edema' class label, were projected to have different class labels. 14 cases with the class label "myocardial rupture" were also predicted as other class labels. 9 instances were projected as different class labels, despite the fact that they all belong to the 'progress of congestive heart failure' class label. 4 examples, all of which fall under the 'thromboembolism' class label, were predicted to have different class labels. 7 instances with the class label 'asystole' were projected as different class labels, while 11 cases with the class label 'ventricular fibrillation' was also predicted as different class labels.

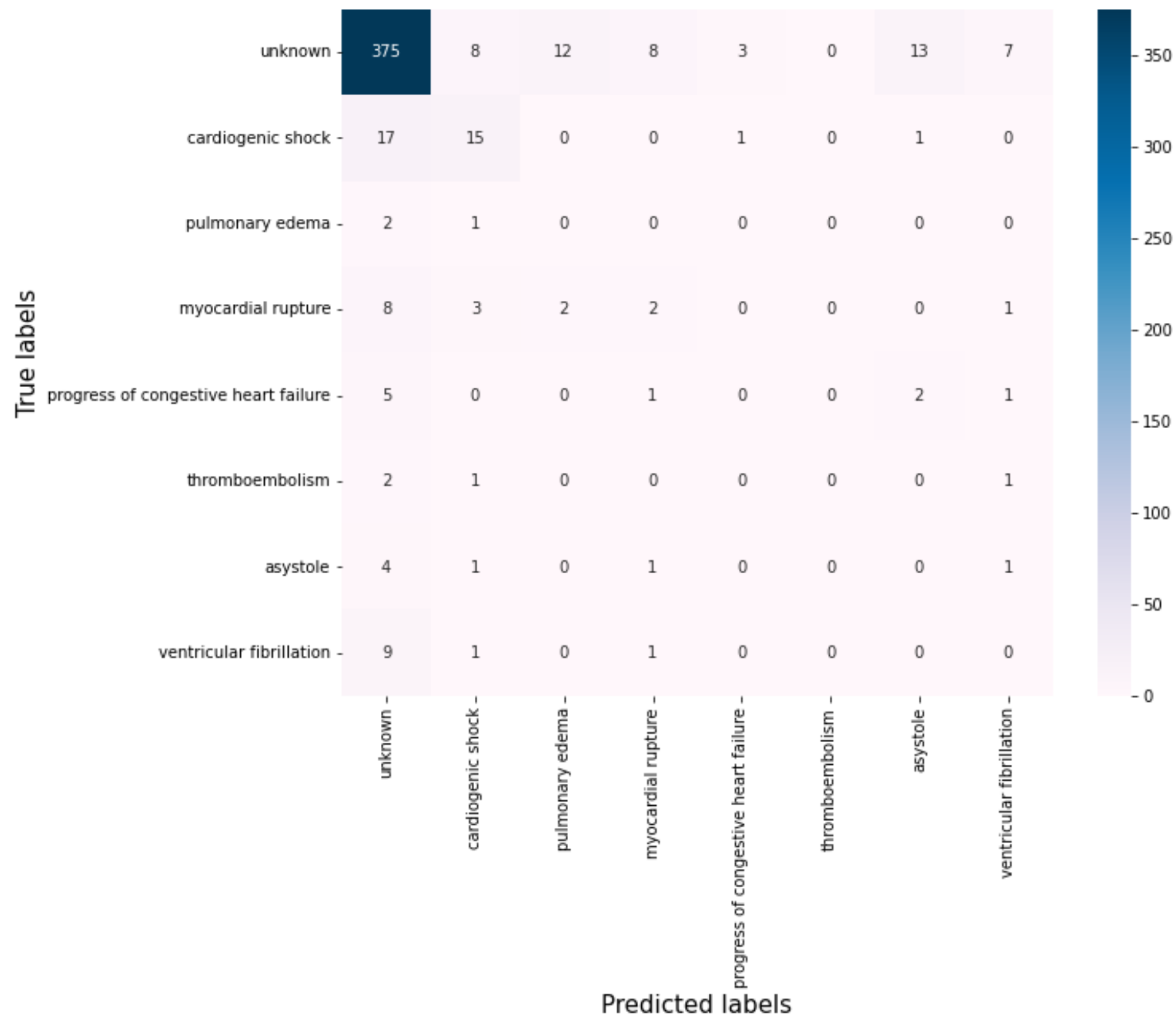

*Figure 5.11 A confusion matrix from LGBM modelling after class imbalance handling (ADASYN)*

The classification report, shown in table 5.5, illustrates the various performance metrics for this multiclass classification task after class imbalance treatment, as well as the precision, recall, and f1-score for each class. Additionally, the weighted average precision, weighted average recall, weighted average f1-score, and accuracy for the measuring overall model performance are 78%, 77%, 77%, and 77%, respectively.

*Table 5.5 The classification report from LGBM modelling after class imbalance handling (ADASYN)*

<table>
<tr><th>Target class label</th><th>precision</th><th>recall</th><th>f1-score</th><th>Weighted avg precision</th><th>Weighted avg recall</th><th>Weighted avg f1-score</th><th>accuracy</th></tr>
<tr><td>unknown</td><td>0.89</td><td>0.88</td><td>0.88</td><td rowspan="4">0.78</td><td rowspan="4">0.77</td><td rowspan="4">0.77</td><td rowspan="4">0.77</td></tr>
<tr><td>cardiogenic shock</td><td>0.50</td><td>0.44</td><td>0.47</td></tr>
<tr><td>pulmonary edema</td><td>0.00</td><td>0.00</td><td>0.00</td></tr>
<tr><td>myocardial rupture</td><td>0.15</td><td>0.12</td><td>0.14</td></tr>
</table>

| progress of congestive heart failure | 0.00 | 0.00 | 0.00 | **0.78** | **0.77** | **0.77** | **0.77** |
|---|---|---|---|---|---|---|---|
| thromboembolism | 0.00 | 0.00 | 0.00 | | | | |
| asystole | 0.00 | 0.00 | 0.00 | | | | |
| ventricular fibrillation | 0.00 | 0.00 | 0.00 | | | | |

The ROC curve shown below in figure 5.12, illustrate how good the model is performing for predicting each class label after class imbalance treatment. Also, the weighted average AUCROC score for the overall model is 75.87%.

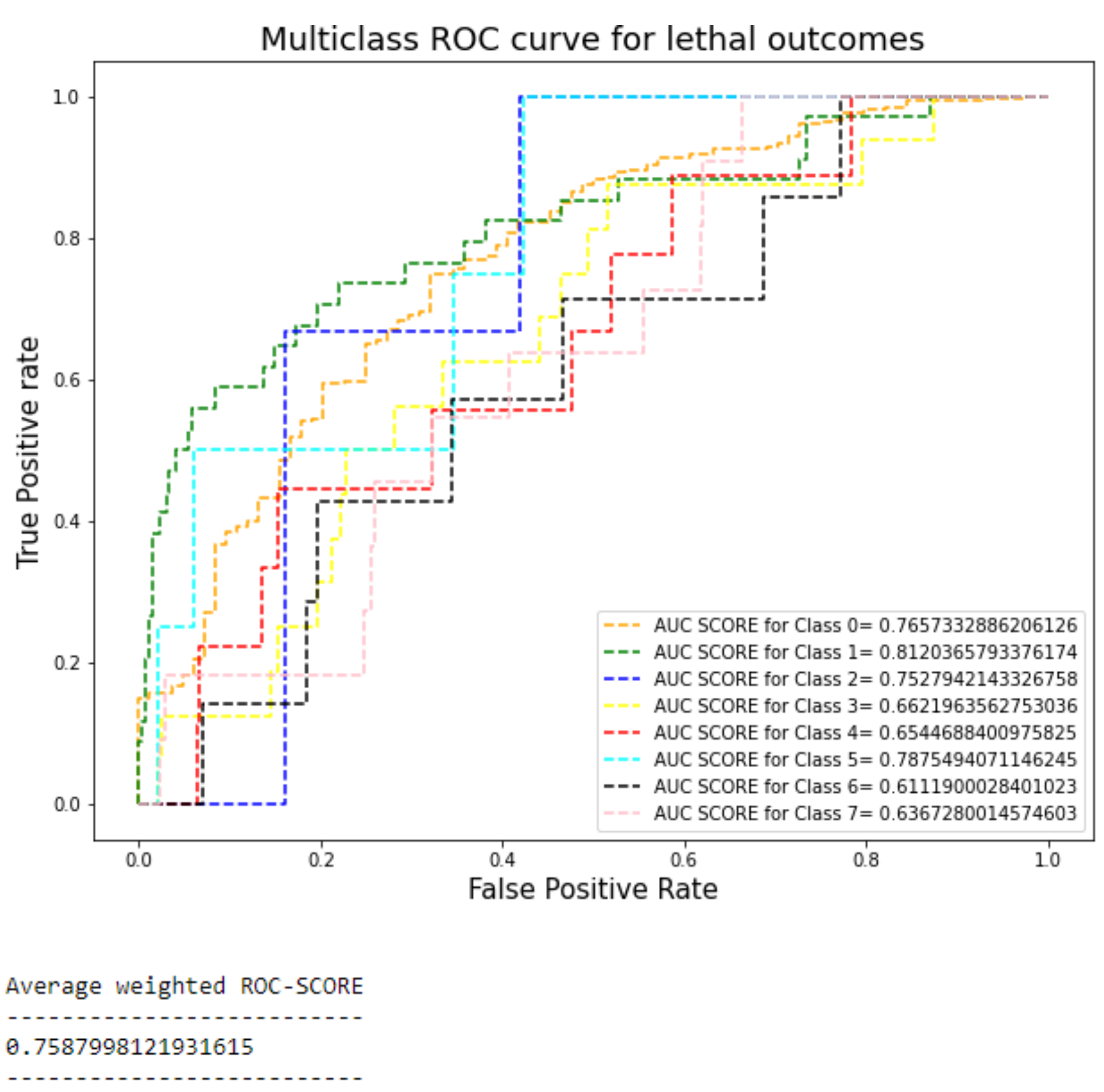

*Figure 5.12 AUCROC curve from LGBM modelling after class imbalance handling (ADASYN)*

#### 5.3.2.3 Evaluation after applying Class Weighted Method

Training dataset before imbalance handling consists of 1,190 records, which was used to train the model with class weighted hyperparameter (class_weight=’balanced’) to handle the imbalance class. Test dataset consists of 510 records which was used to evaluate the model performance. Below figure 5.13 refers to the confusion matrix where rows are representing actual class labels and columns are representing predicted class labels.

As part of the true positive (TP) counts, it can be shown that 419 cases out of 426 were correctly predicted for the 'unknow' class label. For the 'cardiogenic shock' class label, 13 cases out of 34 were correctly predicted. For the 'pulmonary edema' class label, 1 case was correctly predicted out of 3 cases. For the 'myocardial rupture' class label, 1 case was predicted correctly out of 16 cases. For the 'progress of congestive heart failure' class label, nothing (0 cases) was anticipated out of 9 cases. For the 'thromboembolism' class label, nothing (0 cases) was anticipated out of 4 cases. For the 'asystole' class label, nothing (0 case) was anticipated out of 7 cases. Furthermore, nothing (0 cases) was anticipated for the 'ventricular fibrillation' class label out of 11 cases.

As part of the false negative (FN) counts, 7 examples that truly correspond to the 'unknown' class label were forecasted as other class labels. 21 cases were projected as different class labels, despite the fact that they all belong to the 'cardiogenic shock' category. 2 cases, all of which correspond to the 'pulmonary edema' class label, were projected to have different class labels. 15 cases with the class label "myocardial rupture" were also predicted as other class labels. 9 instances were projected as different class labels, despite the fact that they all belong to the 'progress of congestive heart failure' class label. 4 examples, all of which fall under the 'thromboembolism' class label, were predicted to have different class labels. 7 instances with the class label 'asystole' were projected as different class labels, while 11 cases with the class label 'ventricular fibrillation' was also predicted as different class labels.

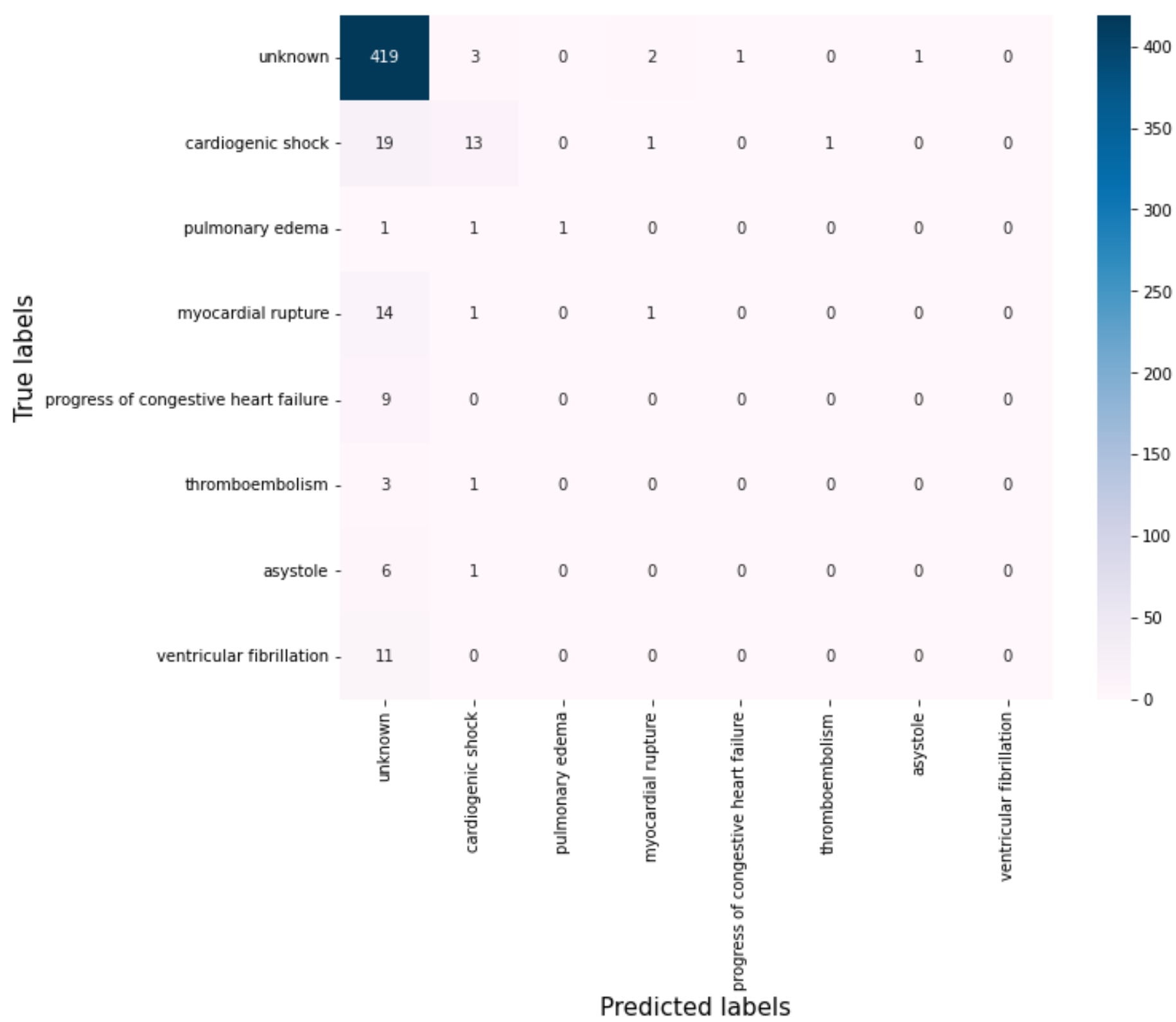

*Figure 5.13 A confusion matrix from LGBM modelling using class weight method*

The classification report, shown in table 5.6, illustrates the various performance metrics for this multiclass classification task with class weighted method of imbalance handling, as well as the precision, recall, and f1-score for each class. Additionally, the weighted average precision, weighted average recall, weighted average f1-score, and accuracy for the measuring overall model performance are 78%, 85%, 81%, and 85%, respectively.

*Table 5.6 The classification report from LGBM modelling with class weighted method*

<table>
<tr><th>Target class label</th><th>precision</th><th>recall</th><th>f1-score</th><th>Weighted avg precision</th><th>Weighted avg recall</th><th>Weighted avg f1-score</th><th>accuracy</th></tr>
<tr><td>unknown</td><td>0.87</td><td>0.98</td><td>0.92</td><td rowspan="4">0.78</td><td rowspan="4">0.85</td><td rowspan="4">0.81</td><td rowspan="4">0.85</td></tr>
<tr><td>cardiogenic shock</td><td>0.65</td><td>0.38</td><td>0.48</td></tr>
<tr><td>pulmonary edema</td><td>1.00</td><td>0.33</td><td>0.50</td></tr>
<tr><td>myocardial rupture</td><td>0.25</td><td>0.06</td><td>0.10</td></tr>
</table>

<table>
<tr><td>progress of congestive heart failure</td><td>0.00</td><td>0.00</td><td>0.00</td><td rowspan="4">0.78</td><td rowspan="4">0.85</td><td rowspan="4">0.81</td><td rowspan="4">0.85</td></tr>
<tr><td>thromboembolism</td><td>0.00</td><td>0.00</td><td>0.00</td></tr>
<tr><td>asystole</td><td>0.00</td><td>0.00</td><td>0.00</td></tr>
<tr><td>ventricular fibrillation</td><td>0.00</td><td>0.00</td><td>0.00</td></tr>
</table>

The ROC curve shown below in figure 5.14, illustrate how good the model is performing for predicting each class label with class weighted method of imbalance handling. Also, the weighted average AUCROC score for the overall model is 85.74%.

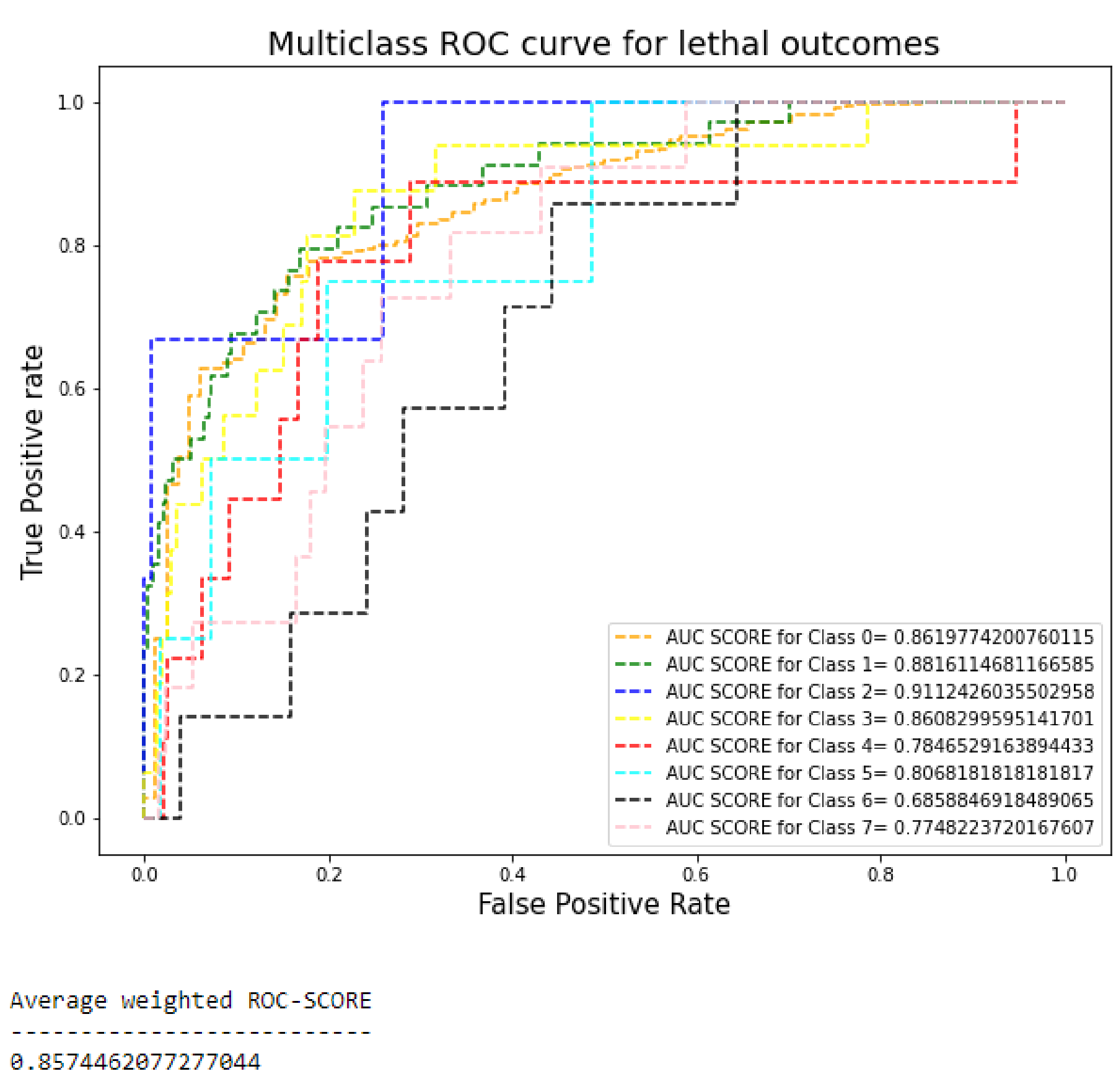

*Figure 5.14 AUCROC curve from LGBM modelling with class weighted method*

### 5.3.3 Random Forest Classifier's Evaluation and Results

For building a classification model, a random forest algorithm was utilised; this classifier was trained twice, once without imbalance handling and once after class imbalance treatment (with ADASYN and class weighted method). The evaluations/results of the classifier on the test set, as well as performance metrics such as the confusion matrix, weighted average F1 score, weighted average precision, weighted average recall, and the AUCROC curve, will be shown in the subsections below. As mentioned earlier, the true positive (TP) counts and false negative (FN) counts are important to predict appropriately from confusion matrix in any of the medical diagnosis or prediction model. Hence the true positive counts and false negative counts will be the main area of focus along with the other performance metrices.

#### 5.3.3.1 Evaluation without Class Imbalance Handling

Training dataset before imbalance handling consists of 1,190 records, which was used to train the model. Test dataset consists of 510 records which was used to evaluate the model performance. Below figure 5.15 refers to the confusion matrix where rows are representing actual class labels and columns are representing predicted class labels.

As part of the true positive (TP) counts, it can be shown that 426 cases out of 426 were correctly predicted for the 'unknow' class label. For the 'cardiogenic shock' class label, 14 cases out of 34 were correctly predicted. For the 'pulmonary edema' class label, nothing (0 cases) was correctly predicted out of 3 cases. For the 'myocardial rupture' class label, nothing (0 cases) out of 16 was correctly predicted. For the 'progress of congestive heart failure' class label, nothing (0 cases) was predicted out of 9 cases. For the 'thromboembolism' class label, nothing (0 cases) was anticipated out of 4 cases. For the 'asystole' class label, nothing (0 cases) was anticipated out of 7 cases. Furthermore, nothing (0 cases) was predicted for the 'ventricular fibrillation' class label out of 11 cases.

As part of the false negative (FN) counts, all the examples that truly correspond to the 'unknown' class label were forecasted correctly. 20 cases were projected as different class labels, despite the fact that they all belong to the 'cardiogenic shock' category. 3 cases, all of which correspond to the 'pulmonary edema' class label, were projected to have different class labels. 16 cases with the class label "myocardial rupture" were also predicted as other class labels. 9 instances were projected as different class labels, despite the fact that they all belong to the 'progress of congestive heart failure' class label. 4 examples, all of which fall under the 'thromboembolism' class label, were predicted to have different class labels. 7 instances with the class label 'asystole' were projected as different class labels, while 11 cases with the class label 'ventricular fibrillation' was also predicted as different class labels.

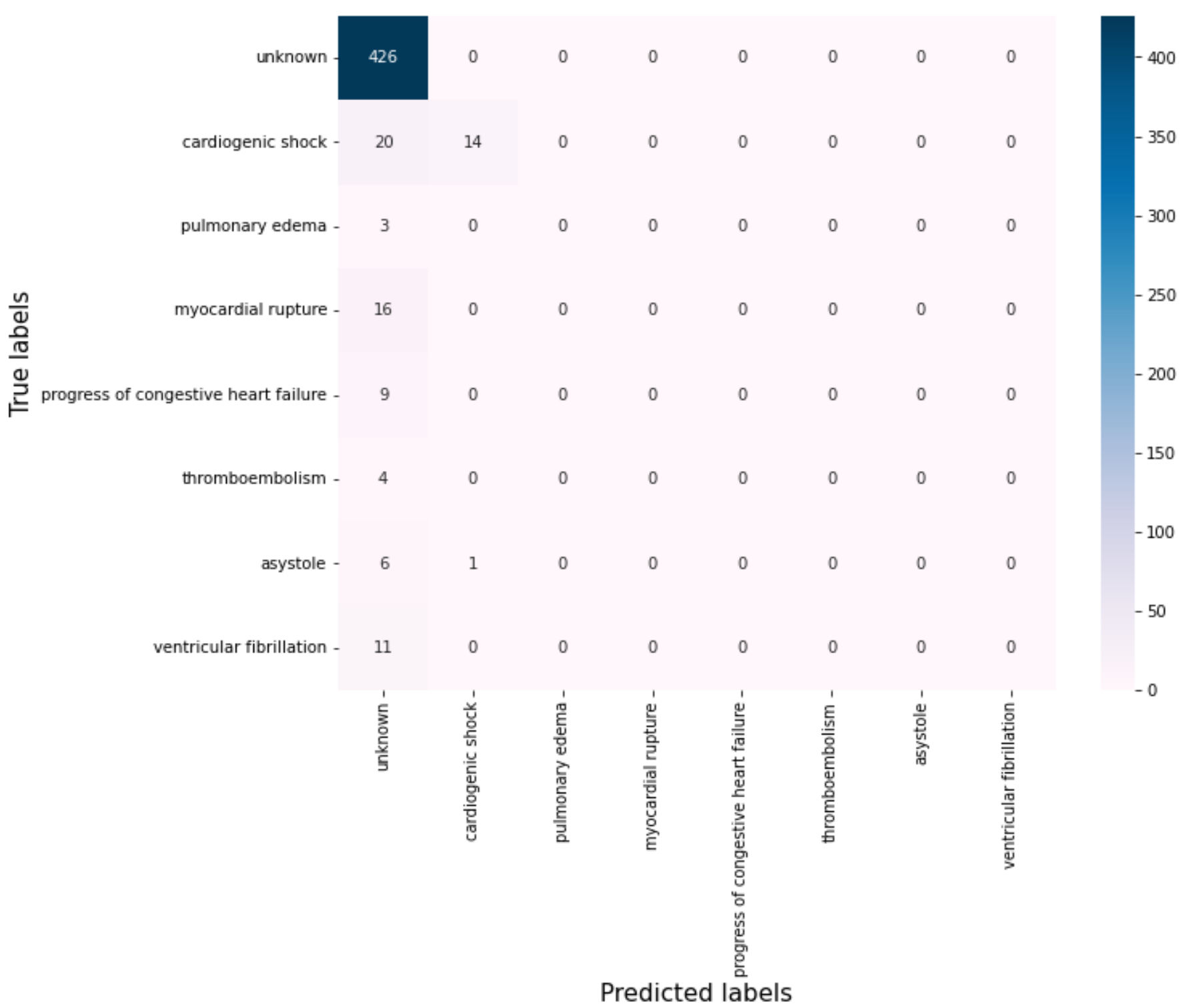

*Figure 5.15 A confusion matrix from random forest modelling without class imbalance handling*

The classification report, shown in table 5.7, illustrates the various performance metrics for this multiclass classification task before class imbalance handling, as well as the precision, recall, and f1-score for each class. Additionally, the weighted average precision, weighted average recall, weighted average f1-score, and accuracy for the measuring overall model performance are 78%, 86%, 81%, and 86%, respectively.

*Table 5.7 The classification report from random forest modelling without class imbalance handling*

| **Target class label** | **precision** | **recall** | **f1-score** | **Weighted avg precision** | **Weighted avg recall** | **Weighted avg f1-score** | **accuracy** |
|---|---|---|---|---|---|---|---|
| unknown | 0.86 | 1.00 | 0.93 | **0.78** | **0.86** | **0.81** | **0.86** |
| cardiogenic shock | 0.93 | 0.41 | 0.57 | | | | |
| pulmonary edema | 0.00 | 0.00 | 0.00 | | | | |
| myocardial rupture | 0.00 | 0.00 | 0.00 | | | | |
| progress of congestive heart failure | 0.00 | 0.00 | 0.00 | **0.78** | **0.86** | **0.81** | **0.86** |
| thromboembolism | 0.00 | 0.00 | 0.00 | | | | |
| asystole | 0.00 | 0.00 | 0.00 | | | | |
| ventricular fibrillation | 0.00 | 0.00 | 0.00 | | | | |

The ROC curve shown below in figure 5.16, illustrate how good the model is performing for predicting each class label before class imbalance handling. Also, the weighted average AUCROC score for the overall model is 80.17%.

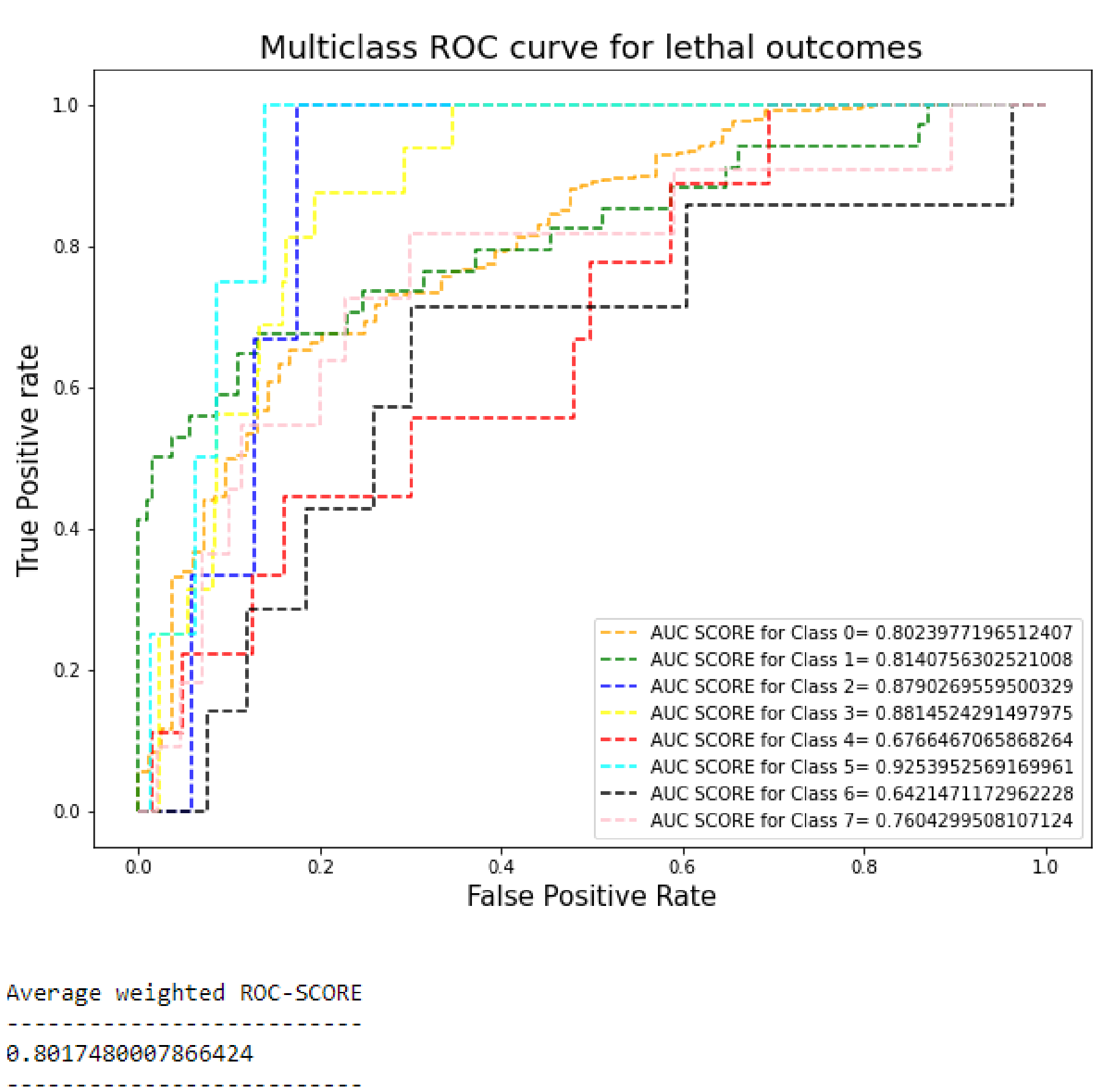

*Figure 5.16 AUCROC curve from random forest modelling without class imbalance handling*

#### 5.3.3.2 Evaluation after Class Imbalance Handling (with ADASYN method)

Training dataset after imbalance handling with ADASYN method consists of 8,010 records, which was used to train the model. Test dataset consists of 510 records which was used to evaluate the model performance. Below figure 5.17 refers to the confusion matrix where rows are representing actual class labels and columns are representing predicted class labels.

As part of the true positive (TP) counts, it can be shown that 372 cases out of 426 were correctly predicted for the 'unknow' class label. For the 'cardiogenic shock' class label, 18 cases out of 34 were correctly predicted. For the 'pulmonary edema' class label, nothing (0 cases) was anticipated out of 3 cases. For the 'myocardial rupture' class label, 2 cases were correctly predicted out of 16 cases. For the 'progress of congestive heart failure' class label, nothing (0 cases) was anticipated out of 9 cases. For the 'thromboembolism' class label, nothing (0 cases) was anticipated out of 4 cases. For the 'asystole' class label, nothing (0 cases) was anticipated out of 7 cases. Furthermore, 1 case was predicted correctly for the 'ventricular fibrillation' class label out of 11 cases.

As part of the false negative (FN) counts, 54 examples that truly correspond to the 'unknown' class label were forecasted as other class labels. 16 cases were projected as different class labels, despite the fact that they all belong to the 'cardiogenic shock' category. 3 cases, all of which correspond to the 'pulmonary edema' class label, were projected to have different class labels. 14 cases with the class label "myocardial rupture" were also predicted as other class labels. 9 instances were projected as different class labels, despite the fact that they all belong to the 'progress of congestive heart failure' class label. 4 examples, all of which fall under the 'thromboembolism' class label, were predicted to have different class labels. 7 instances with the class label 'asystole' were projected as different class labels, while 10 cases with the class label 'ventricular fibrillation' was also predicted as different class labels.

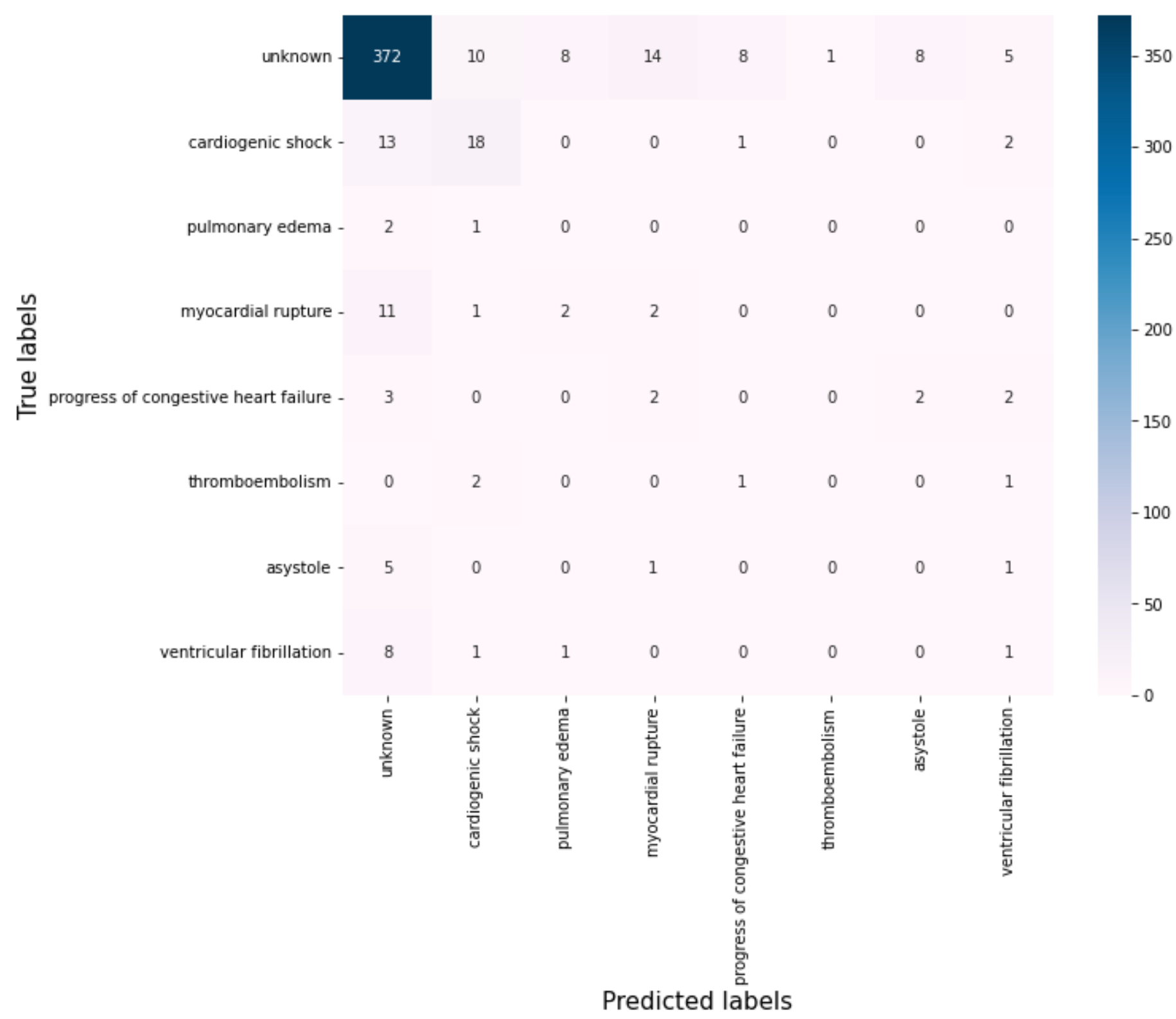

*Figure 5.17 A confusion matrix from random forest modelling after class imbalance handling (ADASYN)*

The classification report, shown in table 5.8, illustrates the various performance metrics for this multiclass classification task after class imbalance treatment, as well as the precision, recall, and f1-score for each class. Additionally, the weighted average precision, weighted average recall, weighted average f1-score, and accuracy for the measuring overall model performance are 79%, 77%, 78%, and 77%, respectively.

*Table 5.8 The classification report from random forest modelling after class imbalance handling (ADASYN)*

| Target class label | precision | recall | f1-score | Weighted avg precision | Weighted avg recall | Weighted avg f1-score | accuracy |
|---|---|---|---|---|---|---|---|
| unknown | 0.90 | 0.87 | 0.89 | **0.79** | **0.77** | **0.78** | **0.77** |
| cardiogenic shock | 0.55 | 0.53 | 0.54 | | | | |
| pulmonary edema | 0.00 | 0.00 | 0.00 | | | | |
| myocardial rupture | 0.11 | 0.12 | 0.11 | | | | |

| progress of congestive heart failure | 0.00 | 0.00 | 0.00 | | | | |
|---|---|---|---|---|---|---|---|
| thromboembolism | 0.00 | 0.00 | 0.00 | **0.79** | **0.77** | **0.78** | **0.77** |
| asystole | 0.00 | 0.00 | 0.00 | | | | |
| ventricular fibrillation | 0.08 | 0.09 | 0.09 | | | | |

The ROC curve shown below in figure 5.18, illustrate how good the model is performing for predicting each class label after class imbalance treatment. Also, the weighted average AUCROC score for the overall model is 77.22%.

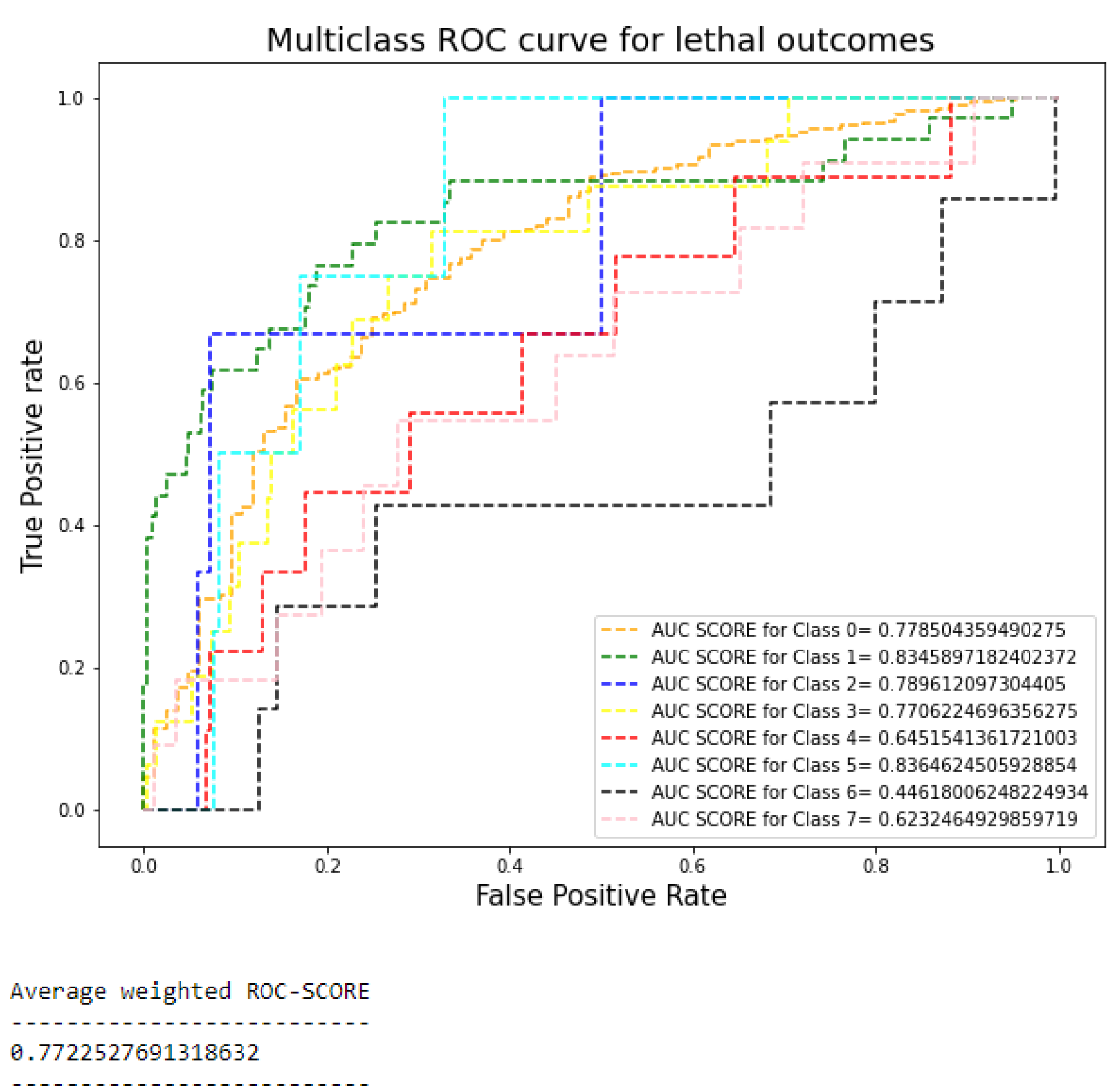

*Figure 5.18 AUCROC curve from random forest modelling after class imbalance handling (ADASYN)*

#### 5.3.3.3 Evaluation after applying Class Weighted Method

Training dataset before imbalance handling consists of 1,190 records, which was used to train the model with class weighted hyperparameter (class_weight='balanced') to handle the imbalanced class. Test dataset consists of 510 records which was used to evaluate the model performance. Below figure 5.19 refers to the confusion matrix where rows are representing actual class labels and columns are representing predicted class labels.

As part of the true positive (TP) counts, it can be shown that 368 cases out of 426 were correctly predicted for the 'unknow' class label. For the 'cardiogenic shock' class label, 19 cases out of 34 were correctly predicted. For the 'pulmonary edema' class label, 1 case was correctly predicted out of 3 cases. For the 'myocardial rupture' class label, 7 case was predicted correctly out of 16 cases. For the 'progress of congestive heart failure' class label, 1 case was correctly predicted out of 9 cases. For the 'thromboembolism' class label, 1 case was correctly predicted out of 4 cases. For the 'asystole' class label, 1 case was correctly predicted out of 7 cases. Furthermore, nothing (0 cases) was anticipated for the 'ventricular fibrillation' class label out of 11 cases.

As part of the false negative (FN) counts, 58 examples that truly correspond to the 'unknown' class label were forecasted as other class labels. 15 cases were projected as different class labels, despite the fact that they all belong to the 'cardiogenic shock' category. 2 cases, all of which correspond to the 'pulmonary edema' class label, were projected to have different class labels. 9 cases with the class label "myocardial rupture" were also predicted as other class labels. 8 instances were projected as different class labels, despite the fact that they all belong to the 'progress of congestive heart failure' class label. 3 examples, all of which fall under the 'thromboembolism' class label, were predicted to have different class labels. 6 instances with the class label 'asystole' were projected as different class labels, while 11 cases with the class label 'ventricular fibrillation' was also predicted as different class labels.

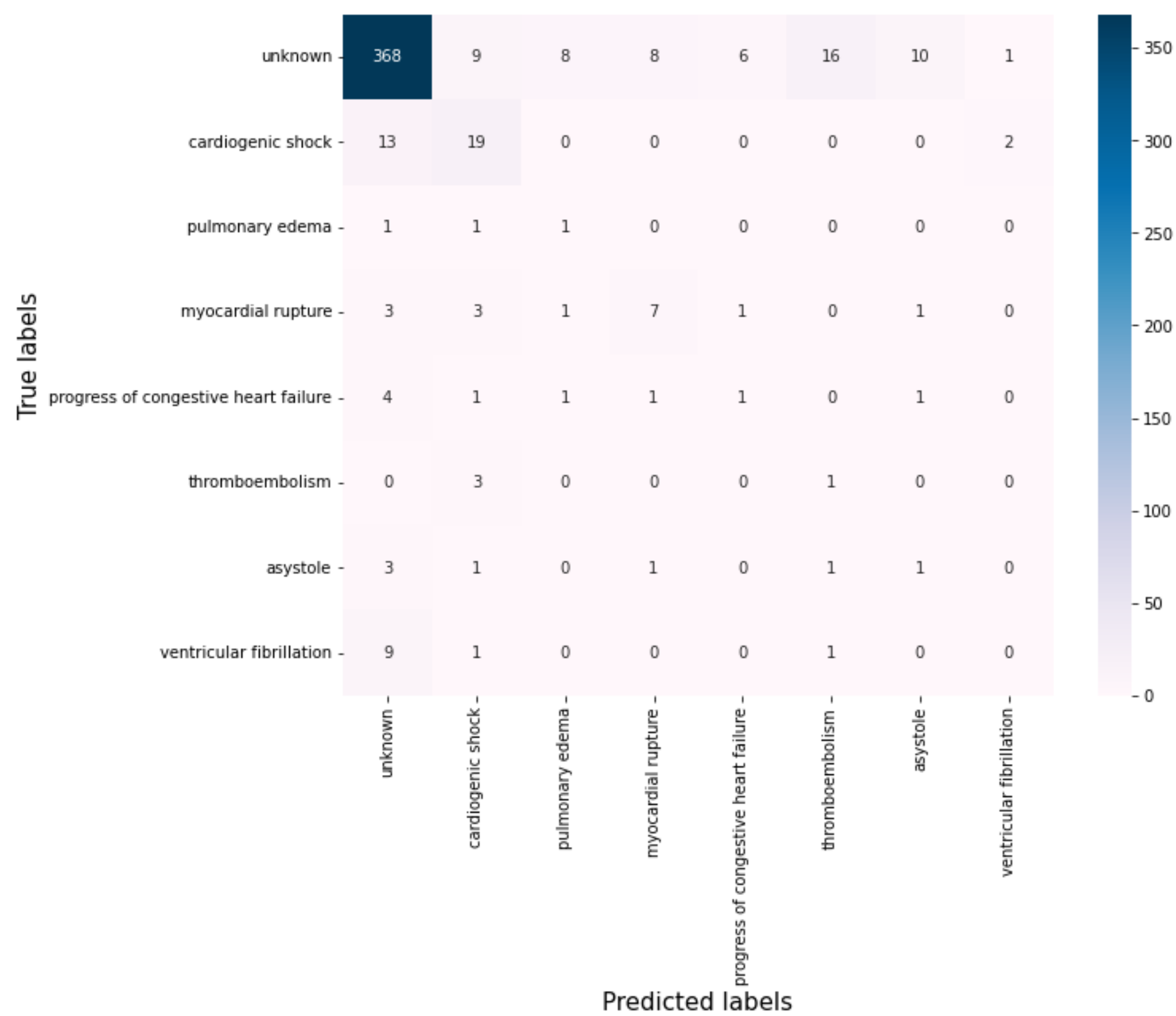

*Figure 5.19 A confusion matrix from random forest modelling using class weight method*

The classification report, shown in table 5.9, illustrates the various performance metrics for this multiclass classification task with class weighted method of imbalance handling, as well as the precision, recall, and f1-score for each class. Additionally, the weighted average precision, weighted average recall, weighted average f1-score, and accuracy for the measuring overall model performance are 82%, 78%, 80%, and 78%, respectively.

*Table 5.9 The classification report from random forest modelling with class weighted method*

<table>
<tr><th>Target class label</th><th>precision</th><th>recall</th><th>f1-score</th><th>Weighted avg precision</th><th>Weighted avg recall</th><th>Weighted avg f1-score</th><th>accuracy</th></tr>
<tr><td>unknown</td><td>0.92</td><td>0.86</td><td>0.89</td><td rowspan="4">0.82</td><td rowspan="4">0.78</td><td rowspan="4">0.80</td><td rowspan="4">0.78</td></tr>
<tr><td>cardiogenic shock</td><td>0.50</td><td>0.56</td><td>0.53</td></tr>
<tr><td>pulmonary edema</td><td>0.09</td><td>0.33</td><td>0.14</td></tr>
<tr><td>myocardial rupture</td><td>0.41</td><td>0.44</td><td>0.42</td></tr>
</table>

<table>
<tr><td>progress of congestive heart failure</td><td>0.12</td><td>0.11</td><td>0.12</td><td rowspan="4">0.82</td><td rowspan="4">0.78</td><td rowspan="4">0.80</td><td rowspan="4">0.78</td></tr>
<tr><td>thromboembolis m</td><td>0.05</td><td>0.25</td><td>0.09</td></tr>
<tr><td>asystole</td><td>0.08</td><td>0.14</td><td>0.10</td></tr>
<tr><td>ventricular fibrillation</td><td>0.00</td><td>0.00</td><td>0.00</td></tr>
</table>

The ROC curve shown below in figure 5.20, illustrate how good the model is performing for predicting each class label with class weighted method of imbalance handling. Also, the weighted average AUCROC score for the overall model is 81.68%.

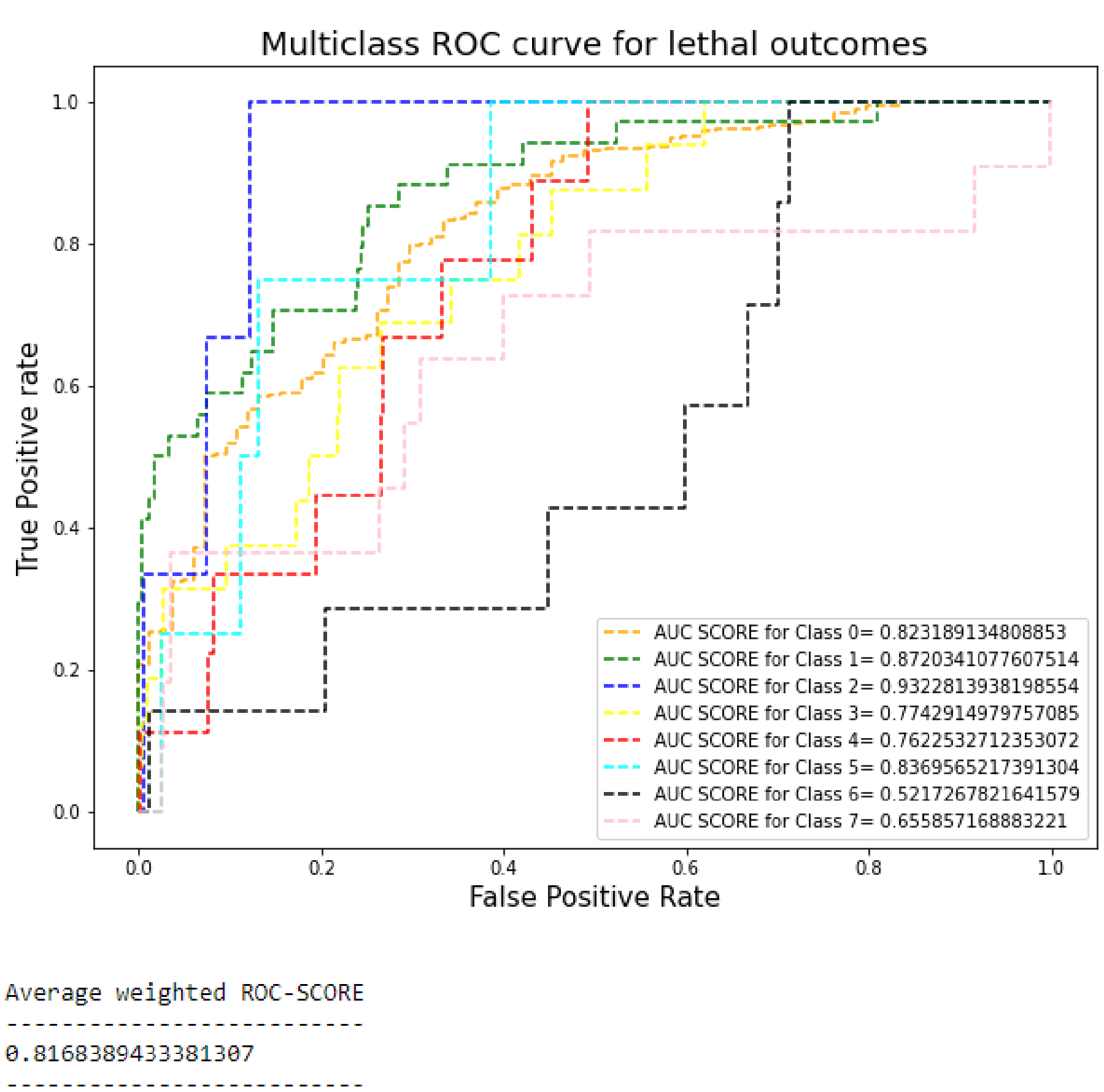

*Figure 5.20 AUCROC curve from random forest modelling with class weighted method*

### 5.3.4 Bagging SVM Classifier’s Evaluation and Results

For building a classification model, a bagging SVM algorithm was utilised; this classifier was trained twice, once without imbalance handling and once after class imbalance treatment (with ADASYN and class weighted method). The evaluations/results of the classifier on the test set, as well as performance metrics such as the confusion matrix, weighted average F1 score, weighted average precision, weighted average recall, and the AUCROC curve, will be shown in the subsections below. As mentioned earlier, the true positive (TP) counts and false negative (FN) counts are important to predict appropriately from confusion matrix in any of the medical diagnosis or prediction model. Hence the true positive counts and false negative counts will be the main area of focus along with the other performance metrices.

#### 5.3.4.1 Evaluation without Class Imbalance Handling

Training dataset before imbalance handling consists of 1,190 records, which was used to train the model. Test dataset consists of 510 records which was used to evaluate the model performance. Below figure 5.21 refers to the confusion matrix where rows are representing actual class labels and columns are representing predicted class labels.

As part of the true positive (TP) counts, it can be shown that 420 cases out of 426 were correctly predicted for the 'unknow' class label. For the 'cardiogenic shock' class label, 15 cases out of 34 were correctly predicted. For the 'pulmonary edema' class label, nothing (0 cases) was correctly predicted out of 3 cases. For the 'myocardial rupture' class label, 1 case out of 16 was correctly predicted. For the 'progress of congestive heart failure' class label, nothing (0 cases) was predicted out of 9 cases. For the 'thromboembolism' class label, nothing (0 cases) was anticipated out of 4 cases. For the 'asystole' class label, 1 case was correctly predicted out of 7 cases. Furthermore, nothing (0 cases) was predicted for the 'ventricular fibrillation' class label out of 11 cases.

As part of the false negative (FN) counts, 6 examples that truly correspond to the 'unknown' class label were forecasted as other class labels. 19 cases were projected as different class labels, despite the fact that they all belong to the 'cardiogenic shock' category. 3 cases, all of which correspond to the 'pulmonary edema' class label, were projected to have different class labels. 15 cases with the class label "myocardial rupture" were also predicted as other class labels. 9 instances were projected as different class labels, despite the fact that they all belong to the 'progress of congestive heart failure' class label. 4 examples, all of which fall under the 'thromboembolism' class label, were predicted to have different class labels. 6 instances with the class label 'asystole' were projected as different class labels, while 11 cases with the class label 'ventricular fibrillation' was also predicted as different class labels.

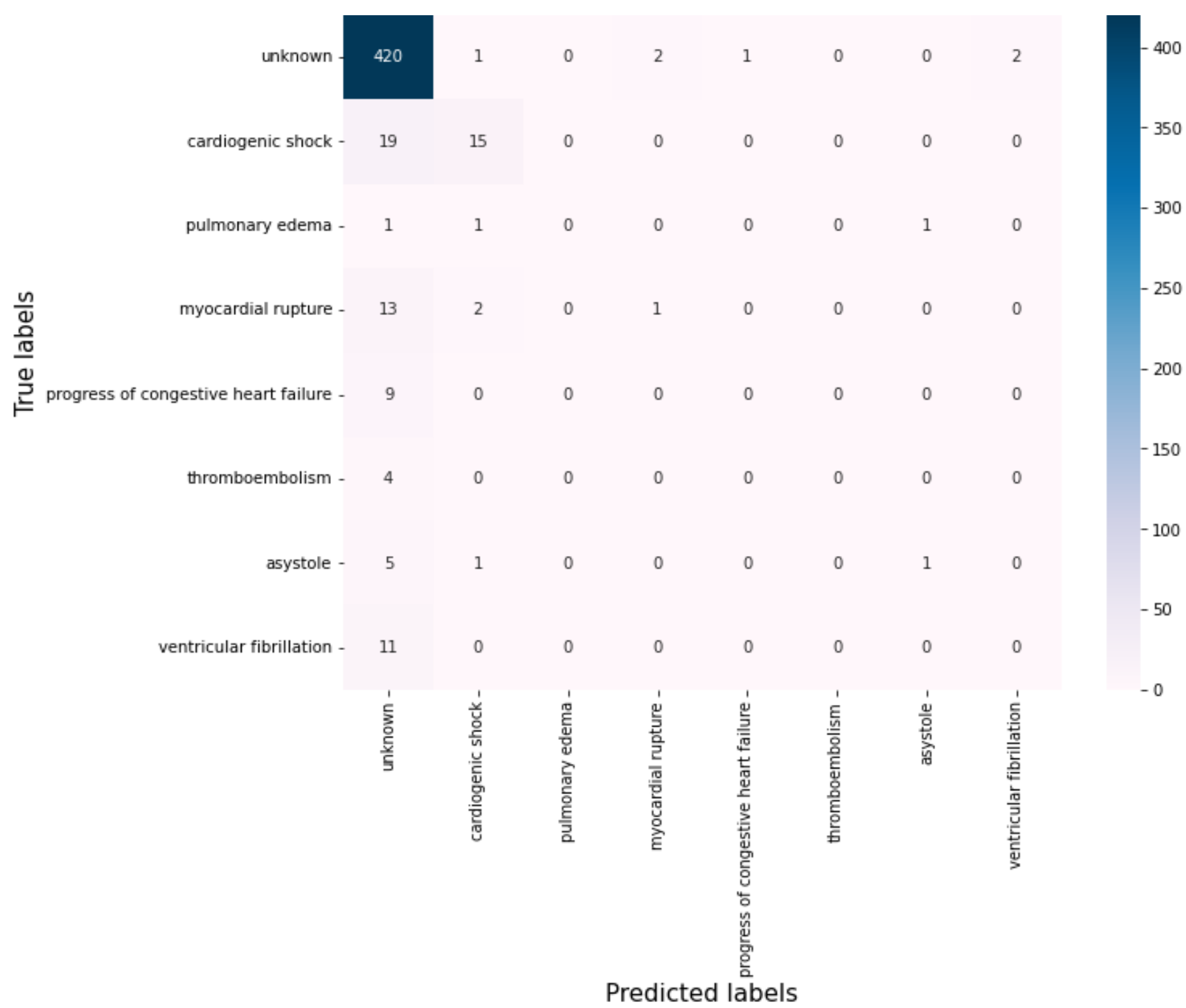

*Figure 5.21 A confusion matrix from bagging SVM modelling without class imbalance handling*

The classification report, shown in table 5.10, illustrates the various performance metrics for this multiclass classification task before class imbalance handling, as well as the precision, recall, and f1-score for each class. Additionally, the weighted average precision, weighted average recall, weighted average f1-score, and accuracy for the measuring overall model performance are 80%, 86%, 82%, and 86%, respectively.

*Table 5.10 The classification report from bagging SVM modelling without class imbalance handling*

| **Target class label** | **precision** | **recall** | **f1-score** | **Weighted avg precision** | **Weighted avg recall** | **Weighted avg f1-score** | **accuracy** |
|---|---|---|---|---|---|---|---|
| unknown | 0.87 | 0.99 | 0.93 | **0.80** | **0.86** | **0.82** | **0.86** |
| cardiogenic shock | 0.75 | 0.44 | 0.56 | | | | |
| pulmonary edema | 0.00 | 0.00 | 0.00 | | | | |
| myocardial rupture | 0.33 | 0.06 | 0.11 | | | | |

| progress of congestive heart failure | 0.00 | 0.00 | 0.00 | **0.80** | **0.86** | **0.82** | **0.86** |
|---|---|---|---|---|---|---|---|
| thromboembolism | 0.00 | 0.00 | 0.00 | | | | |
| asystole | 0.50 | 0.14 | 0.22 | | | | |
| ventricular fibrillation | 0.00 | 0.00 | 0.00 | | | | |

The ROC curve shown below in figure 5.22, illustrate how good the model is performing for predicting each class label before class imbalance handling. Also, the weighted average AUCROC score for the overall model is 77.63%.

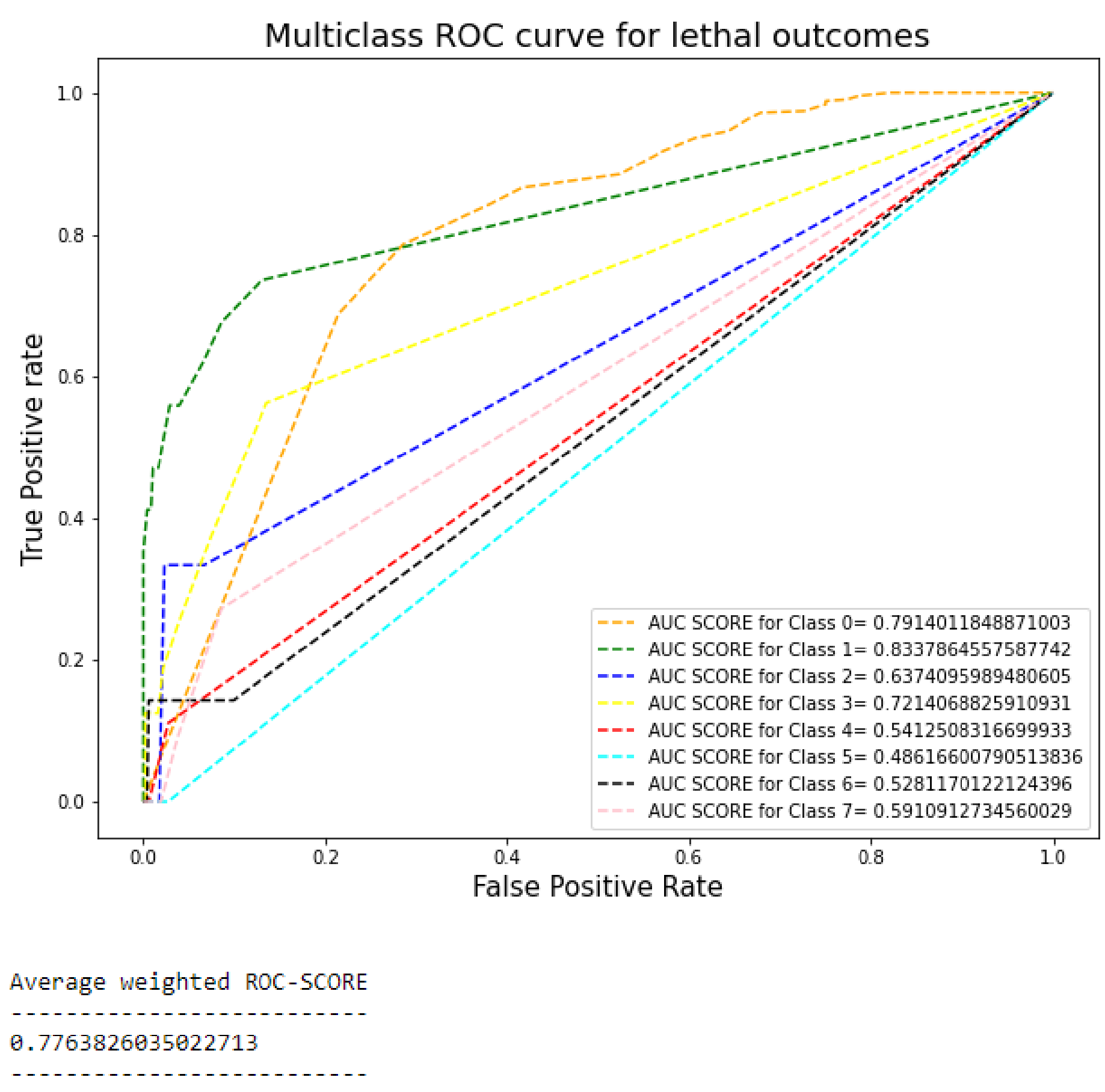

*Figure 5.22 AUCROC curve from bagging SVM modelling without class imbalance handling*

#### 5.3.4.2 Evaluation after Class Imbalance Handling (with ADASYN method)

Training dataset after imbalance handling with ADASYN method consists of 8,010 records, which was used to train the model. Test dataset consists of 510 records which was used to evaluate the model performance. Below figure 5.23 refers to the confusion matrix where rows are representing actual class labels and columns are representing predicted class labels.

As part of the true positive (TP) counts, it can be shown that 392 cases out of 426 were correctly predicted for the 'unknow' class label. For the 'cardiogenic shock' class label, 12 cases out of 34 were correctly predicted. For the 'pulmonary edema' class label, nothing (0 cases) was anticipated out of 3 cases. For the 'myocardial rupture' class label, nothing (0 cases) was anticipated out of 16 cases. For the 'progress of congestive heart failure' class label, nothing (0 cases) was anticipated out of 9 cases. For the 'thromboembolism' class label, nothing (0 cases) was anticipated out of 4 cases. For the 'asystole' class label, 1 case was correctly predicted out of 7 cases. Furthermore, nothing (0 cases) was anticipated for the 'ventricular fibrillation' class label out of 11 cases.

As part of the false negative (FN) counts, 34 examples that truly correspond to the 'unknown' class label were forecasted as other class labels. 22 cases were projected as different class labels, despite the fact that they all belong to the 'cardiogenic shock' category. 3 cases, all of which correspond to the 'pulmonary edema' class label, were projected to have different class labels. 16 cases with the class label "myocardial rupture" were also predicted as other class labels. 9 instances were projected as different class labels, despite the fact that they all belong to the 'progress of congestive heart failure' class label. 4 examples, all of which fall under the 'thromboembolism' class label, were predicted to have different class labels. 6 instances with the class label 'asystole' were projected as different class labels, while 11 cases with the class label 'ventricular fibrillation' was also predicted as different class labels.

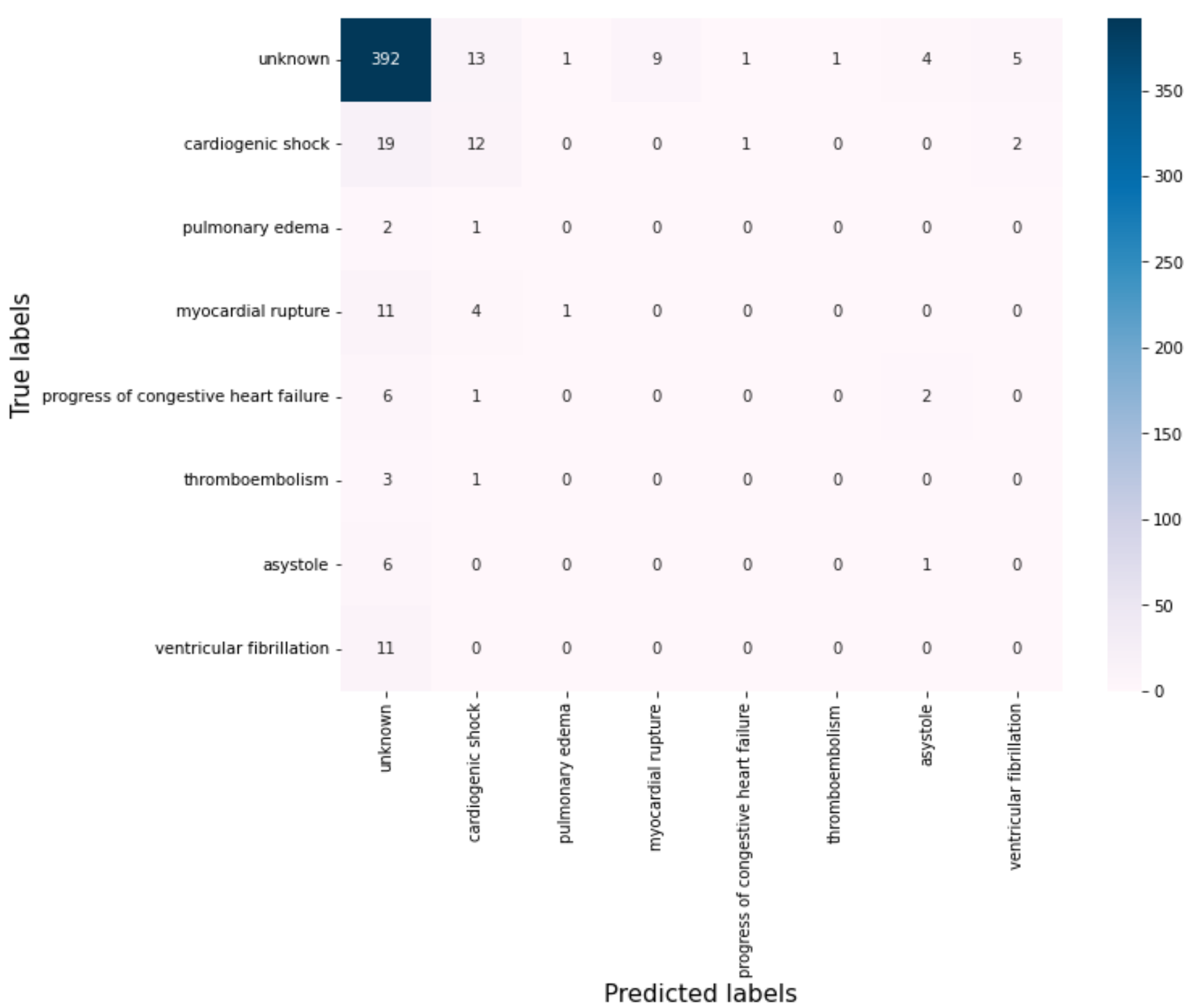

*Figure 5.23 A confusion matrix from bagging SVM modelling after class imbalance handling (ADASYN)*

The classification report, shown in table 5.11, illustrates the various performance metrics for this multiclass classification task after class imbalance treatment, as well as the precision, recall, and f1-score for each class. Additionally, the weighted average precision, weighted average recall, weighted average f1-score, and accuracy for the measuring overall model performance are 79%, 77%, 78%, and 77%, respectively.

*Table 5.11 The classification report from bagging SVM modelling after class imbalance handling (ADASYN)*

| **Target class label** | **precision** | **recall** | **f1-score** | **Weighted avg precision** | **Weighted avg recall** | **Weighted avg f1-score** | **accuracy** |
|---|---|---|---|---|---|---|---|
| unknown | 0.87 | 0.92 | 0.89 | **0.75** | **0.79** | **0.77** | **0.79** |
| cardiogenic shock | 0.38 | 0.35 | 0.36 | | | | |
| pulmonary edema | 0.00 | 0.00 | 0.00 | | | | |
| myocardial rupture | 0.00 | 0.00 | 0.00 | | | | |

| progress of congestive heart failure | 0.00 | 0.00 | 0.00 | **0.75** | **0.79** | **0.77** | **0.79** |
|---|---|---|---|---|---|---|---|
| thromboembolism | 0.00 | 0.00 | 0.00 | | | | |
| asystole | 0.14 | 0.14 | 0.14 | | | | |
| ventricular fibrillation | 0.00 | 0.00 | 0.00 | | | | |

The ROC curve shown below in figure 5.24, illustrate how good the model is performing for predicting each class label after class imbalance treatment. Also, the weighted average AUCROC score for the overall model is 70.70%.

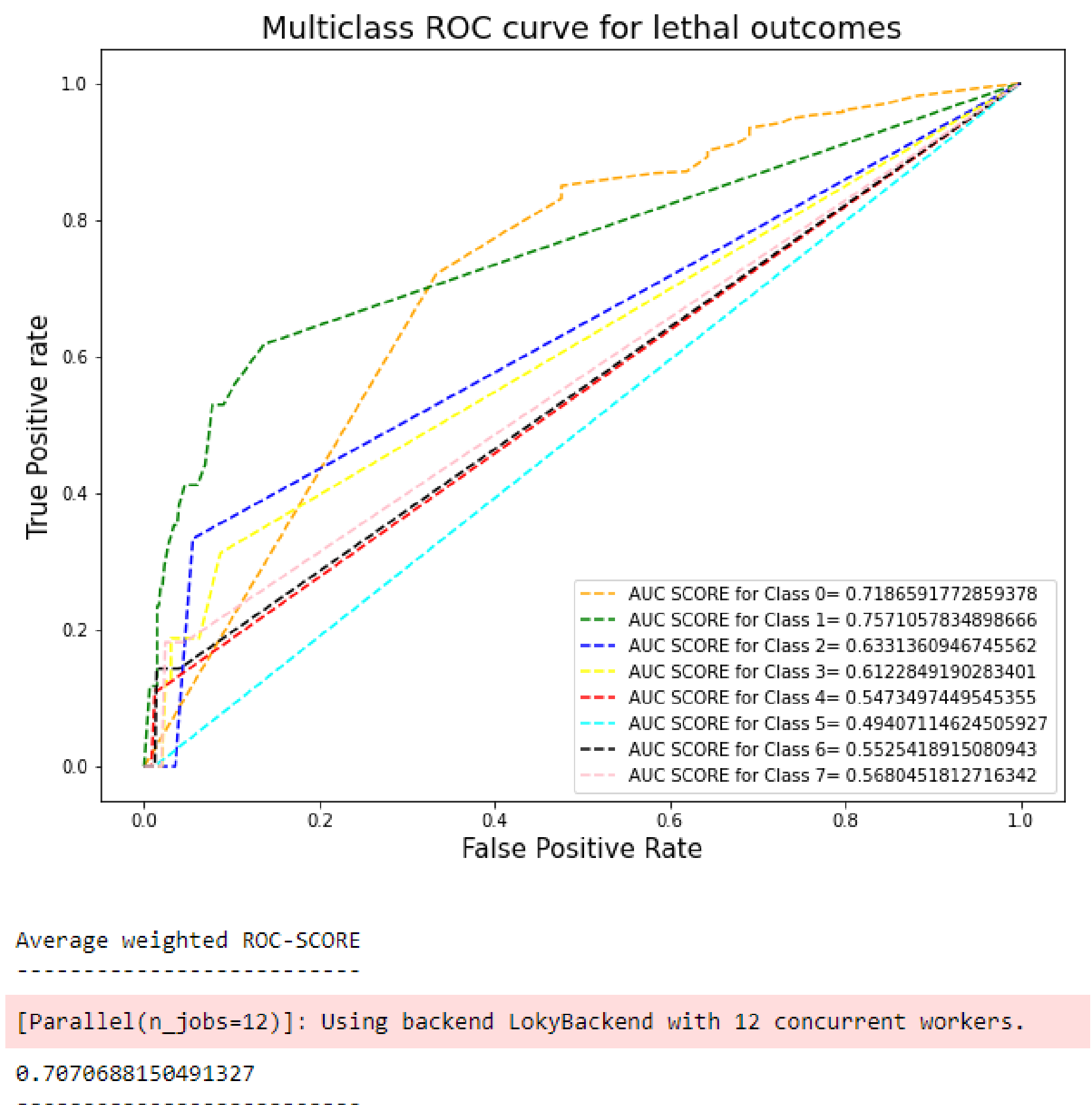

*Figure 5.24 AUCROC curve from bagging SVM modelling after class imbalance handling (ADASYN)*

#### 5.3.4.3 Evaluation after applying Class Weighted Method

Training dataset before imbalance handling consists of 1,190 records, which was used to train the model with class weighted hyperparameter (class_weight=’balanced’) to handle the imbalanced class. Test dataset consists of 510 records which was used to evaluate the model performance. Below figure 5.25 refers to the confusion matrix where rows are representing actual class labels and columns are representing predicted class labels.

As part of the true positive (TP) counts, it can be shown that 404 cases out of 426 were correctly predicted for the 'unknow' class label. For the 'cardiogenic shock' class label, 20 cases out of 34 were correctly predicted. For the 'pulmonary edema' class label, nothing (0 cases) was anticipated out of 3 cases. For the 'myocardial rupture' class label, 3 case was predicted correctly out of 16 cases. For the 'progress of congestive heart failure' class label, nothing (0 cases) was anticipated out of 9 cases. For the 'thromboembolism' class label, nothing (0 cases) was anticipated out of 4 cases. For the 'asystole' class label, 1 case was correctly predicted out of 7 cases. Furthermore, nothing (0 cases) was anticipated for the 'ventricular fibrillation' class label out of 11 cases.

As part of the false negative (FN) counts, 22 examples that truly correspond to the 'unknown' class label were forecasted as other class labels. 14 cases were projected as different class labels, despite the fact that they all belong to the 'cardiogenic shock' category. 3 cases, all of which correspond to the 'pulmonary edema' class label, were projected to have different class labels. 13 cases with the class label "myocardial rupture" were also predicted as other class labels. 9 instances were projected as different class labels, despite the fact that they all belong to the 'progress of congestive heart failure' class label. 4 examples, all of which fall under the 'thromboembolism' class label, were predicted to have different class labels. 6 instances with the class label 'asystole' were projected as different class labels, while 11 cases with the class label 'ventricular fibrillation' was also predicted as different class labels.

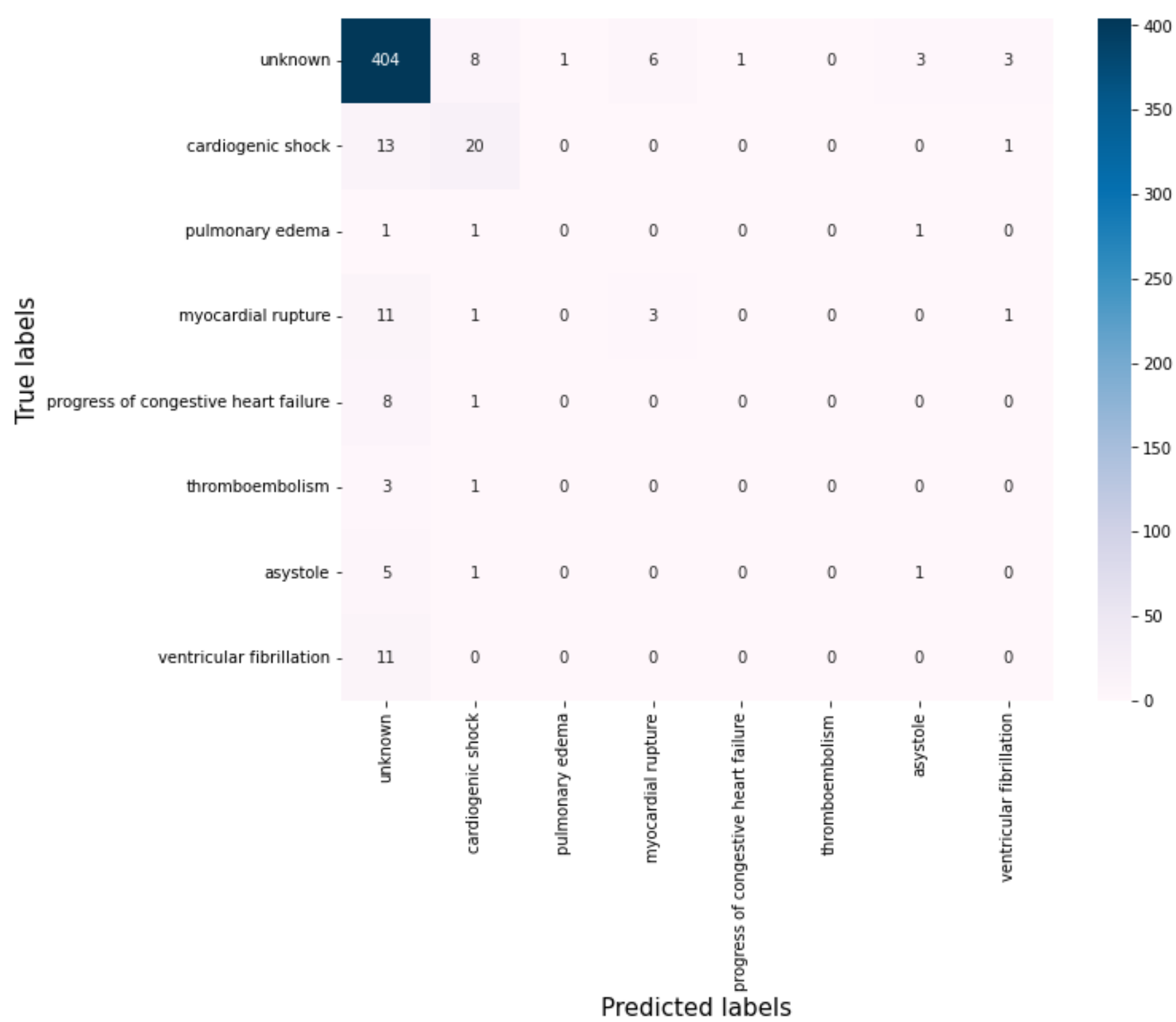

*Figure 5.25  A confusion matrix from bagging SVM modelling using class weight method*

The classification report, shown in table 5.12, illustrates the various performance metrics for this multiclass classification task with class weighted method of imbalance handling, as well as the precision, recall, and f1-score for each class. Additionally, the weighted average precision, weighted average recall, weighted average f1-score, and accuracy for the measuring overall model performance are 79%, 84%, 81%, and 84%, respectively.

*Table 5.12  The classification report from bagging SVM modelling with class weighted method*

| **Target class label** | **precision** | **recall** | **f1-score** | **Weighted avg precision** | **Weighted avg recall** | **Weighted avg f1-score** | **accuracy** |
|---|---|---|---|---|---|---|---|
| unknown | 0.89 | 0.95 | 0.92 | **0.79** | **0.84** | **0.81** | **0.84** |
| cardiogenic shock | 0.61 | 0.59 | 0.60 | | | | |
| pulmonary edema | 0.00 | 0.00 | 0.00 | | | | |
| myocardial rupture | 0.33 | 0.19 | 0.24 | | | | |

<table>
<tr><td>progress of congestive heart failure</td><td>0.00</td><td>0.00</td><td>0.00</td><td rowspan="4">0.79</td><td rowspan="4">0.84</td><td rowspan="4">0.81</td><td rowspan="4">0.84</td></tr>
<tr><td>thromboembolis m</td><td>0.00</td><td>0.00</td><td>0.00</td></tr>
<tr><td>asystole</td><td>0.20</td><td>0.14</td><td>0.17</td></tr>
<tr><td>ventricular fibrillation</td><td>0.00</td><td>0.00</td><td>0.00</td></tr>
</table>

The ROC curve shown below in figure 5.26, illustrate how good the model is performing for predicting each class label with class weighted method of imbalance handling. Also, the weighted average AUCROC score for the overall model is 81.48%.

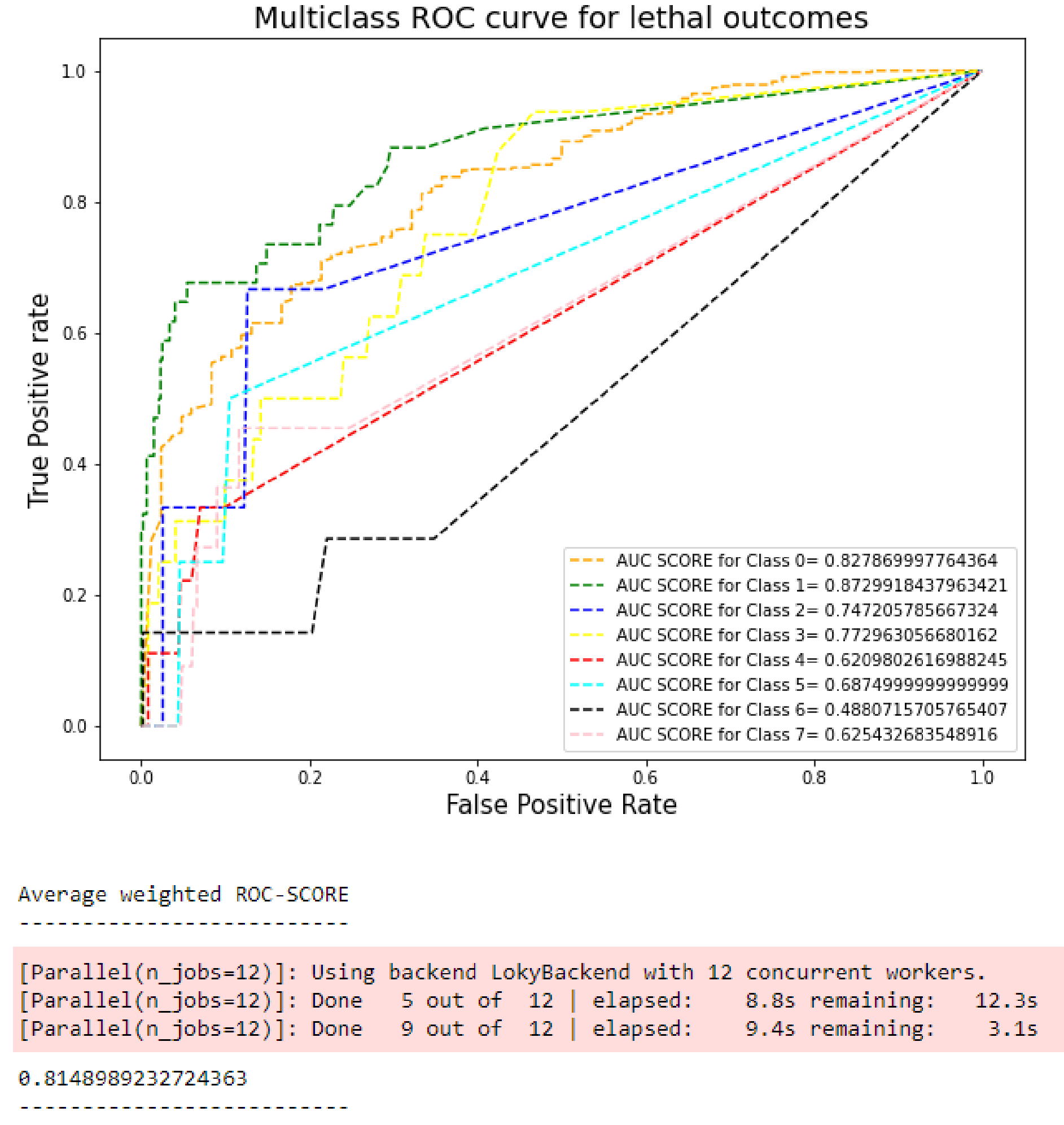

*Figure 5.26 AUCROC curve from bagging SVM modelling with class weighted method*

### 5.3.5 Stacking Blending Classifier's Evaluation and Results

For building a classification model, a stacking blending algorithm was utilised; this classifier was trained twice, once without imbalance handling and once after class imbalance treatment (with ADASYN and class weighted method). The evaluations/results of the classifier on the test set, as well as performance metrics such as the confusion matrix, weighted average F1 score, weighted average precision, weighted average recall, and the AUCROC curve, will be shown in the subsections below. As mentioned earlier, the true positive (TP) counts and false negative (FN) counts are important to predict appropriately from confusion matrix in any of the medical diagnosis or prediction model. Hence the true positive counts and false negative counts will be the main area of focus along with the other performance metrices.

#### 5.3.5.1 Evaluation without Class Imbalance Handling

Training dataset before imbalance handling consists of 1,190 records, which was used to train the model. Test dataset consists of 510 records which was used to evaluate the model performance. Below figure 5.27 refers to the confusion matrix where rows are representing actual class labels and columns are representing predicted class labels.

As part of the true positive (TP) counts, it can be shown that 424 cases out of 426 were correctly predicted for the 'unknow' class label. For the 'cardiogenic shock' class label, 16 cases out of 34 were correctly predicted. For the 'pulmonary edema' class label, nothing (0 case) was correctly predicted out of 3 cases. For the 'myocardial rupture' class label, nothing (0 case) out of 16 was correctly predicted. For the 'progress of congestive heart failure' class label, nothing (0 case) was predicted out of 9 cases. For the 'thromboembolism' class label, again nothing (0 case) was anticipated out of 4 cases. For the 'asystole' class label, nothing (0 case) was correctly predicted out of 7 cases. Furthermore, nothing (0 case) was predicted for the 'ventricular fibrillation' class label out of 11 cases.

As part of the false negative (FN) counts, 2 examples that truly correspond to the 'unknown' class label were forecasted as other class labels. 18 cases were projected as different class labels, despite the fact that they all belong to the 'cardiogenic shock' category. 3 cases, all of which correspond to the 'pulmonary edema' class label, were projected to have different class labels. 16 cases with the class label "myocardial rupture" were also predicted as other class labels. 9 instances were projected as different class labels, despite the fact that they all belong to the 'progress of congestive heart failure' class label. 4 examples, all of which fall under the 'thromboembolism' class label, were predicted to have different class labels. 7 instances with the class label 'asystole' were projected as different class labels, while 11 cases with the class label 'ventricular fibrillation' was also predicted as different class labels.

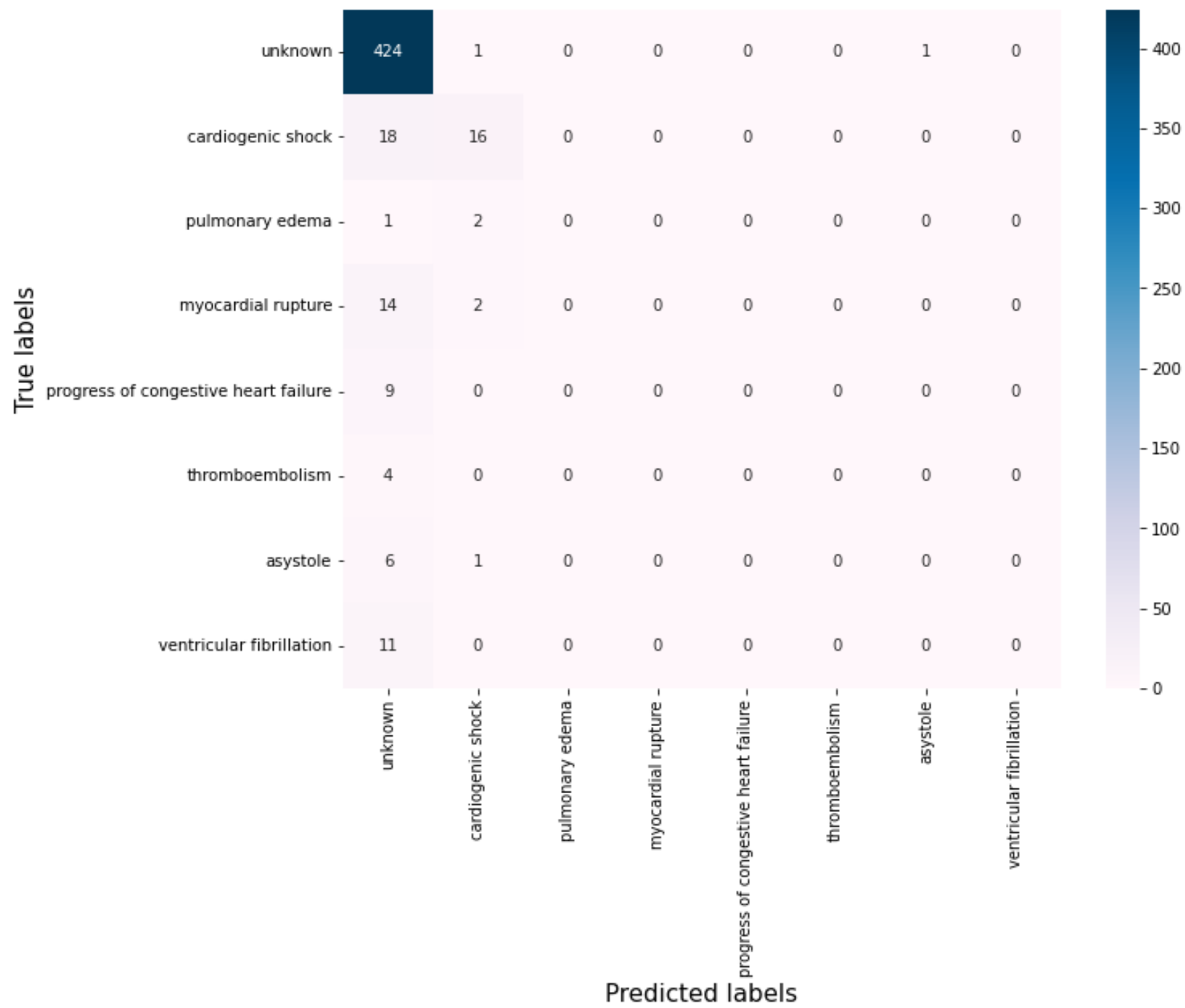

*Figure 5.27 A confusion matrix from stacking blending modelling without class imbalance handling*

The classification report, shown in table 5.13, illustrates the various performance metrics for this multiclass classification task before class imbalance handling, as well as the precision, recall, and f1-score for each class. Additionally, the weighted average precision, weighted average recall, weighted average f1-score, and accuracy for the measuring overall model performance are 78%, 86%, 81%, and 86%, respectively.

*Table 5.13 The classification report from stacking blending without class imbalance handling*

| **Target class label** | **precision** | **recall** | **f1-score** | **Weighted avg precision** | **Weighted avg recall** | **Weighted avg f1-score** | **accuracy** |
|---|---|---|---|---|---|---|---|
| unknown | 0.87 | 1.00 | 0.93 | **0.78** | **0.86** | **0.81** | **0.86** |
| cardiogenic shock | 0.73 | 0.47 | 0.57 | | | | |
| pulmonary edema | 0.00 | 0.00 | 0.00 | | | | |
| myocardial rupture | 0.00 | 0.00 | 0.00 | | | | |
| progress of congestive heart failure | 0.00 | 0.00 | 0.00 | **0.78** | **0.86** | **0.81** | **0.86** |
| thromboembolism | 0.00 | 0.00 | 0.00 | | | | |
| asystole | 0.00 | 0.00 | 0.00 | | | | |
| ventricular fibrillation | 0.00 | 0.00 | 0.00 | | | | |

The ROC curve shown below in figure 5.28, illustrate how good the model is performing for predicting each class label before class imbalance handling. Also, the weighted average AUCROC score for the overall model is 84.03%.

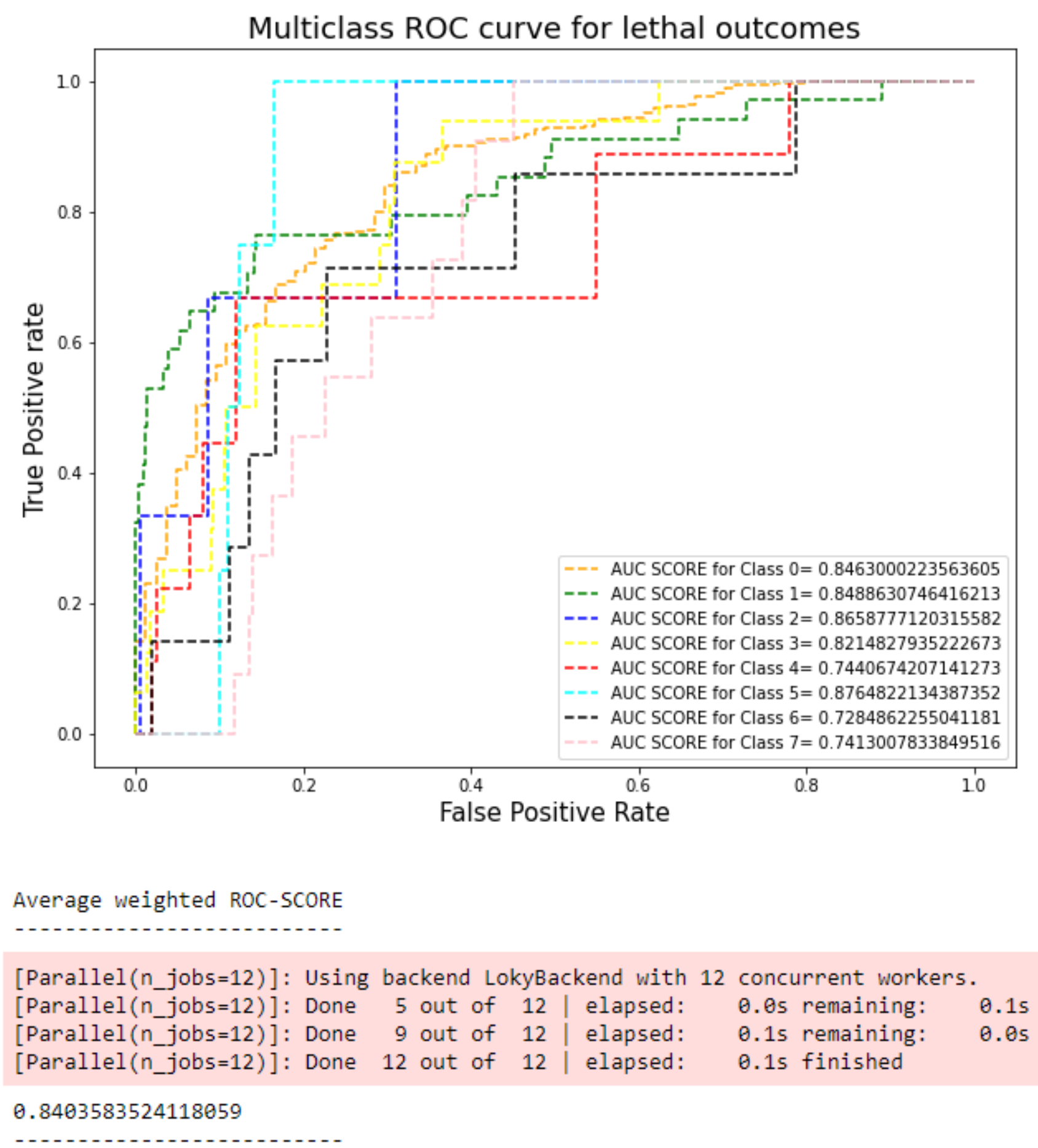

*Figure 5.28 AUCROC curve from stacking blending modelling without class imbalance handling*

#### 5.3.5.2 Evaluation after Class Imbalance Handling (with ADASYN method)

Training dataset after imbalance handling with ADASYN method consists of 8,010 records, which was used to train the model. Test dataset consists of 510 records which was used to evaluate the model performance. Below figure 5.29 refers to the confusion matrix where rows are representing actual class labels and columns are representing predicted class labels.

As part of the true positive (TP) counts, it can be shown that 402 cases out of 426 were correctly predicted for the 'unknow' class label. For the 'cardiogenic shock' class label, 15 cases out of 34 were correctly predicted. For the 'pulmonary edema' class label, nothing (0 cases) was anticipated out of 3 cases. For the 'myocardial rupture' class label, 1 case was correctly predicted out of 16 cases. For the 'progress of congestive heart failure' class label, nothing (0 cases) was anticipated out of 9 cases. For the 'thromboembolism' class label, nothing (0 cases) was anticipated out of 4 cases. For the 'asystole' class label, nothing (0 cases) was anticipated out of 7 cases. Furthermore, nothing (0 cases) was anticipated for the 'ventricular fibrillation' class label out of 11 cases.

As part of the false negative (FN) counts, 24 examples that truly correspond to the 'unknown' class label were forecasted as other class labels. 19 cases were projected as different class labels, despite the fact that they all belong to the 'cardiogenic shock' category. 3 cases, all of which correspond to the 'pulmonary edema' class label, were projected to have different class labels. 15 cases with the class label "myocardial rupture" were also predicted as other class labels. 9 instances were projected as different class labels, despite the fact that they all belong to the 'progress of congestive heart failure' class label. 4 examples, all of which fall under the 'thromboembolism' class label, were predicted to have different class labels. 7 instances with the class label 'asystole' were projected as different class labels, while 11 cases with the class label 'ventricular fibrillation' was also predicted as different class labels.

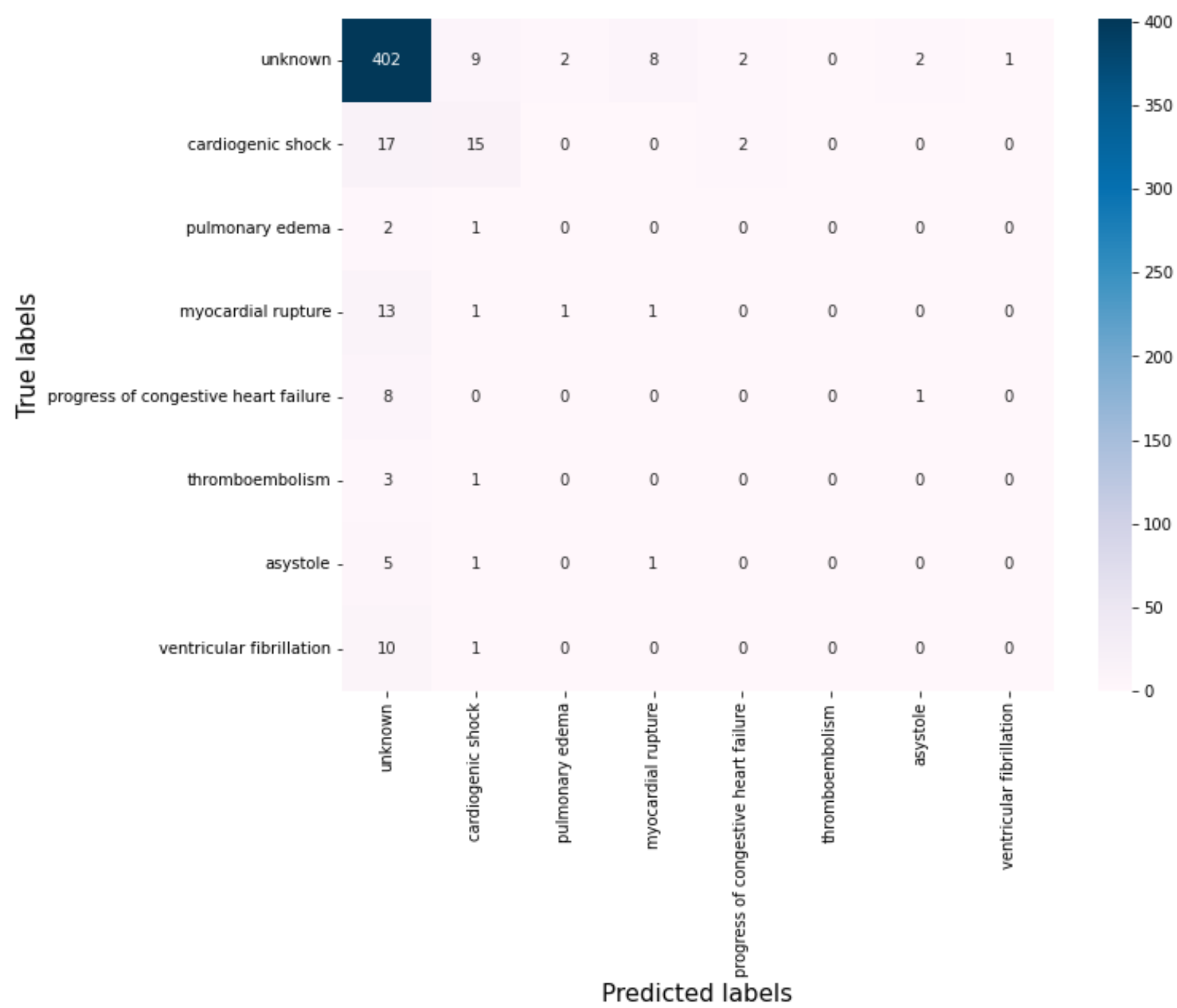

*Figure 5.29 A confusion matrix from stacking blending modelling after class imbalance handling (ADASYN)*

The classification report, shown in table 5.14, illustrates the various performance metrics for this multiclass classification task after class imbalance treatment, as well as the precision, recall, and f1-score for each class. Additionally, the weighted average precision, weighted average recall, weighted average f1-score, and accuracy for the measuring overall model performance are 77%, 82%, 79%, and 82%, respectively.

*Table 5.14 The classification report from stacking blending modelling after class imbalance handling (ADASYN)*

<table>
<tr><th>Target class label</th><th>precision</th><th>recall</th><th>f1-score</th><th>Weighted avg precision</th><th>Weighted avg recall</th><th>Weighted avg f1-score</th><th>accuracy</th></tr>
<tr><td>unknown</td><td>0.87</td><td>0.94</td><td>0.91</td><td rowspan="4">0.77</td><td rowspan="4">0.82</td><td rowspan="4">0.79</td><td rowspan="4">0.82</td></tr>
<tr><td>cardiogenic shock</td><td>0.52</td><td>0.44</td><td>0.48</td></tr>
<tr><td>pulmonary edema</td><td>0.00</td><td>0.00</td><td>0.00</td></tr>
<tr><td>myocardial rupture</td><td>0.10</td><td>0.06</td><td>0.08</td></tr>
</table>

<table>
<tr><td>progress of congestive heart failure</td><td>0.00</td><td>0.00</td><td>0.00</td><td rowspan="4">0.77</td><td rowspan="4">0.82</td><td rowspan="4">0.79</td><td rowspan="4">0.82</td></tr>
<tr><td>thromboembolis m</td><td>0.00</td><td>0.00</td><td>0.00</td></tr>
<tr><td>asystole</td><td>0.00</td><td>0.00</td><td>0.00</td></tr>
<tr><td>ventricular fibrillation</td><td>0.00</td><td>0.00</td><td>0.00</td></tr>
</table>

The ROC curve shown below in figure 5.30, illustrate how good the model is performing for predicting each class label after class imbalance treatment. Also, the weighted average AUCROC score for the overall model is 75.48%.

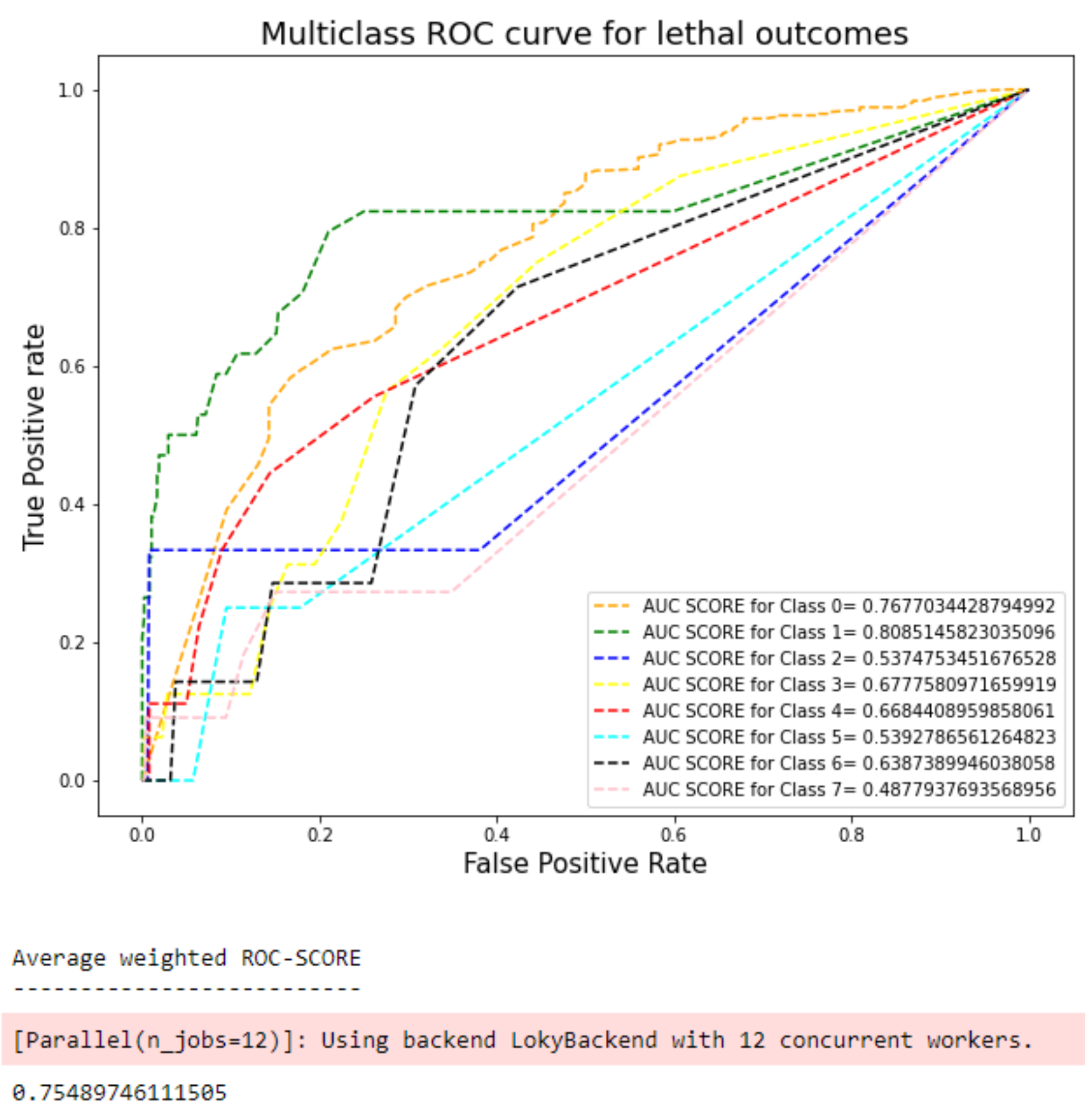

*Figure 5.30 AUCROC curve from stacking blending modelling after class imbalance handling (ADASYN)*

#### 5.3.5.3 Evaluation after applying Class Weighted Method

Training dataset before imbalance handling consists of 1,190 records, which was used to train the model with class weighted hyperparameter (class_weight='balanced') to handle the imbalanced class. Test dataset consists of 510 records which was used to evaluate the model performance. Below figure 5.31 refers to the confusion matrix where rows are representing actual class labels and columns are representing predicted class labels.

As part of the true positive (TP) counts, it can be shown that 288 cases out of 426 were correctly predicted for the 'unknow' class label. For the 'cardiogenic shock' class label, 18 cases out of 34 were correctly predicted. For the 'pulmonary edema' class label, nothing (0 cases) was anticipated out of 3 cases. For the 'myocardial rupture' class label, 4 case was predicted correctly out of 16 cases. For the 'progress of congestive heart failure' class label, 2 cases were correctly predicted out of 9 cases. For the 'thromboembolism' class label, 3 cases were correctly predicted out of 4 cases. For the 'asystole' class label, nothing (0 cases) was anticipated out of 7 cases. Furthermore, 1 case was correctly predicted for the 'ventricular fibrillation' class label out of 11 cases.

As part of the false negative (FN) counts, 138 examples that truly correspond to the 'unknown' class label were forecasted as other class labels. 16 cases were projected as different class labels, despite the fact that they all belong to the 'cardiogenic shock' category. 3 cases, all of which correspond to the 'pulmonary edema' class label, were projected to have different class labels. 12 cases with the class label "myocardial rupture" were also predicted as other class labels. 7 instances were projected as different class labels, despite the fact that they all belong to the 'progress of congestive heart failure' class label. 1 example, all of which fall under the 'thromboembolism' class label, were predicted to have different class labels. 7 instances with the class label 'asystole' were projected as different class labels, while 10 cases with the class label 'ventricular fibrillation' was also predicted as different class labels.

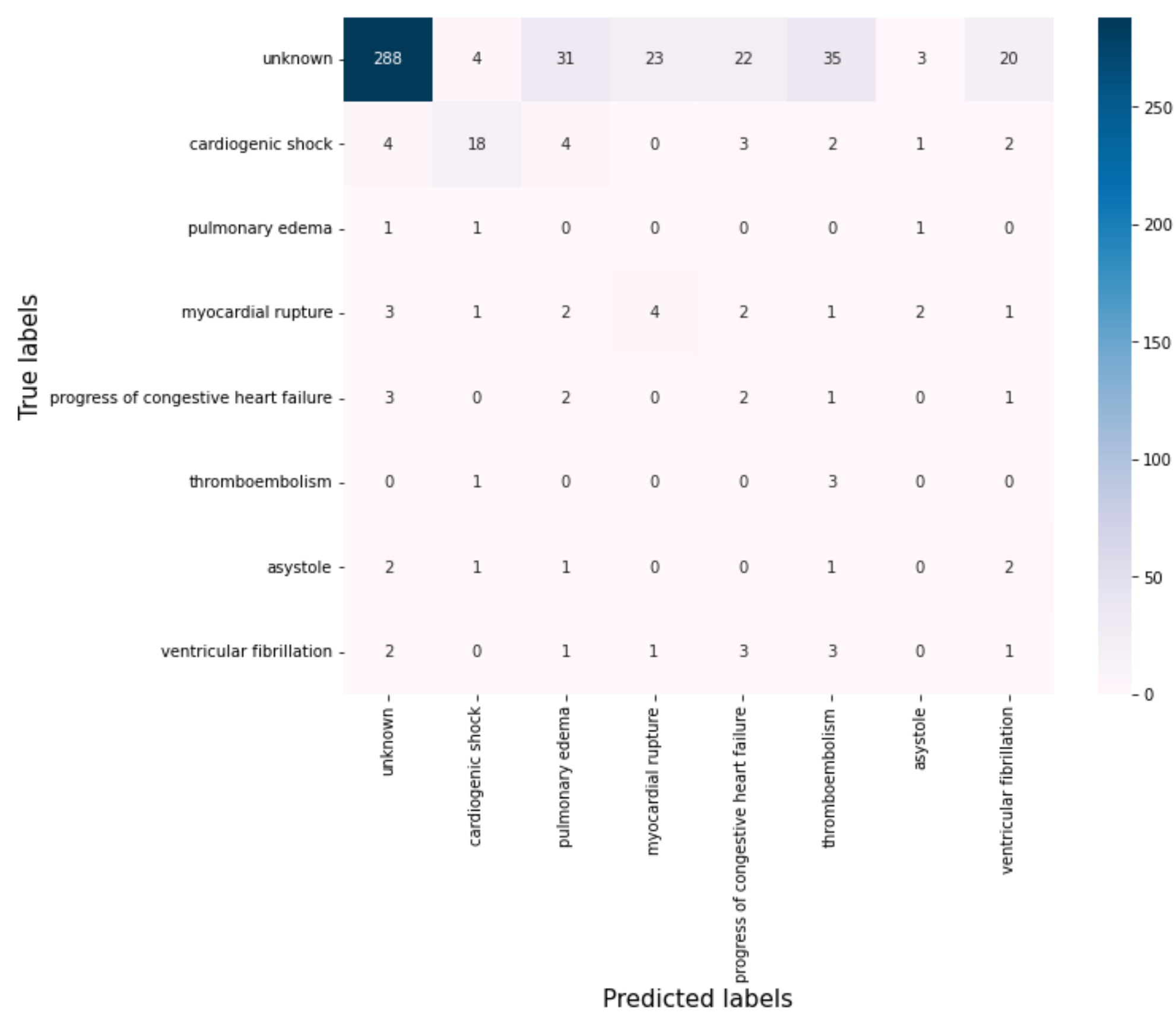

*Figure 5.31 A confusion matrix from stacking blending modelling using class weight method*

The classification report, shown in table 5.15, illustrates the various performance metrics for this multiclass classification task with class weighted method of imbalance handling, as well as the precision, recall, and f1-score for each class. Additionally, the weighted average precision, weighted average recall, weighted average f1-score, and accuracy for the measuring overall model performance are 85%, 62%, 71%, and 62%, respectively.

*Table 5.15 The classification report from stacking blending modelling with class weighted method*

<table>
<tr><th>Target class label</th><th>precision</th><th>recall</th><th>f1-score</th><th>Weighted avg precision</th><th>Weighted avg recall</th><th>Weighted avg f1-score</th><th>accuracy</th></tr>
<tr><td>unknown</td><td>0.95</td><td>0.68</td><td>0.79</td><td rowspan="4">0.85</td><td rowspan="4">0.62</td><td rowspan="4">0.71</td><td rowspan="4">0.62</td></tr>
<tr><td>cardiogenic shock</td><td>0.69</td><td>0.53</td><td>0.60</td></tr>
<tr><td>pulmonary edema</td><td>0.00</td><td>0.00</td><td>0.00</td></tr>
<tr><td>myocardial rupture</td><td>0.14</td><td>0.25</td><td>0.18</td></tr>
</table>

| progress of congestive heart failure | 0.06 | 0.22 | 0.10 | | | | |
|---|---|---|---|---|---|---|---|
| thromboembolism | 0.07 | 0.75 | 0.12 | **0.85** | **0.62** | **0.71** | **0.62** |
| asystole | 0.00 | 0.00 | 0.00 | | | | |
| ventricular fibrillation | 0.04 | 0.09 | 0.05 | | | | |

The ROC curve shown below in figure 5.32, illustrate how good the model is performing for predicting each class label with class weighted method of imbalance handling. Also, the weighted average AUCROC score for the overall model is 81.26%.

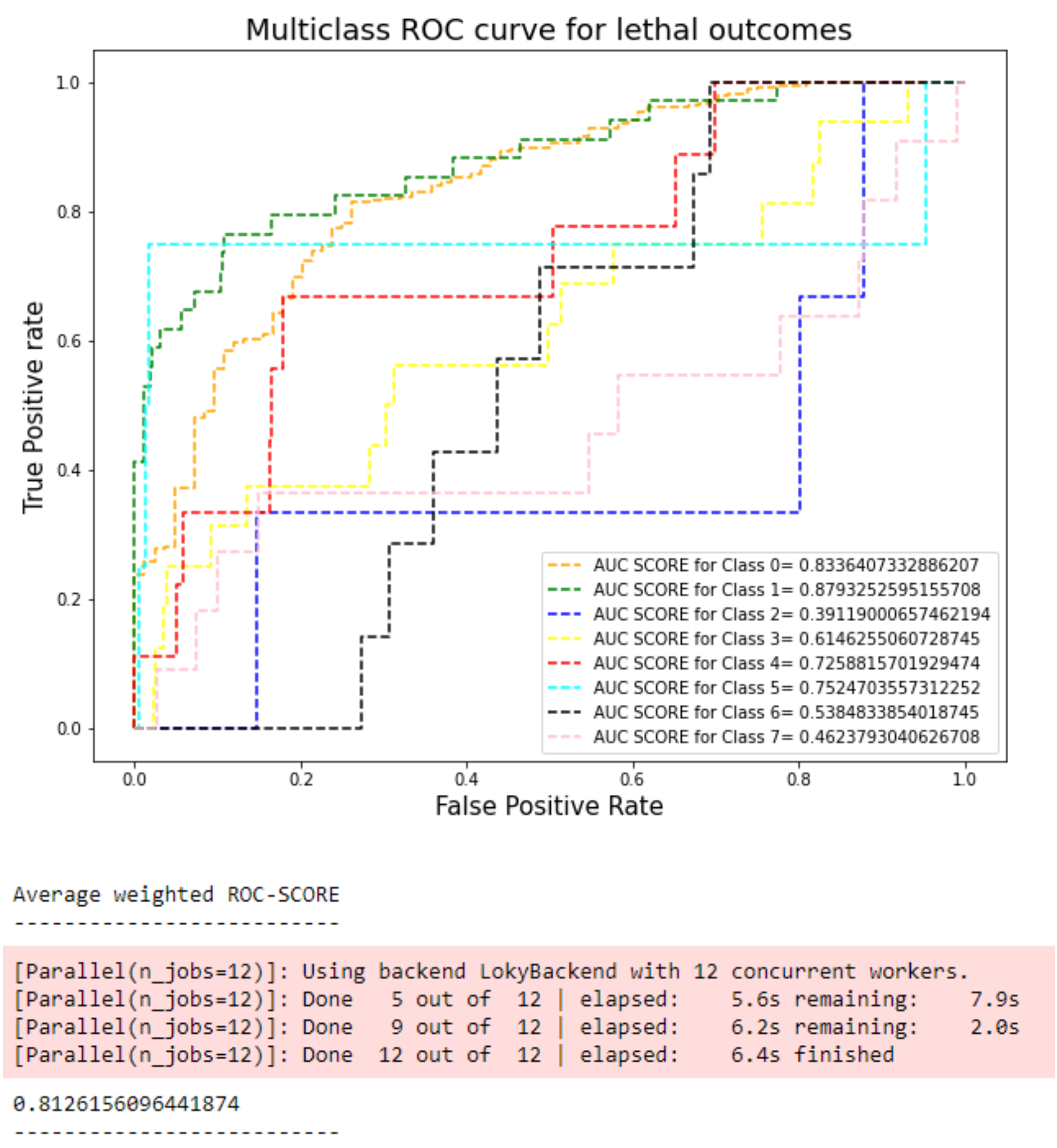

*Figure 5.32 AUCROC curve from stacking blending modelling with class weighted method*

### 5.3.6 Artificial Neural Network Classifier's Evaluation and Results

For building a classification model, a deep artificial neural network algorithm was utilised; this classifier was trained twice, once without imbalance handling and once after class imbalance treatment (with ADASYN and class weighted method). The evaluations/results of the classifier on the test set, as well as performance metrics such as the confusion matrix, weighted average F1 score, weighted average precision, weighted average recall, and the AUCROC curve, will be shown in the subsections below. As mentioned earlier, the true positive (TP) counts and false negative (FN) counts are important to predict appropriately from confusion matrix in any of the medical diagnosis or prediction model. Hence the true positive counts and false negative counts will be the main area of focus along with the other performance metrices.

#### 5.3.6.1 Evaluation without Class Imbalance Handling

Training dataset before imbalance handling consists of 1,190 records, which was used to train the model. Test dataset consists of 510 records which was used to evaluate the model performance. Below figure 5.33 refers to the confusion matrix where rows are representing actual class labels and columns are representing predicted class labels.

As part of the true positive (TP) counts, it can be shown that 423 cases out of 426 were correctly predicted for the 'unknow' class label. For the 'cardiogenic shock' class label, 12 cases out of 34 were correctly predicted. For the 'pulmonary edema' class label, nothing (0 case) was correctly predicted out of 3 cases. For the 'myocardial rupture' class label, nothing (0 case) out of 16 was correctly predicted. For the 'progress of congestive heart failure' class label, nothing (0 case) was predicted out of 9 cases. For the 'thromboembolism' class label, again nothing (0 case) was anticipated out of 4 cases. For the 'asystole' class label, nothing (0 case) was correctly predicted out of 7 cases. Furthermore, nothing (0 case) was predicted for the 'ventricular fibrillation' class label out of 11 cases.

As part of the false negative (FN) counts, 3 examples that truly correspond to the 'unknown' class label were forecasted as other class labels. 22 cases were projected as different class labels, despite the fact that they all belong to the 'cardiogenic shock' category. 3 cases, all of which correspond to the 'pulmonary edema' class label, were projected to have different class labels. 16 cases with the class label "myocardial rupture" were also predicted as other class labels. 9 instances were projected as different class labels, despite the fact that they all belong to the 'progress of congestive heart failure' class label. 4 examples, all of which fall under the 'thromboembolism' class label, were predicted to have different class labels. 7 instances with the class label 'asystole' were projected as different class labels, while 11 cases with the class label 'ventricular fibrillation' was also predicted as different class labels.

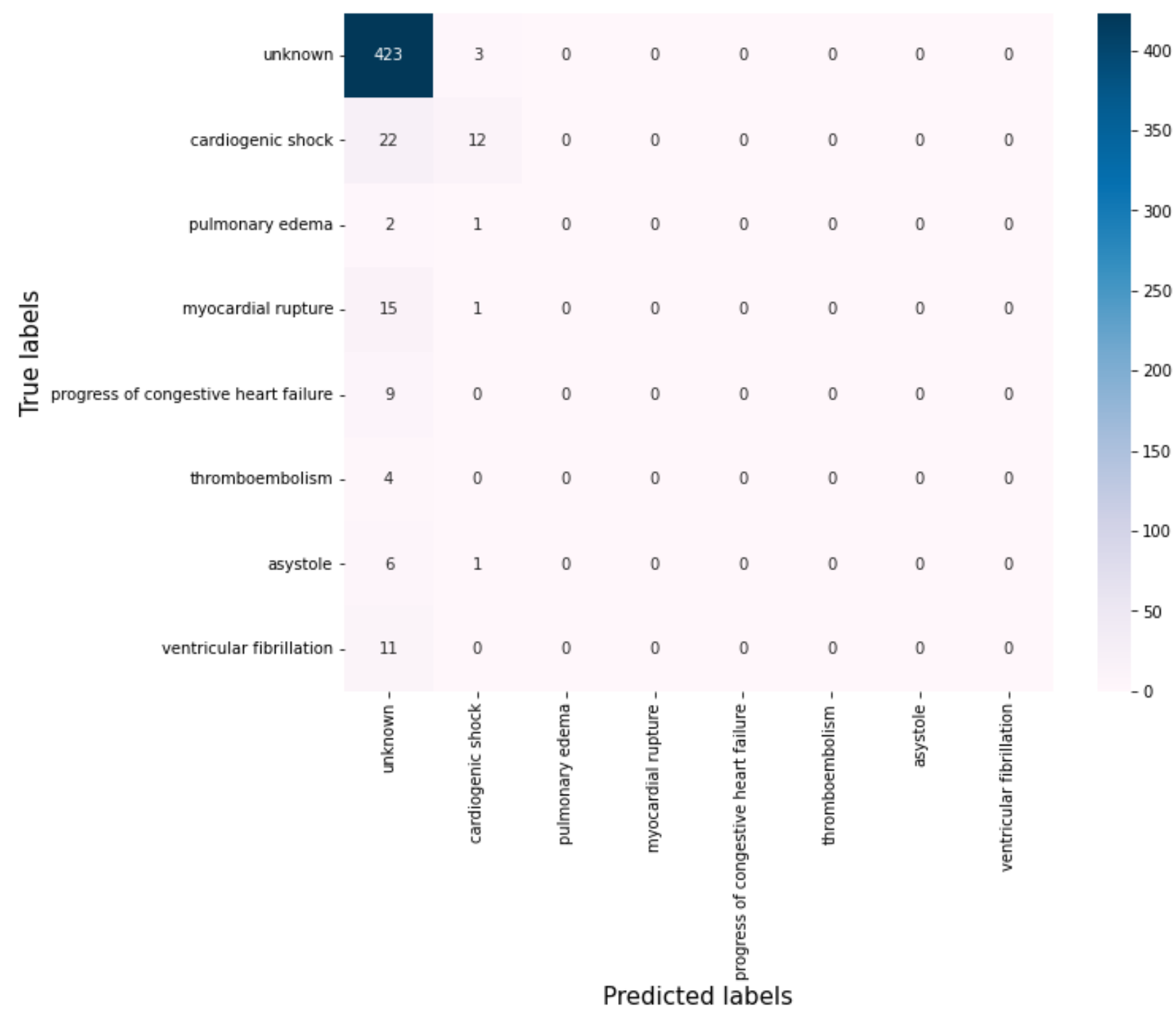

*Figure 5.33 A confusion matrix from ANN modelling without class imbalance handling*

The classification report, shown in table 5.16, illustrates the various performance metrics for this multiclass classification task before class imbalance handling, as well as the precision, recall, and f1-score for each class. Additionally, the weighted average precision, weighted average recall, weighted average f1-score, and accuracy for the measuring overall model performance are 76%, 85%, 80%, and 85%, respectively.

*Table 5.16 The classification report from ANN modelling without class imbalance handling*

| **Target class label** | **precision** | **recall** | **f1-score** | **Weighted avg precision** | **Weighted avg recall** | **Weighted avg f1-score** | **accuracy** |
|---|---|---|---|---|---|---|---|
| unknown | 0.86 | 0.99 | 0.92 | **0.76** | **0.85** | **0.80** | **0.85** |
| cardiogenic shock | 0.67 | 0.35 | 0.46 | | | | |
| pulmonary edema | 0.00 | 0.00 | 0.00 | | | | |
| myocardial rupture | 0.00 | 0.00 | 0.00 | | | | |

<table>
<tr><td>progress of congestive heart failure</td><td>0.00</td><td>0.00</td><td>0.00</td><td rowspan="4">0.76</td><td rowspan="4">0.85</td><td rowspan="4">0.80</td><td rowspan="4">0.85</td></tr>
<tr><td>thromboembolism</td><td>0.00</td><td>0.00</td><td>0.00</td></tr>
<tr><td>asystole</td><td>0.00</td><td>0.00</td><td>0.00</td></tr>
<tr><td>ventricular fibrillation</td><td>0.00</td><td>0.00</td><td>0.00</td></tr>
</table>

The ROC curve shown below in figure 5.34, illustrate how good the model is performing for predicting each class label before class imbalance handling. Also, the weighted average AUCROC score for the overall model is 78.20%.

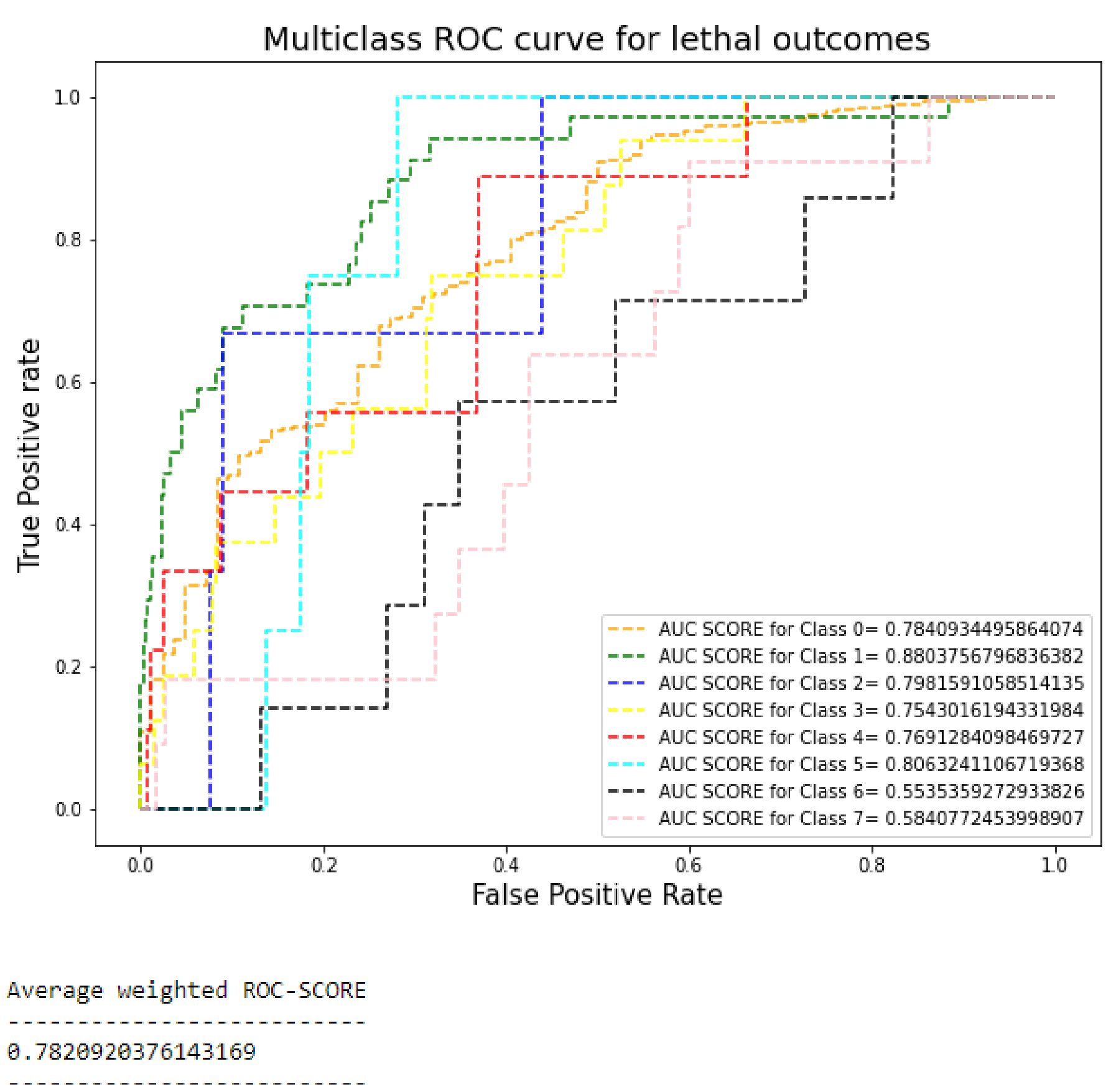

*Figure 5.34 AUCROC curve from ANN modelling without class imbalance handling*

#### 5.3.6.2 Evaluation after Class Imbalance Handling (with ADASYN method)

Training dataset after imbalance handling with ADASYN method consists of 8,010 records, which was used to train the model. Test dataset consists of 510 records which was used to evaluate the model performance. Below figure 5.35 refers to the confusion matrix where rows are representing actual class labels and columns are representing predicted class labels.

As part of the true positive (TP) counts, it can be shown that 396 cases out of 426 were correctly predicted for the 'unknow' class label. For the 'cardiogenic shock' class label, 12 cases out of 34 were correctly predicted. For the 'pulmonary edema' class label, nothing (0 case) was anticipated out of 3 cases. For the 'myocardial rupture' class label, 3 case was correctly predicted out of 16 cases. For the 'progress of congestive heart failure' class label, 1 case was correctly predicted. For the 'thromboembolism' class label, nothing (0 case) was anticipated out of 4 cases. For the 'asystole' class label, nothing (0 case) was anticipated out of 7 cases. Furthermore, 1 case was correctly predicted for the 'ventricular fibrillation' class label out of 11 cases.

As part of the false negative (FN) counts, 30 examples that truly correspond to the 'unknown' class label were forecasted as other class labels. 22 cases were projected as different class labels, despite the fact that they all belong to the 'cardiogenic shock' category. 3 cases, all of which correspond to the 'pulmonary edema' class label, were projected to have different class labels. 13 cases with the class label "myocardial rupture" were also predicted as other class labels. 8 instances were projected as different class labels, despite the fact that they all belong to the 'progress of congestive heart failure' class label. 4 examples, all of which fall under the 'thromboembolism' class label, were predicted to have different class labels. 7 instances with the class label 'asystole' were projected as different class labels, while 10 cases with the class label 'ventricular fibrillation' was also predicted as different class labels.

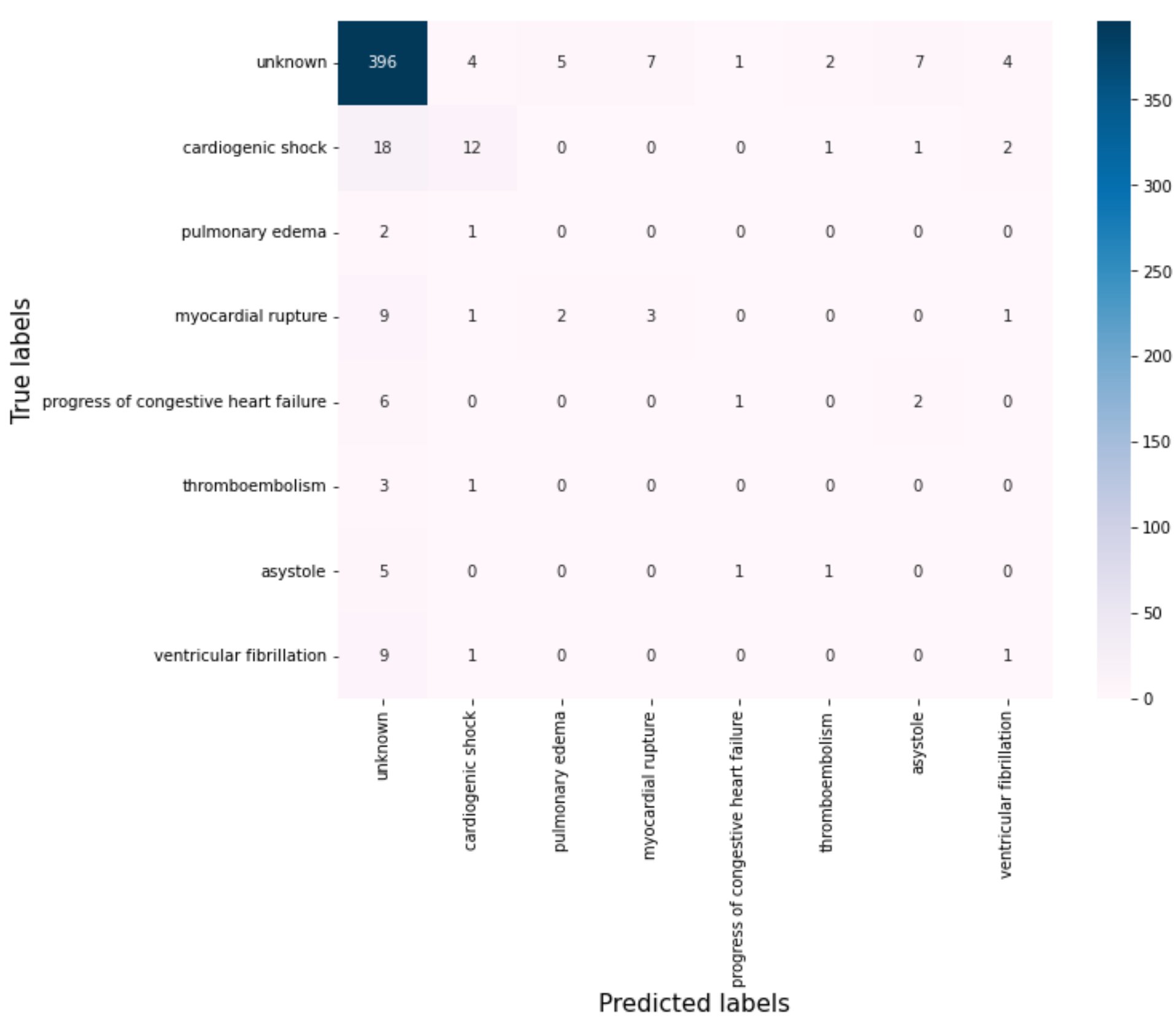

*Figure 5.35 A confusion matrix from ANN modelling after class imbalance handling (ADASYN)*

The classification report, shown in table 5.17, illustrates the various performance metrics for this multiclass classification task after class imbalance treatment, as well as the precision, recall, and f1-score for each class. Additionally, the weighted average precision, weighted average recall, weighted average f1-score, and accuracy for the measuring overall model performance are 80%, 81%, 80%, and 81%, respectively.

*Table 5.17 The classification report from ANN modelling after class imbalance handling (ADASYN)*

| Target class label | precision | recall | f1-score | Weighted avg precision | Weighted avg recall | Weighted avg f1-score | accuracy |
|---|---|---|---|---|---|---|---|
| unknown | 0.88 | 0.93 | 0.91 | **0.80** | **0.81** | **0.80** | **0.81** |
| cardiogenic shock | 0.60 | 0.35 | 0.44 | | | | |
| pulmonary edema | 0.00 | 0.00 | 0.00 | | | | |
| myocardial rupture | 0.30 | 0.19 | 0.23 | | | | |

| progress of congestive heart failure | 0.33 | 0.11 | 0.17 | **0.80** | **0.81** | **0.80** | **0.81** |
|---|---|---|---|---|---|---|---|
| thromboembolism | 0.00 | 0.00 | 0.00 | | | | |
| asystole | 0.00 | 0.00 | 0.00 | | | | |
| ventricular fibrillation | 0.12 | 0.09 | 0.11 | | | | |

The ROC curve shown below in figure 5.36, illustrate how good the model is performing for predicting each class label after class imbalance treatment. Also, the weighted average AUCROC score for the overall model is 74.84%.

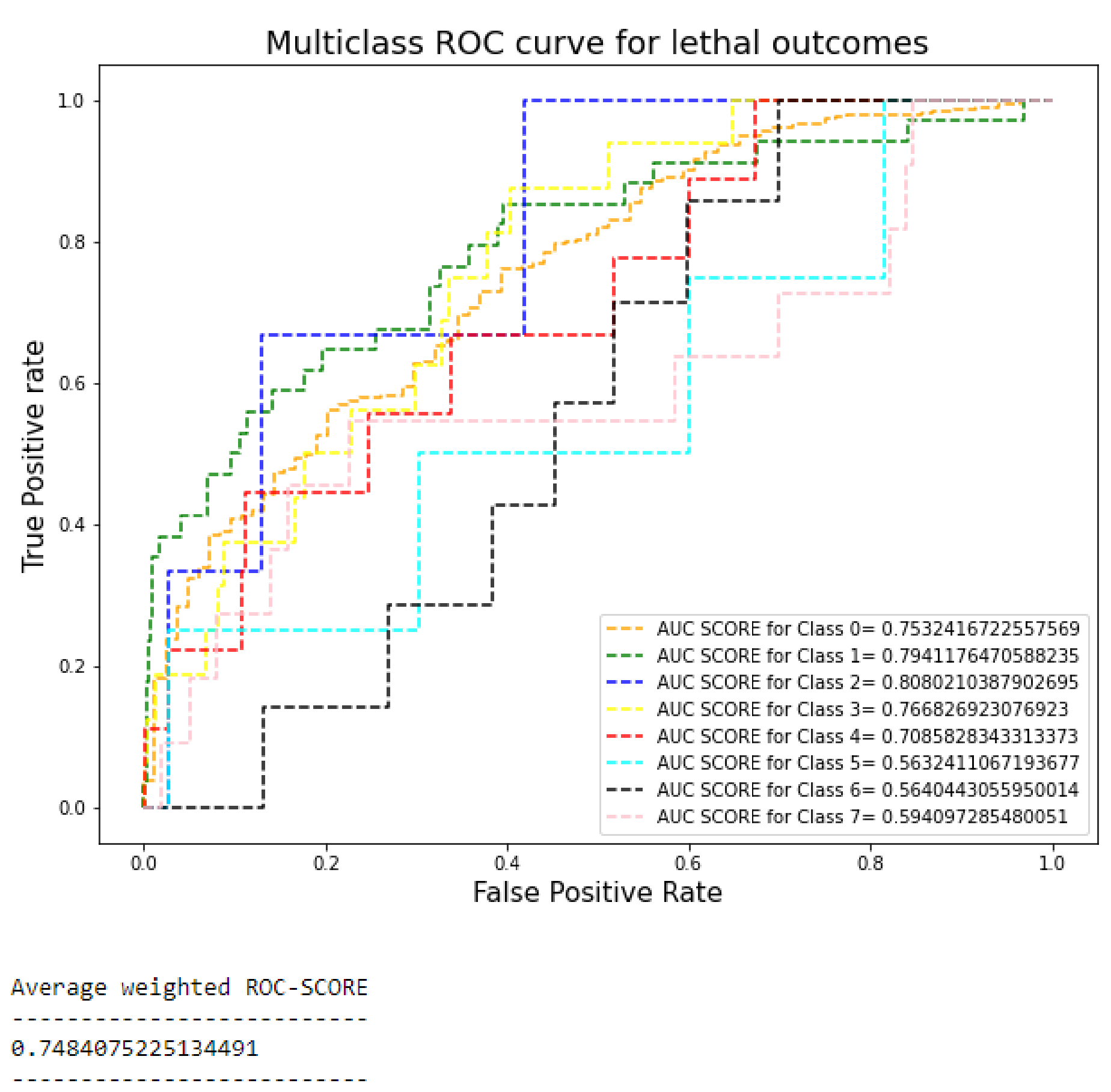

*Figure 5.36 AUCROC curve from ANN modelling after class imbalance handling (ADASYN)*

#### 5.3.6.3 Evaluation after applying Class Weighted Method

Training dataset before imbalance handling consists of 1,190 records, which was used to train the model with class weighted hyperparameter (class_weight='balanced') to handle the imbalanced class. Test dataset consists of 510 records which was used to evaluate the model performance. Below figure 5.37 refers to the confusion matrix where rows are representing actual class labels and columns are representing predicted class labels.

As part of the true positive (TP) counts, it can be shown that 328 cases out of 426 were correctly predicted for the 'unknow' class label. For the 'cardiogenic shock' class label, 21 cases out of 34 were correctly predicted. For the 'pulmonary edema' class label, nothing (0 case) was anticipated out of 3 cases. For the 'myocardial rupture' class label, 7 case was predicted correctly out of 16 cases. For the 'progress of congestive heart failure' class label, 1 case was correctly predicted out of 9 cases. For the 'thromboembolism' class label, nothing (0 case) was anticipated out of 4 cases. For the 'asystole' class label, nothing (0 case) was anticipated out of 7 cases. Furthermore, 2 cases were correctly predicted for the 'ventricular fibrillation' class label out of 11 cases.

As part of the false negative (FN) counts, 98 examples that truly correspond to the 'unknown' class label were forecasted as other class labels. 13 cases were projected as different class labels, despite the fact that they all belong to the 'cardiogenic shock' category. 3 cases, all of which correspond to the 'pulmonary edema' class label, were projected to have different class labels. 9 cases with the class label "myocardial rupture" were also predicted as other class labels. 8 instances were projected as different class labels, despite the fact that they all belong to the 'progress of congestive heart failure' class label. 4 examples, all of which fall under the 'thromboembolism' class label, were predicted to have different class labels. 7 instances with the class label 'asystole' were projected as different class labels, while 9 cases with the class label 'ventricular fibrillation' was also predicted as different class labels.

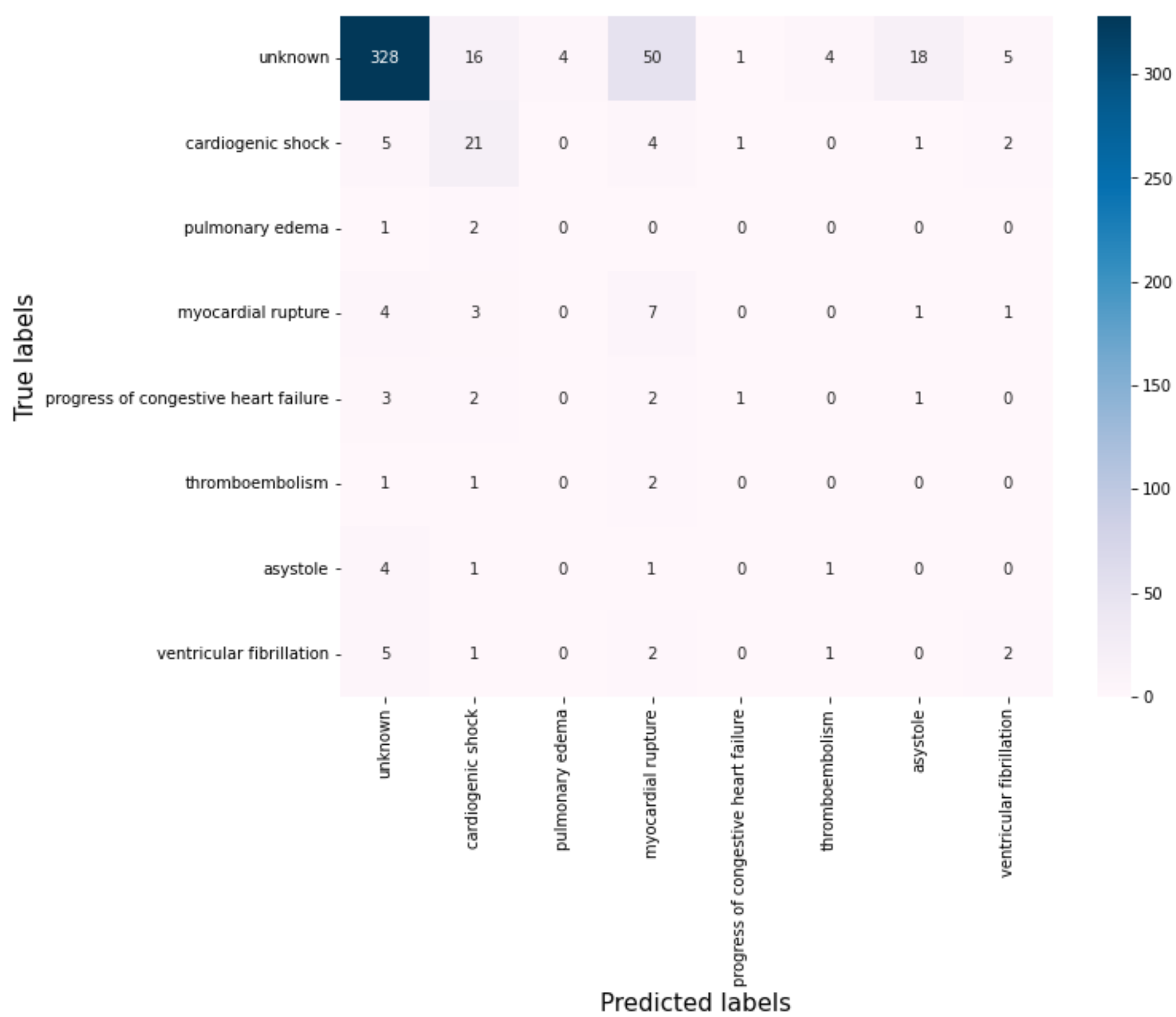

*Figure 5.37 A confusion matrix from ANN modelling using class weight method*

The classification report, shown in table 5.18, illustrates the various performance metrics for this multiclass classification task with class weighted method of imbalance handling, as well as the precision, recall, and f1-score for each class. Additionally, the weighted average precision, weighted average recall, weighted average f1-score, and accuracy for the measuring overall model performance are 82%, 70%, 75%, and 70%, respectively.

*Table 5.18 The classification report from ANN modelling with class weighted method*

| **Target class label** | **precision** | **recall** | **f1-score** | **Weighted avg precision** | **Weighted avg recall** | **Weighted avg f1-score** | **accuracy** |
|---|---|---|---|---|---|---|---|
| unknown | 0.93 | 0.77 | 0.84 | **0.82** | **0.70** | **0.75** | **0.70** |
| cardiogenic shock | 0.45 | 0.62 | 0.52 | | | | |
| pulmonary edema | 0.00 | 0.00 | 0.00 | | | | |
| myocardial rupture | 0.10 | 0.44 | 0.17 | | | | |

<table>
<tr><td>progress of congestive heart failure</td><td>0.33</td><td>0.11</td><td>0.17</td><td rowspan="4">0.82</td><td rowspan="4">0.70</td><td rowspan="4">0.75</td><td rowspan="4">0.70</td></tr>
<tr><td>thromboembolism</td><td>0.00</td><td>0.00</td><td>0.00</td></tr>
<tr><td>asystole</td><td>0.00</td><td>0.00</td><td>0.00</td></tr>
<tr><td>ventricular fibrillation</td><td>0.20</td><td>0.18</td><td>0.19</td></tr>
</table>

The ROC curve shown below in figure 5.38, illustrate how good the model is performing for predicting each class label with class weighted method of imbalance handling. Also, the weighted average AUCROC score for the overall model is 81.86%.

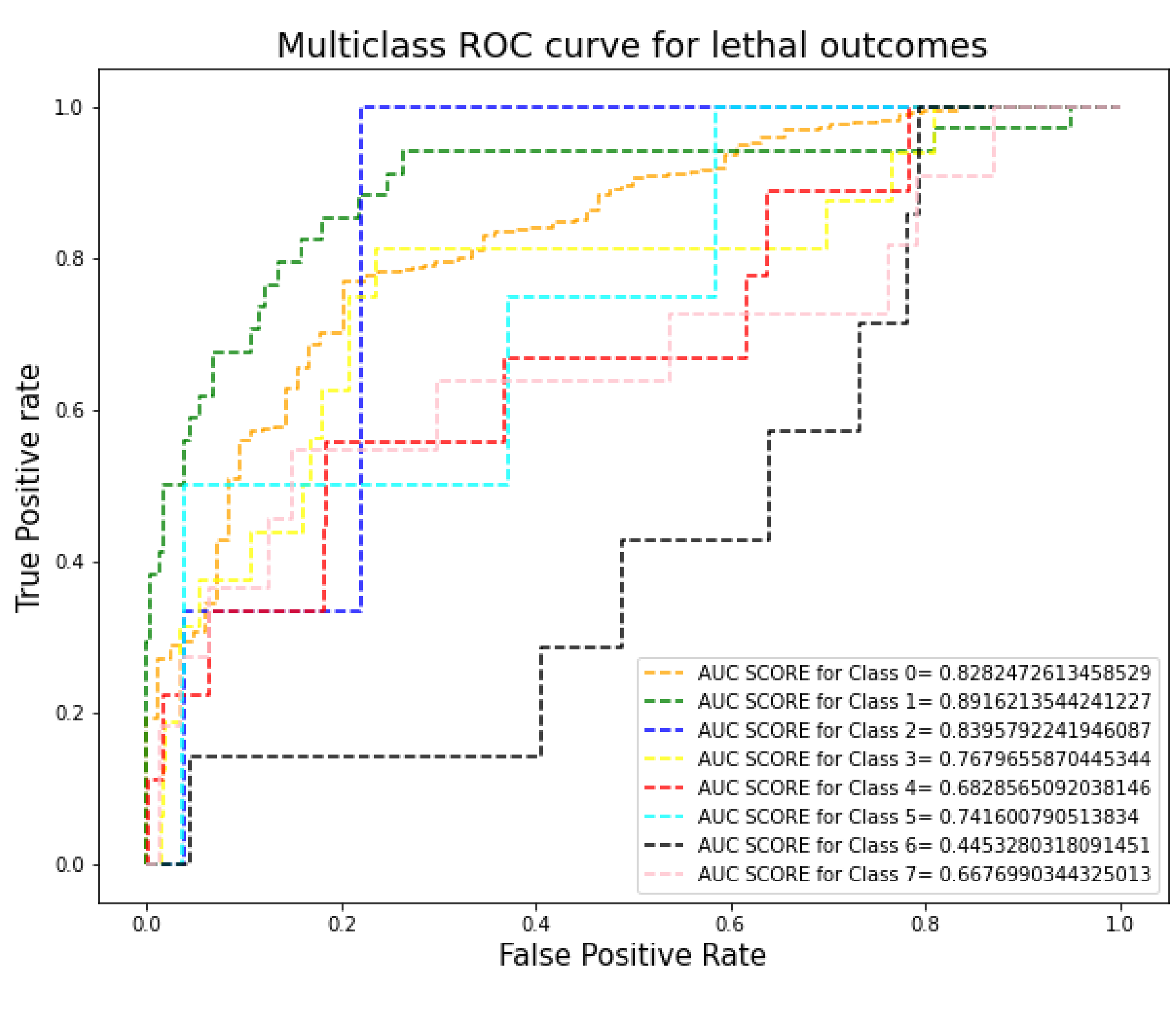

```
Average weighted ROC-SCORE
--------------------------
0.8186838113634531
--------------------------
```

*Figure 5.38 AUCROC curve from ANN modelling with class weighted method*

## 5.4 A Comparison between proposed Classifiers and their Results

The best classification model for the prediction of lethal result of acute myocardial infraction was determined using various performance measures utilized to assess the classifiers. The test dataset, which was used to evaluate the classifiers created on the training set, resulted in the production of classifiers or prediction models with comparable evaluation metric values and no significant overall differences in performances. This demonstrates that all the classifiers performed admirably in the assessment using the test set. However, some of the most popular evaluation criteria used in diagnosing diseases were compared across these classifiers to establish the best or most stable classifier among the six proposed predictive model.

From the below table 5.19, according to the overall weighted average metrices comparison, before class imbalance treatment, bagging SVM classifier (highlighted) performed well on predicting fatal outcome of acute myocardial infarction with weighted average precision, weighted average recall, weighted average f1-score, accuracy, and weighted average ROC-score of 80%, 86%, 82%, 86%, and 77.63%, respectively.

Again, from the below table 5.19, according to the overall weighted average metrices comparison, after class imbalance treatment using ADASYN method, deep artificial neural network (highlighted) performed well on predicting fatal outcome of acute myocardial infarction with weighted average precision, weighted average recall, weighted average f1-score, accuracy, and weighted average ROC-score of 80%, 81%, 80%,81% and 74.84%, respectively.

Also, from the below table 5.19, according to the overall weighted average metrices comparison, with class weighted method of imbalance handling, light gradient boosting machine (LGBM) (highlighted) performed well on predicting fatal outcome of acute myocardial infarction with weighted average precision, weighted average recall, weighted average f1-score, accuracy, and weighted average ROC-score of 78%, 85%, 81%,81% and 85.74%, respectively.

*Table 5.19 All weighted average performance evaluation metrices of proposed classifiers based on balancing methods*

| Balancing methods | Predictive models | Performance Evaluation Metrices (Weighted Average) | | | | |
|---|---|---|---|---|---|---|
| | | Weighted average Precision (%) | Weighted average Recall (%) | Weighted average F1-score (%) | Accuracy (%) | Weighted Average ROC Score (%) |
| **Without Class Imbalance Handling** | Logistic Regression | 78 | 85 | 81 | 85 | 85.29 |
| | LGBM | 79 | 86 | 81 | 86 | 85.62 |
| | Random Forest | 78 | 86 | 81 | 86 | 80.17 |
| | **Bagging SVM** | **80** | **86** | **82** | **86** | **77.63** |
| | Stacking Blending | 78 | 86 | 81 | 86 | 84.03 |
| | ANN | 76 | 85 | 80 | 85 | 78.20 |
| **Class Imbalance Treatment using ADASYN method** | Logistic Regression | 77 | 75 | 76 | 75 | 73.75 |
| | LGBM | 78 | 77 | 77 | 77 | 75.87 |
| | Random Forest | 79 | 77 | 78 | 77 | 77.22 |
| | Bagging SVM | 75 | 79 | 77 | 79 | 70.70 |
| | Stacking Blending | 77 | 82 | 79 | 82 | 75.48 |
| | **ANN** | **80** | **81** | **80** | **81** | **74.84** |
| **Class Weighted Method of Imbalance Handling** | Logistic Regression | 80 | 65 | 71 | 65 | 78.94 |
| | **LGBM** | **78** | **85** | **81** | **85** | **85.74** |
| | Random Forest | 82 | 78 | 80 | 78 | 81.68 |
| | Bagging SVM | 79 | 84 | 81 | 84 | 81.48 |
| | Stacking Blending | 85 | 62 | 71 | 62 | 81.26 |
| | ANN | 82 | 70 | 75 | 70 | 81.86 |

From the below table 5.20, according to the performance metrices comparison for predicting each fatal outcomes (each target class) of acute myocardial infarction, before class imbalance treatment, bagging SVM classifier (highlighted) performed well, where it has predicted four classes namely 'unknown' with precision - 87%, recall - 99%, f1-score - 93%, 'cardiogenic shock' with precision - 75%, recall - 44%, f1-score - 56%, 'myocardial rupture' with precision - 33%, recall - 6%, f1-score - 11%, and 'asystole' with precision - 50%, recall – 14%, f1-score - 22%. This model hasn't predicted anything for the rest four classes.

Again from the below table 5.20, according to the performance metrices comparison for predicting each fatal outcomes (each target class) of acute myocardial infarction, after class imbalance treatment using ADASYN method, deep artificial neural network (highlighted) performed well, where it has predicted five classes namely 'unknown' with precision - 88%, recall - 93%, f1-score - 91%, 'cardiogenic shock' with precision - 60%, recall - 35%, f1-score - 44%, 'myocardial rupture' with precision - 30%, recall - 19%, f1-score - 23%, 'progress of congestive heart failure' with precision - 33%, recall - 11%, f1-score - 17%, and finally, 'ventricular fibrillation' with precision - 12%, recall - 9%, f1-score - 11%. This model hasn't predicted anything for the rest three classes.

Again from the below table 5.20, according to the performance metrices comparison for predicting each fatal outcomes (each target class) of acute myocardial infarction, with class weighted method of imbalance handling, random forest (highlighted) performed well, where it has predicted seven classes namely 'unknown' with precision - 92%, recall - 86%, f1-score - 89%, 'cardiogenic shock' with precision - 50%, recall - 56%, f1-score - 53%, 'myocardial rupture' with precision - 41%, recall - 44%, f1-score - 42%, 'progress of congestive heart failure' with precision - 12%, recall - 11%, f1-score - 12%, 'asystole' with precision - 8%, recall -14%, f1-score - 10%, 'thromboembolism' with precision - 5%, recall - 25%, f1-score - 9%, and finally, 'pulmonary edema' with precision - 9%, recall - 33%, f1-score - 14%. This model hasn't predicted anything for the rest one class namely 'ventricular fibrillation'.

*Table 5.20 All performance evaluation metrices of proposed classifiers based on balancing methods for each target class*

| Balancing methods | Predictive models | Target Class Label | Performance Evaluation Metrices | | |
|---|---|---|---|---|---|
| | | | Precision (%) | Recall (%) | F1-Score (%) |
| **Without Class Imbalance Handling** | Logistic Regression | unknown | 87 | 98 | 92 |
| | | cardiogenic shock | 56 | 44 | 49 |
| | | pulmonary edema | 0 | 0 | 0 |
| | | myocardial rupture | 33 | 6 | 11 |
| | | progress of congestive heart failure | 0 | 0 | 0 |
| | | thromboembolism | 0 | 0 | 0 |
| | | asystole | 0 | 0 | 0 |
| | | ventricular fibrillation | 0 | 0 | 0 |
| | LGBM | unknown | 87 | 100 | 93 |
| | | cardiogenic shock | 80 | 35 | 49 |
| | | pulmonary edema | 100 | 33 | 50 |
| | | myocardial rupture | 33 | 6 | 11 |
| | | progress of congestive heart failure | 0 | 0 | 0 |
| | | thromboembolism | 0 | 0 | 0 |
| | | asystole | 0 | 0 | 0 |
| | | ventricular fibrillation | 0 | 0 | 0 |
| | Random Forest | unknown | 86 | 100 | 93 |
| | | cardiogenic shock | 93 | 41 | 57 |
| | | pulmonary edema | 0 | 0 | 0 |
| | | myocardial rupture | 0 | 0 | 0 |
| | | progress of congestive heart failure | 0 | 0 | 0 |
| | | thromboembolism | 0 | 0 | 0 |
| | | asystole | 0 | 0 | 0 |
| | | ventricular fibrillation | 0 | 0 | 0 |
| | **Bagging SVM** | **unknown** | **87** | **99** | **93** |
| | | **cardiogenic shock** | **75** | **44** | **56** |
| | | **pulmonary edema** | **0** | **0** | **0** |
| | | **myocardial rupture** | **33** | **6** | **11** |
| | | **progress of congestive heart failure** | **0** | **0** | **0** |
| | | **thromboembolism** | **0** | **0** | **0** |
| | | **asystole** | **50** | **14** | **22** |
| | | **ventricular fibrillation** | **0** | **0** | **0** |
| | Stacking Blending | unknown | 87 | 100 | 93 |
| | | cardiogenic shock | 73 | 47 | 57 |
| | | pulmonary edema | 0 | 0 | 0 |
| | | myocardial rupture | 0 | 0 | 0 |
| | | progress of congestive heart failure | 0 | 0 | 0 |
| | | thromboembolism | 0 | 0 | 0 |
| | | asystole | 0 | 0 | 0 |
| | | ventricular fibrillation | 0 | 0 | 0 |
| | ANN | unknown | 86 | 99 | 92 |
| | | cardiogenic shock | 67 | 35 | 46 |
| | | pulmonary edema | 0 | 0 | 0 |
| | | myocardial rupture | 0 | 0 | 0 |
| | | progress of congestive heart failure | 0 | 0 | 0 |
| | | thromboembolism | 0 | 0 | 0 |
| | | asystole | 0 | 0 | 0 |
| | | ventricular fibrillation | 0 | 0 | 0 |
| | Logistic Regression | unknown | 88 | 86 | 87 |
| | | cardiogenic shock | 44 | 35 | 39 |
| | | pulmonary edema | 0 | 0 | 0 |
| | | myocardial rupture | 0 | 0 | 0 |
| **Class Imbalance Treatment using ADASYN method** | | progress of congestive heart failure | 18 | 22 | 20 |
| | | thromboembolism | 0 | 0 | 0 |
| | | asystole | 9 | 14 | 11 |
| | | ventricular fibrillation | 5 | 9 | 7 |
| | LGBM | unknown | 89 | 88 | 88 |
| | | cardiogenic shock | 50 | 44 | 47 |
| | | pulmonary edema | 0 | 0 | 0 |
| | | myocardial rupture | 15 | 12 | 14 |
| | | progress of congestive heart failure | 0 | 0 | 0 |
| | | thromboembolism | 0 | 0 | 0 |
| | | asystole | 0 | 0 | 0 |
| | | ventricular fibrillation | 0 | 0 | 0 |
| | Random Forest | unknown | 90 | 87 | 89 |
| | | cardiogenic shock | 55 | 53 | 54 |
| | | pulmonary edema | 0 | 0 | 0 |
| | | myocardial rupture | 11 | 12 | 11 |
| | | progress of congestive heart failure | 0 | 0 | 0 |
| | | thromboembolism | 0 | 0 | 0 |
| | | asystole | 0 | 0 | 0 |
| | | ventricular fibrillation | 8 | 9 | 9 |
| | Bagging SVM | unknown | 87 | 92 | 89 |
| | | cardiogenic shock | 38 | 35 | 36 |
| | | pulmonary edema | 0 | 0 | 0 |
| | | myocardial rupture | 0 | 0 | 0 |
| | | progress of congestive heart failure | 0 | 0 | 0 |
| | | thromboembolism | 0 | 0 | 0 |
| | | asystole | 14 | 14 | 14 |

| | | ventricular fibrillation | 0 | 0 | 0 |
|---|---|---|---|---|---|
| | Stacking Blending | unknown | 87 | 94 | 91 |
| | | cardiogenic shock | 52 | 44 | 48 |
| | | pulmonary edema | 0 | 0 | 0 |
| | | myocardial rupture | 10 | 6 | 8 |
| | | progress of congestive heart failure | 0 | 0 | 0 |
| | | thromboembolism | 0 | 0 | 0 |
| | | asystole | 0 | 0 | 0 |
| | | ventricular fibrillation | 0 | 0 | 0 |
| | **ANN** | **unknown** | **88** | **93** | **91** |
| | | **cardiogenic shock** | **60** | **35** | **44** |
| | | **pulmonary edema** | **0** | **0** | **0** |
| | | **myocardial rupture** | **30** | **19** | **23** |
| | | **progress of congestive heart failure** | **33** | **11** | **17** |
| | | **thromboembolism** | **0** | **0** | **0** |
| | | **asystole** | **0** | **0** | **0** |
| | | **ventricular fibrillation** | **12** | **9** | **11** |
| **Class Weighted Method of Imbalance Handling** | Logistic Regression | unknown | 92 | 73 | 82 |
| | | cardiogenic shock | 37 | 44 | 40 |
| | | pulmonary edema | 0 | 0 | 0 |
| | | myocardial rupture | 7 | 25 | 11 |
| | | progress of congestive heart failure | 0 | 0 | 0 |
| | | thromboembolism | 0 | 0 | 0 |
| | | asystole | 4 | 14 | 6 |
| | | ventricular fibrillation | 4 | 9 | 6 |
| | LGBM | unknown | 87 | 98 | 92 |
| | | cardiogenic shock | 65 | 38 | 48 |
| | | pulmonary edema | 100 | 33 | 50 |
| | | myocardial rupture | 25 | 6 | 10 |
| | | progress of congestive heart failure | 0 | 0 | 0 |
| | | thromboembolism | 0 | 0 | 0 |
| | | asystole | 0 | 0 | 0 |
| | | ventricular fibrillation | 0 | 0 | 0 |
| | **Random Forest** | **unknown** | **92** | **86** | **89** |
| | | **cardiogenic shock** | **50** | **56** | **53** |
| | | **pulmonary edema** | **9** | **33** | **14** |
| | | **myocardial rupture** | **41** | **44** | **42** |
| | | **progress of congestive heart failure** | **12** | **11** | **12** |
| | | **thromboembolism** | **5** | **25** | **9** |
| | | **asystole** | **8** | **14** | **10** |
| | | **ventricular fibrillation** | **0** | **0** | **0** |
| | Bagging SVM | unknown | 89 | 95 | 92 |
| | | cardiogenic shock | 61 | 59 | 60 |
| | | pulmonary edema | 0 | 0 | 0 |
| | | myocardial rupture | 33 | 19 | 24 |
| | | progress of congestive heart failure | 0 | 0 | 0 |
| | | thromboembolism | 0 | 0 | 0 |
| | | asystole | 20 | 14 | 17 |
| | | ventricular fibrillation | 0 | 0 | 0 |
| | Stacking Blending | unknown | 95 | 68 | 79 |
| | | cardiogenic shock | 69 | 53 | 60 |
| | | pulmonary edema | 0 | 0 | 0 |
| | | myocardial rupture | 14 | 25 | 18 |
| | | progress of congestive heart failure | 6 | 22 | 10 |
| | | thromboembolism | 7 | 75 | 12 |
| | | asystole | 0 | 0 | 0 |
| | | ventricular fibrillation | 4 | 9 | 5 |
| | ANN | unknown | 93 | 77 | 84 |
| | | cardiogenic shock | 45 | 62 | 52 |
| | | pulmonary edema | 0 | 0 | 0 |
| | | myocardial rupture | 10 | 44 | 17 |
| | | progress of congestive heart failure | 33 | 11 | 17 |
| | | thromboembolism | 0 | 0 | 0 |
| | | asystole | 0 | 0 | 0 |
| | | ventricular fibrillation | 20 | 18 | 19 |

According to the above analysis and performance comparison, if predicting each target class or each lethal outcome of acute myocardial infarction is important, among the predictive models, random forest algorithm with class weighted method of imbalance handling, performed quite well in predicting almost all target classes (seven classes of lethal outcomes), with the exception of one class, ‘ventricular fibrillation’. Although LGBM with class weighted method of imbalance handling has given a better overall weighted performance than random forest, but it is unable to predict four classes: 'progress of congestive heart failure', 'thromboembolism', 'asystole', and 'ventricular fibrillation'. If focus is to predict each target class or each lethal outcome of acute myocardial infarction, LGBM did not performed well. On the other hand, if overall weighted average metrices is only to be considered instead of performance based on predicting each target class then bagging SVM classifier performed best when the model was generated from the dataset without class balance handling and ANN classifier performed best when the model was generated from the dataset after class balance treatment with ADASYN method.

## 5.5 Summary

This chapter summarised and analysed the findings of the entire study, as well as how the categorization methods findings were interpreted. Only the key risk factors that were discovered to play a vital role in offering better insights into the myocardial infarction dataset were included in section 5.2 (Significant biomarkers from visualisation and feature selection). The interrelationship among the independent factors and the interactions between the variables with the target lethal outcome (LET_IS) were extensively discussed. These important biomarkers will provide physicians with a quick overview of the key parameters associated with the fatal outcome of acute myocardial infraction, and for cardiac patients, this will serve as a reference in understanding the impact of certain risk variables while predicting or analysing the cause of death due to acute myocardial infarction.

The classification algorithms' confusion matrices were explained on the basis of recall, precision, f1-score, accuracy, and ROC score, which were built using test datasets and various balancing strategies. For each classifier, a quality assessment table was provided, and the balancing strategies were reviewed and contrasted. The results revealed that among the six classifiers (Logistic regression, LGBM, Random forest, Bagging SVM, Stacking blending, and ANN), Bagging SVM classifier performed best when the model was generated from the dataset without class balance handling, ANN classifier performed best when the model was generated from the dataset after class balance treatment with ADASYN method, and finally random forest classifier performed best when the model was generated from the dataset with class weighted method of handling imbalance dataset.

The weighted average values, as well as each class's recall, precision, and f1-score for each classifier for these two balancing approaches, were then evaluated to arrive at a single balancing method. Based on this, the class weighted method was chosen as the final imbalance handling technique across all six classification models, because the combination of the random forest algorithm and the class weighted method of imbalance handling was able to predict most of the target class or most of the lethal outcome of acute myocardial infarction better than other methods. On the other hand, the ADASYN class balancing method was also found to be suitable for myocardial infarction datasets, as the proportion of the majority and minority classes could be generated even with synthetic samples, and the deep neural network based on this balancing method produced better overall weighted average performance measures than other methodologies. As a result, the prediction of the deadly outcome of acute myocardial infarction will be more precise and less prone to biases.

# CHAPTER 6

# CONCLUSION AND FUTURE RECOMENDATION

## 6.1 Introduction

This will be followed by the conclusion drawn to portray whether the aim of this study has been achieved or not. Under the contribution of study, implementation of different methods on the data or looking into the study from a different angle to suggest ways to improve the proposed model or future recommendation or to determine a better model than the one in this study. Following that, a conclusion will be made to show if the study's goal has been accomplished or not. Implementing other methodologies on the data or examining this study from a different perspective to offer ways to improve the proposed model or to discover a better model than the one in this study falls under the future recommendations.

## 6.2 Summary of the Study

The myocardial dataset was used in this investigation, which included 1,700 records of patients, their demographics, and risk factors who were admitted to the hospital after suffering from an acute myocardial infarction. This dataset had 112 variables, including all risk factors for myocardial infarction as well as 12 complications of the condition, one of which was chosen as a dependent/target variable for this investigation, namely Lethal outcome (cause) (LET IS). To make the analysis and modelling process efficient and effective, unrelated variables like patient 'ID' were removed, 21 categorical variables were turned into numerical variables using dummification, and 4 numerical features were turned into categorical columns as part of data pre-processing and transformation. Around 110 features containing missing data were found and imputed using appropriate missing value imputation, namely, iterative and mode imputation technique provide a better result, as part of missing value analysis. Additionally, it has been discovered that four characteristics, notably IBS NASL, S AD KBRIG, D AD KBRIG, and KFK BLOOD, have more than a 45% null value. Imputing more than 45% of missing data can inject bias into the model and impair the efficiency of the classification result, so dropping those features would be a better approach.

Following that, univariate analysis was utilized to understand the distribution of data in each variable as part of the exploratory data analysis on the cardiac dataset with imputed missing values. The count plot or percentage plot for a few significant variables, as well as the overall statistical analysis, were thoroughly examined. As part of the bivariate and correlation analysis, the percentage analysis for each target class, as well as the variable relationships, were checked and reported. Each visualisation report revealed if the predictor variable has any influence on the occurrence of any of the fatal outcomes of acute myocardial infarction. The report and graphical representation also highlight the link between the numerical aspects that decide if the combined relationship has any effect on the target variable, LET_IS (lethal outcome).

The prediction capacity of a machine learning model is directly proportional to the number of characteristics or features used to train it. The performance of ML models will degrade if irrelevant features are used to construct them. Feature selection is the process of reducing the number of unneeded attributes, or dimensionality reduction, based on any statistical relationship, either automatically or manually. This could improve the performance or predictability of machine learning models. As a result, on the presented dataset, the wrapping and embedding methods of feature selection, namely RFE (recursive feature elimination) and extra tree classifier, were used to choose the most appropriate and relevant feature for classification model building. The top 70 features were chosen as the most important elements in predicting the target class, and they were later used to develop the classification models.

The interactions and visualisations, which are based on selected key features, provide a critical and deeper insight into the myocardial infraction dataset that was examined and processed, providing a clear image of all the variables in the dataset as well as a summary of their relationships. Some of the important biomarkers that are identified from this study are: age, systolic blood pressure, diastolic blood pressure, gender, atrial fibrillation (irregular ECG rhythm), serum potassium content, serum AlAT concentration, coronary heart disease in the previous weeks, subjected to acetylsalicylic acid in the ICU, subjected to anticoagulants (heparin) in the ICU, Use of NSAIDs and complete RBBB on ECG at the time of admission to hospital. These key biomarkers or variables will give doctors a quick overview of the key factors linked to the fatal consequences of acute myocardial infarction, as well as serve as a guide in understanding the impact of certain risk factors when predicting the cause of death from acute myocardial infarction.

Using the default split, the original dataset with imputed missing values was divided into two sets, with the training set accounting for 70% and the test set for 30%. Because the offered dataset is very imbalanced in nature and as already discussed earlier in section 4.3 (subsection 4.3.7.1), that SVM-SMOTE class imbalance treatment method has already been discarded from this study due to its inability to equally balance the target class in the presented myocardial infarction dataset, hence class balancing approach, namely ADASYN, and a class weighted method, were used to solve the issue of class imbalance on the training dataset. The six classifiers were tuned using the grid Search cross validation technique with five folds. These six classifiers (Logistic regression, LGBM, Bagging SVM, Random Forest, Stacking Blending, and ANN) were trained on both the imbalanced and balanced training datasets before being validated/evaluated on the test dataset.

There after the classification algorithms' confusion matrices were explained based on recall, precision, f1-score, accuracy, and ROC score, which were built using test datasets and various balancing strategies. The results revealed that among the six classifiers (Logistic regression, LGBM, Random Forest, Bagging SVM, stacking blending, and ANN), Bagging SVM classifier performed best when the model was generated from the dataset without any class balancing strategies with weighted average precision, weighted average recall, weighted average f1-score, accuracy, and weighted average ROC-score of 80%, 86%, 82%, 86%, and 77.63%, respectively. ANN classifier performed best when the model was generated from the dataset after class imbalance treatment with ADASYN method which provided weighted average precision, weighted average recall, weighted average f1-score, accuracy, and weighted average ROC-score of 80%, 81%, 80%,81% and 74.84%, respectively. Finally, random forest classifier performed best when the model was generated from the dataset with class weighted method of handling imbalance dataset with weighted average precision, weighted average recall, weighted average f1-score, accuracy, and weighted average ROC-score of 78%, 85%, 81%,81% and 85.74%, respectively.

For these two balancing approaches, the weighted average values, as well as each class's recall, precision, and f1-score for each classifier, were then analysed to arrive at a single balance strategy. Because the combination of the random forest algorithm and the class weighted method of imbalance handling was able to predict most of the target class or most of the lethal outcome of acute myocardial infarction better than other methods, where it has predicted seven classes namely 'unknown' with precision - 92%, recall - 86%, f1-score - 89%, 'cardiogenic shock' with precision - 50%, recall - 56%, f1-score - 53%, 'myocardial rupture' with precision - 41%, recall - 44%, f1-score - 42%, 'progress of congestive heart failure' with precision - 12%, recall - 11%, f1-score - 12%, 'asystole' with precision - 8%, recall -14%, f1-score - 10%,

'thromboembolism' with precision - 5%, recall - 25%, f1-score - 9%, and finally, 'pulmonary edema' with precision - 9%, recall - 33%, f1-score - 14%. The class weighted method was chosen as the final imbalance handling technique across all six classification models.
The ADASYN class balancing method, on the other hand, was found to be suitable for myocardial infarction datasets because the proportion of majority and minority classes could be generated even with synthetic samples, and the deep neural network based on this balancing method produced better overall weighted average performance measures than other methodologies. As a result, predicting the fatal outcome of an acute myocardial infarction will be more exact and less prone to prejudice.

## 6.3 Future Recommendations

In this study, there is room for improvement that can be addressed as part of future recommendations, notably, further effort can be done to improve class imbalance handling strategies. Hybrid imbalance handling methods (which can be a combination of oversampling and under sampling) like SMOTE Tomek method or Spread sample method along with SMOTE can be used for better class imbalance treatment, rather than only generating synthetic data to balance the minority class with the majority, as an improvement to ADASYN or class weighted method of imbalance handling.

This myocardial dataset contains only 1,700 data or observations for cardiac patients, which is inadequate to train a classification model or learn more diverse information about this condition. More observations or data from more cardiac patients suffering from acute myocardial infarction from other health institutions or hospitals needs to be collected.

Iterative imputation was employed in this thesis as part of the missing value imputation methodology, however the KNN imputation method of missing value imputation can be use, and the classification model performance can then be compared.

Along with the wrapper-based and embedded selection strategies employed in this investigation, a more robust feature selection technique incorporating information value, statistical testing such as ANOVA, Chi-Square, and other techniques can be applied to improve the study. Information value, or IV, will determine feature importance based on weight of evidence and rank each feature independently of weight or cover value or node importance, like the feature importance of any tree model. Statistical testing, on the other hand, will determine the influence of individual predictors on the dependent variable. The most significant features that can produce more accurate and stable model performance include whether the independent feature is a good separator of the target class, whether the variance or mean value of the predictor attributes are equal or not among the classes, and so on.

As previously stated in this thesis, the higher the true positive rate, recall, or sensitivity of an ideal disease prediction model, the better the diseased patient's forecast. This is the goal of any disease prediction model. Higher true negative rates or specificity are also desired, but they are less important than true positive rates or recall. It's vital that the prediction model accurately diagnoses the condition and prevents misdiagnosis. While predicting each target class (multiclass) in this thesis, it was discovered that recall or sensitivity for any of the six classification models is lower than any other performance metrices during model development on the test set. As a result, while predicting each target class, an appropriate output class probability threshold (preferably lower threshold limit) can be set for each classification model, resulting in a higher recall score and a lower chance of misdiagnosis.